\documentclass[english]{article}

\makeatletter
\def\input@path{{LatexCommands/}}
\makeatother

\usepackage[utf8]{inputenc}
\usepackage[a4paper,margin=1in]{geometry}
\usepackage{amsmath,amssymb,amsthm,mathtools}
\usepackage{enumitem}
\usepackage{microtype}
\usepackage{hyperref}
 
\usepackage[english]{babel} %gestion of english characters
\usepackage[utf8]{inputenc}  % The package translates various standard and other input encodings into a ‘LaTeX internal language’.
\usepackage[T1]{fontenc}  % The package allows the user to select font encodings
\usepackage{lmodern} %vectorial characters
\usepackage{relsize} %have better character sizes

\usepackage{fancyhdr} %to have a fancy presentation of the article (headers, titles...)
\usepackage{geometry} %to have a good page layout
\usepackage{setspace} %used to set spaces between lines
\usepackage{xspace} % tool to help not having undesired spaces in the document
\usepackage{enumitem} %to make enumerations, itemize, etc

\usepackage{soul} %basic stuff
\usepackage[normalem]{ulem} %more precise entries

\usepackage{xr-hyper} % gestion of refs between different files
\usepackage{subfiles} %packages to manage files and subfiles
\usepackage{hyperref} % gestion of hypertext links
\hypersetup{unicode,colorlinks,linkcolor=blue,urlcolor=blue,runcolor=blue,breaklinks=true} %initialises hyperref package
\usepackage{jobname-suffix}

\usepackage{float} % most flexible package to make figures out of nothing
\usepackage{graphicx} % used to insert images (the easiest tool to use)
\usepackage{wrapfig} % makes figures in any way wanted (very precise precision etc) 
\usepackage{caption, subcaption} % use to caption figures
\usepackage{multirow} % used to modify tables with high flexibility
\usepackage{media9} %used to have animated 3D objects, sounds, etc
\usepackage{xargs, xparse} % handles multiple optional parameters 

\usepackage{lastpage} %new command to address last page labels

\usepackage{blindtext} %creates random text
\usepackage{comment} %to make commentaries

\usepackage{ifthen} %to be able to use boolean propositions in the latex code
\usepackage{expl3} %to make more complex boolean logic operations
\usepackage{multido} %be able to make code loops

\usepackage{pdfcomment}

\usepackage{amsmath,amssymb,mathrsfs,mathtools, amsthm, amsbsy} % packages of mathematical symbols
\usepackage{empheq} %extension of amsmath
\usepackage{esint} %integral symbols
\usepackage{siunitx} %write numbers in scientific writing
\usepackage{cancel} %to bar some expressions
\usepackage{cases} % to make case by case equations
\usepackage{mdframed} % allows to frame equations
\usepackage{bm, bbm} %to type bold math fonts
\usepackage{upgreek} % to type non italic versions of greek letters
\usepackage{mleftright} % makes a better displaying for parentheses

\usepackage{tikz}%automatically draw math functions
\usepackage{pgfplots}%automatically draw math functions
\usetikzlibrary{plotmarks} %initiates tikz
\pgfplotsset{compat=1.14} %initiates pgfplot
\usetikzlibrary{math} %to perform math operations in latex
\usetikzlibrary{decorations.pathreplacing,shapes,shapes.misc} %to make shapes and decorations

\usepackage{algorithm, algpseudocode} %to make algorithms

\usepackage{cleveref} % make nice references

\newcommand\newname[1]{\textit{#1}}

\newcommand{\safemath}[2]{\newcommand{#1}{\ensuremath{#2}\xspace}}

\newcommand{\dynamicBiblio}{%
  \ifthenelse{\equal{\jobname}{\detokenize{main}}}
    {}
    {\bibliographystyle{IEEEtranS}
    \bibliography{Bibliography/library.bib}}%
}
\newtheorem{theorem}{Theorem}[section]

\newtheorem{lemma}[theorem]{Lemma}
\newtheorem{corollary}[theorem]{Corollary}
\newtheorem{definition}{Definition}[section]

\newtheorem*{claim*}{Claim}

\newtheoremstyle{flexith}{}{}{\mdseries}{}{\itshape}{}{.5em}{\textit{\textbf{Theorem}}  \thmnumber{#2}: (#3).}
\theoremstyle{flexith}

\newtheoremstyle{flexide}{}{}{\mdseries}{}{\itshape}{}{.5em}{\textit{\textbf{Definition}}  \thmnumber{#2}: (#3).}
\theoremstyle{flexide}

\newtheoremstyle{flexile}{}{}{\mdseries}{}{\itshape}{}{.5em}{\textit{\textbf{Lemma}}  \thmnumber{#2}: (#3).}
\theoremstyle{flexile}

\newcommand\bfI{\ensuremath{{\bf I}}}

\bmdefine{\bmualphad}{\upalpha}
\bmdefine{\bmubetad}{\upbeta}
\bmdefine{\bmuchid}{\upchi}
\bmdefine{\bmudeltad}{\updelta}
\bmdefine{\bmuepsilond}{\upepsilon}
\bmdefine{\bmuvarepsilond}{\upvarepsilon}
\bmdefine{\bmuetad}{\upeta}
\bmdefine{\bmugammad}{\upgamma}
\bmdefine{\bmuiotad}{\upiota}
\bmdefine{\bmukappad}{\upkappa}
\bmdefine{\bmulambdad}{\uplambda}
\bmdefine{\bmumud}{\upmu}
\bmdefine{\bmunud}{\upnu}
\bmdefine{\bmuomegad}{\upomega}
\bmdefine{\bmuphid}{\upphi}
\bmdefine{\bmuvarphid}{\upvarphi}
\bmdefine{\bmupid}{\uppi}
\bmdefine{\bmuvarpid}{\upvarpi}
\bmdefine{\bmupsid}{\uppsi}
\bmdefine{\bmurhod}{\uprho}
\bmdefine{\bmuvarrhod}{\upvarrho}
\bmdefine{\bmusigmad}{\upsigma}
\bmdefine{\bmuvarsigmad}{\upvarsigma}
\bmdefine{\bmutaud}{\uptau}
\bmdefine{\bmuthetad}{\uptheta}
\bmdefine{\bmuvarthetad}{\upvartheta}
\bmdefine{\bmuupsilond}{\upupsilon}
\bmdefine{\bmuxid}{\upxi}
\bmdefine{\bmuzetad}{\upzeta}

\bmdefine{\bmiad}{a}
\bmdefine{\bmibd}{b}
\bmdefine{\bmicd}{c}
\bmdefine{\bmidd}{d}
\bmdefine{\bmied}{e}
\bmdefine{\bmifd}{f}
\bmdefine{\bmigd}{g}
\bmdefine{\bmihd}{h}
\bmdefine{\bmiid}{i}
\bmdefine{\bmijd}{j}
\bmdefine{\bmikd}{k}
\bmdefine{\bmild}{l}
\bmdefine{\bmimd}{m}
\bmdefine{\bmind}{n}
\bmdefine{\bmiod}{o}
\bmdefine{\bmipd}{p}
\bmdefine{\bmiqd}{q}
\bmdefine{\bmird}{r}
\bmdefine{\bmisd}{s}
\bmdefine{\bmitd}{t}
\bmdefine{\bmiud}{u}
\bmdefine{\bmivd}{v}
\bmdefine{\bmiwd}{w}
\bmdefine{\bmixd}{x}
\bmdefine{\bmiyd}{y}
\bmdefine{\bmizd}{z}

\bmdefine{\bmialphad}{\alpha}
\bmdefine{\bmibetad}{\beta}
\bmdefine{\bmichid}{\chi}
\bmdefine{\bmideltad}{\delta}
\bmdefine{\bmiepsilond}{\epsilon}
\bmdefine{\bmivarepsilond}{\varepsilon}
\bmdefine{\bmietad}{\eta}
\bmdefine{\bmigammad}{\gamma}
\bmdefine{\bmiiotad}{\iota}
\bmdefine{\bmikappad}{\kappa}
\bmdefine{\bmivarkappad}{\varkappa}
\bmdefine{\bmilambdad}{\lambda}
\bmdefine{\bmimud}{\mu}
\bmdefine{\bminud}{\nu}
\bmdefine{\bmiomegad}{\omega}
\bmdefine{\bmiphid}{\phi}
\bmdefine{\bmivarphid}{\varphi}
\bmdefine{\bmipid}{\pi}
\bmdefine{\bmivarpid}{\varpi}
\bmdefine{\bmipsid}{\psi}
\bmdefine{\bmirhod}{\rho}
\bmdefine{\bmivarrhod}{\varrho}
\bmdefine{\bmisigmad}{\sigma}
\bmdefine{\bmivarsigmad}{\varsigma}
\bmdefine{\bmitaud}{\tau}
\bmdefine{\bmithetad}{\theta}
\bmdefine{\bmivarthetad}{\vartheta}
\bmdefine{\bmiupsilond}{\upsilon}
\bmdefine{\bmixid}{\xi}
\bmdefine{\bmizetad}{\zeta}

\bmdefine{\bmuDeltad}{\Updelta}
\bmdefine{\bmuGammad}{\Upgamma}
\bmdefine{\bmuLambdad}{\Uplambda}
\bmdefine{\bmuOmegad}{\Upomega}
\bmdefine{\bmuPhid}{\Upphi}
\bmdefine{\bmuPid}{\Uppi}
\bmdefine{\bmuPsid}{\Uppsi}
\bmdefine{\bmuSigmad}{\Upsigma}
\bmdefine{\bmuThetad}{\Uptheta}
\bmdefine{\bmuUpsilond}{\Upupsilon}
\bmdefine{\bmuXid}{\Upxi}

\bmdefine{\bmiAd}{A}
\bmdefine{\bmiBd}{B}
\bmdefine{\bmiCd}{C}
\bmdefine{\bmiDd}{D}
\bmdefine{\bmiEd}{E}
\bmdefine{\bmiFd}{F}
\bmdefine{\bmiGd}{G}
\bmdefine{\bmiHd}{H}
\bmdefine{\bmiId}{I}
\bmdefine{\bmiJd}{J}
\bmdefine{\bmiKd}{K}
\bmdefine{\bmiLd}{L}
\bmdefine{\bmiMd}{M}
\bmdefine{\bmiOd}{N}
\bmdefine{\bmiPd}{O}
\bmdefine{\bmiQd}{P}
\bmdefine{\bmiRd}{R}
\bmdefine{\bmiSd}{S}
\bmdefine{\bmiTd}{T}
\bmdefine{\bmiUd}{U}
\bmdefine{\bmiVd}{V}
\bmdefine{\bmiWd}{W}
\bmdefine{\bmiXd}{X}
\bmdefine{\bmiYd}{Y}
\bmdefine{\bmiZd}{Z}

\bmdefine{\bmiDeltad}{\Delta}
\bmdefine{\bmiGammad}{\Gamma}
\bmdefine{\bmiLambdad}{\Lambda}
\bmdefine{\bmiOmegad}{\Omega}
\bmdefine{\bmiPhid}{\Phi}
\bmdefine{\bmiPid}{\Pi}
\bmdefine{\bmiPsid}{\Psi}
\bmdefine{\bmiSigmad}{\Sigma}
\bmdefine{\bmiThetad}{\Theta}
\bmdefine{\bmiUpsilond}{\Upsilon}
\bmdefine{\bmiXid}{\Xi}

\newcommand\Bc{\mathcal{B}}
\newcommand\Cc{\mathcal{C}}
\newcommand\Dc{\mathcal{D}}

\newcommand\Fc{\mathcal{F}}

\newcommand\Hc{\mathcal{H}}

\newcommand\Lc{\mathcal{L}}
\newcommand\Mc{\mathcal{M}}
\newcommand\Nc{\mathcal{N}}
\newcommand\Oc{\mathcal{O}}

\newcommand\Rc{\mathcal{R}}
\newcommand\Sc{\mathcal{S}}

\newcommand\Wc{\mathcal{W}}
\newcommand\Xc{\mathcal{X}}
\newcommand\Yc{\mathcal{Y}}

\newcommand\Db{\mathbb{D}}

\newcommand\Nb{\mathbb{N}}

\newcommand\Rb{\mathbb{R}}

\newcommand\Zb{\mathbb{Z}}

\newcommand\N{\Nb}
\newcommand\No{\ensuremath{{\Nb_0}}}
\newcommand\Z{\Zb}

\newcommand\R{\Rb}

\newcommand\Reals{\ensuremath{\mathbb R}}

\newcommand\Naturals{\ensuremath{\mathbb N}}

\NewDocumentCommand\listIntegers{O{none} O{0}}{ % set of consecutive integers
  \ensuremath{\left\{#2,\ldots, #1 \right\}} 
}

\NewDocumentCommand\funcdef{ m m m m m O{rrcl}}{ %definition of a function
  \ensuremath{
    \begin{array}{#6}
      #1 : & #2 & \to & #3\\
           & #4 & \mapsto & #5
    \end{array}
  }
}

\renewcommand\sup[1][]{
    \ifthenelse{\equal{#1}{}}{\ensuremath{\mathop{\operatorname{sup}}}}{
        \ensuremath{\mathop{\operatorname{sup}}}\left(#1\right)
    }
}

\newcommand{\limi}[1]{\ensuremath{\underset{#1 \to \infty}{\lim}}} %limit to infinity with 1 argument, that is the variable of the limit

\newcommand\bigOx[2][1]{
    \ifthenelse{\equal{#1}{1}}{\underset{|#1| \to \infty}{\mathcal{O}}(1)}{\underset{#2 \to \infty}{\mathcal{O}}(#2#1)}} %big O notations
    
\newcommand\bigOw[2][1]{
    \ifthenelse{\equal{#1}{1}}{\underset{|#1| \to \infty}{\mathcal{O}}(1)}{\underset{|#2| \to \infty}{\mathcal{O}}(|#2|#1)}} %big O notations on the length of a sequence

\makeatletter
\newcommand{\oset}[3][0ex]{%this command is a flexible way to put something on top of an element in an equation
  \mathrel{\mathop{#3}\limits^{
    \vbox to#1{\kern-2\ex@
    \hbox{$\scriptstyle #2$}\vss}}}}
\makeatother

\safemath\CantorSpace{\{0,1\}^\Naturals}
\safemath\SigmaCantorSpace{\Sigma^\Naturals}
\safemath\BaireSpace{\Naturals^\Naturals}

\newcommand\len[1]{\ensuremath{\ell\left(#1\right)}}
\NewDocumentCommand\namingSystem{ O{} O{} }{ % naming system
    \ifthenelse{\equal{#1}{} \AND \equal{#2}{}}{
        \ensuremath{\delta}
    }{
        \ifthenelse{\equal{#2}{}}{
            \ensuremath{\delta_{#1}}
        }{
            \ifthenelse{\equal{#1}{}}{
                \ensuremath{\delta\left(#2\right)}
            }{
                \ensuremath{\delta_{#1}\left(#2\right)}
            }
        }
    }
}

\NewDocumentCommand\baseRepresentation{O{} O{}}{ % base representation naming system
    \namingSystem[{\Reals_{#1}}][#2]
}
\newcommand\base{\ensuremath{\beta}} % standard base notation
\NewDocumentCommand\setBetaExpansions{ O{} O{\base}}{
    \ifthenelse{\equal{#1}{}}{\ensuremath{\Omega_{#2}}}{\ensuremath{\Omega_{#2}(#1)}
    }
}

\NewDocumentCommand\eval{O{} O{\base}}{
    \ifthenelse{\equal{#1}{}}{\ensuremath{\delta_{#2}}}{
        \ensuremath{\delta_{#2}\left(#1\right)}
    }
}

\newcommand\betaint{0}

\NewDocumentCommand\ndig{ O{\base} O{\betaint}}{
    \ifthenelse{
        \equal{#2}{0}
    }{
        \ensuremath{\lceil #1 \rceil}
    }{
        \ensuremath{\the\numexpr #2 + 1 \relax}
    }
} 

\NewDocumentCommand\mdig{ O{\base} O{\betaint}}{
    \ifthenelse{
        \equal{#2}{0}
    }{
        \ensuremath{\lceil #1 \rceil - 1}}
    {
        {\ensuremath{ #2 }}
    }
} 

\NewDocumentCommand\digs{ O{\base} O{\betaint}}{
    \ensuremath{\listIntegers[{\mdig[#1][#2]}][0]}
    } 

\NewDocumentCommand\rangeEvalMax{ O{\base} O{\betaint}}{
    \ifthenelse{
        \equal{#2}{0}
    }{
        \ensuremath{\frac{\mdig[#1]}{#1 - 1}}
    }{
        \ensuremath{ \frac{#2}{#1 - 1}}
    }
} 

\NewDocumentCommand\rangeEval{O{\base} O{\betaint}}{
    \left[0, \rangeEvalMax[\base][\betaint] \right]
} 

\NewDocumentCommand\betadicSolver{O{} O{\integern} O{\base}}{
    \ifthenelse{\equal{#1}{}}{
        \ensuremath{\mathcal{S}_{#3,#2}}
    }{
        \ensuremath{\mathcal{S}_{#3,#2}}\left(#1\right)
    }
} 

\NewDocumentCommand\betaPrefixSolver{O{} O{\integern} O{\base}}{
    \ifthenelse{\equal{#1}{}}{
        \ensuremath{\mathcal{S}^{\operatorname{lex}}_{#3,#2}}
    }{
        \ensuremath{\mathcal{S}^{\operatorname{lex}}_{#3,#2}\left(#1\right)}
    }
}

\NewDocumentCommand\indexBetadicSolver{ O{} O{\integern} O{\base}}{
    \ifthenelse{\equal{#1}{}}{
        \ensuremath{I_{#3,#2}}
    }{
        \ensuremath{I_{#3,#2}}\left(#1\right)
    }
}

\NewDocumentCommand\indexBetaPrefixSolver{O{} O{\integern} O{\base}}{
    \ifthenelse{\equal{#1}{}}{
        \ensuremath{I^{\operatorname{lex}}_{#3,#2}}
    }{
        \ensuremath{I^{\operatorname{lex}}_{#3,#2}\left(#1\right)}
    }
}

\NewDocumentCommand\setBetaPrefixes{ O{} O{\integern} O{\base}}{ 
    \ifthenelse{\equal{#1}{} \AND \equal{#2}{}}{\ensuremath{W_{#3}}}{
        \ifthenelse{\equal{#2}{}}{
            \ensuremath{W_{#3}^\ast(#1)}
        }{\ensuremath{W_{#3,#2}(#1)}}
    }
}

\NewDocumentCommand\setBetadics{ O{} O{\integern} O{\base}}{
\ifthenelse{\equal{#1}{} \AND \equal{#2}{}}{\ensuremath{\mathbb{Q}_{#3}}}{
    \ifthenelse{\equal{#2}{}}{
        \ensuremath{\mathbb{Q}_{#3}(#1)}
        }{
            \ifthenelse{\equal{#1}{}}{
            \ensuremath{\mathbb{Q}_{#3,#2}}
            }{\ensuremath{\mathbb{Q}_{#3,#2}(#1)}}
        }
    }
}

\NewDocumentCommand\setGreedyBetaPrefixes{O{} O{\base}}{
\ifthenelse{
    \equal{#1}{}}{\ensuremath{\widehat{W}_{#2}}}{
        \ensuremath{\widehat{W}_{#2,#1}}
    }
}

\NewDocumentCommand\setGreedyBetadics{O{} O{\base}}{
    \ifthenelse{
        \equal{#1}{}}{\ensuremath{\widehat{\mathbb{Q}}_{#2}}}{
            \ensuremath{\widehat{\mathbb{Q}}_{#2,#1}}
        }
}

\NewDocumentCommand\setGreedyBetadicsWithOne{O{} O{\base}}{
    \ifthenelse{
        \equal{#1}{}}{\ensuremath{\widehat{\mathbb{Q}}^+_{#2}}}{
            \ensuremath{\widehat{\mathbb{Q}}^+_{#2,#1}}
        }
} 

\NewDocumentCommand\numberGreedyPrefixes{O{} O{\base}}{ 
    \ifthenelse{
    \equal{#1}{}}{\ensuremath{N_{#2}}}{
        \ensuremath{N_{#2}({#1})}
    }
}

\NewDocumentCommand\indicesGreedyPrefixes{O{} O{\base}}{
            \ensuremath{\listIntegers[{\numberGreedyPrefixes[#1][#2]-1}][0]}
}

\NewDocumentCommand\enumerationBetadics{O{\integeri} O{\integern} O{\base}}{ 
    \ifthenelse{
    \equal{#1}{}}{\ensuremath{q_{#3,#2}}}{
        {\ensuremath{q_{#3,#2}(#1)}}
    }
}

\NewDocumentCommand\enumerationGreedyPrefixes{O{\integeri} O{\integern} O{\base}}{
    \ifthenelse{
    \equal{#1}{}}{\ensuremath{w_{#3,#2}}}{
        {\ensuremath{w_{#3,#2}(#1)}}
    }
}

\NewDocumentCommand\indexBetadics{O{\wordw} O{\integern} O{\base}}{
    \ifthenelse{
    \equal{#1}{}}{\ensuremath{\operatorname{ind}_{#3,#2}}}{
        {\ensuremath{\operatorname{ind}_{#3,#2}(#1)}}
    }
}

\NewDocumentCommand\variationSetGreedyBetadics{O{} O{} O{\base}}{
\ifthenelse{
    \equal{#1}{} \OR \equal{#2}{}}{\setGreedyBetadics[#2][#3]}{
        \ensuremath{\partial_#1\setGreedyBetadics[#2][#3]}
    }
}

\NewDocumentCommand\variationSetGreedyPrefixes{O{} O{} O{\base}}{
\ifthenelse{
    \equal{#1}{} \OR \equal{#2}{}}{\setGreedyBetaPrefixes[#2][#3]}{
        \ensuremath{\partial_#1\setGreedyBetaPrefixes[#2][#3]}
    }
}

\newcommand\realx{\ensuremath{x}} % real variable x
\newcommand{\wordw}{\ensuremath{w}} % word w

\newcommand\indexFunc{\iota} % to denote index injections
\newcommand\proj{\pi} % to denote the projection associated to an index injection
\newcommand\prfx[1]{\ensuremath{\mkern1mu\overline{\mkern-1mu#1\mkern-1mu}\mkern1mu}} % to denotes the prefix-free encoding function
\newcommand\timeBrackets[1]{\ifthenelse{\equal{#1}{}}{[\cdot]}{\ensuremath{[#1]}}} % time brackets

\newcommand\setAffine[2]{\ensuremath{\mathcal{A}_{#1,#2}}} % to denote the set of affine maps of a given input dimension and output dimension

\newcommand{\ind}[1]{\mathbbm{1}_{\{#1\}}} % to denote the truth function which takes on the value 1 if the statement inside { } is true and equals 0 otherwise
\newcommand{\RNN}{RNN\xspace} % recurrent neural network
\newcommand{\RNNs}{RNNs\xspace} % recurrent neural networks

\newcommand\ReLU{\ensuremath{\rho}} % Rectified Linear Unit
\newcommand\nn{\ensuremath{\Psi}} % neural network
\newcommand\nns[2]{\ensuremath{\Nc_{#1,#2}}} % set of neural nets
\newcommand\nnInDim[1]{{d_i(#1)}} % to denote the input dimension of a neural network
\newcommand\nnOutDim[1]{{d_o(#1)}} % to denote the output dimension of a neural network
\newcommand\nnDepth[1]{{\Lc(#1)}} % to denote the depth of a neural network
\newcommand\nnWidth[1]{{\Wc(#1)}} % to denote the width of a neural network
\newcommand\nnWeights[1]{{\Omega(#1)}} % to denote the set of weights of a neural network
\newcommand\nnMag[1]{{\Mc(#1)}} % to denote the weight magnitude of a neural network

\newcommand\nnAncestor[2]{\ensuremath{#1^{\le #2}}} % to denote the ancestor subnetwork of a neural network
\newcommand\rnnHidDim[1]{\ensuremath{m(#1)}} % hidden dimension of the RNN

\newcommand{\TMNU}{TMNU\xspace} % Turing machine with neural units
\newcommand{\TMNUs}{TMNUs\xspace} % Turing machines with neural units

\newcommand{\workSymbols}{\ensuremath{\Gamma}} % set of work symbols
\newcommand{\blanksymb}{\ensuremath{\square}} % blank symbol

\newcommand{\tm}{\ensuremath{M}} % Turing machine
\newcommand\nStates{\ensuremath{n}} % number of states
\newcommand\states{\ensuremath{Q}} % set of states
\newcommand\ntapes{\ensuremath{k}} % number of tapes
\newcommand\transFunc{\ensuremath{\delta}} % transition function
\newcommand{\transFuncState}{\transFunc^q} % to denote the state component of the transition function
\newcommand{\transFuncSymb}{\transFunc^\symb} % to denote the symbol component of the transition function
\newcommand{\transFuncMove}{\transFunc^m} % to denote the move component of the transition function
\newcommand\reluSet[1]{\ensuremath{\mathcal{R}_{#1}}} % to denote the set of ReLU functions

\newcommand{\symb}{\ensuremath{\sigma}} % symbol
\newcommand\state{\ensuremath{q}} % state
\newcommand{\tape}{\ensuremath{\tau}} % tape
\newcommand{\move}{\ensuremath{m}} % move
\newcommand{\config}{c} % configuration

\newcommand{\readOp}{R} % to denote the read operation
\newcommand{\writeOp}{W} % to denote the write operation
\newcommand{\shiftOp}{S} % to denote the shift operation
\newcommand\indexFuncState{\indexFunc_\state}

\newcommand{\tmnuM}{\ensuremath{{\bf M}}} % Turing machine with neural units M
\newcommand{\tmnuN}{\ensuremath{{\bf N}}} % Turing machine with neural units N
\newcommand\neurDim{\ensuremath{d}} % neural dimension
\newcommand\neurSpace{\ensuremath{\R^{\neurDim}}} % neural space
\newcommand\commFunc{\ensuremath{\kappa}} % command function

\newcommand{\nFunctions}[1]{\ensuremath{\nu_{#1}}} % to denote the number of functions of a TMNU
\newcommand\neurState{\ensuremath{\omega}} % neural state
\newcommand\indexFuncNeur{\indexFunc_\neurState}

\newcommand\encDya[1]{u^{(#1)}} % to denote the dyadic encoding of a real number in [-1,1]
\newcommand\encPoly[2]{\ensuremath{u_{#1,#2}}} % to denote the encoding of a polynomial with coefficients in [-1,1]
\newcommand\encCheb[2]{\ensuremath{u^{\mathsf T}_{#1,#2}}} % encoding of a finite native Chebyshev sum
\newcommand\encChebSeq[2]{\ensuremath{\langle #1,#2\rangle_{\mathsf T}}} % encoding of a sequence of native Chebyshev sums

\newcommand\approxRea[2]{#1\hspace{-3.5pt}\downharpoonright_{#2}} % to denote the dyadic approximation of a real number in [-1,1] with precision 2^{-#2}
\newcommand\approxPoly[2]{#1\hspace{-3.5pt}\downharpoonright_{#2}} % to denote the approximation of a polynomial with precision 2^{-#2}

\newcommand\Sign{\ensuremath{{\pmb{{\pm}}}}} % TMNU for the sign function
\newcommand\Scale{\ensuremath{{\bm{\mathsf{S}}}}} % TMNU for the scaling function
\newcommand\Contr{\ensuremath{{\bm{\mathsf{Hom}}}}} % TMNU for the contraction function
\newcommand\Times{\ensuremath{{\pmb{\times}}}} % TMNU for the multiplication function
\newcommand\UpPoly{\ensuremath{\bm{\mathsf{\pi}}}} % TMNU for micro-polynomial updates
\newcommand\Poly{\ensuremath{{\bm{\mathsf{\Pi}}}}} % TMNU for polynomial functions
\newcommand\Continuous{\ensuremath{{\bm{\mathsf{C}}}}} % TMNU for the continuous function
\newcommand\ChebStep{\ensuremath{{\bm{\mathsf{TStep}}}}} % TMNU for one native Chebyshev recurrence step
\newcommand\ChebSum{\ensuremath{{\bm{\mathsf{TSum}}}}} % TMNU for finite native Chebyshev sums
\newcommand\ChebContinuous{\ensuremath{{\bm{\mathsf{TC}}}}} % TMNU for successive native Chebyshev sums

\newcommand\tapeR[1]{#1^+} % to denote the right part of a tape
\newcommand\tapeL[1]{#1^-} % to denote the left part of a tape
\newcommand\cantorMap{\eta} % to denote the Cantor map
\newcommand\cantorSet{\mathcal{C}} % to denote the Cantor set
\newcommand\oneHot[2]{\mathbf{1}_{#1,#2}} % to denote the one-hot encoding of k in n
\newcommand\configMap[1]{\gamma_{#1}} % to denote the configuration mapping of a TMNU
\newcommand\kroen[2]{\delta_{#1,#2}} % to denote the Kronecker symbol

\newcommand\transFuncSymbMove{\transFunc^\tape} % to denote the transition function of a TMNU that gives the direction of the head movement

\newcommand\firstNN{\mathcal{N}_f} % to denote the neural network computing the first element of a sequence
\newcommand\firstNNmat[1]{A_f^{#1}} % to denote the matrix of the first layer of \firstNN
\newcommand\firstNNvec[1]{b_f^{#1}} % to denote the vector of the first layer of \firstNN

\newcommand\id{\ensuremath{\operatorname{id}}} % identity function

\usetikzlibrary{ipe} 

\newcommand{\SetLineSpacing}[1]{\setstretch{#1}}
\SetLineSpacing{1.0}

\title{Recurrent neural networks approximate continuous functions}
\author{Valentin Abadie, Clemens Hutter and Helmut Bölcskei}
\date{\today}

\begin{document}

\maketitle
\begin{abstract}
Classical approximation theorems ask for a new neural network whenever the target accuracy is improved. This paper studies the opposite possibility: can the network be chosen once and for all, and can accuracy be bought only by letting it run longer? We prove that this is possible for every continuous function on $[-1,1]$. More precisely, each such function is uniformly approximated by the time evolution of a single ReLU recurrent neural network with fixed weights and fixed hidden dimension. The mechanism behind the construction is a new intermediate model, the Turing machine with neural units (\TMNU). This model retains the algorithmic freedom needed to implement polynomial approximation schemes, while remaining rigid enough to be simulated by RNNs with explicit bounds on hidden dimension and weight magnitude. The resulting convergence rates reflect the underlying polynomial approximation rates. We complement the construction with minimax lower bounds showing that runtime is not merely a proof artifact, but an unavoidable resource in this fixed-network approximation paradigm.
\end{abstract}

% \tableofcontents

\section{Introduction}\label{sec:introduction}

The work \cite{hutter2025quantifierRNN} introduced an approximation paradigm for recurrent neural networks (RNNs), which asks whether RNNs can approximate a function $f : [-1,1] \to \R$ in a sense that we will detail below. The starting point for this approximation paradigm is the classical universal approximation theorem for feedforward neural networks \cite{cybenko89,funahashi89,hornik89}, which asserts that every continuous function on a compact domain can be approximated arbitrarily well by a shallow neural network with sigmoidal activation function. Subsequent quantitative results relating the smoothness of the target function and the prescribed approximation error to the size of the approximating network were later obtained in \cite{Barron_1993,Barron_1994}. 

In the past two decades the rectified linear unit (ReLU) has become the dominant activation
function in theory and practice. Beginning with \cite{yarotskyErrorBoundsApproximations2017},
quantitative approximation results for deep ReLU networks have been developed
\cite{telgarskyNeuralNetworksRational2017,
schmidthieberKolmogorovArnoldRepresentation2021, siegel2023optimal},
culminating in \cite{deepAT2019}, which shows that deep ReLU networks approximate a wide
range of function classes in metric-entropy–optimal manner.

The quantitative approximation results in the literature typically take the following form: for a given function $f$ and a given approximation error $\epsilon > 0$, there exists a neural network $\nn$ that approximates $f$ to within error $\epsilon$,
formalized as
\begin{equation}\label{eq:paradigm1}
	% \tag{Paradigm 1}
	\forall f: \forall \epsilon: \exists\, \nn \textrm{ such that $\nn$ approximates $f$ to within error $\epsilon$.}
\end{equation}
Thus the network architecture and weights depend on the chosen value of $\epsilon$.
If a smaller error is later required, a new (typically larger) network must be instantiated.
For instance, \cite[Proposition~III.5]{deepAT2019} shows that for every polynomial $f$ and every $\epsilon>0$, there exists a deep ReLU network of $\epsilon$-independent width and depth $\Oc(\log{(\epsilon^{-1})})$ that approximates $f$ to within error $\epsilon$.

In \cite{hutter2025quantifierRNN}, the authors introduced an approximation framework in which the approximating network is no longer chosen as a function of the target tolerance $\epsilon$. The guiding idea is to exchange the usual quantifier order $\forall \epsilon, \exists\, \nn$ for a statement of the form
%in \eqref{eq:paradigm1} to yield
\begin{equation}\label{eq:paradigm2}
	% \tag{Paradigm 1}
	\forall f: \exists\, \nn, \textrm{such that } \forall \epsilon: \textrm{ $\nn$ approximates $f$ to within error $\epsilon$.}
\end{equation}
In practice, the accuracy is improved by iterating one and the same neural network, whose architecture and weights remain fixed throughout the process. The paper \cite{hutter2025quantifierRNN} realizes this idea with Recurrent Neural Networks (RNNs) \cite{Goodfellow2016}; in the case where $f : [-1,1]\to\R$ is a polynomial, the iterates yield errors that tend to zero, and moreover do so at an exponential rate in the number of compositions.

In this paper, we generalize this result by showing that \eqref{eq:paradigm2} holds for every continuous function $f : [-1,1] \to \R$. Note that this result is not a mere consequence of the density of polynomials in the space of continuous functions, as in \cite{hutter2025quantifierRNN}, the RNN that approximates a given polynomial function $f$ has a hidden state size that grows linearly with the degree of $f$. A density-type result would thus yield an RNN with infinite hidden state size, which goes out of our formalism. In fact, our proof relies on a completely different approach. 

The idea is to introduce a new model that has the same approximation properties as RNNs, but that one can manipulate more easily than RNNs. The intuition is to use the well-known fact that RNNs are Turing-complete, i.e., they can simulate Turing machines \cite[Chapter 3.1]{sipser2012introduction} in some sense to be made precise later. Turing machines are easier to manipulate than RNNs, because they can be specified simply by writing an algorithm \cite[Chapter 3.3]{sipser2012introduction}, whereas RNNs are specified by writing matrices and vectors, which is more cumbersome. 

The key principle of the proof is to create an RNN whose hidden state is separated in two parts: one part that simulates the computation of a Turing machine, and one part that manipulates the input $x \in \R$ according to the instructions of the Turing machine. To make explicit this separation of the hidden state into two parts, we in fact introduce a new model of computation, which we call the \TMNU (for "Turing Machine with Neural Units"), that is essentially made of a Turing machine and of a real computation unit that applies piecewise-linear transformations to an input $x \in \R$ according to the instructions of the Turing machine. 

\begin{figure}[t]
    \hspace*{-1.3cm}%
    \scalebox{0.6}{
    \input{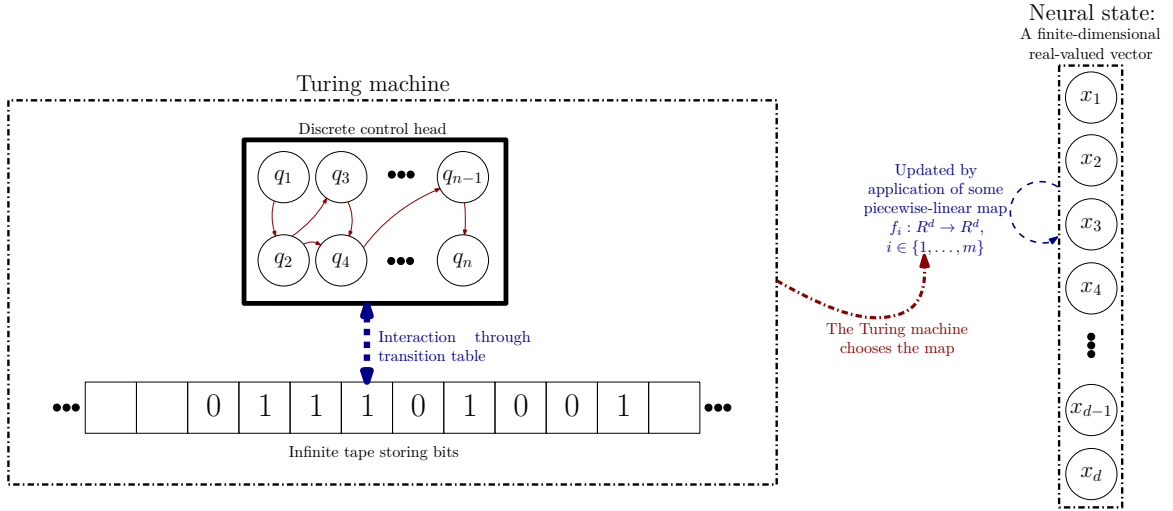}}

	\caption{Schematic representation of a \TMNU.}
\end{figure}

% IMAGE TMNU

\TMNUs have the same flexibility of design as Turing machines, but they also incorporate the ability to manipulate real numbers through piecewise-linear transformations, which belongs to the realm of RNNs. However, we will show that \TMNUs are essentially equivalent to RNNs, and that showing \eqref{eq:paradigm2} can be reduced to the problem of approximating continuous functions with \TMNUs, which is much simpler to solve, because \TMNUs can be designed by writing algorithms, and notably we can design simple \TMNUs to be used later as subroutines in the design of more complex \TMNUs. This flexibility of design is crucial to the proof, and is not available when working directly with RNNs.

The idea of the \TMNU detour comes from two observations: (i) \RNNs are known to be Turing-complete \cite{siegelmann1992computational}, and (ii) the theory of computable analysis \cite{weihrauch2000computable} studies the approximation of functions by Turing machines, in a paradigm that is very similar to \eqref{eq:paradigm2}. The functions that can be approximated by Turing machines in the paradigm of computable analysis are called computable functions, and include most functions of interest in analysis, such as polynomials, trigonometric functions, exponential functions, etc. In fact, it is known that every computable function $f : [-1,1] \to \R$ is continuous \cite[Proposition 6.1]{brattka2008tutorial} and ``\textit{Often, a function that is not computable [...] is so for purely topological reasons; that is, it is not continuous}'' \cite[p452]{brattka2008tutorial}. Therefore, computable functions make a good proxy for continuous functions. A first naive approach would be to use observations (i) and (ii) to conclude that RNNs can approximate any computable function according to \eqref{eq:paradigm2}. However, this is not the case because the simulation of Turing machines by RNNs use encodings of the input to the Turing machine in a way that is not compatible with the approximation paradigm \eqref{eq:paradigm2}. We explain this in greater detail in Section \ref{sec:tmnu-model}. The \TMNU model is designed to circumvent this issue.

We conclude this section by mentioning that the \TMNU model has to be viewed as a proof tool that is specifically designed to prove the main result of the paper, but that can certainly be used in other proofs regarding the approximation properties of RNNs. We next formalize the RNN approximation paradigm, state the main results, and then develop the \TMNU model used in the proofs.

\section{The RNN Approximation Paradigm}
\label{sec:rnn-approximation-paradigm}

We proceed to give a formal definition of RNNs, and to operationalize the approximation paradigm \eqref{eq:paradigm2} in terms of RNNs.

\begin{definition}[RNN \cite{elman90,Goodfellow2016,hutter2025quantifierRNN}]\label{def:elman_rnn}
	We denote by $\ReLU : \R \to \R, \ReLU(x) :=  max(0, x)$ the \newname{ReLU} function which acts component-wise, i.e., $\ReLU(x_1, \dots, x_m) := (\ReLU(x_1), \dots, \ReLU(x_m))$.
	An RNN is an ordered set $\Rc := (d,m,d';A_h,b_h,A_x,A_o,b_o)$ where $d\in\N$ is the input dimension, $d'\in\N$ is the output dimension, $m\in\N$ is the hidden state size, $A_h \in \R^{m \times m}$, $A_x \in \R^{m \times d}$, $A_o \in \R^{d' \times m}$, and $b_h \in \R^{m}$, $b_o \in \R^{d'}$.
	To $\Rc$ we associate the {hidden state operator} $\Hc: \left(\R^d\right)^{\No} \rightarrow \left(\R^m\right)^{\No}$ mapping an input sequence $(x[t])_{t \in \N_0}$ recursively to the sequence of hidden states $(h[t])_{t \in \N_0}$ according to
	\begin{align}
		h[-1] & := 0 \in \R^m\\
		(\Hc x)[t]= h[t]  & = \ReLU(A_h h[t-1] + A_x x[t] + b_h).
	\end{align}
    We finally view $\Rc$ as a map $\Rc : \left(\R^d\right)^{\No} \rightarrow \left(\R^{d'}\right)^{\No}$ defined by
	\[
		(\Rc x)[t] = A_o(\Hc x)[t]+ b_o, \quad t\in\No, \quad (x[t])_{t \in \No} \in \left(\R^d\right)^{\No}.
	\]
    We also let $\rnnHidDim{\Rc} := m$ denote the hidden state size of $\Rc$ and $\nnMag{\Rc}$ denote the magnitude of the weights of $\Rc$, i.e., $\nnMag{\Rc} := \max\{\|A_h\|_\infty, \|b_h\|_\infty, \|A_x\|_\infty, \|A_o\|_\infty, \|b_o\|_\infty\}$.
\end{definition}

Following \cite{hutter2025quantifierRNN}, to conform with \eqref{eq:paradigm2}, the aim is, for a given function $f$, to find an RNN that achieves any arbitrarily small approximation error $\epsilon$ provided we let it run sufficiently long. To this end, if we desire an approximation of the function $f$ at the point $x \in \R$, we take the input sequence of the RNN as $\tilde x[t]=x \ind{t=0}, t \in \N_0$. Here, $\ind{\cdot}$ denotes the truth function which takes on the value $1$ if the statement inside $\{\cdot\}$ is true and equals $0$ otherwise. To formalize this, the following operator was introduced in \cite{hutter2025quantifierRNN}. 

\begin{definition}\label{def:spreadInput}
	The mapping $\Dc: \R^d \rightarrow \left(\R^d \right)^{\No}$ is defined according to
	\[
		(\Dc x)[t] = x \ind{t=0}, \qquad t \in \N_0.
	\]
\end{definition}
The corresponding output sequence of the RNN then produces increasingly accurate approximations of $f(x)$ as time $t$ evolves. The approximation paradigm \eqref{eq:paradigm2} has been operationalized in terms of RNNs in \cite{hutter2025quantifierRNN} as
\begin{align}
	\begin{split}
	\label{eq:paradigm3}
	\forall f: \exists \, \Rc: \forall \epsilon: \exists t_0: 
	\sup_{t\geq t_0} \sup_{x}  |(\Rc \Dc x)[t] - f(x)| < \epsilon.
	\end{split} 
\end{align}
Every approximation theorem fitting this paradigm must have the size, topology, and weights of the approximating RNN $\Rc$ be independent of the approximation error $\epsilon$, simply by virtue of the quantifier order in \eqref{eq:paradigm3}. Only the runtime required to achieve the desired approximation error $\epsilon$ will depend on $\epsilon$. This approximation paradigm exhibits interesting practical properties as storing the fixed RNN on digital devices requires little memory and the approximation error can be controlled simply by adjusting the runtime of the RNN.

Before proceeding to the main results, we want to make clear the necessity of allowing the \RNN to compute for an arbitrary long time to reach arbitrary precision $\varepsilon$. Indeed, let $f : x \mapsto x^2$, $\Rc$ be an RNN, $\varepsilon > 0$ and suppose that there exists $t_0 \in \No$ such that
\begin{equation}
    \sup_{t \geq t_0} \sup_{x \in [-1,1]} |(\Rc \Dc x)[t] - f(x)| < \varepsilon.
\end{equation}
Then, in particular, we have $\sup_{x \in [-1,1]} |(\Rc \Dc x)[t_0] - f(x)| < \varepsilon$. Note that $(\Rc \Dc x)[t_0]$ is a piecewise-linear function of $x$, and, in fact, can be identified to a deep ReLU neural network of depth $t_0+1$ and width $\rnnHidDim{\Rc}$. As shown in Appendix \ref{subsec:fixed-time-lower-bound}, it has at most $(\rnnHidDim{\Rc}+1)^{t_0+1} - 1$ breakpoints, while any piecewise-linear approximation of $f$ within precision $\varepsilon$ requires at least $(2\varepsilon)^{-1/2} - 1$ breakpoints, hence we obtain the lower bound
\begin{equation}\label{eq:lower-bound-on-hidden-state-size}
    \rnnHidDim{\Rc} \geq \left(\frac{1}{2\varepsilon}\right)^{\frac1{2(t_0+1)}} - 1 \underset{\varepsilon \to 0} {\longrightarrow} \infty.
\end{equation}
Therefore, in order to comply with the approximation paradigm \eqref{eq:paradigm3}, we need to allow the RNN to compute for an arbitrary long time, which is the only way to possibly reach arbitrary precision $\varepsilon$ while keeping the size and weights of the RNN fixed. The core of this paper is to show that, indeed, the approximation paradigm \eqref{eq:paradigm3} can be achieved for a large class of functions $f$.

\section{Main Results}
\label{sec:main-results}

We now state the main result of the paper, which shows that every continuous function $f : [-1,1] \to \R$ can be approximated by an RNN in the sense of \eqref{eq:paradigm3}, when the RNN is allowed to compute for an arbitrary long time.
\begin{theorem}\label{thm:main-theorem-continuous}
    Let $f : [-1,1] \to \R$ be a continuous function. Then, there exists an RNN $\Rc$ such that
    \begin{equation}
        \limi{t} \sup_{x \in [-1,1]} |\Rc \Dc x[t] - f(x)| = 0.
    \end{equation}
    Moreover, the hidden state dimension and the magnitude of the weights of $\Rc$ satisfy \[
        \rnnHidDim{\Rc} \leq 5320, \quad \text{and} \quad\nnMag{\Rc} \leq 5 \left(1+ \|f\|_{L^\infty([-1,1])}\right).
    \]
\end{theorem}
\begin{proof}[Sketch of the proof]
    This theorem is a direct consequence of combining the link between RNNs and \TMNUs given by Theorem \ref{thm:simulation-of-a-tmnu-by-an-rnn} with the construction of a \TMNU that approximates $f$ given by Theorem \ref{thm:continuous-approximation-tmnu-main}. The explicit bounds on the hidden state dimension and the weight magnitude of $\Rc$ are obtained by plugging the quantities appearing in Lemma \ref{lem:constructed-tmnus-command-magnitudes} into the bounds given by Theorem \ref{thm:simulation-of-a-tmnu-by-an-rnn}.
\end{proof}
Surprisingly perhaps, and in sharp contrast to the main result of \cite{hutter2025quantifierRNN}, the hidden state dimension of the RNN $\Rc$ \textit{does not} depend on the function $f$. Still, we are able to recover exponential decay of the approximation error as a function of the runtime $t$ for the class of polynomial functions. This is stated in the following result.
\begin{theorem}\label{thm:main-theorem-poly}
    Let $N\in\N$, and $a_0, \dots, a_N \in \R$. Then, there exists an RNN $\Rc$ satisfying
    \[
        \rnnHidDim{\Rc} \leq 5320, \quad \text{and} \quad \nnMag{\Rc} \leq 5 \left(1+ \|a\|_1\right)
    \]
    such that
    \begin{equation}
        \sup_{x \in [-1,1]} \left|\Rc \Dc x[t] - \sum_{i=0}^N a_i x^i\right| \leq 2\cdot 2^{-t/t_a}, \quad t \geq \tau_a,
    \end{equation}
    where $t_a := 72(N+1)$ and $\tau_a := t_a(6\|a\|_1 + N + 17)$.
\end{theorem}
\begin{proof}[Sketch of the proof]
    The proof of this theorem follows exactly that of Theorem \ref{thm:main-theorem-continuous}, and the exponential rate is specifically obtained by application of Theorem \ref{th:convergence-rate-polynomial} that construct a \TMNU that approximates the target polynomial with an exponential rate.
\end{proof}
Note that the exponential rate is optimal, in the sense that there are some polynomials for which this exponential rate cannot be beaten. Indeed, \eqref{eq:lower-bound-on-hidden-state-size} can be reformulated as a lower bound on the approximation error of $x \mapsto x^2$ as a function of the runtime $t$ as
\begin{equation}
    \varepsilon := \sup_{x \in [-1,1]} |\Rc \Dc x[t] - f(x)| \geq \left(\frac1{2(\rnnHidDim{\Rc}+1)^t}\right)^2 = \frac142^{-2t \log_2(\rnnHidDim{\Rc}+1)}, \quad t \in \N.
\end{equation}
Therefore, among the class of polynomial functions, the function $x \mapsto x^2$ has an approximation rate that decays exponentially with $t$, and cannot be approximated with a faster rate. In a minimax sense, this shows that the class of polynomial functions has an approximation rate that decays exponentially with $t$.

More generally, we link the decay of the approximation error as a function of the runtime $t$ to the rate of convergence of the Chebyshev series of $f$. Given a continuous function $f : [-1,1] \to \R$, its Chebyshev coefficients are defined as
\begin{equation}\label{eq:chebyshev-coefficients}
    c^{(f)}_0 := \frac{1}{\pi} \int_{-1}^1 \frac{f(x) T_0(x)}{\sqrt{1-x^2}} \mathrm{d}x, \qquad c^{(f)}_n := \frac{2}{\pi} \int_{-1}^1 \frac{f(x) T_n(x)}{\sqrt{1-x^2}} \mathrm{d}x, \quad n \in \N,
\end{equation}
and the Chebyshev partial sums are defined as
\begin{equation}\label{eq:chebyshev-partial-sums}
    S_n^{(f)}(x) := \sum_{k=0}^n c^{(f)}_k T_k(x), \quad n \in \N, \quad x \in [-1,1],
\end{equation}
where $T_n$ is the $n$-th Chebyshev polynomial given by
\begin{equation}\label{eq:chebyshev-polynomials}
    T_0(x) = 1, \quad T_1(x) = x, \quad T_{n+1}(x) = 2x T_n(x) - T_{n-1}(x), \quad n \in \N, \quad x \in [-1,1].
\end{equation}
In accordance with our approximation paradigm \eqref{eq:paradigm3}, we are interested in functions for which the Chebyshev partial sums converge uniformly to $f$ on $[-1,1]$, i.e., whether the quantity
\begin{equation}\label{eq:rate-of-approximation-by-tchebychev-series}
    S(f,n) := \sup_{m \geq n}\sup_{x \in [-1,1]} |f(x) - S_m^{(f)}(x)|, \quad n \in \N,
\end{equation}
converges to $0$, in which case we say that $f$ has a Chebyshev series. In fact, Chebyshev series essentially reduce to Fourier series under the change of variable $x = \cos(\theta)$, because $T_n(\cos \theta) = \cos(n \theta)$, and therefore the Chebyshev partial sums $S_n^{(f)}$ converge uniformly to $f$ on $[-1,1]$ if and only if the Fourier partial sums of the function $\theta \mapsto f(\cos(\theta))$ converge uniformly to $\theta \mapsto f(\cos(\theta))$ on $[0,\pi]$. It is known that the Fourier partial sums of a continuous function converge uniformly to the function if the function satisfy the property of being Dini-Lipschitz continuous \cite[Theorem 3.10]{gil2007numerical}, which definition is given as follows.
\begin{definition}
[Dini-Lipschitz continuity] \label{def:dini-continuous-function}
    A continuous function $f : [-1,1] \to \R$ is said to be Dini-Lipschitz continuous if
    \[
        \lim_{\varepsilon \to 0} \omega_f(\varepsilon)\log(\varepsilon) = 0,
    \]
    where $\omega_f$ is the modulus of continuity of $f$ defined by
    \[
        \omega_f(\varepsilon) := \sup_{x,y\in[-1,1], |x-y|\leq\varepsilon} |f(x)-f(y)|, \quad \varepsilon \in [0,2].
    \]
\end{definition}
% i.e., if and only if the modulus of continuity of $f$ defined by
% \[
%     \omega_f(\varepsilon) := \sup_{x,y\in[-1,1], |x-y|\leq\varepsilon} |f(x)-f(y)|, \quad \varepsilon > 0,
% \]
% satisfies \[\int_0^1 \frac{\omega_f(\varepsilon)}{\varepsilon} \mathrm{d}\varepsilon < \infty.\]
Therefore, we restrict our attention to the vast class of Dini-Lipschitz continuous functions, which includes for example all Lipschitz and Hölder-continuous functions. One may be interested in the approximation of functions that are not Dini-Lipschitz continuous, but this would require going beyond Chebyshev series, which we reserve for future work.

% The Chebyshev series of a Lipschitz function $f : [-1,1] \to \R$ is given by
% \begin{equation}
%     f(x) = \sum_{n=0}^\infty a_n T_n(x), \quad x \in [-1,1],
% \end{equation}
% where $T_n$ is the $n$-th Chebyshev polynomial given by
% \begin{equation}
%     T_0(x) = 1, \quad T_1(x) = x, \quad T_{n+1}(x) = 2x T_n(x) - T_{n-1}(x), \quad n \in \N, \quad x \in [-1,1],
% \end{equation}
% and where the coefficients $a_n$ are uniquely determined by $f$. By denoting by $S_n^{(f)}$ the $n$-th partial sum of the Chebyshev series of $f$, i.e., $S_n^{(f)}(x) := \sum_{k=0}^n a_k T_k(x)$ for every $n \in \N$ and $x \in [-1,1]$.

The rate of approximation $S(f,n)$ can be linked to smoothness properties of $f$. We consider the following normalized function classes:
\begin{itemize}
    \item For $\alpha\in(0,1]$, the class $\Hc^\alpha$ of $\alpha$-H\"older functions $f$ satisfying $\|f\|_{L^\infty([-1,1])}\leq1$ and \[\sup_{x\neq y}\frac{|f(x)-f(y)|}{|x-y|^\alpha}\leq1.\]
    Note that for $\alpha=1$, the class $\Hc^1$ coincides with the class $\operatorname{Lip}$ of $1$-Lipschitz functions.
    \item For $k\in\N$, the class $\Cc^k\Hc^\alpha$ of $k$-times continuously differentiable functions $f$ with derivatives up to order $k$ satisfying $\|f^{(j)}\|_{L^\infty([-1,1])}\leq1$, and $f^{(k)} \in \Hc^\alpha$.
    \item The classes $\Sc$ and $\Sc_1$ of smooth functions with derivatives of order $n$ satisfying $\|f^{(n)}\|_{L^\infty([-1,1])}\leq n!$ and $\|f^{(n)}\|_{L^\infty([-1,1])}\leq1$, respectively.
\end{itemize}
We summarize the corresponding Chebyshev approximation rates in Table \ref{tab:link-smoothness-and-rate-of-approximation-by-tchebychev-series}. Such approximation rates are classical and can be found in standard references on numerical approximation \cite{trefethen2019approximation} completed with references on approximation with Fourier series \cite{stepanets2001uniform}.
\begin{table}[ht]
	    \centering
\begin{tabular}{c|c}
	    \hline
	    Name of the class & Rate of approximation $S(f,n)$ \\
	    \hline
	        $\Hc^\alpha$ & $\Oc(n^{-\alpha} \log(n))$ \\
	    \hline
	    $\Cc^k \Hc^\alpha$ & $\Oc(n^{-(k+\alpha)}\log(n))$ \\
	    \hline
	    $\Sc$ & $\exp(-\Omega(n))$ \\
	    \hline
	    $\Sc_1$ & $\exp(-\Omega(n\log n))$ \\
	    \hline
\end{tabular}
    \caption{Link between the smoothness of $f$ and the rate of approximation of $f$ by its Chebyshev series.}
    \label{tab:link-smoothness-and-rate-of-approximation-by-tchebychev-series}
\end{table}

Our third main result shows that there is a direct link between the decay of the approximation error as a function of the runtime $t$ and the rate of approximation of $f$ by its Chebyshev series. Specifically, the approximation error decays as the sum of two terms: the first term that accounts for RNN-specific computation errors, and a second term that accounts for the error of approximating $f$ by its Chebyshev partial sums. The tradeoff between these two terms is controlled by the choice of a nondecreasing function $\eta : [1,\infty) \to \N$, which, depending on the rate of approximation of $f$ by its Chebyshev series, can be chosen to get the fastest possible decay of the approximation error as a function of $t$. The choices of $\eta$ for the different function classes are summarized in Table \ref{tab:link-smoothness-and-approximation-error-decay}, together with the resulting approximation error decay as a function of $t$.
\begin{theorem}\label{thm:main-theorem-tchebychev}
    Let $f : [-1,1] \to \R$ be a Dini-Lipschitz continuous function. Let
    $\eta:[1,\infty)\to\N$ be nondecreasing, with
    $\eta(t)\to\infty$ and $t/\eta(t)$ nondecreasing. Assume that there exists $A>0$ such that
    \[
        \eta(t)\log_2(\eta(t)+1)\leq At,
        \qquad t\geq1.
    \]
    Then, there exist a constant $\tau := \tau_{f,A}>0$ and an RNN $\Rc$ satisfying
    \[
        \rnnHidDim{\Rc} \leq 6700, \quad \text{and} \quad\nnMag{\Rc} \leq 5 \left(3+ S(f,0) + \|f\|_{L^\infty([-1,1])}\right)
    \]
    such that for every $t\geq \tau$,
    \begin{equation}\label{eq:link-approximation-error-decay-to-tchebychev-series}
        \sup_{x \in [-1,1]} |\Rc \Dc x[t] - f(x)|
        \leq
        2\cdot
        2^{-\frac{t/\tau}{\eta(t/\tau)}}
        +
        S\left(f,\eta(t/\tau)\right).
    \end{equation}
\end{theorem}
\begin{proof}[Sketch of the proof]
    Once again, this theorem follows from combining the link between RNNs and \TMNUs given by Theorem \ref{thm:simulation-of-a-tmnu-by-an-rnn} with the construction of a \TMNU that computes Chebyshev partial sums of $f$ specified in Theorem \ref{th:convergence-rate-chebychev-series}. The explicit bounds on the hidden state dimension and the weight magnitude of $\Rc$ are also obtained with Lemma \ref{lem:constructed-tmnus-command-magnitudes}.
\end{proof}
Theorem \ref{thm:main-theorem-tchebychev} gives an upper bound on the convergence speed of the RNN computations to the target function $f$. The natural follow-up question is whether this convergence speed is optimal. To answer this question, we show that, for fixed hidden dimension and fixed weight magnitude, the approximation error of RNNs cannot decay faster than a certain rate that depends on some information-theoretic properties of the class of target functions. In particular, this shows that the approximation rates given in Table \ref{tab:link-smoothness-and-approximation-error-decay} are almost optimal, up to logarithmic factors. The intuition behind this result is that, for fixed hidden dimension and fixed weight magnitude, the set of functions that can be approximated by RNNs after a fixed number of time steps $t$ must have limited metric entropy \cite{kolmogorov1959epsilon}, a concept that we now recall the definition of. In the following, we denote by $C([-1,1])$ the set of continuous functions from $[-1,1]$ to $\R$, equipped with the uniform norm $\|\cdot\|_{L^\infty([-1,1])}$.
\begin{definition}[Covering number]\label{def:covering-number}
    Let $\Xc\subset C([-1,1])$ be compact and $\varepsilon>0$. An $\varepsilon$-covering of $\Xc$ is a finite collection of functions $g_1,\ldots,g_N\in C([-1,1])$ such that for every $f\in\Xc$, there exists $j\in\{1,\ldots,N\}$ with $\|f-g_j\|_{L^\infty([-1,1])}\leq\varepsilon$. We denote by $\mathcal N(\varepsilon,\Xc)$ the size of the smallest $\varepsilon$-covering of $\Xc$. The metric entropy of $\Xc$ is defined as $\varepsilon \mapsto \log_2 \mathcal N(\varepsilon,\Xc)$.
\end{definition}
The function classes that we have introduced above have well-known metric entropy estimates \cite{kolmogorov1959epsilon}, that we summarize in Table \ref{tab:function-classes}.
\begin{table}[ht]
    \centering
\begin{tabular}{c|c}
    \hline
    Name of the class $\Xc$ & Metric entropy $\log_2 \mathcal N(\varepsilon,\Xc)$ \\
    \hline
        $\Hc^\alpha$ & $\asymp(\varepsilon^{-1/\alpha})$
    \\
    \hline
    $\Cc^k \Hc^\alpha$ & $\asymp(\varepsilon^{-1/(k+\alpha)})$
    \\
    \hline
    $\Sc$ & $\asymp(\log(1/\varepsilon)^2)$
    \\
    \hline
    $\Sc_1$ & $\asymp\left(\frac{\log(1/\varepsilon)^2}{\log \log (1/\varepsilon)}\right)$
    \\
    \hline
\end{tabular}
    \caption{Link between the smoothness of $f$ and the metric entropy of the corresponding normalized function class.}
    \label{tab:function-classes}
\end{table}

We now turn these metric entropy estimates into lower bounds on the approximation error of RNNs with fixed hidden dimension and fixed weight magnitude. We denote by $\mathfrak R_{m,B}$ the set of RNNs with hidden dimension at most $m$ and weight magnitude at most $B$. Then, for compact $\Xc\subset C([-1,1])$ and $t\in\No$, we investigate the worst-case approximation error at time $t$ of functions in $\Xc$ by RNNs in $\mathfrak R_{m,B}$, defined as
\[
        \mathcal E_t(\Xc;m,B)
        :=
        \sup_{f\in\Xc}
        \inf_{\Rc\in\mathfrak R_{m,B}}
        \sup_{x\in[-1,1]}
        \left|(\Rc\Dc x)[t]-f(x)\right|.
\]
We find the following lower bound on $\mathcal E_t(\Xc;m,B)$ in function of the metric entropy of $\Xc$.

\begin{theorem}\label{thm:main-minimax-lower-bound}
    Let $\Xc\subset C([-1,1])$ be compact, let $m\in\N$, and let $B\geq1$.
    Let $h:(0,1]\to(0,\infty)$ be nonincreasing and assume that
    \[
        \lim_{\varepsilon \to 0} \frac{\log(\varepsilon^{-1})}{h(\varepsilon)}=0,
        \qquad
        \text{and} \qquad \log\mathcal N(\varepsilon,\Xc)\geq h(\varepsilon), \quad \varepsilon \in (0,1].
    \]
    Then there exist constants $K>0$ and $t_0\in\N$, depending only on
    $m,B$ and $h$, such that, for every $t\geq t_0$,
    \[
        \mathcal E_t(\Xc;m,B)
        \geq
        \frac12 h^{-1}(K(t+1)),
    \]
    where $h^{-1}(T)
        :=
        \inf\left\{
            \varepsilon\in(0,1]
            \,:\,
            h(\varepsilon)\leq T
        \right\}$, for $T > 0$.
\end{theorem}
\begin{proof}[Sketch of the proof]
    The intuition behind the result is that the metric entropy of the set if functions that can be realized by an RNN in at most $t$ time steps grows as $O(t)$. Moreover, to every function of a set $\Xc$ with precision $\varepsilon$, that set of approximation functions must have a metric entropy of the same order as the metric entropy of $\Xc$ at precision $\varepsilon$. Therefore, we must have $h(\varepsilon) = O(t)$, which gives the desired lower bound on $\varepsilon$ as a function of $t$. The full proof is given in Section \ref{subsec:minimax-lower-bound-notation}.
\end{proof}
We can now apply this general lower bound to the function classes that we have introduced above, by plugging in the corresponding metric entropy estimates. For example, for the class $\Hc^\alpha$, we have $h(\varepsilon)\asymp\varepsilon^{-1/\alpha}$, and therefore $h^{-1}(t)\asymp t^{-\alpha}$. By applying Theorem \ref{thm:main-minimax-lower-bound}, we get that $\mathcal E_t(\Hc^\alpha;m,B) = \Omega(t^{-\alpha})$. We summarize the resulting lower bounds in Table \ref{tab:link-smoothness-and-approximation-error-decay}.

\begin{table}[H]
    \centering
    \small
\begin{tabular}{c|c|c|c}
    \hline
    Name of the class & Choice of $\eta(t)$ & \begin{tabular}{c}
    RNN approximation rate\\
    (Upper bound)
    \end{tabular} & \begin{tabular}{c}
    RNN approximation rate\\
    (Lower bound)
    \end{tabular} \\
    \hline
        $\Hc^\alpha$ & $\asymp t/\log t$ & $\Oc\left(\frac{(\log t)^{1+\alpha}}{t^{\alpha}}\right)$ & $\Omega\left(\frac1{t^{\alpha}}\right)$ \\
    \hline
    %     $\operatorname{Lip}$ & $\asymp t/\log t$ & $\Oc((\log t)^2/t)$ & $\Omega(t^{-1})$ \\
    % \hline
    $\Cc^k\Hc^\alpha$ & $\asymp t/\log t$ & $\Oc\left(\frac{(\log t)^{1+k+\alpha}}{t^{k+\alpha}}\right)$ & $\Omega\left(\frac1{t^{k+\alpha}}\right)$ \\
    \hline
    $\Sc$ & $\asymp \sqrt t$ & $\exp(-\Omega(\sqrt{t}))$ & $\exp(-\Oc(\sqrt t))$ \\
    \hline
    $\Sc_1$ & $\asymp \sqrt{t/\log t}$ & $\exp(-\Omega(\sqrt{t\log t}))$ & $\exp(-\Oc(\sqrt{t\log t}))$ \\
    \hline
\end{tabular}
    \caption{Link between the smoothness of $f$, the choice of $\eta$, and the approximation error decay as a function of the runtime $t$. The lower bounds are minimax lower bounds for fixed hidden dimension and fixed weight magnitude. Note that the upper bounds are almost tight with the lower bounds, up to logarithmic factors. For the sets $\Sc$ and $\Sc_1$, the upper and lower bounds are tight at the exponential scale.}
    \label{tab:link-smoothness-and-approximation-error-decay}
\end{table}

The rest of the paper is devoted to proving these statements. The main technical step is to introduce \TMNUs as an intermediate computational model: they are rigid enough to be simulated by \RNNs with controlled dimension and weight magnitude, but flexible enough to implement the polynomial and Chebyshev computations appearing above. We first define this model in Section \ref{sec:tmnu-model}, show its simulation by \RNNs in Section \ref{sec:tmnu-rnn-simulation}, and then build the machines that realize the approximation schemes and the corresponding convergence estimates in Section \ref{sec:tmnu-constructions}.

\section{Turing machines, \TMNUs, and the approximation paradigm for \TMNUs}\label{sec:tmnu-model}

In this section, we introduce formally the \TMNU model. Before doing so, we recall the definition of a Turing machine \cite[Chapter 3.1]{sipser2012introduction} and related concepts, which will be used to define the \TMNU model. We will also explain in greater details why the classical correspondence between RNNs and \TMNUs is not enough to solve the approximation paradigm \eqref{eq:paradigm3}, and why we need to introduce the \TMNU model as an intermediate computational model to solve the approximation paradigm \eqref{eq:paradigm3}.

\subsection{Turing machines}

A Turing machine is a mathematical model of computation that informally consists of three components:
\begin{enumerate}
    \item a finite number $\nStates \in \N$ of internal states,
    \item a tape of cells, which cells can either be empty or having the symbol $0$ or the symbol $1$ written on them. For the sake of mathematical convenience, we assume that the tape is infinite in both directions, i.e. that the tape is a bi-infinite sequence of cells, which we can model as a function $\tape : \Z \to \workSymbols := \{-1,0,1\}$, where $\tape(i)$ is the symbol written on the $i$-th cell of the tape, and where $\tape(i) = -1$ means that the $i$-th cell is empty, $\tape(i) = 1$ means that the $i$-th cell has the symbol $1$ written on it, and $\tape(i) = 0$ means that the $i$-th cell has the symbol $0$ written on it. For sake of clarity, we let $\blanksymb := -1$ denote the symbol that represents an empty cell. Depending on the context, we might write either $\tape(i)$ or $\tape_i$ to denote the symbol written on the $i$-th cell of the tape.
    \item A control head that decides of an internal state to transition to, can read and write symbols on the cell numbered $0$ of the tape, and that can decide to shift the tape to the left, to the right or leave it as it is. Specifically, if the head decides to write a symbol $\symb \in \workSymbols$, then the tape is updated by changing the symbol written on the cell numbered $0$ to $\symb$, i.e., $\tape(0) \gets \symb$. If the head decides to shift the tape to the right, then the tape is updated by shifting all the symbols on the tape one cell to the right, i.e., $\tape(i) \gets \tape(i-1)$ for every $i \in \Z$. If the head decides to shift the tape to the left, then the tape is updated by shifting all the symbols on the tape one cell to the left, i.e., $\tape(i) \gets \tape(i+1)$ for every $i \in \Z$.
\end{enumerate}
In general, the Turing machine model can be made more general, i.e., the cells of the tape may contain more than three symbols, there might be multiple tapes, etc. However, we here consider only the simple version of the Turing machine model described above, which is sufficient for our purposes and mathematically lighter. This informal description of a Turing machine can be made formal as follows.
\begin{definition}(Turing machine)
    A Turing machine is a pair $M := (\nStates, \transFunc)$, such that $\nStates \in \N$ is the number of states, and $\transFunc : \{1,\ldots,\nStates\} \times \workSymbols \to \{1,\ldots,\nStates\} \times \workSymbols \times \{-1,0,1\}$ is the transition function, that given an internal state $\state \in \{1,\ldots,\nStates\}$ and a symbol $\symb \in \workSymbols$ read by the head, specifies a triple $(\state', \symb', m) := \transFunc(\state, \symb)$, where $\state'$ is the internal state to transition to, $\symb'$ is the symbol to write on the tape, and $m$ is the shift of the tape to be effected, with $-1$ denoting a left shift, $0$ denoting no shift and $1$ denoting a right shift. We conveniently define $\transFuncState$, $\transFuncSymb$ and $\transFuncMove$ to be such that $\transFunc = (\transFuncState, \transFuncSymb, \transFuncMove)$, i.e., $\transFuncState(\state, \symb) = \state'$, $\transFuncSymb(\state, \symb) = \symb'$ and $\transFuncMove(\state, \symb) = m$ for every $\state \in \{1,\ldots,\nStates\}$ and $\symb \in \workSymbols$.
\end{definition}
A precise description of the evolution of a Turing machine can be given as follows. We first describe the actions that the control head of a Turing machine can perform on the tape. We define the reading operation $\readOp : \workSymbols^\Z \to \workSymbols$, the writing operations $\writeOp_\symb : \workSymbols^\Z \to \workSymbols^\Z$ for $\symb \in \workSymbols$, and the shifting operation $\shiftOp_m : \workSymbols^\Z \to \workSymbols^\Z$ for $m \in \{-1,0,1\}$ by
\begin{equation}
    \readOp\tape := \tape(0),
\end{equation}
\begin{equation}
    \writeOp_\symb\tape(i) := \begin{cases}
        \symb, & i = 0, \\
        \tape(i), & i \in \Z \setminus \{0\},
    \end{cases}
\end{equation}
and 
\begin{equation}
    \shiftOp_m\tape(i) := \tape(i-m), \quad i \in \Z,
\end{equation}
for every $\tape \in \workSymbols^\Z$. The evolution of a Turing machine is then defined as follows. Assuming that $M$ has some internal state $\state \in \{1, \ldots,\nStates\}$ and some tape $\tape \in \workSymbols^\Z$, we update the internal state of $M$ to $\state'$ and the tape of $M$ to $\shiftOp_m \writeOp_{\symb}\tape$, where $(\state', \symb, m) := \transFunc(\state, \readOp\tape)$ is the triple specified by the transition function $\transFunc$ given the current internal state $\state$ and the symbol $\readOp = \tape(0)$ read by the head. This can be formulated more compactly by using the formalism of configurations \cite[Chapter 3.1]{sipser2012introduction}, that here again, we define in a slightly different way as in the classical literature, in order to make it more convenient for our purposes. The definition of configuration and the update of configuration are given as follows.
\begin{definition}(Configuration of a Turing machine)\label{def:configuration-of-turing-machine-and-update-of-configuration}
    Given a Turing machine $M := (\nStates, \transFunc)$, we call configuration of $M$ any pair $\config := (\state; \tape)$ such that $\state \in \{1, \ldots, \nStates\}$ and $\tape \in \workSymbols^\Z$. We let $\Cc_M := \{1, \ldots, \nStates\} \times \workSymbols^\Z$ be the set of configurations of $M$. We give a meaning of $M$ as a map from $\Cc_M$ to $\Cc_M$ by letting
    \begin{equation}
        M(\state; \tape) := (\transFuncState(\state, \readOp \tape), \shiftOp_{\transFuncMove(\state, \readOp \tape)} \writeOp_{\transFuncSymb(\state, \readOp \tape)}\tape),
    \end{equation}
    for every configuration $(\state; \tape) \in \Cc_M$.
\end{definition}
The formalism above can be used to define a computation of a Turing machine $M := (\nStates, \transFunc)$ starting from some initial configuration $\config \in \Cc_M$ as the sequence of configurations $(M^t(c))_{t \in \No}$ obtained by iterating the map $M : \Cc_M \to \Cc_M$ starting from $\config$. Finally, we say that $M$ \newname{halts} on configuration $\config$ if there exists $t \in \No$ such that $(\state_t; \tape_t) := M^t(c)$ satisfies $q_t = \nStates$. We call $\nStates$ the halting state of $M$.

Often, the initial configuration of a Turing machine is of a specific form. Specifically, we will often assume that the initial tape $\tape$ is empty on its left part, i.e., $\tape(i) = \blanksymb$ for every $i < 0$. In order to condense notations, we introduce the following convention. We denote by $\{0,1\}^\ast$ the set of finite binary sequences, by $\{0,1\}^\N$ the set of infinite binary sequences, and define $\{0,1\}^\# := \{0,1\}^\ast \cup \{0,1\}^\N$ as the set of all binary sequences. For every $u \in \{0,1\}^\ast$, $\len u$ denotes the length of $u$, and for every $u \in \{0,1\}^\N$, we let $\len u := \infty$. We will write \[\tape = |u\] to mean that $\tape(i) = u_{i+1}$ for every $i \in \{0, \ldots, \len u - 1\}$, and that $\tape(i) = \blanksymb$ for every $i < 0$ or $i > \len u - 1$, for some $u \in \{0,1\}^\ast \cup \{0,1\}^\N$. For instance, if $u = 101$, then $\tape = |u$ means that $\tape(0) = 1$, $\tape(1) = 0$, $\tape(2) = 1$, and $\tape(i) = \blanksymb$ for every $i < 0$ and every $i > 2$. We now explain why the classical correspondence between RNNs and Turing machines is not enough to solve the approximation paradigm \eqref{eq:paradigm3}.

\subsection{The necessity of the \TMNU detour}

In Section \ref{sec:introduction}, we have mentioned that it is not enough to use the simulation of Turing machines by RNNs to solve the approximation paradigm \eqref{eq:paradigm3}. Here, we show more formally why this is the case. We say Turing machine $\tm$ computes a function $f : \{0,1\}^\ast \to \{0,1\}^\ast$, if upon being started in configuration $c_0 := (1;|u)$, the computation reaches the halting configuration $c := (\nStates;|f(u))$ after some finite number of computation steps, for every $u \in \{0,1\}^\ast$. It was notably proven in \cite{siegelmann1992computational} that if a Turing machine $\tm$ computes a function $f : \{0,1\}^\ast \to \{0,1\}^\ast$ in the aforementioned sense, then there is an RNN that computes $f$ in the following way. One can give a reformulation of $f$ in the form of a function $\hat f : \cantorSet \to \cantorSet$, where $\cantorSet \subseteq [-1,1]$ is a Cantor set, i.e., a set of real numbers that is totally disconnected. Specifically, by defining the map $\eta_{\text{Cantor}} : \{0,1\}^\ast \to [-1,1]$ by 
\[
    \eta_{\text{Cantor}}(u) = \sum_{i=1}^{\len u} (2u_i+1) 4^{-i} \quad \text{for every } u \in \{0,1\}^\ast,
\]
the set $\cantorSet$ is defined as $\cantorSet := \eta_{\text{Cantor}}(\{0,1\}^\ast)$, and the function $\hat f : \cantorSet \to \cantorSet$ is defined to be such that as $\hat f(\eta_{\text{Cantor}}(u)) := \eta_{\text{Cantor}}(f(u))$ for every $u \in \{0,1\}^\ast$. Then, the result by \cite{siegelmann1992computational} states that for every Turing machine $\tm$ that computes a function $f : \{0,1\}^\ast \to \{0,1\}^\ast$, there exists an RNN $\Rc$ such that for every $u \in \{0,1\}^\ast$, there exists $t_0 \in \No$ such that for all $t \geq t_0$, $\Rc \Dc \eta_{\text{Cantor}}(u)[t] = \hat f(\eta_{\text{Cantor}}(u))$ for every $t \in \No$, thus establishing the Turing completeness of RNNs.

One could be tempted to use this bridge between Turing machines and RNNs to solve the approximation paradigm \eqref{eq:paradigm3}. Indeed, one could try to exploit the theory of computable analysis \cite{weihrauch2000computable,brattka2008tutorial}, that studies the properties of functions $f : [-1,1] \to \R$ by means of an underlying Turing machine that computes it. Specifically, this theory stipulates a natural representation of real numbers as binary sequences, which is called the Cauchy representation \cite[Definition 4.4]{brattka2008tutorial}, that we denote here $\eta_{\text{Cauchy}} : \{0,1\}^\N \to \R$. The choice of such a representation is far from arbitrary, and is motivated by a well-established theory to compare representations of real numbers in the context of computable analysis. Then, given a function $f : [-1,1] \to \R$, we say that $f$ is computed by a Turing machine $M$ if for every $x \in [-1,1]$, if we initialize the tape of $M$ with an infinite binary sequence $u \in \{0,1\}^\N$ such that $\eta_{\text{Cauchy}}(u) = x$, then, in the long run, the machine generates a sequence $v \in \{0,1\}^\N$ that satisfies $\eta_{\text{Cauchy}}(v) = f(x)$ \cite[Definition 4.9]{brattka2008tutorial}. A function $f : [-1,1] \to \R$ that can be computed in such a way by some Turing machine is said to be computable. As mentioned in Section \ref{sec:introduction}, computable functions form a very large class that includes basically every function of interest, and solving the approximation paradigm \eqref{eq:paradigm3} for computable functions would be a significant leap towards solving the approximation paradigm for continuous functions.

The temptation here is to think that this is possible to design an RNN that, given an input $x \in [-1,1]$, can somehow extract some prefix $p$ of an infinite binary sequence $u \in \{0,1\}^\N$ satisfying $\eta_{\text{Cauchy}}(u) = x$, encode it as $\eta_{\text{Cantor}}(p)$ as in the approach in \cite{siegelmann1992computational} and manipulate this sequence in the same way as some Turing machine $M$ does. The RNN in question would then extract increasingly long prefixes of $u$ and manipulate them in the same way as $M$ does to generate increasingly long prefixes of some sequence $v \in \{0,1\}^\N$ such that $\eta_{\text{Cauchy}}(v) = f(x)$, and therefore to approximate $f$. However, the major obstruction to this approach is precisely the fact that given a real number $x \in [-1,1]$, the RNN in question should be able to extract some prefix $p$ of an infinite binary sequence $u \in \{0,1\}^\N$ such that $\eta_{\text{Cauchy}}(u) = x$, and encode it as $\eta_{\text{Cantor}}(p)$. Therefore, the RNN should be able to realize, in a finite number of computation steps, a map $g : [-1,1] \to \cantorSet$. However, such a function $g$ cannot be realized by an RNN with ReLU activation function, because $g$ is not continuous, and every function that can be realized by an RNN in a finite number of time steps is continuous. To see why $g$ is not continuous, note that $[-1,1]$ connected set, while $\cantorSet$ is a totally disconnected set, and therefore there is no continuous function from $[-1,1]$ to $\cantorSet$ by the intermediate value theorem \cite[Theorem 4.7]{lee2011IntroductionTopologicalManifolds}. Therefore, going in this direction would require to circumvent this obstruction, which does not appear to be evident. In other words, it is not clear how to use the theory of computable analysis to show that every function $f : [-1,1] \to \R$ that is computed by a Turing machine can be approximated by an RNN. This is precisely here that the need to investigate other models of computations arise, especially models that can manipulate directly real-valued variables.

Such models are already present in the literature. To the best of our knowledge, there exist three main models of computation that can manipulate real-valued variables. Firstly, there is the Blum-Shub-Smale (BSS) machine \cite{blum2012complexity}, which is a natural extension of Turing machines to manipulate real-valued variables. However, the BSS machine model has a fundamental mismatch with RNNs, because at its core is the use of discontinuous functions: during its computation, a BSS machine has the ability to apply the function $f : \R \to \R$ defined by $f(x) = 0$ if $x < 0$, and $f(x) = 1$ if $x \geq 0$, which is not continuous. Therefore, the BSS machine model cannot be used to study RNNs with ReLU activation function, which are fundamentally continuous. 

Secondly, there is the General Purpose Analog Computer (GPAC) \cite{shannon1941mathematical}, which is a model of computation that can manipulate real-valued variables through the use of ordinary differential equations. Such models have been proven to approximate continuous functions in a sense that is similar to paradigm \eqref{eq:paradigm3} for computable functions \cite{bournez2007polynomial}. This model is probably the closest to RNNs, because it manipulates real-valued variables through the use of continuous transformations. However, the computations of the GPAC model are based on a continuous-time dynamics, while the computations of RNNs are based on a discrete-time dynamics and use update functions that are not piecewise-linear but rather polynomial, which leaves no direct simulation of the GPAC model by RNNs. However, we strongly believe that there exist deep connections between the two models, and we leave the study of these interactions as future work. 

Finally, there is a class of models of computations called hybrid models \cite{henzinger1996theory}, that are constituted of a discrete controller that has access to some continuous variables that can be manipulated. The discrete controller may decide to apply some function to the continuous variables, based on its internal state and on the value of the continuous variables. When the discrete controller does not depend on the continuous variables, we say that the hybrid model is in open-loop. The \TMNU model that we introduce in this paper can be seen as a particular instance of open-loop hybrid models, where the discrete controller is a Turing machine, and where the continuous variables can only be manipulated through the use of piecewise-linear transformations. To the best of our knowledge, there is no literature investigating the approximation capabilities of such models in the sense of paradigm \eqref{eq:paradigm3}. Moreover, in order to keep the proof direct, we do not study the \TMNU model in the context of hybrid models, but we believe that significant generalizations may be possible.

\subsection{\TMNUs and the approximation paradigm for \TMNUs}

Informally speaking, a \TMNU is a Turing machine that has an extra component, which we call the "neural state", that is a vector of real numbers $\neurState \in \R^\neurDim$ of some fixed dimension $\neurDim \in \N$, that we call neural dimension. The neural state is updated at each step by application of some function $f : \R^d \to \R^d$ that depends on the current internal state and the symbol read by the head. The functions that can be applied to the neural state are restricted to be of the form $f(x) = Ax + b$ for some matrix $A \in \R^{d \times d}$ and some vector $b \in \R^d$, or $f(x) = \ReLU_{S}(x)$ where $S \subseteq \{1, \ldots, d\}$ and $\ReLU_S(x)_i := \ReLU(x_i)$ if $i \in S$ and $\ReLU_S(x)_i := x_i$ if $i \notin S$ for every $x = (x_1, \ldots, x_d) \in \R^d$, where $\ReLU(x) := \max(x,0)$ is the ReLU activation function. For $d \in \N$, we let $\reluSet{d} := \{ \ReLU_S : S \subseteq \{1,\ldots,d\}\}$, and $\setAffine{d}{d}$ the set of affine functions from $\R^d$ to $\R^d$. We cast this informal description of \TMNUs into the following definition.
\begin{definition}\label{def:tmnu}
    A \TMNU is a quadruple $\tmnuM := (\nStates, \transFunc, \neurDim, \commFunc)$, where $(\nStates, \transFunc)$ is a Turing machine, $\neurDim \in \N$ is the neural dimension, and $\commFunc : \nStates \times \workSymbols \to \setAffine{d}{d} \cup \Rc_d$ is the command function which specifies a function $f := \commFunc(\state, \symb)$ to be applied to the neural state given an internal state $\state$ and a symbol $\symb$ read by the head. We define
    \begin{equation}
        \Fc_\tmnuM := \{\commFunc(\state, \symb) : \state \in \{1,\ldots,\nStates\}, \symb \in \workSymbols\} \subseteq \setAffine{d}{d} \cup \reluSet{d}, \quad \text{and} \quad \nFunctions{\tmnuM} := \#(\Fc_\tmnuM).
    \end{equation}
\end{definition}
Given an internal state $\state$, a symbol $\symb$ read by the head, and a neural state $\neurState$, the \TMNU $\tmnuM := (\nStates, \transFunc, \neurDim, \commFunc)$ updates its internal state and tape according to the transition function $\transFunc$ given $\state$ and $\symb$, and updates its neural state according to the function specified by the command function $\commFunc$ given $\state$ and $\symb$. More precisely, if we denote by $f := \commFunc(\state, \symb)$ the function specified by the command function $\commFunc$ given the internal state $\state$ and the symbol $\symb$ read by the head, then the neural state is update to $f(\neurState)$. Again, this can be formulated more compactly by using the formalism of configurations, as follows.
\begin{definition}\label{def:configuration-of-tmnu-and-update-of-tmnu}
    Given a \TMNU $\tmnuM := (\nStates, \transFunc, \neurDim, \commFunc)$, we call configuration of $\tmnuM$ any triple $\config := (\state; \tape; \neurState)$ such that $\state \in \{1,\ldots,\nStates\}$, $\tape : \Z \to \workSymbols$ and $\neurState \in \R^\neurDim$. We let $\Cc_\tmnuM := \{1,\ldots,\nStates\} \times \workSymbols^\Z \times \R^\neurDim$ be the set of configurations of $\tmnuM$. We give a meaning of $\tmnuM$ as a map from $\Cc_\tmnuM$ to $\Cc_\tmnuM$ by letting
    \begin{equation}
        \tmnuM(\state; \tape; \neurState) := (M(\state,\tape); \commFunc(\state,\readOp\tape)(\neurState)),
    \end{equation}
    for every configuration $(\state; \tape; \neurState) \in \Cc_\tmnuM$, where $M := (\nStates, \transFunc)$ is the Turing machine component of $\tmnuM$.
\end{definition}
As for Turing machines, the formalism above can be used to define a computation of a \TMNU $\tmnuM := (\nStates, \transFunc, \neurDim, \commFunc)$ starting from some initial configuration $\config \in \Cc_\tmnuM$ as the sequence of configurations $(\tmnuM^t(c))_{t \in \No}$ obtained by iterating the map $\tmnuM : \Cc_\tmnuM \to \Cc_\tmnuM$ starting from $\config$. In order to ease notations, we define projections on the state, tape and neural state of $\tmnuM$ as follows. Given a configuration $\config := (\state; \tape; \neurState)$ of $\tmnuM$, we let 
\begin{equation}\label{eq:projections-on-state-tape-and-neural-state}
    \proj_\state(\config) := \state, \quad \proj_\tape(\config) := \tape, \quad \proj_\neurState(\config) := \neurState.
\end{equation}

As a very important remark, note that the evolution of the internal state $\proj_\state \tmnuM^t(c)$ and tape $\proj_\tape \tmnuM^t(c)$ of $\tmnuM$ is governed entirely by the Turing machine component of $\tmnuM$, i.e., 
\[
   ( \proj_\state\tmnuM^t(c), \proj_\tape\tmnuM^t(c)) = M^t(\proj_\state c, \proj_\tape c) \quad \text{for every } t \in \No.
\]
% and therefore depends only on the initial state and tape, i.e., for every two configurations $\config := (\state; \tape; \neurState)$ and $\config' := (\state'; \tape'; \neurState')$ of $\tmnuM$ such that $\state = \state'$ and $\tape = \tape'$, we have $\proj_\state \tmnuM^t(\config) = \proj_\state \tmnuM^t(\config')$ and $\proj_\tape \tmnuM^t(\config) = \proj_\tape \tmnuM^t(\config')$ for every $t \in \No$. In other words, the evolution of the internal state and tape of $\tmnuM$ they can be considered as the evolution of the Turing machine $M := (\nStates,\transFunc)$. 
This fact makes the \TMNU model easy to manipulate, because one can first establish the evolution of the internal state and tape of $\tmnuM$ by reasoning in terms of Turing machines, and then establish the evolution of the neural state of $\tmnuM$ computing the successive neural states that are obtained by applying the piecewise-linear transformations specified by the command function $\commFunc$, given the evolution of the internal state and tape. 

As mentioned above, beyond their ease of manipulation, the other interest of \TMNUs is that they correspond, in a sense to be made clear in the later sections, to the computations of an RNN. Hence, to show the existence of an RNN that approximates a function $f : [-1,1] \to \R$, it is enough to show the existence of a \TMNU that approximates $f$, in a sense that is consistent with paradigm \eqref{eq:paradigm3}, that we now describe. We will say that a \TMNU $\tmnuM := (\nStates, \transFunc, \neurDim, \commFunc)$ approximates a function $f : [-1,1] \to \R$ if there exists some $u \in \{0,1\}^\#$ such that for every $x \in [-1,1]$, if we start the computation of $\tmnuM$ from the initial configuration $c := (q;\tape;\neurState)$ where $q =1$, $\tape = |u$ and $\neurState$ is the vector of $\R^\neurDim$ whose first coordinate is $x$ and whose other coordinates are $0$, then the last coordinate of the neural state of $\tmnuM$ at time $t$ converges to $f(x)$ as $t \to \infty$, in a uniform way with respect to $x$. To formalize this notion of approximation, we introduce a notation that describes the evolution of the last coordinate of the neural state of $\tmnuM$ at time $t$ when we start the computation of $\tmnuM$ from the initial configuration $c$ as described above.
\begin{definition}\label{def:tmnu-as-operator}
    Given a \TMNU $\tmnuM := (\nStates, \transFunc, \neurDim, \commFunc)$, $x \in \R$ and $u \in \{0,1\}^\#$, we let $\tmnuM^u x : \No \to \R$ be the sequence defined by
    \begin{equation}
        \tmnuM^u x [t] := \proj_\neurDim \circ \proj_\neurState(\tmnuM^t(1;|u; x, 0, \ldots, 0)),
    \end{equation}
    for every $t \in \No$, where $\proj_\neurDim : \R^\neurDim \to \R$ is the projection on the last coordinate of $\R^\neurDim$, and where $\proj_\neurState$ is the projection on the neural state defined in \eqref{eq:projections-on-state-tape-and-neural-state}.
\end{definition}
In the remainder of the paper, we show that given a function $f : [-1,1] \to \R$, there exists a \TMNU $\tmnuM$ and a sequence $u \in \{0,1\}^\#$ such that
\begin{equation}\label{eq:paradigm4}
    \lim_{t \to \infty} \sup_{x \in [-1,1]} |\tmnuM^ux[t] - f(x)| = 0,
\end{equation}
which is consistent with the paradigm \eqref{eq:paradigm3} for RNNs, and we will study the convergence rate of $\sup_{x \in [-1,1]} |\tmnuM^ux[t] - f(x)|$ as $t \to \infty$ for some specific functions $f : [-1,1] \to \R$ with particular smoothness properties. To sum up, the proof of the main result is based on the two main steps:
\begin{enumerate}[label =(\alph*)]
    \item Showing that, under some mild conditions, the paradigm \eqref{eq:paradigm4} for \TMNUs is essentially equivalent to paradigm \eqref{eq:paradigm3} for RNNs, in a sense to be made precise in Section \ref{sec:tmnu-rnn-simulation}.
    \item Showing that \TMNUs can approximate every continuous function $f : [-1,1] \to \R$ in the sense described above, and studying the convergence rate of the approximation for some specific functions $f : [-1,1] \to \R$. This is done in Section \ref{sec:tmnu-constructions}.
\end{enumerate}
The combination of these two steps delivers the main results of the paper.

% \subsection{Organization of the paper}

% \subsection{Notation and useful results}

\section{Simulation of TMNU by RNNs}

\label{sec:tmnu-rnn-simulation}

% \section{Simulation of TMNUs with RNNs}

In this section, we show that RNNs can simulate the computation of \TMNUs, under some boundedness assumption of the \TMNU. Specifically, we introduce a notion of magnitude of a configuration of a \TMNU, and we show that if the configuration of a \TMNU remains bounded during the computation, then there exists an RNN that simulates the computation of this \TMNU. Specifically, given a \TMNU $\tmnuM$ and a real number $C > 0$ we define the set of configurations of $\tmnuM$ with neural state bounded by $C$ as the set given by
\begin{equation}\label{eq:bounded-configurations}
    \Bc_\tmnuM(C) := \{ c \in \Cc_\tmnuM : \| c\| \leq C\},
\end{equation}
where
\begin{equation}\label{eq:magnitude-configuration}
    \|c\| := \|\proj_\neurState(c)\|_\infty,
\end{equation}
and we define the set of initial configurations whose trajectory under $\tmnuM$ has neural state bounded by $C$ as the set given by
\begin{equation}\label{eq:bounded-trajectories}
    \Bc^\infty_\tmnuM(C) := \{ c \in \Cc_\tmnuM : \tmnuM^t(c) \in \Bc_\tmnuM(C) \ \text{for all} \ t \in \No\}.
\end{equation}

% \begin{definition}\label{def:magnitude-configuration}
%     Let $\tmnuM := (\nStates, \neurDim, \transFunc, \commFunc)$ be a \TMNU. Define $\Fc_\tmnuM$ to be the set of functions of $\tmnuM$, i.e.,
%     \begin{equation}
%         \Fc_\tmnuM := \{ \commFunc(\state,\symb) : \state \in \{1,\ldots, \nStates\}, \symb \in \workSymbols\}.
%     \end{equation}
%     We define the magnitude of a configuration $c := (\state; \tape; \neurState)$ of $\tmnuM$ as the quantity given by
%     \begin{equation}\label{eq:magnitude-configuration}
%         \|c\| := \|\neurState\|_\infty,
%     \end{equation}
%     its \newname{potential magnitude} as
%     \begin{equation}\label{eq:potential-magnitude}
%         \| c\|_\tmnuM := \max_{f \in \Fc_\tmnuM} \| f(\neurState) \|_\infty,
%     \end{equation}
%     and its \newname{potential magnitude horizon} by
%     \begin{equation}
%         \|c\|^\infty_\tmnuM := \sup_{t \in \No} \|\tmnuM^t(c)\|_\tmnuM.
%     \end{equation}
%     Given $C > 0$, we define the set of configurations of $\tmnuM$ with potential magnitude horizon bounded by $C$ as the set given by
%     \begin{equation}
%         \Bc_\tmnuM(C) := \{ c \in \Cc_\tmnuM : \|c\|^\infty_\tmnuM \leq C\}.
%     \end{equation}
% \end{definition}

In this section, we will then show that given a \TMNU $\tmnuM := (\nStates, \neurDim, \transFunc, \commFunc)$ and $C > 0$, there exists an RNN $\Rc := \Rc_{\tmnuM,C}$, a natural number $t_\tmnuM \in \No$ and a mapping $\configMap{\tmnuM} : \Cc_\tmnuM \to \R^{n + 2 + d}$, such that for every configuration $c \in \Bc^\infty_\tmnuM(C)$ and every $t \in \No$, we have
\begin{equation}\label{eq:simulation-tmnu-by-rnn}
    \Rc \Dc \configMap{\tmnuM}(c) [t] = \configMap{\tmnuM}(\tmnuM^{t//t_\tmnuM} (c)),
\end{equation}
where $//$ denotes the integer division. The design of $\Rc$ and $\configMap{\tmnuM}$ is largely inspired by the construction in \cite{siegelmann1992computational} of an RNN that simulates a Turing machine, by means of simulating an intermediate machine called a stack machine. Our construction here does not rely on such an intermediate machine, and directly simulates the \TMNU. The construction is effected in two steps. First, we show that there exists a continuous piecewise-linear function $F := F_{\tmnuM,C}$ that implements $\tmnuM$ under some boundedness condition on its neural state, in the sense that
\begin{equation}\label{eq:simulation-tmnu-by-piecewise-linear-function}
    \configMap{\tmnuM} \circ \tmnuM(c) = F \circ \configMap{\tmnuM}(c), \quad c \in \Bc_\tmnuM(C).
\end{equation}
Then, we show how to use $F$ to design an RNN $\Rc$ that simulates the computation of $\tmnuM$ as in \eqref{eq:simulation-tmnu-by-rnn}.

We first explain how $\configMap{\tmnuM}$ is designed. Given a configuration $\config := (\state; \tape; \neurState)$, we want to encode it as a vector $x^c \in \R^{\nStates + 2 + \neurDim}$, precisely of the form
\[
    x^c := (\oneHot{\nStates}{\state}, x^\tape, \neurState),
\]
where $\oneHot{\nStates}{\state}$ is the one-hot encoding of the state $\state$, specifically given by
\begin{equation}\label{eq:one-hot-encoding-state}
    \oneHot{\nStates}{\state} := (\kroen{1}{\state}, \kroen{2}{\state}, \ldots, \kroen{\nStates}{\state}) \in \{0,1\}^\nStates,
\end{equation}
such that $\kroen{i}{\state} = 1$ if $i = \state$ and $\kroen{i}{\state} = 0$ otherwise is the usual Kronecker symbol, and $x^\tape \in [-1,1]^2$ is a pair of real numbers that encodes the tape $\tape$ that we define as follows. $x_1^\tape$ encodes the right part of the tape, and $x_2^\tape$ encodes its left part, i.e., by denoting
\begin{equation}\label{eq:right-left-tape}
    \tapeR{\tape} := (\tape_{i-1})_{i \in \N}, \quad \text{and} \quad \tapeL{\tape} := (\tape_{-i})_{i \in \N},
\end{equation}
we have that $x_1^\tape$ encodes $\tapeR{\tape}$ and $x_2^\tape$ encodes $\tapeL{\tape}$. Specifically, we define the encoding of a sequence $(u_i)_{i\in\N} \in \workSymbols^\N$ as the real number given by
\begin{equation}\label{eq:encoding-sequence}
    \begin{array}{rrlc}
        \tilde \cantorMap : & \workSymbols^\N & \to & [-1,1]\\
        & (u_i)_{i \in \N} & \mapsto & 4 \sum_{i=1}^\infty u_i 4^{-i}.
    \end{array}
\end{equation}
\begin{figure}[t]
    \centering
    \begin{tikzpicture}[baseline=(current bounding box.center), every node/.style={font=\small}, >=Latex]
      \node (c) at (0,0) {$\Bc_\tmnuM(C)$};
      \node (mc) at (4.8,0) {$\Cc_\tmnuM$};
      \node (gammac) at (0,-2.2) {$\R^m$};
      \node (fgammac) at (4.8,-2.2) {$\R^m$};

      \draw[->, thick] (c) -- node[above] {$\tmnuM$} (mc);
      \draw[->, thick] (c) -- node[left] {$\gamma_\tmnuM$} (gammac);
      \draw[->, thick] (mc) -- node[right] {$\gamma_\tmnuM$} (fgammac);
      \draw[->, thick, blue!70!black] (gammac) -- node[below] {$F_{\tmnuM,C}$} (fgammac);

    %   \node[draw, rounded corners, fill=blue!5, inner sep=4pt, align=center] at (8.4,-1)
    %   {commutes for every $\config \in \mathcal C_\tmnuM$\\ with $\|\neurState\|_\infty \leq C$};
    \end{tikzpicture}
    \caption{Principle of the simulation of $\tmnuM$ with a piecewise-linear function $F_{\tmnuM,C}$ under some boundedness condition on its neural state. The simulation is expressed by the fact that the above diagram commutes. One challenge is ensure sure that $\tmnuM$ maps $\Bc_\tmnuM(C)$ to itself, so that one can iterate the simulation.}
\end{figure}
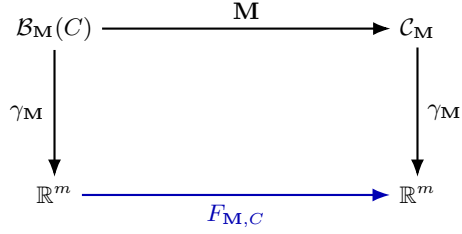
The interest of such an encoding resides in the fact that one can read the first symbol $u_1$ of a sequence $(u_i)_{i \in \N}$ by application of a continuous piecewise-linear function to $\tilde \cantorMap((u_i)_{i \in \N})$, which is a key ingredient for the simulation of the reading operation of the tape by a neural network. We can then define the encoding of a tape $\tape$ as the pair of real numbers given by $\tilde \cantorMap(\tapeR{\tape})$ and $\tilde \cantorMap(\tapeL{\tape})$, i.e., we define the encoding of a tape $\tape$ as the vector given by
\begin{equation}\label{eq:encoding-tape}
    \begin{array}{rrlc}
        \cantorMap : & \workSymbols^\Z & \to & [-1,1]^2\\
        & \tape & \mapsto & (\tilde \cantorMap(\tapeR{\tape}), \tilde \cantorMap(\tapeL{\tape})).
    \end{array}
\end{equation}
To sum up, we have defined the mapping $\configMap{\tmnuM}$ as follows. Given a configuration $\config := (\state; \tape; \neurState)$, we have
\begin{equation}\label{eq:encoding-configuration}
    \begin{array}{rrlc}
        \configMap{\tmnuM} : & \states \times \workSymbols^\Z \times \neurSpace & \to & \R^{\nStates + 2 + \neurDim}\\
        & (\state; \tape; \neurState) & \mapsto & (\oneHot{\nStates}{\state}, \cantorMap(\tape), \neurState).
    \end{array}
\end{equation}

We now explain the two steps of the construction of the RNN $\Rc$ that simulates the computation of $\tmnuM$. We only sketch the main ideas of the construction here, but the exact details can be found in Appendix \ref{sec:tmnu-rnn-simulation-appendix}. The first step is to show that there exists a piecewise-linear function $F$ such that \eqref{eq:simulation-tmnu-by-piecewise-linear-function} holds. Following the spirit of \cite{siegelmann1992computational}, we simply notice that every component of $\tmnuM$ can be realized by a piecewise-linear function. Specifically, the reading, writing and shifting operations on the tape can be implemented by continuous piecewise-linear functions over the encoding of the tape. Moreover, the transition function $\transFunc$ being simply a lookup table, it can be implemented by a continuous piecewise-linear function by linear interpolation. The command function $\commFunc$ can be seen as choosing a function $f$ among the finite set of functions $\Fc_\tmnuM$ depending on the state and the symbol read on the tape, which can also be formulated as lookup table and hence be implemented by a continuous piecewise-linear function by linear interpolation. Finally, applying the function $f$ chosen by the command function $\commFunc$ is the more delicate part, and the reason for the boundedness assumption. The strategy is the following. First, upon noting that the functions $f_i \in \Fc_\tmnuM$, $i \in \{1,\ldots, \nFunctions{\tmnuM}\}$ are piecewise-linear, the function
\[
   \neurState \in \R^\neurDim \mapsto (f_1(\neurState), f_2(\neurState), \ldots, f_{\nFunctions{\tmnuM}}(\neurState)) \in \R^{\nFunctions{\tmnuM}d}.
\]
Now, assuming that the command function has chosen a function $f_i$ among the functions of $\tmnuM$, we want to design a piecewise-linear function that takes as input the vector $(f_1(\neurState), f_2(\neurState), \ldots, f_{\nFunctions{\tmnuM}}(\neurState))$ and returns $f_i(\neurState)$. This can be done by remarking that
\begin{equation}\label{eq:linear-combination-functions}
    f_i(\neurState) = \sum_{j=1}^{\nFunctions{\tmnuM}} \kroen{i}{j} f_j(\neurState),
\end{equation}
and that, by extending slightly \cite[Lemma 4.1]{siegelmann1992computational}, one has the following property.
\begin{lemma}\label{lem:linear-combination-of-functions-with-relu}
    Let $M > 0$, $z \in [0,M]$ and $b \in \{0,1\}$. Then, we have $\ReLU(z + M(b-1)) = b z$.
\end{lemma}
\begin{proof}
    If $b = 0$, then we have $\ReLU(z + M(b-1)) = \ReLU(z - M) = 0$. If $b = 1$, then we have $\ReLU(z + M(b-1)) = \ReLU(z) = z$. This concludes the proof.
\end{proof}
Note that for every $j \in \{1, \ldots, \nFunctions{\tmnuM}\}$, $f_j$ is continuous, and hence for every $C > 0$, we have
\begin{equation}
    \sup_{\neurState \in [-C,C]^d} \|f_j(\neurState)\|_\infty < \infty.
\end{equation}
Accordingly, we define
\begin{equation}
    \|\tmnuM\|_C := \max_{j \in \{1, \ldots, \nFunctions{\tmnuM}\}} \sup_{\neurState \in [-C,C]^d} \|f_j(\neurState)\|_\infty,
\end{equation}
and by application of Lemma \ref{lem:linear-combination-of-functions-with-relu} with $M = \|\tmnuM\|_C$, we can reformulate \eqref{eq:linear-combination-functions} as
\begin{equation}\label{eq:linear-combination-functions-relu}
    f_i(\neurState) = \sum_{j=1}^{\nFunctions{\tmnuM}} \ReLU(f_j(\neurState) + \|\tmnuM\|_C(\kroen{i}{j} - 1)),
\end{equation}
which is a piecewise-linear function. Combining the piecewise-linear functions defined above results in establishing the existence of a piecewise-linear function $F$ such that \eqref{eq:simulation-tmnu-by-piecewise-linear-function} holds. In fact, in Appendix \ref{subsec:construction-of-a-neural-network-that-simulates-a-tmnu:appendix}, we get more quantitative about the shape of this piecewise-linear function $F$. In fact, the piecewise-linear function $F$ that we construct is a deep ReLU neural network, defined as follows.
\begin{definition}(ReLU Neural network)\label{def:neural network}
    We call ReLU neural network an ordered sequence
    \begin{equation}\label{eq:definition-of-neural-network}
        \nn := (N_0, N_1, \ldots, N_L; A^1, b^1, A^2, b^2, \ldots, A^L, b^L),
    \end{equation}
    where $L \in \N$ and $N_0, N_1, \ldots, N_L \in \N$, $A^\ell = (A^\ell_{jk}) \in \R^{N_\ell \times N_{\ell-1}}$ and $b^\ell = (b^\ell_j) \in \R^{N_\ell}$ for $\ell \in \{1, \ldots, L\}$. We consider the ReLU neural network $\nn$ as a function $\nn : \R^{N_0} \to \R^{N_L}$ defined by
    \begin{equation}\label{eq:function-represented-by-a-neural-network}
        \nn(\realx) := A^L \ReLU\left( A^{L-1} \ReLU\left( \ldots \ReLU\left( A^1 \realx + b^1 \right) + b^{L-1} \right) + b^L \right) \text{ for all } \realx \in \R^{N_0},
    \end{equation}
    where $\ReLU$ is applied coordinate-wise. We let 
    \[
        \nnWidth{\nn} := \max_{\ell \in \{0, \ldots, L\}} N_\ell, \quad \nnDepth{\nn} := L, \quad \nnMag{\nn} := \max_{\ell \in \{1,\ldots,L\}} \|A^\ell\|_\infty \vee \|b^\ell\|_\infty.
    \]
    We let $\nns{d}{k}$ be the set of ReLU neural networks with input dimension $N_0 = d$ and output dimension $N_L = k$.
\end{definition}
We generally say that a neural network $\nn$ is \newname{deep} if $\nnDepth{\nn} > 2$. Our final simulation theorem stipulates that we can simulate the iterations of a $\tmnuM$ by the iterations of a deep ReLU neural network (of depth $4$), and that the width and the magnitude of the weights and biases of this neural network can be upper bounded by a quantity that depends on $\nStates$, $\neurDim$, $\nFunctions{\tmnuM}$ and $\|M\|_C$.
\begin{theorem}\label{thm:transition-function-as-neural-network}
    Let $\tmnuM := (\nStates, \neurDim, \transFunc, \commFunc)$ be a \TMNU and $C > 0$. Then, there exists a ReLU neural network $\nn \in \nns{\nStates + 2 + \neurDim}{\nStates + 2 + \neurDim}$ such that 
    \begin{equation}
        \nnDepth{\nn} = 4, \quad \nnWidth{\nn} \leq \max\{3\nStates, \nStates + \nFunctions{\tmnuM} + 9\} + 2\neurDim\nFunctions{\tmnuM} + 18, \quad \text{and} \quad \nnMag{\nn} \leq \max\{4, \|M\|_C\},
    \end{equation}
    satisfying
    \begin{equation}
        \nn(\configMap{\tmnuM}(c)) = \configMap{\tmnuM}(M(c)), \quad c \in \Bc_\tmnuM(C).
    \end{equation}
\end{theorem}
\begin{proof}
    The proof is carried out by designing concrete ReLU neural networks that implement the continuous piecewise-linear functions defined above, and combining them into a larger neural network. For the full proof, see the proof of Theorem \ref{thm:transition-function-as-neural-network:appendix}.
\end{proof}

We now convert this result into the existence of an RNN $\Rc$ that simulates the computation of $\tmnuM$ as in \eqref{eq:simulation-tmnu-by-rnn}. Specifically, we show that the iterations of a deep ReLU neural network can be simulated by an RNN.

\begin{theorem}\label{lem:simulation-of-cpwl-function-iterations-with-an-rnn}
    Let $n \in \N$ and $\nn \in \nns{n}{n}$ be a ReLU neural network. Then, there exists an RNN $\Rc$ such that
    \begin{equation}
        \rnnHidDim{\Rc} \leq (2\nnDepth{\nn}+2)(2n \vee \nnWidth{\nn}+1), \quad \text{and} \quad \nnMag{\Rc} \leq \max\{4, \nnMag{\nn}\},
    \end{equation}
    satisfying
    \[\Rc \Dc x[t] = \nn^{t//\nnDepth{\nn}}(x), \quad t \in \N, \quad x \in \R^n.\]
\end{theorem}
\begin{proof}[Sketch of the proof]
    The proof relies on designing an RNN $\Rc_f$ that reflects the structure of $\nn$. Essentially, at each time step, the RNN performs one of the affine transforms $A^\ell$, adds the corresponding bias $b^\ell$, and applies the ReLU function. After $L := \nnDepth{\nn}$ time steps, the RNN will have computed $\nn(x)$, and then it can repeat this process to compute $\nn(\nn(x))$, and so on. This construction comes with some technicalities. For example, the RNN cannot hold negative values in its hidden state, so we need to design the RNN to hold both the positive and negative parts of the intermediate computations, and then combine them appropriately to recover the correct output. Moreover, we need a mechanism to retain the value of $\nn^i(x)$ for $L$ iterations, while the RNN is performing the computations for $\nn^{i+1}(x)$, and replace this value with $\nn^{i+1}(x)$ after $L$ iterations. This can be achieved by using some additional coordinates in the hidden state of the RNN to keep track of the current iteration and to store the intermediate values. The details of this construction can be found in Appendix \ref{subsection:rnn-simulates-neural-network-iterations:appendix}.
\end{proof}

We finally conclude that the approximation paradigm \eqref{eq:paradigm3} for RNNs corresponds to the approximation paradigm \eqref{eq:paradigm4} for \TMNUs, given that a certain boundedness condition holds. Given a \TMNU $\tmnuM$ and $C>0$, we say that $\tmnuM$ has uniformly $C$-bounded trajectories at $u \in \{0,1\}^\N$ if
\begin{equation}
    \{c_x = (1; |u; x, 0, \ldots, 0) : x \in [-1,1]\} \subseteq \Bc^\infty_\tmnuM(C). 
\end{equation}
\begin{theorem}\label{thm:simulation-of-a-tmnu-by-an-rnn}
    Let $\tmnuM := (\nStates, \neurDim,\transFunc,\commFunc)$ be a \TMNU, $C>0$ and $u \in \{0,1\}^\N$ such that $\tmnuM$ has uniformly $C$-bounded trajectories at $u$. Then, there exists an RNN $\Rc$ such that
    \begin{equation}
        \rnnHidDim{\Rc} \leq 10 (\max\{3\nStates, \nStates + \nFunctions{\tmnuM} + 9\} + 2 \neurDim \nFunctions{\tmnuM} + 19) + 1, \quad \text{and} \quad \nnMag{\Rc} \leq \max \{4, \|\tmnuM\|_C\},
    \end{equation}
    satisfying
    \begin{equation}
        \Rc \Dc x[t] = \tmnuM^ux[t//4], \quad t \in \No, \quad x \in [-1,1].
    \end{equation}
\end{theorem}
\begin{proof}[Sketch of the proof]
    Let $\tmnuM = (\nStates, \neurDim, \transFunc, \commFunc)$ be a \TMNU, $C>0$ and $u \in \{0,1\}^\N$ such that $\tmnuM$ is uniformly $C$-bounded at $u$. Upon invoking Theorems \ref{thm:transition-function-as-neural-network} and \ref{lem:simulation-of-cpwl-function-iterations-with-an-rnn} together, we get that there exists an RNN $\Rc$ such that
    \[
        \rnnHidDim{\Rc} \leq 10 (\max\{3\nStates, \nStates + \nFunctions{\tmnuM} + 9\} + 2 \neurDim \nFunctions{\tmnuM} + 19), \quad \text{and} \quad \nnMag{\Rc} \leq \max \{4, \|\tmnuM\|_C\},
    \]
    satisfying
    \begin{equation}        \Rc \Dc(\configMap{\tmnuM}(c))[t] = \configMap{\tmnuM}(\tmnuM^{t//4}(c)), \quad t \in \No, \quad c \in \Bc^\infty_\tmnuM(C).
    \end{equation}
    Now, note that since $\tmnuM$ is uniformly $C$-bounded at $u$, we have that for every $x \in [-1,1]$, $c_x := (1; |u; x, 0, \ldots, 0)\in\Bc^\infty_\tmnuM(C)$, so that for every $x \in [-1,1]$ and $t \in \No$, we have
    \begin{equation}
        \Rc \Dc \configMap{\tmnuM}(c_x)[t] = \configMap{\tmnuM}(\tmnuM^{t//4}(c_x)).
    \end{equation}
    Moreover, for every $x \in [-1,1]$, we have
    \[
        \configMap{\tmnuM}(c_x) = (\oneHot{\nStates}{1}, \cantorMap(|u), x, 0, \ldots, 0) = (0_{\nStates+2}, x, 0_{d-1})x + (\oneHot{\nStates}{1}, \cantorMap(|u), 0, \ldots, 0) =: Ax + b,
    \]
    and moreover,
    \[
        \tmnuM^u x[t] = \proj_d \proj_\neurState \tmnuM^{t}(c_x) = \proj_{\nStates+2+\neurDim} \configMap{\tmnuM}(\tmnuM^{t}(c_x)) =: A' \configMap{\tmnuM}(\tmnuM^{t}(c_x)).
    \]
    We show in Lemmata \ref{lem:linear-transform-of-the-output-of-an-rnn} and \ref{lem:affine-transform-of-the-input-of-an-rnn} that RNNs are essentially stable through affine transformations of their inputs and outputs, and that we can design an RNN $\Rc'$ such that
    \begin{equation}
        \rnnHidDim{\Rc'} = \rnnHidDim{\Rc} + 1 \quad \text{and} \quad \nnMag{\Rc'} = \max\{4, \|M\|_C\},
    \end{equation}
    satisfying
    \begin{equation}        \Rc' \Dc x[t] = A'\Rc \Dc (Ax + b)[t], \quad t \in \No, \quad x \in [-1,1].
    \end{equation}
    Therefore, we have
    \begin{equation}        
        \Rc' \Dc x[t] = A' \Rc \Dc (Ax + b)[t] = A' \configMap{\tmnuM}(\tmnuM^{t//4}(c_x)) = \proj_d \proj_\neurState \configMap{\tmnuM}(\tmnuM^{t//4}(c_x)) = \tmnuM^u x[t//4].
    \end{equation}
    For more details, see the proof of Theorem \ref{thm:simulation-of-a-tmnu-by-an-rnn:app}. This concludes the proof.
\end{proof}

We hence have reduced approximation paradigm \eqref{eq:paradigm3} for RNNs to approximation paradigm \eqref{eq:paradigm4} for \TMNUs. In the next section, we will solve the approximation paradigm \eqref{eq:paradigm4} for continuous functions, and hence conclude that approximation paradigm \eqref{eq:paradigm3} for RNNs can be solved.

\section{TMNU constructions}

\label{sec:tmnu-constructions}

In this section, we explain the construction of a \TMNU that approximates every continuous function $f : [-1,1] \to \R$ in the sense of paradigm \eqref{eq:paradigm4}. The mathematical strategy is supplied by the Weierstrass approximation theorem: every continuous function on $[-1,1]$ is the uniform limit of a sequence of polynomials. By truncating the binary expansions of their coefficients, we obtain a sequence of finitely encodable dyadic polynomials $(P_i)_{i\in\No}$ that still converges uniformly to $f$. Thus, at an abstract level, it suffices to design a machine that reads the successive finite encodings of each $P_i$ from its tape, evaluates the polynomials successively at the input $x$, and outputs increasingly accurate approximations of $f(x)$. The construction below may therefore be viewed as an implementation of the Weierstrass approximation theorem by a \TMNU.

The key question is how one fixed machine can implement this increasingly complex sequence of polynomial computations. Our answer is to treat \TMNUs as programs assembled from reusable subroutines. Rather than defining the final machine $\Continuous$ directly and analyzing one large transition system, we first construct machines performing elementary arithmetic operations. We then embed them as subroutines of progressively more expressive machines. This subroutine mechanism is not merely a device for making the proof manageable; it is also the intuition behind the main theorem of the paper. It suggests that one fixed \TMNU can implement the Weierstrass approximation procedure for every continuous function by composing a finite collection of elementary operations according to the information encoded on its tape. Since a uniformly bounded \TMNU computation can be transferred to an RNN by Theorem \ref{thm:simulation-of-a-tmnu-by-an-rnn}, this gives a conceptual reason to expect that a fixed-size RNN can approximate arbitrary continuous functions.

At the technical level, the subroutine mechanism transfers the complete trajectory of a smaller machine into every larger machine that calls it. Specifically, if a machine $\tmnuM$ contains another machine $\tmnuN$ as a subroutine, then on prescribed states and neural coordinates, $\tmnuM$ follows exactly the trajectory of $\tmnuN$, while its remaining neural coordinates are carried along unchanged. The \emph{shadow} of an ambient configuration extracts the configuration seen by $\tmnuN$, while the \emph{lift} inserts the resulting subroutine configuration back into $\tmnuM$. Schematically, throughout a call to the subroutine, 
\[
    \text{trajectory in }\tmnuM
    =
    \text{lift}\bigl(\text{trajectory in }\tmnuN\bigr).
\]
This identity means that a lemma proved for $\tmnuN$ immediately describes the corresponding portion of the trajectory of $\tmnuM$. Consequently, the input-output behavior, running time, approximation error, and trajectory bound of a machine can all be reused at the next level of the construction. This makes it possible to control the two quantitative properties needed later: the convergence rate of $\Continuous$ and the uniform boundedness of its complete trajectory. The latter is precisely the hypothesis required by Theorem \ref{thm:simulation-of-a-tmnu-by-an-rnn} to transfer the final \TMNU construction to an RNN. 

\subsection{Overview of the \TMNU constructions for the main result}

The construction is organized as the following hierarchy of subroutine calls:
\[
    \Sign,\ \Contr^+ \longrightarrow \Contr,
    \qquad
    \Contr,\ \Times \longrightarrow \UpPoly,
    \qquad
    \UpPoly,\ \Scale \longrightarrow \Poly,
    \qquad
    \Poly \longrightarrow \Continuous.
\]
The machine $\Scale$ reads a block $1^k0$ and sends $x$ to $2^kx$. The machines $\Sign$ and $\Contr^+$ respectively apply a sign and multiply by a nonnegative dyadic number $a \in [0,1)$, and are combined as subroutines of $\Contr$ to implement $x \mapsto ax$ for any dyadic $a \in (-1,1)$. Finally, $\Times$ reads a computational parameter $1^n0$ and approximates $(x,y) \mapsto xy$ up to an error of order $2^{-2n}$. These elementary machines form the reusable arithmetic instructions of the construction.

At the next level, $\UpPoly$ calls $\Contr$ and $\Times$ to perform one monomial-accumulation update. If the neural state contains the input $x$, a current approximation $y$ of the monomial $x^i$, and a current partial sum $z$, while the tape contains a dyadic coefficient $a_i$ together with a multiplication parameter $n$, then $\UpPoly$ approximately sends
\[
    (x,y,z) \mapsto (x,yx,z+a_i y).
\]
Starting from $y=1$ and $z=0$, the machine $\Poly$ repeatedly calls the $\UpPoly$ subroutine to compute the successive monomial contributions $a_i x^i$ and add them to the partial sum. For the boundedness reasons explained below, it actually evaluates a rescaled polynomial $P^\ast$ and then calls $\Scale$ to recover the corresponding approximation of $P$. Thus, for a dyadic polynomial $P$ and a multiplication parameter $n$, $\Poly$ produces an approximation $F_{P,n}(x)$ of $P(x)$. Its behavior and trajectory bound follow by composing the previously established properties of $\UpPoly$ and $\Scale$.

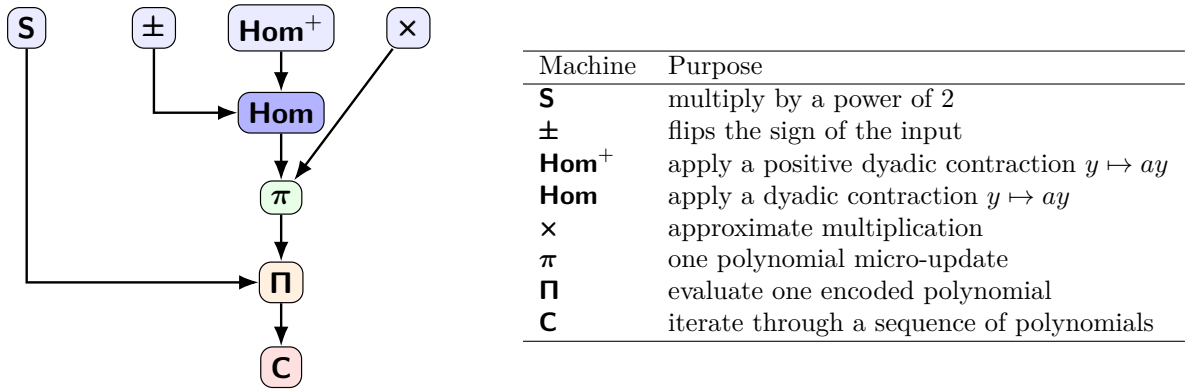
\begin{figure}[b]
\centering
\begin{minipage}{0.35\textwidth}
  \centering
        \resizebox{\linewidth}{!}{%
        \begin{tikzpicture}[>=Latex]
    \node[draw,rounded corners,fill=blue!8] (s) at (0,0) {\Scale};
    \node[draw,rounded corners,fill=blue!8] (si) at (1.5,0) {\Sign};
    \node[draw,rounded corners,fill=blue!8] (c+) at (3,0) {$\Contr^+$};
    \node[draw,rounded corners,fill=blue!8] (t) at (4.5,0) {\Times};
    \node[draw,rounded corners,fill=blue!30] (c) at (3,-1) {\Contr};
    \node[draw,rounded corners,fill=green!10] (u) at (3,-2) {\UpPoly};
    \node[draw,rounded corners,fill=orange!12] (p) at (3,-3) {\Poly};
    \node[draw,rounded corners,fill=red!12] (cont) at (3,-4) {\Continuous};

    \draw[->,thick] (c+) -- (c);
    \draw[->,thick] (si) |- (c);
    \draw[->,thick] (c) -- (u);
    \draw[->,thick] (t) -- (u);
    \draw[->,thick] (u) -- (p);
    \draw[->,thick] (s) |- (p);
    \draw[->,thick] (p) -- (cont);
  \end{tikzpicture}%
  }
\end{minipage}\hfill
\begin{minipage}{0.6\textwidth}
  \centering
         \begin{tabular}{ll}
    \hline
    Machine & Purpose \\
    \hline
    \Scale & multiply by a power of 2 \\
    \Sign & flips the sign of the input \\
    $\Contr^+$ & apply a positive dyadic contraction $y \mapsto a y$\\
    \Contr & apply a dyadic contraction $y \mapsto ay$ \\
    \Times & approximate multiplication \\
    \UpPoly & one polynomial micro-update \\
    \Poly & evaluate one encoded polynomial \\
    \Continuous & iterate through a sequence of polynomials \\
    \hline
  \end{tabular}
\end{minipage}
\caption{Subroutine hierarchy for evaluating dyadic polynomials in the monomial basis.}
\label{fig:monomial-tmnu-hierarchy}
\end{figure}

At the top of the hierarchy, $\Continuous$ repeatedly calls $\Poly$. Its tape is an infinite encoding of dyadic polynomials $P_i$ and multiplication parameters $n_i$, and during its $i$-th cycle the $\Poly$ subroutine computes $F_{P_i,n_i}(x)$, which $\Continuous$ stores as its current output. If $P_i$ converges uniformly to a continuous function $f$ and the parameters $n_i$ are chosen so that $F_{P_i,n_i}$ is close to $P_i$, then these successive outputs converge to $f$. Once again, the subroutine lemma turns the analysis of each complete cycle of $\Continuous$ into the already established analysis of $\Poly$.

The main additional difficulty is uniform boundedness. Convergence of the outputs alone would establish an approximation result for $\Continuous$, but would not yet allow us to obtain the desired RNN: the correspondence result of Theorem \ref{thm:simulation-of-a-tmnu-by-an-rnn} applies only when the simulated \TMNU computation is uniformly bounded. Although each individual computation of $\Poly$ is bounded by a finite quantity depending on the $\ell^1$-norm of its coefficients,
\[
    \|P_i\|_1 := \sum_{j=0}^{\deg(P_i)} |a_j|,
\]
the sequence $(\|P_i\|_1)_{i\in\No}$ need not be uniformly bounded. To avoid transferring this possible growth to the neural state, $\Poly$ first evaluates a rescaled polynomial $P_i^\ast$ satisfying $\|P_i^\ast\|_1<1$, so that all monomials and partial sums remain bounded by one, and then calls $\Scale$ to recover the desired output. Consequently, the $i$-th cycle is bounded in terms of $1 \vee |F_{P_i,n_i}(x)|$, rather than $\|P_i\|_1$. Since the outputs remain close to $f$, the complete trajectory of $\Continuous$ is uniformly bounded in terms of $1 \vee \|f\|_{L^\infty([-1,1])}$. This verifies the hypothesis needed to apply Theorem \ref{thm:simulation-of-a-tmnu-by-an-rnn} and transfer the construction from $\Continuous$ to an RNN.

By composing the subroutine lemmas along the hierarchy above, we obtain the following main \TMNU approximation statement, which is the basis for Theorem \ref{thm:main-theorem-continuous}.
\begin{theorem}\label{thm:continuous-approximation-tmnu-main}
    For every continuous function $f : [-1,1] \to \R$, there exists $u \in \{0,1\}^\N$ such that the machine $\Continuous$ is uniformly $C_f$-bounded at $u$ and satisfies
    \[
        \lim_{t \to \infty} \|\Continuous^ux[t] - f(x)\|_{L^{\infty}([-1,1])} = 0,
    \]
    % and 
    % \[
    %     (1;|u;x,0_{16}) \in \Bc_\Continuous(C_f).
    % \]
    where $C_f := 1 + \|f\|_{L^\infty([-1,1])}$.
\end{theorem}
\begin{proof}
    See Theorem \ref{thm:continuous-approximation-tmnu}.
\end{proof}

\subsection{Convergence rates for polynomials}
The same modular analysis also gives quantitative rates. Each subroutine lemma records its running time and approximation error; composing these estimates gives the cost and error of every cycle of $\Continuous$. Importantly, the output produced during cycle $i$ is reached only after all preceding polynomials $P_0,\ldots,P_i$ have been read and evaluated. The relevant time cost is therefore the cumulative time
\[
    N_i \asymp \sum_{j=0}^{i} \operatorname{cost}(P_j),
\]
rather than the cost of evaluating $P_i$ alone. To obtain a sharp rate in the machine time $t$, the sequence $(P_i)_{i\in\No}$ must be chosen so that its approximation error decreases rapidly compared with this cumulative cost.

Consider first a fixed polynomial $P$ with real coefficients. A naive first choice would be to let cycle $i$ evaluate the dyadic polynomial obtained by truncating the coefficients of $P$ after $i$ binary digits. This gives an error of order $2^{-i}$ and, since the degree of $P$ is fixed, a cost of order $i$ for cycle $i$. However, the cumulative time then satisfies
\[
    N_i \asymp \sum_{j=0}^{i} j \asymp i^2,
\]
so an error of order $2^{-i}$ becomes only an error of order $2^{-\sqrt{t}}$ when expressed in terms of machine time.

To avoid this loss, cycle $i$ instead uses a dyadic polynomial $P_i$ obtained by truncating the coefficients of $P$ after a number of bits of order $2^i$. Its approximation error is then of order $2^{-2^i}$, while its encoding and evaluation cost is of order $2^i$. Moreover, the geometric growth makes the cost of evaluating the latest polynomial comparable with the cost of evaluating the entire sequence up to that point:
\[
    \operatorname{cost}(P_i) \asymp 2^i
    \qquad\text{and}\qquad
    N_i \asymp \sum_{j=0}^{i} 2^j \asymp 2^i.
\]
Thus, by the time $t$ is of order $2^i$, the output error is of order $2^{-2^i}$, which becomes an error of order $2^{-t}$. In this sense, the exponential truncation schedule is optimal up to constant factors for this sequential strategy: evaluating $P_i$ alone already costs order $2^i$, and evaluating all earlier approximations adds only a comparable amount of work. This strategy results in the following exponential convergence rate for polynomials.
\begin{theorem}\label{th:convergence-rate-polynomial}
    Let $N \in \N$, and let $P(x) := \sum_{i=0}^N a_i x^i$ for some $a_0, \ldots, a_N \in \R$. Then, there exists $u \in \{0,1\}^\N$ such that $\Continuous$ is uniformly $C_P$-bounded at $u$, where $C_P := 1 + \|P\|_{L^\infty([-1,1])}$, and satisfying
    \[
        \left|\Continuous^ux[t] - P(x)\right| \leq 2 \cdot 2^{-\frac{t}{t_P}}, \quad \text{for all} \ x \in [-1,1], \ t \geq \tau_P,
    \]
    where $t_P := 18(N+1)$, $\tau_P := t_P(6\|a\|_1 + N + 11) + 1$.
\end{theorem}
\begin{proof}
    See Theorem \ref{thm:convergence-rate-polynomial} in Appendix \ref{sec:tmnu-construction-appendix}.
\end{proof}

\subsection{Convergence rates for general continuous functions}

For more general continuous functions, the strategy is to specifically choose the approximating polynomials $P_i$ to be the Chebyshev truncations of $f$. However, the monomial construction above is not well-suited for this choice, because its would require first expanding the Chebyshev truncation in powers of $x$ and then evaluating the resulting monomial sum. The coefficients of the monomial expansion grow exponentially with the degree, so the the downscaling-upscaling scheme implemented in the machine $\Poly$ to keep the trajectory bounded would cause an additional time cost that leads to poor convergence rates. To avoid this issue, the construction no longer
passes through the monomial machine $\Poly$. Instead, it uses the Chebyshev
basis directly.
The relevant hierarchy of subroutines is
\[
    \Times \longrightarrow \ChebStep,
    \qquad
    \Contr,\ \ChebStep,\ \Scale \longrightarrow \ChebSum,
    \qquad
    \ChebSum \longrightarrow \ChebContinuous.
\]
Here, upon receiving a parameter $n \in \N$, $\ChebStep$ is used to compute approximations of the Chebyshev polynomials by calling the machine $\Times$ to approximate the recurrence relation $T_{k+2}(x)=2xT_{k+1}(x)-T_k(x)$, with a precision of order $2^{-2n}$. Upon being iterated, $\ChebStep$ hence compute approximations $\tilde T_{k,n}(x)$ of the Chebyshev polynomials $T_k(x)$ with a uniform error of $k^2 2^{-2n-1}$, which constitutes a mild blows up with $k$. This mild blow-up is key in the obtained convergence rates, and is established by a careful stability analysis of the Chebyshev recurrence in Lemma \ref{lem:native-chebychev-stability}, which notably relies on the use of Chebyshev polynomials of the second kind.

The machine $\ChebSum$ evaluates one finite Chebyshev sum with dyadic coefficients. Given dyadic
coefficients $\mathbf a=(a_0,\ldots,a_d)$ and a precision parameter $n$, it
successively forms the Chebyshev values by repeated calls to
$\ChebStep$, uses $\Contr$ to multiply the current value by the next dyadic
coefficient, and adds the result to an accumulator. As in the monomial
construction, it first works with a rescaled coefficient vector
$\mathbf a^\ast$ satisfying $2\|\mathbf a^\ast\|_1<1$, so that the accumulator
stays bounded during the computation, and finally calls $\Scale$ to undo the
rescaling. The output is a function denoted $H_{\mathbf a,n}$, that satisfies
\[
    \left|
    H_{\mathbf a,n}(x)-\sum_{k=0}^d a_kT_k(x)
    \right|
    \leq
    (1 + 2\|f\|_{L^\infty([-1,1])})d^2 2^{-2n-1}.
\]

\begin{figure}[b]
\centering
\begin{minipage}{0.35\textwidth}
  \centering
        \resizebox{\linewidth}{!}{%
        \begin{tikzpicture}[>=Latex]
    \node[draw,rounded corners,fill=blue!8] (s) at (0,0) {\Scale};
    \node[draw,rounded corners,fill=blue!8] (si) at (1.5,0) {\Sign};
    \node[draw,rounded corners,fill=blue!8] (c+) at (3,0) {$\Contr^+$};
    \node[draw,rounded corners,fill=blue!8] (t) at (4.5,0) {\Times};
    \node[draw,rounded corners,fill=green!10] (cs) at (4.5,-1) {\ChebStep};
    \node[draw,rounded corners,fill=blue!30] (c) at (3,-1) {\Contr};
    \node[draw,rounded corners,fill=orange!12] (cS) at (3,-2) {\ChebSum};
    \node[draw,rounded corners,fill=red!12] (cC) at (3,-3) {\ChebContinuous};

    \draw[->,thick] (c+) -- (c);
    \draw[->,thick] (si) |- (c);
    \draw[->,thick] (c) -- (cS);
    \draw[->,thick] (t) -- (cs);
    \draw[->,thick] (cs) -- (cS);
    \draw[->,thick] (cS) -- (cC);
    \draw[->,thick] (s) |- (cS);
  \end{tikzpicture}%
  }
\end{minipage}\hfill
\begin{minipage}{0.6\textwidth}
  \centering
         \begin{tabular}{ll}
    \hline
    Machine & Purpose \\
    \hline

    \Scale & multiply by a power of 2 \\
    \Sign & flips the sign of the input \\
    $\Contr^+$ & apply a positive dyadic contraction $y \mapsto a y$\\
    \Contr & apply a dyadic contraction $y \mapsto ay$ \\
    \Times & approximate multiplication \\
    \ChebStep & computes one step of the Chebyshev recursion \\
    \ChebSum & computes a dyadic Chebychev partial sum \\
    \ChebContinuous & computes a Chebyshev expansion \\
    \hline
  \end{tabular}
\end{minipage}
\caption{Subroutine hierarchy for evaluating dyadic Chebyshev sums.}
\label{fig:chebyshev-tmnu-hierarchy}
\end{figure}
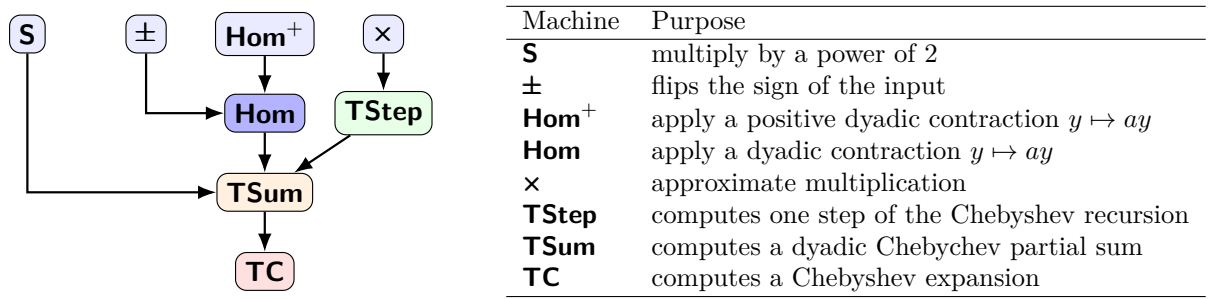

Finally, given a function $f : [-1,1] \to \R$, $\ChebContinuous$ repeatedly calls $\ChebSum$ on successive degrees $d_i$, precision parameters $n_i$, and dyadic coefficient vectors $\mathbf{ \tilde c}_i := (\tilde c_{i,0}, \ldots, \tilde c_{i,d_i}) \in \R^{d_i+1}$ that approximate the Chebyshev coefficient vectors $\mathbf c_i^{(f)} := (c_0^{(f)}, \ldots, c_{d_i}^{(f)})$ of $f$ to precision $p_i$. Therefore, during its $i$-th cycle, $\ChebContinuous$ computes $H_{\mathbf{\tilde c}_i,n_i}(x)$, which by the previous estimate gives an approximation of the Chebyshev truncation
\begin{align*}
    \left|\ChebContinuous^ux[t] - f(x)\right| &
\leq \left|H_{\mathbf{\tilde c}_i,n_i}(x) - \sum_{k=0}^{d_i} \tilde c_{i,k} T_k(x)\right| + 
    \left|\sum_{k=0}^{d_i} \tilde c_{i,k} T_k(x) - S^{(f)}_{d_i}(x)\right|
    +
    \left|S^{(f)}_{d_i}(x) - f(x)\right|\\
    & \leq (1 + 2\|f\|_{L^\infty([-1,1])})d_i^2 2^{-2n_i-1} + d_i 2^{-p_i} + S(f,d_i),
\end{align*}
while requiring a time of order $T_i \simeq d_i(q_i+n_i + \log (d_i))$ to read the coefficients and perform the calls to $\ChebStep$. We choose the parameters $d_i$, $p_i$, and $n_i$ to balance the three terms in the error estimate above, and to make $T_i$ of order $2^i$, so that the cumulative time up to cycle $i$ is also of order $2^i$. Specifically, given a function $\eta:[1,\infty)\to\N$, we set
\begin{equation}
    d_i:=\eta(2^i), \qquad p_i \simeq n_i \simeq 2^i/d_i + O( \log_2(d_i)).
\end{equation}
With these choices, we have $T_i = O(2^i)$ and the error estimate becomes
\[
    \left|\ChebContinuous^ux[t] - f(x)\right|
    \simeq 2^{-2^i/\eta(2^i)}+S(f,\eta(2^i)).
\]
Since the cumulative time up $N_i$ to cycle $i$ is of order $O( 2^i)$, this gives a convergence rate of order
\[
    2^{-\frac{t/\tau}{\eta(t/\tau)}}+S(f,\eta(t/\tau)),
\]
for some constant $\tau>0$, which is the intended tradeoff between the Chebyshev truncation error and the machine error, expressed in terms of machine time $t$.

\begin{theorem}\label{th:convergence-rate-chebychev-series}
    Let $f : [-1,1] \to \R$ be continuous. Let
    $\eta:[1,\infty)\to\N$ be nondecreasing, with $\eta(t)\to\infty$ and
    $t/\eta(t)$ nondecreasing. Assume that there exists $A>0$ such that
    \[
        \eta(t)\log_2(\eta(t)+1)\leq At,
        \qquad t\geq1.
    \]
    Define
    \[
        \gamma_f:=\left\lceil
        \log_2\left(2\|f\|_{L^\infty([-1,1])}+1\right)
        \right\rceil
    \]
    and
    \[
        \tau := \tau_{f,A}:=
        4\bigl(10+\gamma_f+A(80+6\gamma_f)\bigr).
    \]
    Then there exists $u \in \{0,1\}^\N$ such that $\ChebContinuous$ is
    uniformly $C_f$-bounded at $u$ with
    \[
        C_f:=3+ S(f,0) + \|f\|_{L^\infty([-1,1])},
    \]
     and for every $x\in[-1,1]$ and $t\geq\tau$,
    \[
        \left|\ChebContinuous^{u}x[t]-f(x)\right|
        \leq
        2\cdot
        2^{-\frac{t/\tau}{\eta(t/\tau)}}
        +
        S\left(f,\eta(t/\tau)\right).
    \]
\end{theorem}
\begin{proof}
    See Theorem \ref{thm:native-chebychev-tradeoff} in Appendix
    \ref{sec:tmnu-construction-appendix}.
\end{proof}

\appendix

\section{Notation and definitions}

\label{sec:notation-appendix}

\begin{enumerate}[label=(\alph*)]
    \item Given $A \subseteq \R$, we let $\pm A := A \cup \{-x : x \in A\}$. 
    \item Let $A \in \R^{n \times m}$. We say that $A$ is a \emph{left-selector matrix} if $A \in \{0,1\}^{n \times m}$ and every row of $A$ contains at most one entry equal to $1$. We say that $A$ is a \emph{right-selector matrix} if $A \in \{0,1\}^{n \times m}$ and every column of $A$ contains at most one entry equal to $1$. \label{item:selector-matrix}
    \item Let $A \in \R^{m \times n}$ and $b \in \R^m$. We define
    \begin{equation}
        \nnWeights{A} := \{ A_{jk} : j \in \{1, \ldots, m\}, k \in \{1, \ldots, n\} \}, \quad \nnWeights{b} := \{ b_j : j \in \{1, \ldots, m\} \}.
    \end{equation}
    Given a ReLU neural network $\nn := (N_0, N_1, \ldots, N_L; A_1, b_1, \ldots, A_L, b_L)$, we define
    \begin{equation}
        \nnWeights{\nn} := \bigcup_{\ell = 1}^L \nnWeights{A_\ell} \cup \nnWeights{b_\ell}.
    \end{equation}
    Given an affine map $W : \R^n \to \R^m$, note that $W$ is uniquely defined by $W(x) = Ax + b$ for all $x \in \R^n$, for some $A \in \R^{m \times n}$ and $b \in \R^m$. We then define 
    \begin{equation}
        \nnWeights{W} := \nnWeights{A} \cup \nnWeights{b}.
    \end{equation}
    Given a \TMNU $\tmnuM := (\nStates, \neurDim, \transFunc, \commFunc)$, we define
    \begin{equation}
        \nnWeights{\tmnuM} := \bigcup_{\state \in \states, \symb \in \workSymbols} \nnWeights{\commFunc(\state,\symb)}.
    \end{equation}
    \item We define the lexicographical ordering on $\R^m$ by 
    \begin{equation}\label{eq:lexicographical-ordering-R-m}
        x <_L y \iff \exists i \in \{1, \ldots, m\} \text{ such that } x_i <_L y_i \text{ and } x_j = y_j \text{ for all } j \in \{1, \ldots, i-1\}.
    \end{equation}
    We define the lexicographical ordering on $\R^{n \times m}$ by
    \begin{equation}\label{eq:lexicographical-ordering-R-n-m}
        A <_L B \iff \exists i \in \{1, \ldots, n\} \text{ such that } A_i <_L B_i \text{ and } A_j = B_j \text{ for all } j \in \{1, \ldots, i-1\},
    \end{equation}
    where $A_i$ and $B_i$ denote the $i$-th rows of $A$ and $B$, respectively. 
    We define the lexicographical ordering on $\setAffine{n}{m}$ by
    \begin{equation}\label{eq:lexicographical-ordering-affine-maps}
         f <_L g \iff A <_L A' \text{ or } A = A' \text{ and } b <_L b',
    \end{equation}
    where $f(x) = A x + b$ and $g(x) = A' x + b'$ for all $x \in \R^n$. Given $A \subseteq \{1,\ldots,n\}$, wew define $\chi_A := (\chi_A(1), \ldots, \chi_A(n)) \in \R^n$, where $\chi_A(i) := 1$ if $i \in A$ and $\chi_A(i) := 0$ otherwise. We define the lexicographical ordering on $\reluSet{n}$ by 
    \begin{equation}\label{eq:lexicographical-ordering-relu-maps}
         \ReLU_A <_L \ReLU_B \iff \chi_A <_L \chi_B.
    \end{equation}
    Finally, we define the lexicographical ordering on $\setAffine{n}{n} \cup \reluSet{n}$ by
    \begin{equation}\label{eq:lexicographical-ordering-affine-relu-maps}
         f <_L g \iff f,g \in \setAffine{n}{n} \text{ and } f <_L g, \text{ or } f,g \in \reluSet{n} \text{ and } f <_L g, \text{ or } f \in \setAffine{n}{n} \text{ and } g \in \reluSet{n}.
    \end{equation}Note that the lexicographical ordering defined in this way is a total order.
\end{enumerate}

\section{Construction of an RNN that simulates a TMNU}

\label{sec:tmnu-rnn-simulation-appendix}

This appendix is devoted to the detailed construction of an RNN that simulates a \TMNU. The construction will be mainly divided into three parts: first, we show that the update of a \TMNU can be implemented by a continuous piecewise-linear (CPWL) function (CPWL), then we will construct a ReLU neural network that implements this CPWL function, and finally we will show how to use this ReLU neural network to construct an RNN that simulates the \TMNU.

\subsection{Construction of a piecewise-linear function that implements the transition function of a \TMNU}
\label{subsection:tmnu-update-as-cpwl-function:appendix}

This section is devoted to showing that the transition function of a \TMNU can be implemented by a continuous piecewise-linear function. We divide the analysis in 3 parts: first, we show how the actions on the tape can be implemented by CPWL functions, then we show how the transition and command functions can be implemented by CPWL functions, and finally we show how to combine these CPWL functions to obtain a CPWL function that implements the transition function of the \TMNU.

\subsubsection{Reading, writing and shifting operations as continuous piecewise-linear functions over the encoding of the tape}

In this part, we show that the reading, writing and shifting operations on the tape can be implemented by continuous piecewise-linear functions over the encoding of the tape.
The first lemma isolates these three elementary tape operations at the level of the Cantor encoding.
\begin{lemma}\label{lem:reading-writing-shifting-operations}
    There exists continuous piecewise-linear functions $r : \R^2 \to \R$, $w_\symb : \R^2 \to \R^2$ for $\symb \in \{-1,0,1\}$ and $s_\move : \R^2 \to \R^2$ for $\move \in \{-1,0,1\}$ such that for every tape $\tape$, we have
    \begin{equation}\label{eq:reading-writing-shifting-operations}
        r(\cantorMap(\tape)) = \readOp\tape, \quad w_\symb(\cantorMap(\tape)) = \writeOp_\symb\tape, \quad s_\move(\cantorMap(\tape)) = \shiftOp_\move\tape.
    \end{equation}
\end{lemma}
\begin{proof}
    % Here is presented a sketch of the proof, but the exact proof can be found in APPENDIX. 
    First note that
    \[
        \tilde \cantorMap(x) \in u_1 + [-1/3,1/3], \quad u \in \workSymbols^\N.
    \]
    Therefore, depending on the value of $u_1$, the value of $\cantorMap(u)$ belongs to one of the three disjoint intervals $I_{-1} := [-4/3,-2/3]$, $I_0 := [-1/3,1/3]$ and $I_1 := [2/3,4/3]$. We can then design a piecewise-linear function that is equal to $-1$ on $I_{-1}$, $0$ on $I_0$ and $1$ on $I_1$ can be used to read the first element of $u$. We denote by $f$ such a function, i.e., $f : \R \to \R$ satisfies
    \begin{equation}\label{eq:piecewise-linear-function-read-first-symbol:lem:reading-writing-shifting-operations}
        f(\tilde \cantorMap(u)) = u_1, \quad u \in \workSymbols^\N.
    \end{equation}
    A graphical representation of $f$ is given on Figure \ref{fig:piecewise-linear-function-read-first-symbol}, which can be easily cast into the following piecewise-linear function defined by
    \begin{equation}\label{eq:piecewise-linear-function-read-first-symbol:explicit}
        f(x) := \begin{cases}
            -1 & \text{if } x \leq -2/3,\\
            3x + 1 & \text{if } -2/3 < x < -1/3,\\
            0 & \text{if } -1/3 \leq x \leq 1/3,\\
            3x - 1 & \text{if } 1/3 < x < 2/3,\\
            1 & \text{if } x \geq 2/3.
        \end{cases}
    \end{equation}
\begin{figure}[H]
    \centering
    \begin{tikzpicture}
    \begin{axis}[
    axis lines=middle,
    xmin=-2, xmax=2,
    ymin=-1.9, ymax=1.9,
    domain=-2:2,
    samples=2,
    grid=both,
    width=9cm, height=7cm,
    xlabel={$x$}, ylabel={$f(x)$},
    ]
    % Example piecewise-linear function via vertices (edit the points!)
    \addplot[thick] coordinates {
        (-2,-1)
        (-2/3, -1)
        ( -1/3,0)
        (1/3, 0)
        ( 2/3, 1)
        ( 2, 1)
    };

    % height where the interval arrow sits
        \def\yinterval{-1.6}

    % dashed guides to x-axis
        \draw[dashed,gray] (axis cs:-1/3,0) -- (axis cs:-1/3,\yinterval);
        \draw[dashed,gray] (axis cs:1/3,0)  -- (axis cs:1/3,\yinterval);

        % interval arrow
        \draw[<->,thick,red]
        (axis cs:-1/3,\yinterval) -- (axis cs:1/3,\yinterval)
        node[midway,above] {$I_0$};

        % dashed guides to x-axis
        \draw[dashed,gray] (axis cs:-4/3,0) -- (axis cs:-4/3,\yinterval);
        \draw[dashed,gray] (axis cs:-2/3,0)  -- (axis cs:-2/3,\yinterval);

        % interval arrow
        \draw[<->,thick,red]
        (axis cs:-4/3,\yinterval) -- (axis cs:-2/3,\yinterval)
        node[midway,above] {$I_{-1}$};

        % dashed guides to x-axis
        \draw[dashed,gray] (axis cs:2/3,0) -- (axis cs:2/3,\yinterval);
        \draw[dashed,gray] (axis cs:4/3,0)  -- (axis cs:4/3,\yinterval);

        % interval arrow
        \draw[<->,thick,red]
        (axis cs:2/3,\yinterval) -- (axis cs:4/3,\yinterval)
        node[midway,above] {$I_1$};
    \end{axis}
    \end{tikzpicture}
    \caption{A piecewise-linear function that is equal to $-1$ on $I_{-1}$, $0$ on $I_0$ and $1$ on $I_1$.}
    \label{fig:piecewise-linear-function-read-first-symbol}
\end{figure}
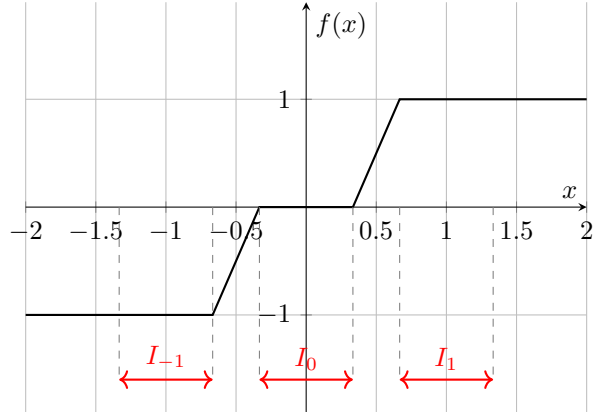

Now, we can use this function $f$ to realize the reading, writing and shifting operations. Namely, we define the reading operation $r : \R^2 \to \R$ by 
\begin{equation}\label{eq:reading-writing-shifting-operations:reading}
    r(x,y) := f(x), \quad x,y \in \R,
\end{equation}
the writing operations $w_\symb : \R^2 \to \R^2$ for $\symb \in \{-1,0,1\}$ by 
\begin{equation}\label{eq:reading-writing-shifting-operations:writing}
    w_\symb(x,y) := (x + \symb - f(x), y), \quad x,y \in \R,
\end{equation}
and the shifting operations $s_m : \R^2 \to \R^2$ for $m \in \{-1,0,1\}$ by 
\begin{equation}\label{eq:reading-writing-shifting-operations:shifting}
    s_m(x,y) := \begin{cases}
        (x,y) & \text{if } m = 0,\\
        (4(x - f(x)), f(x) + y/4) & \text{if } m = 1,\\
        (f(y) + x/4, 4(y - f(y))) & \text{if } m = -1,
    \end{cases} \quad x,y \in \R.
\end{equation}
First note that we have defined $r$, $w_\symb$ and $s_m$ as linear combinations of $x$, $y$ and $f(x)$, and hence these functions are continuous piecewise-linear. To see that these functions indeed implement the claimed operations, fix a tape $\tape \in \workSymbols^\Z$. Then, first have
\[
    r(\cantorMap(\tape)) \overset{\eqref{eq:encoding-tape}}= r(\tilde \cantorMap(\tapeR{\tape}), \tilde \cantorMap(\tapeL{\tape})) \overset{\eqref{eq:reading-writing-shifting-operations:reading}}= f(\tilde \cantorMap(\tapeR{\tape})) \overset{\eqref{eq:piecewise-linear-function-read-first-symbol:lem:reading-writing-shifting-operations}} = \tapeR{\tape}_1 \overset{\eqref{eq:right-left-tape}}= \tape_0 = \readOp\tape.
\]
Second, fix $\symb \in \{-1,0,1\}$, and define $\tape_\symb^+ \in \workSymbols^\N$ by $\tape_\symb^+(1) := \symb$ and $\tape_\symb^+(i) := \tapeR{\tape}(i)$ for every $i \geq 2$. Note that $(\writeOp_\symb \tape)^+ = \tape_\symb^+$ and $(\writeOp_\symb\tape)^- = \tapeL{\tape}$. Then, we have
\[
    w_\symb(\cantorMap(\tape)) \overset{\eqref{eq:encoding-tape}}= w_\symb(\tilde \cantorMap(\tapeR{\tape}), \tilde \cantorMap(\tapeL{\tape})) \overset{\eqref{eq:reading-writing-shifting-operations:writing}}= (\tilde \cantorMap(\tapeR{\tape}) + \symb - f(\tilde \cantorMap(\tapeR{\tape})), \tilde \cantorMap(\tapeL{\tape})) \overset{\eqref{eq:piecewise-linear-function-read-first-symbol:lem:reading-writing-shifting-operations}}= (\tilde \cantorMap(\tapeR{\tape}) + \symb - \tapeR{\tape}_1, \tilde \cantorMap(\tapeL{\tape})).
\]
 Note that $(\writeOp_\symb\tape)^+ = (\tape_\symb^+, \tapeL{\tape})$. Moreover, we have
\[
    \cantorMap(\tapeR{\tape}) + \symb - \tapeR{\tape}_1 \overset{\eqref{eq:encoding-sequence}}= 4 \sum_{i=1}^\infty \tapeR{\tape}_i 4^{-i} + \symb - \tapeR{\tape}_1 = 4 \cdot \symb \cdot 4^{-1} + 4 \sum_{i=2}^\infty \tapeR{\tape}_i 4^{-i} = \tilde \cantorMap(\tape_\symb^+).
\]
and hence 
\[
    w_\symb(\cantorMap(\tape)) = (\tilde \cantorMap(\tape_\symb^+), \tilde \cantorMap(\tapeL{\tape})) = (\tilde \cantorMap((\writeOp_\symb\tape)^+), \tilde \cantorMap((\writeOp_\symb\tape)^-)) \overset{\eqref{eq:encoding-tape}}= \cantorMap(\writeOp_\symb\tape).
\]
Finally, for $m =0$, we have $s_0(\cantorMap(\tape)) = \cantorMap(\tape) = \cantorMap(\shiftOp_0\tape)$. For $m = 1$, we have
\begin{align*}
    s_1(\cantorMap(\tape)) &\overset{\eqref{eq:reading-writing-shifting-operations:shifting}}= (4(\tilde \cantorMap(\tapeR{\tape}) - f(\cantorMap(\tapeR{\tape}))), f(\cantorMap(\tapeR{\tape})) + \tilde \cantorMap(\tapeL{\tape})/4) \overset{\eqref{eq:piecewise-linear-function-read-first-symbol:lem:reading-writing-shifting-operations}}= (4(\tilde \cantorMap(\tapeR{\tape}) - \tapeR{\tape}_1), \tapeR{\tape}_1 + \tilde \cantorMap(\tapeL{\tape})/4).
\end{align*}
Note that
\begin{align*}
    4(\cantorMap(\tapeR{\tape}) - \tapeR{\tape}_1) &\overset{\eqref{eq:encoding-sequence}}= 4 \sum_{i=2}^\infty \tapeR{\tape}_{i} 4^{-i+1} = 4 \sum_{i=2}^\infty \tape_{i-1} 4^{-i+1} = 4 \sum_{i=1}^\infty \tape_{i-2} 4^{-i} = 4 \sum_{i=1}^\infty \shiftOp(\tape,1)^+_i 4^{-i} = \tilde \cantorMap((\shiftOp_1\tape)^+).
\end{align*}
and, similarly, 
\begin{equation}
        \tapeR{\tape}_1 + \tilde \cantorMap(\tapeL{\tape})/4 = \tilde \cantorMap((\shiftOp_1\tape)^-).
\end{equation}
Hence, we have 
\[s_1(\cantorMap(\tape)) = (\tilde \cantorMap((\shiftOp_1\tape)^+), \tilde \cantorMap((\shiftOp_1\tape)^-)) =\cantorMap(\shiftOp_1\tape).\] 
We can analogously show that $s_{-1}(\cantorMap(\tape)) = \cantorMap(\shiftOp_{-1}\tape)$. This concludes the proof.
\end{proof}

In order to condense the incoming notation, we combine $\transFuncMove$ and $\transFuncSymb$ into a single function $\transFuncSymbMove : \states \times \workSymbols \to \{1,\ldots,9\}$ defined by
\begin{equation}\label{eq:combined-transition-function}
    \transFuncSymbMove(\state,\symb) := 3 \cdot \transFuncMove(\state,\symb)  + \transFuncSymb(\state,\symb) + 5, \quad \state \in \states, \symb \in \workSymbols.
\end{equation}
Note that \eqref{eq:combined-transition-function} implicitly associates a number $k \in \{1,\ldots,9\}$ to each pair $\symb_k, \move_k \in \{-1,0,1\}^2$, given by
\begin{equation}
    k = 3 \cdot \move_k + \symb_k + 5.
\end{equation}
Accordingly, for $k \in \{1,\ldots,9\}$, the map $U_k : \workSymbols^\Z \to \workSymbols^\Z$ defined by
\begin{equation}\label{eq:combined-transition-function-operations}
    U_{3m + \symb + 5} \tape := \shiftOp_{\move} \writeOp_{\symb} \tape, \ \quad \tape \in \workSymbols^\Z, m \in \{-1,0,1\}, \symb \in \{-1,0,1\}.
\end{equation}
We then define by $u_k : \R^2 \to \R^2$ the continuous piecewise-linear function given by
\begin{equation}
    u_k(x,y) := s_{\move_k}(w_{\symb_k}(x,y)), \quad x,y \in \R,
\end{equation}
and we have by Lemma \ref{lem:reading-writing-shifting-operations} that
\begin{equation}\label{eq:combined-transition-function-operations-piecewise-linear}
    u_k(\cantorMap(\tape)) = \cantorMap(U_k \tape) \quad \tape \in \workSymbols^\Z, k \in \{1,\ldots,9\}.
\end{equation}
We also let $u : \R^2 \to \R^{18}$ be the continuous piecewise-linear function given by
\begin{equation}\label{eq:combined-transition-function-piecewise-linear}
    u(x,y) := (u_1(x,y), u_2(x,y), \ldots, u_9(x,y)), \quad x,y \in \R.
\end{equation}
We now move to show that we can also implement the transition function $\transFunc$ and command function $\commFunc$ of $\tmnuM$ by continuous piecewise-linear functions.

\subsubsection{Transition function as a continuous piecewise-linear function}

In this section, we show that the transition function $\transFunc$ and command function $\commFunc$ of $\tmnuM$ can be implemented by continuous piecewise-linear functions. Before proceeding, note that we can reformulate the definition of the command function $\commFunc$ as follows.
We begin by fixing a convenient enumeration of the neural commands used by the machine.
\begin{definition}\label{def:functions-tmnu}
    Let $\tmnuM := (\nStates, \neurDim, \transFunc, \commFunc)$ be a \TMNU. We define $f_1, \ldots, f_{\nFunctions{\tmnuM}} : \R^\neurDim \to \R^\neurDim$ be an enumeration $\Fc_\tmnuM$ in increasing order according to the ordering $<_L$ over $\setAffine{d}{d} \cup \reluSet{d}$ defined in \eqref{eq:lexicographical-ordering-affine-relu-maps}, and let
\[
    f_\tmnuM := (f_1, f_2, \ldots, f_{\nFunctions{\tmnuM}}) : \R^d \to \R^{\nFunctions{\tmnuM}d}.
\]
Further, we define the function $\tilde \commFunc : \states \times \workSymbols \to \{1, \ldots, \nFunctions{\tmnuM}\}$ by 
\[
    \tilde \commFunc(\state,\symb) := i \quad \iff \commFunc(\state,\symb) = f_i, \quad \state \in \states, \symb \in \workSymbols.
\]
\end{definition}

Under the above reformulation of the command function, we can proceed to simulate the transition and command functions by continuous piecewise-linear functions.
The next lemma shows that the discrete transition choices can be recovered by CPWL selectors.
\begin{lemma}\label{lem:transition-functions-as-cpwl-functions}
    Let $\tmnuM := (\nStates, \neurDim, \transFunc, \commFunc)$ be a \TMNU, and $\gamma \in \{\state, \tape, \neurState\}$. Then, there exists continuous piecewise-linear functions $\Delta_\gamma : \R^{\nStates + 2} \to \R^{n_\gamma}$, where $n_\state := \nStates$, $n_\tape := 9$ and $n_\neurState := \nFunctions{\tmnuM}$, such that for every $\state \in \{1,\ldots, \nStates\}$ and $\tape \in \workSymbols^\Z$, we have
    \begin{align}\label{eq:transition-functions-as-cpwl-functions}
        \Delta_\gamma(\oneHot{\nStates}{\state}, \cantorMap(\tape)) &= \oneHot{n_\gamma}{h^\gamma(\state,\readOp\tape)},
    \end{align}
    where $h^\state := \transFuncState$, $h^\tape := \transFuncSymbMove$ as defined by \eqref{eq:combined-transition-function} and $h^\neurState := \tilde \commFunc$ as defined in Definition \ref{def:functions-tmnu}.
\end{lemma}
\begin{proof}
    We make the proof for $\gamma = \state$ only, as the remaining cases are exactly analogous. For every $\state \in \{1,\ldots, \nStates\}$ and $\symb \in \workSymbols$, we let 
    \begin{equation}
        x_{\state,\symb} := (\oneHot{\nStates}{\state}, \symb) \in \R^{\nStates + 1}, \quad y_{\state,\symb} := \oneHot{\nStates}{\transFuncState(\state,\symb)} \in \R^{\nStates}.
    \end{equation}
    Note that the set defined by
    \begin{equation}
        \left\{ (x_{\state,\symb}, y_{\state,\symb}) : \state \in \{1,\ldots, \nStates\}, \symb \in \workSymbols\right\} \subseteq \R^{\nStates + 1} \times \R^{\nStates}
    \end{equation}
    is finite (actually has at most $3n$ elements), so by linear interpolation, there exists a continuous piecewise-linear function $f_\state : \R^{\nStates + 1} \to \R^{\nStates}$ such that $f_\state(x_{\state,\symb}) = y_{\state,\symb}$ for every $\state \in \{1,\ldots, \nStates\}$ and $\symb \in \workSymbols$. Now, we can define $\Delta_\state : \R^{\nStates + 2} \to \R^{\nStates}$ by $\Delta_\state(x,y) := f_\state(x, r(y))$, where $r : \R^2 \to \R$ is the continuous piecewise-linear function defined in Lemma \ref{lem:reading-writing-shifting-operations}. Note that $\Delta_\state$ is continuous piecewise-linear as a composition of continuous piecewise-linear functions. Moreover, for every $\state \in \{1,\ldots, \nStates\}$ and $\tape \in \workSymbols^\Z$, we have
    \begin{align*}
        \Delta_\state(\oneHot{\nStates}{\state}, \cantorMap(\tape)) &= f_\state(\oneHot{\nStates}{\state}, r(\cantorMap(\tape))) \overset{\eqref{eq:reading-writing-shifting-operations:reading}}= f_\state(\oneHot{\nStates}{\state}, \readOp\tape) = f_\state(x_{\state, \readOp\tape}) = y_{\state, \readOp \tape} = \oneHot{\nStates}{\transFuncState(\state,\readOp(\tape))}.
    \end{align*}
    This concludes the proof.
\end{proof}

We next combine the results of Lemmata \ref{lem:reading-writing-shifting-operations} and \ref{lem:transition-functions-as-cpwl-functions} to design a continuous piecewise-linear function that implements $\tmnuM$.

\subsubsection{\texorpdfstring{Definition of a continuous piecewise-linear function that implements $\tmnuM$}{Definition of a continuous piecewise-linear function that implements a TMNU}}

Before proceeding, we need the following Lemma, which shows that we can use continuous piecewise-linear functions to select some coordinates of a vector.
This selection device will be used to choose the update prescribed by the current state and scanned symbol.
\begin{lemma}\label{lem:selection-function}
    Let $n,d \in \N$. Then, there exists a continuous piecewise-linear function $g_{n,d} : \R^n \times \R^{nd} \to \R^d$ such that for every $M > 0$, $\ell \in \{1,\ldots, n\}$ and $z_1, \ldots, z_n \in [0,M]^d$, we have
    \begin{equation}
        g_{n,d}(M(\oneHot{n}{\ell} - 1_n), (z_1, \ldots, z_n)) = z_\ell.
    \end{equation}
\end{lemma}
\begin{proof}
    % By Lemma
    % We first establish the following simple claim, that is inspired from \cite[Lemma 4.1]{siegelmann1992computational}.
    % \begin{claim*}
    %     Let $M > 0$, $z \in [0,M]$ and $b \in \{0,1\}$. Then, we have $\ReLU(z + M(b-1)) = b z$.
    % \end{claim*}
    % \begin{proof}[Proof of the claim]
    %     If $b = 0$, then we have $\ReLU(z + M(b-1)) = \ReLU(z - M) = 0$. If $b = 1$, then we have $\ReLU(z + M(b-1)) = \ReLU(z) = z$. This concludes the proof.
    % \end{proof}
    Let $n,d \in \N$. We first define $g_{n,d} : \R^n \times \R^{nd} \to \R^d$ by \[g_{n,d}(x,z_1, \ldots, z_n) := \sum_{i=1}^n \ReLU(z_i + (x_i - 1) 1_d)\] for every $x \in \R^n$, $z_1, \ldots, z_n \in \R^d$. Note that $g_{n,d}$ is continuous piecewise-linear as a sum of continuous piecewise-linear functions. Moreover, for every $M>0$, $\ell \in \{1,\ldots, n\}$ and $z_1, \ldots, z_n \in [0,M]^d$, we have
    \begin{align}\label{eq:selection-function:proof:positive}
        g_{n,d}(M(\oneHot{n}{\ell}-1_n), z_1, \ldots, z_n)_i & = \sum_{i=1}^n \ReLU(z_i + M (\oneHot{n}{\ell}(i) - 1) 1_d) \overset{(a)}= \sum_{i=1}^n z_i \oneHot{n}{\ell}(i)
         \overset{\eqref{eq:one-hot-encoding-state}} \sum_{i=1}^n z_i \kroen{\ell}{i}
         = z_\ell
    \end{align}
    where (a) follows from Lemma \ref{lem:linear-combination-of-functions-with-relu}.
    % We now define $g_{M,n,d} : \R^n \times \R^{nd} \to \R^d$ by 
    % \[
    %     g_{M,n,d}(x,z_1, \ldots, z_n) := g^+_{M,n,d}(x,\ReLU(z_1), \ldots, \ReLU(z_n)) - g^+_{M,n,d}(x,\ReLU(-z_1), \ldots, \ReLU(-z_n)).
    % \]
    % Note that $g_{M,n,d}$ is continuous piecewise-linear as a combination of continuous piecewise-linear functions. Moreover, for every $\ell \in \{1,\ldots, n\}$ and $z_1, \ldots, z_n \in [-M,M]^d$, we have
    % \begin{align*}
    %     g_{M,n,d}(\oneHot{n}{\ell}, z_1, \ldots, z_n) & = g^+_{M,n,d}(\oneHot{n}{\ell},\ReLU(z_1), \ldots, \ReLU(z_n)) - g^+_{M,n,d}(\oneHot{n}{\ell},\ReLU(-z_1), \ldots, \ReLU(-z_n))\\
    %     &\overset{\eqref{eq:selection-function:proof:positive}}= \ReLU(z_\ell) - \ReLU(-z_\ell) = z_\ell.
    % \end{align*}
    This concludes the proof.
\end{proof}

We can now exploit the results of Lemmata \ref{lem:reading-writing-shifting-operations}, \ref{lem:transition-functions-as-cpwl-functions} and \ref{lem:selection-function} to design a continuous piecewise-linear function that implements $\tmnuM$.
The theorem below assembles the three ingredients into a single CPWL transition map.

\begin{theorem}\label{thm:simulation-of-tmnu-by-cpwl-function}
    Let $\tmnuM := (\nStates, \neurDim, \transFunc, \commFunc)$ be a \TMNU, and $C > 0$. Then, there exists a continuous piecewise-linear function $F := F_{\tmnuM,C} : \R^{\nStates + 2 + \neurDim} \to \R^{\nStates + 2 + \neurDim}$ such that
    \begin{equation}
        F(\configMap{\tmnuM}(c)) = \configMap{\tmnuM}(\tmnuM(c)), \quad c \in \Bc_\tmnuM(C).
    \end{equation}
\end{theorem}
\begin{proof}
    We will define three piecewise-linear functions $F_\state : \R^{\nStates + 2 + \neurDim} \to \R^{\nStates}$, $F_\tape : \R^{\nStates + 2 + \neurDim} \to \R^2$ and $F_\neurState : \R^{\nStates + 2 + \neurDim} \to \R^\neurDim$ such that for every configuration $c := (\state; \tape; \neurState)$ of $\tmnuM$, we have
    \begin{align}
        F_\state(\configMap{\tmnuM}(c)) &= \oneHot{\nStates}{\proj_\state \tmnuM(c)},\quad
        F_\tape(\configMap{\tmnuM}(c)) = \cantorMap(\proj_\tape\tmnuM(c)), \quad
        F_{\neurState,C}(\configMap{\tmnuM}(c)) = \proj_\neurState\tmnuM(c).
    \end{align}
    We then define $F : \R^{\nStates + 2 + \neurDim} \to \R^{\nStates + 2 + \neurDim}$ by $F(x) := (F_\state(x), F_\tape(x), F_{\neurState,C}(x))$ for every $x \in \R^{\nStates + 2 + \neurDim}$. Note that $F$ is continuous piecewise-linear as a combination of continuous piecewise-linear functions. Moreover, for every configuration $c$ of $\tmnuM$, we have
    \begin{align*}
        F(\configMap{\tmnuM}(c)) & = (F_\state(\configMap{\tmnuM}(c)), F_\tape(\configMap{\tmnuM}(c)), F_{\neurState,C}(\configMap{\tmnuM}(c))) = (\oneHot{\nStates}{\proj_\state \tmnuM(c)}, \cantorMap(\proj_\tape\tmnuM(c)),\proj_\neurState \tmnuM(c)) \\
        &= \configMap{\tmnuM}(\proj_\state \tmnuM(c), \proj_\tape \tmnuM(c), \proj_\neurState \tmnuM(c)) = \configMap{\tmnuM}(\tmnuM(c)).
    \end{align*}
    We now move to the definition of the functions $F_\state$, $F_\tape$ and $F_\neurState$. 
    \begin{enumerate}[label=(\alph*)]
        \item We let $F_\state(x,y,z) := \Delta_\state(x,y)$ for every $x \in \R^{\nStates}$, $y \in \R^2$ and $z \in \R^\neurDim$, where $\Delta_\state$ is as in Lemma \ref{lem:transition-functions-as-cpwl-functions}. Then,
        \begin{equation}
            F_\state(\configMap{\tmnuM}(c)) = \Delta_\state(\oneHot{\nStates}{\state}, \cantorMap(\tape)) \overset{\eqref{eq:transition-functions-as-cpwl-functions}}= \oneHot{\nStates}{\transFuncState(\state,\readOp\tape)} \overset{(a)}= \oneHot{\nStates}{\proj_\state\tmnuM(c)},
        \end{equation}
        where (a) is by Definition \ref{def:configuration-of-turing-machine-and-update-of-configuration}.
        \item We here proceed to the definition of $F_\tape$. We let
        \begin{equation}
            F_\tape(x,y,z) := g_{9,2}(\Delta_\tape(x,y), u(y)), \quad x \in \R^{\nStates}, y \in \R^2, z \in \R^\neurDim,
        \end{equation}
        where $\Delta_\tape$ is as in Lemma \ref{lem:transition-functions-as-cpwl-functions}, $u$ is defined by \eqref{eq:combined-transition-function-piecewise-linear} and $g_{9,2}$ is as in Lemma \ref{lem:selection-function}. Note that $F_\tape$ is continuous piecewise-linear as a combination of continuous piecewise-linear functions. Moreover, for every configuration $c := (\state; \tape; \neurState)$ of $\tmnuM$, we have
        \begin{align*}
            F_\tape(\configMap{\tmnuM}(c)) & = g_{9,2}(\Delta_\tape(\oneHot{\nStates}{\state}, \cantorMap(\tape)), u(\cantorMap(\tape)))\overset{(a)}= g_{9,2}(\oneHot{9}{\transFuncSymbMove(\state,\readOp\tape)}, u_1(\cantorMap(\tape)), \ldots, u_9(\cantorMap(\tape)))\\
            &\overset{(b)}=  u_{\transFuncSymbMove(\state,\readOp\tape)}(\cantorMap(\tape)) \overset{\eqref{eq:combined-transition-function-operations-piecewise-linear}}= \cantorMap(U_{\transFuncSymbMove(\state,\readOp\tape)}\tape) \overset{\eqref{eq:combined-transition-function}}= \cantorMap(U_{3\transFuncMove(\state,\readOp\tape) + \transFuncSymb(\state,\readOp\tape) + 5}\tape)\\
            &\overset{\eqref{eq:combined-transition-function-operations}}= \cantorMap(\shiftOp_{\transFuncMove(\state,\readOp\tape)} \writeOp_{\transFuncSymb(\state,\readOp\tape)}\tape) \overset{(c)}= \cantorMap(\proj_\tape\tmnuM(c)),
        \end{align*}
        where (a) follows from Lemma \eqref{eq:transition-functions-as-cpwl-functions}, (b) is by Lemma \ref{lem:selection-function} and (c) is by Definition \ref{def:configuration-of-turing-machine-and-update-of-configuration}.

        \item We here proceed to the definition of $F_{\neurState,C}$. We let $f_\tmnuM = (f_1, \ldots, f_{\nFunctions{\tmnuM}})$ and $\tilde \commFunc$ be as in Definition \ref{def:functions-tmnu}, and we define
        \begin{align}
            F_{\neurState,C}(x,y,z) := & \ g_{\nFunctions{\tmnuM},\neurDim}(\|\tmnuM\|_C(\Delta_\commFunc(x,y) - 1_{\nFunctions{\tmnuM}}), \ReLU(f_\tmnuM(z)))\nonumber\\
            & \,  - g_{\nFunctions{\tmnuM},\neurDim}(\|\tmnuM\|_C(\Delta_\commFunc(x,y) - 1_{\nFunctions{\tmnuM}}), \ReLU(-f_\tmnuM(z))),\label{eq:definition-of-F-neural-state}
        \end{align}
        where $g_{\nFunctions{\tmnuM},\neurDim}$ is as in Lemma \ref{lem:selection-function}, and $\Delta_\commFunc$ is as in Lemma \ref{lem:transition-functions-as-cpwl-functions}. Note that $F_{\neurState,C}$ is continuous piecewise-linear as a combination of continuous piecewise-linear functions. Now note that, for every configuration $c \in \Bc_\tmnuM(C)$, we have
        \begin{equation}\label{eq:definition-of-F-neural-state:boundedness}
            \|\neurState\| \leq C \quad \Rightarrow \quad \|f_\tmnuM(\neurState)\|_\infty \leq \|\tmnuM\|_C \quad \Rightarrow \quad \ReLU(f_\tmnuM(\neurState)), \ReLU(-f_\tmnuM(\neurState)) \in [0,\|\tmnuM\|_C]^{\neurDim \nFunctions{\tmnuM}}.
        \end{equation}
        Therefore, for $c := (\state, \tape, \neurState) \in \Bc_\tmnuM(C)$, we have
        \begin{align*}
            g_{\nFunctions{\tmnuM},\neurDim}(\|\tmnuM\|_C(\Delta_\neurState(\oneHot{\nStates}{\state},\cantorMap(\tape)) - 1_{\nFunctions{\tmnuM}}), \ReLU(f_\tmnuM(\neurState))) & \overset{(a)}= g_{\nFunctions{\tmnuM},\neurDim}(\|\tmnuM\|_C(\oneHot{\nFunctions{\tmnuM}}{\tilde \commFunc(\state,\readOp\tape)} - 1_{\nFunctions{\tmnuM}}), \ReLU(f_\tmnuM(\neurState)))\\
            & \overset{(b)}= \ReLU(f_{\tilde \commFunc(\state,\readOp\tape)}(\neurState)) \overset{(c)}= \ReLU(\commFunc(\state,\readOp\tape)(\neurState)),
        \end{align*}
        where (a) follows from Lemma \ref{lem:transition-functions-as-cpwl-functions}, (b) is by Lemma \ref{lem:selection-function} and \eqref{eq:definition-of-F-neural-state:boundedness}, and (c) is by Definition \ref{def:functions-tmnu}, and similarly, we have
        \begin{align*}
            g_{\nFunctions{\tmnuM},\neurDim}(\|\tmnuM\|_C(\Delta_\neurState(\oneHot{\nStates}{\state},\cantorMap(\tape)) - 1_{\nFunctions{\tmnuM}}), \ReLU(-f_\tmnuM(\neurState))) = \ReLU(-\commFunc(\state,\readOp\tape)(\neurState)),
        \end{align*}
        so that
        \begin{align*}
            F_{\neurState,C}(\configMap{\tmnuM}(c)) & = \ReLU(\commFunc(\state,\readOp\tape)(\neurState)) - \ReLU(-\commFunc(\state,\readOp\tape)(\neurState)) \overset{(a)}= \commFunc(\state,\readOp\tape)(\neurState) = \proj_\neurState\tmnuM(c),
        \end{align*}
        where (a) is by $\ReLU(x) - \ReLU(-x) = x$ for every $x \in \R$. This concludes the proof.
    \end{enumerate} 
\end{proof}

\subsection{Construction of a neural network that simulates a \TMNU}

\label{subsec:construction-of-a-neural-network-that-simulates-a-tmnu:appendix}

In Section \ref{subsection:tmnu-update-as-cpwl-function:appendix}, we have established that given a \TMNU $\tmnuM$ and $C >0$, there exists a CPWL function $F_{\tmnuM,C}$ such that 
\[
        F_{\tmnuM,C}(\configMap{\tmnuM}(c)) = \configMap{\tmnuM}(\tmnuM(c)), \quad \config \in \Bc_\tmnuM(C).
\]
Here, we construct explicitly a ReLU neural network that realizes this function $F_{\tmnuM,C}$. We first introduce useful notation for ReLU neural networks. Let $\nn := (N_0, N_1, \ldots, N_L; A^1, b^1, A^2, b^2, \ldots, A^L, b^L)$ be a ReLU neural network as in \eqref{eq:definition-of-neural-network}.
\begin{enumerate}[label=(\alph*)]
    \item For $\ell \in \{1, \ldots, L\}$, we let $\nn_\ell : \R^{N_0} \to \R^{N_\ell}$ be the function defined by
    \begin{equation}\label{eq:function-represented-by-ancestor-subnetwork}
        \nn_\ell(\realx) := A^\ell \ReLU\left( A^{\ell-1} \ReLU\left( \ldots \ReLU\left( A^1 \realx + b^1 \right) + b^{\ell-1} \right) + b^\ell \right) \text{ for all } \realx \in \R^{N_0}.
    \end{equation}
    We call $\nn_\ell$ the $\ell$-th ancestor subnetwork of $\nn$. By convention, we let $\nn_0(x) = x$ for every $x \in \R^{N_0}$.
    \item We let 
    \begin{equation}\label{eq:input-output-dim-neural-network}
        \nnInDim{\nn} := N_0, \quad \nnOutDim{\nn} := N_L.
    \end{equation}
    \item We say that $\nn$ is a \emph{left-selector} if $A^1$ is a left-selector matrix and $b^1 = 0$, and we say that $\nn$ is a \emph{right-selector} if $A^L$ is a right-selector matrix and $b^L = 0$.
\end{enumerate}

This section is organized as follows. First, we establish some useful properties of ReLU neural networks. Then we reproduce the construction scheme from Section \ref{subsection:tmnu-update-as-cpwl-function:appendix} by designing explicit networks for the tape operations, transition and command functions, and neural-state update.

\subsubsection{Auxiliary results on ReLU neural networks}

This part is devoted to the proof of some auxiliary results on ReLU neural networks that will be useful for the construction of a neural network that simulates a \TMNU.

\newcommand\nnId{\bfI}
We first need an identity network whose depth can be prescribed.
\begin{lemma}\label{lemma:identity-neural-network}
    Let $n,d\in\N$. Then, there exists a left- and right-selector neural network $\nnId_{n,d}$ with $\nnWidth{\nnId_{n,d}} = n$, $\nnDepth{\nnId_n} = d$ and $\nnWeights{\nnId_n} \subseteq \{0,1\}$ such that $\nnId_n(x) = \ReLU(x)$ for all $x \in \R^n$.
\end{lemma}
\begin{proof}
    For every $x \in \R^n$, we have $\nnId_{n,1}(x) = I_n x = x$. Now, for $d \in \N$, we define 
    \[\nnId_{n,d} := (n, \ldots, n; I_n, 0, \ldots, I_n, 0),\]
    where $I_n$ is the $n \times n$ identity matrix and there are $d$ layers. 
    Note that indeed, $\nnDepth{\nnId_{n,d}} = d$, $\nnWidth{\nnId_{n,d}} = n$ and $\nnWeights{\nnId_{n,d}} = \{0,1\}$. Now, let $x \in \R^n$, and note that by definition of $\nnId_n$, we have
    \[
        \nnId_{n,d}(x) = \underbrace{\ReLU \circ \ldots \circ \ReLU}_{d - 1 \ \text{times}}(x)= \ReLU(x).
    \]
    Note also that $I_n$ is a left-selector matrix and a right-selector matrix, and $0$ is the zero vector, so $\nnId_{n,d}$ is indeed a left- and right-selector neural network. This concludes the proof.
\end{proof}

The next lemma asserts that neural networks are closed under composition.
\begin{lemma}\label{lemma:composition-of-neural-networks}
    Let $\nn_1,\nn_2$ be two neural networks such that $\nnOutDim{\nn_1} = \nnInDim{\nn_2}$ and either $\nn_1$ is a right-selector or $\nn_2$ is a left-selector. Then, there exists a neural network $\nn$ such that $\nnDepth{\nn} = \nnDepth{\nn_1} + \nnDepth{\nn_2} - 1$, $\nnWidth{\nn} \leq \max\{\nnWidth{\nn_1}, \nnWidth{\nn_2}\}$, $\nnWeights{\nn} \subseteq \nnWeights{\nn_1} \cup \nnWeights{\nn_2}$ and $\nn(x) = \nn_2(\nn_1(x))$ for all $x \in \R^{\nnInDim{\nn_1}}$.
\end{lemma}
\begin{proof}
    Suppose that \[\nn_1 = (N_0^{(1)}, N_1^{(1)}, \ldots, N_{L_1}^{(1)}; A^1_{(1)}, b^1_{(1)}, A^2_{(1)}, b^2_{(1)}, \ldots, A^{L_1}_{(1)}, b^{L_1}_{(1)})\] and \[\nn_2 = (N_0^{(2)}, N_1^{(2)}, \ldots, N_{L_2}^{(2)}; A^1_{(2)}, b^1_{(2)}, A^2_{(2)}, b^2_{(2)}, \ldots, A^{L_2}_{(2)}, b^{L_2}_{(2)}).\] Note that by assumption, we have $N_{L_1}^{(1)} = N_0^{(2)}$. We define 
    \begin{align*}
        \nn := & (N^{(1)}_0, N^{(1)}_1, \ldots, N^{(1)}_{L_1-1}, N^{(2)}_{1}, \ldots, N^{(2)}_{L_2};\\
        & A^1_{(1)}, b^1_{(1)}, \ldots, A^{L_1-1}_{(1)}, b^{L_1-1}_{(1)}, A^1_{(2)} A^{L_1}_{(1)}, A^1_{(2)} b^{L_1}_{(1)} + b^1_{(2)}, A^2_{(2)}, b^2_{(2)}, \ldots, A^{L_2}_{(2)}, b^{L_2}_{(2)}).
    \end{align*}
    Note that indeed, $\nnDepth{\nn} = \nnDepth{\nn_1} + \nnDepth{\nn_2} - 1$, $\nnWidth{\nn} \leq \max\{\nnWidth{\nn_1}, \nnWidth{\nn_2}\}$. Moreover, if $\nn_1$ is a right-selector, then $A^{L_1}_{(1)}$ is a right-selector matrix and $b^{L_1}_{(1)} = 0$, so $\nnWeights{A^1_{(2)} A^{L_1}_{(1)}} \subseteq \nnWeights{A^1_{(2)}}$ and $\nnWeights{A^1_{(2)} b^{L_1}_{(1)} + b^1_{(2)}} \subseteq \nnWeights{b^1_{(2)}}$. If $\nn_2$ is a left-selector, then $A^1_{(2)}$ is a left-selector matrix and $b^1_{(2)} = 0$, so $\nnWeights{A^1_{(2)} A^{L_1}_{(1)}} \subseteq \nnWeights{A^{L_1}_{(1)}}$ and $\nnWeights{A^1_{(2)} b^{L_1}_{(1)} + b^1_{(2)}} \subseteq \nnWeights{b^{L_1}_{(1)}}$. In either case, we have $\nnWeights{\nn} \subseteq \nnWeights{\nn_1} \cup \nnWeights{\nn_2}$. Now, let $x \in \R^{\nnInDim{\nn_1}}$, and note that by definition of $\nn$, we have
    \[
        \nnAncestor{\nn}{L_1}(x) = A_{(2)}^1 \nn_1 (x) + b^1_{(2)},
    \]
    and therefore
    \[
        \nn(x) = \nn_2(\nn_1(x)).
    \] This concludes the proof.
\end{proof}

We shall also use a parallelization device to run several networks on selected input coordinates.
\begin{lemma}\label{lemma:parallelization-of-neural-networks}
    Let $m,d \in \N$, $\nn_1,\nn_2, \ldots, \nn_m$ be $m$ neural networks such that $\nnDepth{\nn_1} = \ldots = \nnDepth{\nn_m}$ and let $\iota_i : \{1, \ldots, \nnInDim{\nn_i}\} \to \{1,\ldots,d\}$, $i \in \{1,\ldots,m\}$. Then, there exists a neural network $\nn$ such that $\nnDepth{\nn} = \nnDepth{\nn_1}$, $\nnWidth{\nn} = d \vee \sum_{i=1}^m \nnWidth{\nn_i}$, $\nnWeights{\nn} \subseteq \bigcup_{i=1}^m \nnWeights{\nn_i} \cup \{0\}$ and \[\nn(x) = (\nn_1(\proj_{\iota_1}x),\nn_2(\proj_{\iota_2}x), \ldots, \nn_m(\proj_{\iota_m}x))\] for all $x \in \R^d$.
\end{lemma}
\begin{proof}
    Suppose that \[\nn_i = (N_0^{(i)}, N_1^{(i)}, \ldots, N_{L}^{(i)}; A^1_{(i)}, b^1_{(i)}, A^2_{(i)}, b^2_{(i)}, \ldots, A^{L}_{(i)}, b^{L}_{(i)}),\] for every $i \in \{1,\ldots, m\}$.
    We define \[\nn := (d, N_1, \ldots, N_L; A^1, b^1, \ldots, A^L, b^L),\] where $N_\ell = \sum_{i=1}^m N_\ell^{(i)}$ for $\ell \in \{1, \ldots, L\}$ and $A^\ell = \begin{pmatrix}
    A^\ell_{(1)} & 0 & \ldots & 0\\
    0 & A^\ell_{(2)} & \ldots & 0\\
    \vdots & \vdots & \ddots & \vdots\\
    0 & 0 & \ldots & A^\ell_{(m)}
\end{pmatrix}$ and $b^\ell = \begin{pmatrix}
    b^\ell_{(1)}\\
    b^\ell_{(2)}\\
    \vdots\\
    b^\ell_{(m)}
\end{pmatrix}$ for $\ell \in \{2, \ldots, L\}$, and 
\[
    A^1 = \begin{pmatrix}
        A^1_{(1)} \proj_{\iota_1}\\
        A^1_{(2)} \proj_{\iota_2}\\
        \vdots\\
        A^1_{(m)} \proj_{\iota_m}
    \end{pmatrix}, \quad b^1 = \begin{pmatrix}
        b^1_{(1)}\\
        b^1_{(2)}\\
        \vdots\\
        b^1_{(m)}
    \end{pmatrix}.
\]
    Note that indeed, $\nnDepth{\nn} = L$, $\nnWidth{\nn} = d \vee \sum_{i=1}^m \nnWidth{\nn_i}$ and $\nnWeights{\nn} \subseteq \bigcup_{i=1}^m \nnWeights{\nn_i} \cup \{0\}$. Now, let $x \in \R^d$, and note that by definition of $\nn$, we have
    \[       \nn(x) = (\nn_1(\proj_{\iota_1}x),\nn_2(\proj_{\iota_2}x), \ldots, \nn_m(\proj_{\iota_m}x)).
    \] This concludes the proof.
\end{proof}

\subsubsection{Neural networks for writing and shifting operations}

This part is devoted to the construction of neural networks that realize the reading, writing and shifting operations of a \TMNU. To this end, we construct a neural network that realizes the CPWL function that implements the reading operation.

The following lemma gives an explicit ReLU realization of the one-symbol reader.
\begin{lemma}\label{lemma:first-element-reading}
    Let $f : \R \to \R$ be the function defined by \eqref{eq:piecewise-linear-function-read-first-symbol:explicit}. Then, there exists a neural network $\firstNN$ with $\nnWidth{\firstNN} = 4$, $\nnDepth{\firstNN} = 2$ and $\nnWeights{\firstNN} = \{-2,-1,0,1,2,3\}$,
     such that $\firstNN(x) = f(x)$ for all $x \in \R$.
\end{lemma}
\begin{proof}
    We define $\firstNN : \R \to \R$ by $\firstNN := (1, 4, 1; \firstNNmat 1, \firstNNvec 1, \firstNNmat 2, \firstNNvec 2)$, where
    \[
        \firstNNmat 1 := \begin{pmatrix}
            3\\
            3\\
            3\\
            3
        \end{pmatrix}, \quad \firstNNvec 1 := \begin{pmatrix}
            2\\
            1\\
            -1\\
            -2
        \end{pmatrix}, \quad \firstNNmat 2 := \begin{pmatrix}
            1 & -1 & 1 & -1
        \end{pmatrix}, \quad \firstNNvec 2 := -1.
    \]
Note that indeed, $\nnDepth{\firstNN} = 2$, $\nnWidth{\firstNN} = 4$ and $\nnWeights{\firstNN} = \{0,1,2,3,-1,-2\}$. Now, let $x \in \R$, and note that by definition of $\firstNN$, we have
\[
    \firstNN(x) = \ReLU(3x+2) - \ReLU(3x+1) + \ReLU(3x-1) - \ReLU(3x-2) - 1 = f(x).
\]
% \begin{enumerate}[label=(\alph*)]
%     \item If $x \leq -2/3$, then
%     \[
%         \firstNN(x) = 0 - 0 + 0 - 0 - 1 = -1 = f(x).
%     \]
%     \item If $-2/3 < x < -1/3$, then
%     \[
%         \firstNN(x) = 3x+2 - 0 + 0 - 0 - 1 = 3x + 1 = f(x).
%     \]
%     \item If $-1/3 \leq x < 1/3$, then
%     \[
%         \firstNN(x) = 3x+2 - (3x+1) + 0 - 0 - 1 = 0 = f(x).
%     \]
%     \item If $1/3 \leq x < 2/3$, then
%     \[
%         \firstNN(x) = 3x+2 - (3x+1) + (3x-1) - 0 - 1 = 3x - 1 = f(x).
%     \]
%     \item If $x \geq 2/3$, then
%     \[       \firstNN(x) = 3x+2 - (3x+1) + (3x-1) - (3x-2) - 1 = 1 = f(x).
%     \]
% \end{enumerate}
% Therefore, $\firstNN(x) = f(x)$ for all $x \in \R$. This concludes the proof.
\end{proof}

Once the reader is available, the combined write-and-shift candidates can be realized by a single shallow network.
\begin{lemma}\label{lemma:writing-and-shifting-operations-as-neural-network}
    Let $u : \R^2 \to \R^{18}$ be the function defined by \eqref{eq:combined-transition-function-piecewise-linear}. Then, there exists a neural network $\nn$ satisfying $\nnWidth{\nn} = 18$, $\nnDepth{\nn} = 2$, $\nnWeights{\nn} \subseteq \pm\{0,1/4,1,2,3,4\}$ such that $\nn(x) = u(x)$ for all $x \in \R^2$.
\end{lemma}
\begin{proof}
First, let 
\[
    A^1 := \begin{bmatrix}
        1 & 0\\
        0 & 1\\
        \firstNNmat 1 & 0\\
        0 & \firstNNmat 1
    \end{bmatrix}, \quad b^1 := \begin{bmatrix}
        0\\
        0\\
        \firstNNvec 1\\
        \firstNNvec 1
    \end{bmatrix}, \quad A^2 := \begin{bmatrix}
        1 & 0 & 0 & 0\\
        0 & 1 & 0 & 0\\
        0 & 0 & \firstNNmat 2 & 0\\
        0 & 0 & 0 & \firstNNmat 2
    \end{bmatrix}, \quad b^2 := \begin{bmatrix}
        0\\
        0\\
        \firstNNvec 2\\
        \firstNNvec 2
    \end{bmatrix},
\]
where $\firstNNmat 1, \firstNNvec 1, \firstNNmat 2, \firstNNvec 2$ are as in the proof of Lemma \ref{lemma:first-element-reading}. Note that for every $x,y \in \R_+$, we have
\begin{align}\label{eq:definition-of-f-symb-move-0}
    A^2 \ReLU(A^1(x,y) + b^1) + b^2 & = (x,y, f(x), f(y)),
\end{align}where $f$ is the function defined by \eqref{eq:piecewise-linear-function-read-first-symbol:explicit}. Now, consider, for $\symb \in \{-1,0,1\}$ and $\move \in \{-1,0,1\}$, the functions $f_{\symb,-1},f_{\symb,0},f_{\symb,1} : \R^4 \to \R^2$ defined by
\begin{equation}\label{eq:definition-of-f-symb-move-1}
    f_{\symb,-1}(x) := \left(x_4 + \frac{x_1 + \symb - x_3}{4}, 4(x_2 - x_4)\right) = \begin{pmatrix}
        \frac{1}{4} & 0 & -\frac{1}{4} & 1\\
        0 & 4 & 0 & -4
     \end{pmatrix} x + \begin{pmatrix}
         \symb/4\\
         0
     \end{pmatrix} =: A_{\symb,-1} x + b_{\symb,-1},
\end{equation}
\begin{equation}\label{eq:definition-of-f-symb-move-2}
    f_{\symb,0}(x) := \left(x_1 + \symb - x_3, x_2\right) = \begin{pmatrix}
        1 & 0 & -1 & 0\\
        0 & 1 & 0 & 0
     \end{pmatrix} x + \begin{pmatrix}
         \symb\\
         0
     \end{pmatrix} =: A_{\symb,0} x + b_{\symb,0},
\end{equation}
and
\begin{equation}\label{eq:definition-of-f-symb-move-3}
    f_{\symb,1}(x) := \left(4(x_1 - x_3), \symb + x_2/4\right) = \begin{pmatrix}
        4 & 0 & -4 & 0\\
        0 & \frac{1}{4} & 0 & 0
     \end{pmatrix} x + \begin{pmatrix}
         0\\
         \symb
     \end{pmatrix} =: A_{\symb,1} x + b_{\symb,1},
\end{equation}
for every $x \in \R^4$. Note that for every $\symb \in \{-1,0,1\}$, $\move \in \{-1,0,1\}$ and $x,y \in \R$, we have
\begin{equation}\label{eq:definition-of-f-symb-move-4}
     f_{\symb,\move}(x,y,f(x),f(y)) = s_m(w_\symb(x,y)) = u_{3 \move + \symb + 5}(x,y).
\end{equation}
Now, define 
% $u : \R^4 \to \R^{18}$ by
\[
     A := \begin{bmatrix}
        A_{-1,-1}\\
        \vdots\\
        A_{1,1}
     \end{bmatrix}, \quad b := \begin{bmatrix}
         b_{-1,-1}\\
         \vdots\\
         b_{1,1}
     \end{bmatrix}
\]
% for all $x \in \R^4$, let 
% \[
%      \tilde A_2 := A \times A_2, \quad \tilde b_2 := A b^2 + b,
% \]
and $\nn := (2,4,18; A^1, b^1, A \times A_2, A b^2 + b)$. Note that indeed, $\nnDepth{\nn} = 2$, $\nnWidth{\nn} = 18$ and $\nnWeights{\nn} \subseteq \pm\{0,1/4,1,2,3,4\}$. Moreover, for every $x,y \in \R_+$, we have
\begin{align*}
    \nn(x,y) & = A \times A_2 \ReLU(A^1(x,y) + b^1) + A b^2 + b = A (A_2\ReLU(A^1(x,y) + b^1) + b^2) + b\\
    & \overset{\eqref{eq:definition-of-f-symb-move-0}}= A(x,y,f(x),f(y)) + b \overset{\eqref{eq:definition-of-f-symb-move-1},\eqref{eq:definition-of-f-symb-move-2},\eqref{eq:definition-of-f-symb-move-3}}= (f_{-1,-1}(x,y,f(x),f(y)), \ldots, f_{1,1}(x,y,f(x),f(y)))\\
    &\overset{\eqref{eq:definition-of-f-symb-move-4}}= (u_1(x,y), \ldots, u_9(x,y)) = u(x,y).
\end{align*}
% Therefore, for every $\tape \in \workSymbols^\Z$, we have
% \begin{align*}
%     \nn(\cantorMap(\tape)) & = (s_{m_1}(w_{s_1}(\cantorMap(\tape))), \ldots, s_{m_9}(w_{s_9}(\cantorMap(\tape)))) = \cantorMap(U_{\symb_1,\move_1}(\tape), \ldots, U_{\symb_9,\move_9}(\tape)) = \cantorMap (U \tape).
% \end{align*}
% Note that also, $\nnWeights{\nn} \subseteq \pm \{0,1/4,1,2,3,4\}$. This concludes the proof.
\end{proof}

\subsubsection{Simulation of the transition and command functions}
\newcommand{\writeMoveOp}{SW}
% We defined the operation of exhaustive write-moves, which is the operation that, given a tape, computes the result of applying all possible write-moves to the tape. We define this operation formally as $\writeMoveOp : \workSymbols^\Z \to (\workSymbols^\Z)^9$ by
% \begin{align}\label{eq:write-move-operation}
%     \writeMoveOp(\tape) &:= (\shiftOp(\writeOp(\tape,-1),-1), \shiftOp(\writeOp(\tape,-1),0), \shiftOp(\writeOp(\tape,-1),1),\\
%     & \quad \quad \shiftOp(\writeOp(\tape,0),-1), \shiftOp(\writeOp(\tape,0),0), \shiftOp(\writeOp(\tape,0),1),\nonumber\\
%     & \quad \quad \shiftOp(\writeOp(\tape,1),-1), \shiftOp(\writeOp(\tape,1),0), \shiftOp(\writeOp(\tape,1),1)).\nonumber
% \end{align}

% \subsection{Simulation of the transition and command functions}

\newcommand\symbEncNN{\mathcal{N}_{\mathrm{symb}}}
\newcommand\symbEncNNmat[1]{A^{#1}_{\mathrm{symb}}}
\newcommand\symbEncNNvec[1]{b^{#1}_{\mathrm{symb}}}

We next build a network that encodes the currently scanned symbol together with the current state.
\begin{lemma}\label{lemma:pattern-making-neural-network}
    Let $n \in \N$. There exists a right-selector neural network $\nn$ with $\nnWidth{\nn} = 3n$, $\nnDepth{\nn} = 3$ and $\nnWeights{\nn} \subseteq \pm\{0,1,2,3\}$ such that
    \[
        \nn(\oneHot{n}{q},\cantorMap(\tape)) = \oneHot{3n}{3(q-1)+\readOp \tape +2},
    \]
    for every $q \in \{1,\ldots,n\}$ and $\tape \in \workSymbols^\Z$.
\end{lemma}
\begin{proof}
    First, let use consider 
    \[
        A^2 := \begin{bmatrix}
            -1 & 1 & 0 & 0\\
            1 & -1 & -1 & 1\\
            0 & 0 & 1 & -1
        \end{bmatrix}, \quad b^2 := \begin{bmatrix}
            1\\
            0\\
            0
        \end{bmatrix}.
    \]
    Let $g : \R \to \R^3$ be the function defined by $g(x) := A^2 \ReLU(\firstNNmat 1 x + \firstNNvec 1) + b^2$, where $\firstNNmat 1$ and $\firstNNvec 1$ are as in the proof of Lemma \ref{lemma:first-element-reading}. Let $f : \R \to \R$ be the function defined by \eqref{eq:piecewise-linear-function-read-first-symbol:explicit}.
    \begin{enumerate}
        \item For $x \leq -2/3$, we have $g(x) = (1,0,0) = \oneHot{3}{1} = \oneHot{3}{2+f(x)}$.
        \item For $-1/3 < x < -1/3$, we have $g(x) = (0,1,0) = \oneHot{3}{2} = \oneHot{3}{2+f(x)}$.
        \item For $x \geq 2/3$, we have $g(x) = (0,0,1) = \oneHot{3}{3} = \oneHot{3}{2+f(x)}$.
    \end{enumerate}
    Therefore, in particular,
    \[
        g(\tilde \cantorMap(\tape^+)) = \oneHot{3}{2+f(\tilde \cantorMap(\tape^+))} = \oneHot{3}{2 + R\tape},
    \]
    for every $\tape \in \workSymbols^\Z$. Now, let
    $\tilde A^1 := \begin{bmatrix}
        I_n & 0 \\
        0 & \firstNNmat 1
     \end{bmatrix} \in \R^{(n+4) \times (n+1)}$, $\tilde b^1 := \begin{bmatrix}
        0\\
        \firstNNvec 1
    \end{bmatrix} \in \R^{n +4}$, $\tilde A^2 \in \R^{3n \times (n+4)}$ be defined by
    \[
    \tilde A^2 := \begin{bmatrix}
        \tilde A_1\\
        \tilde A_2\\
        \vdots\\
        \tilde A_n
    \end{bmatrix} \quad \text{ where } \tilde A_\ell := \begin{bmatrix}
            \begin{array}{c}
                \oneHot{n}{\ell}^T\\
                \oneHot{n}{\ell}^T\\
                \oneHot{n}{\ell}^T
            \end{array} & A^2
    \end{bmatrix}  \in \R^{3 \times (n+4)}, \quad \text{and} \quad \tilde b^2 = \begin{bmatrix}
        b^2-1_3 \\
        b^2-1_3\\
        \vdots\\
        b^2-1_3
    \end{bmatrix} \in \R^{3n}.
\] Define the neural network $\nn := (n+1, n+4, 3n, 3n; \tilde A^1, \tilde b^1, \tilde A^2, \tilde b^2, I_{3n}, 0_{3n})$. Note that indeed, $\nn$ is a right-selector, $\nnDepth{\nn} = 3$, $\nnWidth{\nn} = 3n$ and $\nnWeights{\nn} \subseteq \pm\{0,1,2,3\}$. Moreover, for every $q,\ell \in \{1,\ldots,n\}$, $\tape \in \workSymbols^\Z$, we have
\begin{align*}
    \ReLU\left(\tilde A_\ell \ReLU(\tilde A^1 (\oneHot{n}{q}, \tilde \cantorMap(\tape^+)) + \tilde b^1) + \tilde b^2 -1_3\right) & = \ReLU\left(
        \begin{bmatrix}
            \oneHot{n}{\ell}^T\\
            \oneHot{n}{\ell}^T\\
            \oneHot{n}{\ell}^T
        \end{bmatrix} \ReLU(I_n \oneHot{n}{q} + 0) +
        A^2 \ReLU(\firstNNmat 1 \oneHot{n}{q}\cantorMap(\tape) + \firstNNvec 1) + b^2 - 1_3\right)\\
        & = \ReLU\left(\begin{bmatrix}
            \oneHot{n}{\ell}^T\\
            \oneHot{n}{\ell}^T\\
            \oneHot{n}{\ell}^T
        \end{bmatrix}\oneHot{n}{q} + g(\tilde \cantorMap(\tape^+)) - 1_3 \right)\\
        & = \ReLU\left(\delta_{q,\ell} 1_3 + \oneHot{3}{2 + R\tape} - 1_3\right) = \delta_{q,\ell} \oneHot{3}{2 + R\tape}.
\end{align*}
Therefore, for every $q \in \{1,\ldots,n\}$ and $\tape \in \workSymbols^\Z$, we have
\begin{align*}
    \nn(\oneHot{n}{q}, \tilde \cantorMap(\tape^+)) & = \tilde A^2 \ReLU(\tilde A^1 (\oneHot{n}{q}, \tilde \cantorMap(\tape^+)) + \tilde b^1) + \tilde b^2\\
    & = (\delta_{q,1} \oneHot{3}{2 + R\tape}, \ldots, \delta_{q,n} \oneHot{3}{2 + R\tape}) = \oneHot{3n}{3(q-1)+2 + R\tape}.
\end{align*}
Finally, by Lemma \ref{lemma:parallelization-of-neural-networks} applied to $m = 1$, $d=n+2$, $\nn_1 = \nn$ and $\iota_1 : \{1, \ldots, n+1\} \to \{1, \ldots, n+2\}, i \mapsto i$, there exists a neural network $\tilde \nn$ with $\nnWidth{\tilde \nn} = 3n$, $\nnDepth{\tilde \nn} = 3$ and $\nnWeights{\tilde \nn} = \nnWeights{\nn} \cup \{0\} = \pm\{0,1,2,3\}$ such that
\[
    \tilde \nn(\oneHot{n}{q}, \cantorMap(\tape)) = \nn(\proj_\iota(\oneHot{n}{q}, \cantorMap(\tape))) = \nn(\oneHot{n}{q}, \tilde \cantorMap(\tape^+)) = \oneHot{3n}{3(q-1)+2 + R\tape},
\]
for every $q \in \{1,\ldots,n\}$ and $\tape \in \workSymbols^\Z$. This concludes the proof.
\end{proof}

\newcommand\patternNN[1]{\mathcal{N}^{#1}_{\mathrm{pat}}}
\newcommand\patternNNmat[2]{A^{#1,#2}_{\mathrm{pat}}}
\newcommand\patternNNvec[2]{b^{#1,#2}_{\mathrm{pat}}}
\newcommand\diracNotation[1]{\delta_{#1}}

\newcommand\transStateMat[1]{W^{#1,\state}}
\newcommand\transSymbMat[1]{W^{#1,\symb}}
\newcommand\transMoveMat[1]{W^{#1,\move}}
\newcommand\transSymbMoveMat[1]{W^{#1,\tape}}
\newcommand\transCommMat[1]{W^{#1,\commFunc}}
\newcommand\patternMatchingNNMat[1]{W^{#1,\mathrm{match}}}
\newcommand\patternMatchingNNvec[1]{b^{#1,\mathrm{match}}}
\newcommand\patternMatchingNN[1]{\nn^{#1,\mathrm{match}}}
\newcommand\transNN[1]{\mathcal{N}^{#1}_{\mathrm{trans}}}
\newcommand\funGraph{\operatorname{graph}}
\newcommand\characteristicFunc[1]{\operatorname{\chi}_{#1}}
% \newcommand\transFuncSymbMove{\transFunc^{\symb \move}}

% Given a \TMNU $\tmnuM := (\nStates, \ntapes, \neurDim, \transFunc, \commFunc)$, we let $\Fc_\tmnuM := \{ \commFunc(\state,\symb) : \state \in \{1, \dots, \nStates\}, \symb \in \workSymbols\}$ be the set of all possible outputs of the command function $\commFunc$ of $\tmnuM$, and we let $\nFunctions{\tmnuM} := \# \Fc_\tmnuM$. We order the elements of $\Fc_\tmnuM$ as $f_1 <_L f_2 <_L \dots <_L f_{\nFunctions{\tmnuM}}$. We define $\tilde \commFunc : \{1, \dots, \nStates\} \times \workSymbols \to \{1, \ldots, \nFunctions{\tmnuM}\}$ to be such that 
% \[
%     f_{\tilde \commFunc(\state,\symb)} = \commFunc(\state,\symb), \quad \state \in \{1, \dots, \nStates\}, \symb \in \workSymbols.
% \]

 Given a set $S \subseteq A$, we denote by $\characteristicFunc{S} : A \to \{0,1\}$ the characteristic function of $S$, defined by $\characteristicFunc{S}(a) := 1$ if $a \in S$ and $\characteristicFunc{S}(a) := 0$ otherwise.

\newcommand\funcTMNUindex{\ell}
\newcommand\symbMove{j}
The pattern network allows us to realize the state, tape, and command selectors by ordinary affine layers.
\begin{lemma}\label{lem:transition-functions-as-neural-networks-1}
    Let $\tmnuM := (\nStates, \ntapes, \neurDim, \transFunc, \commFunc)$ be a \TMNU, $M >0$, and let the functions $\Delta_\state : \R^{\nStates + 1} \to \R^{\nStates}$, $\Delta_{\tape} : \R^{\nStates + 1} \to \R^9$, and $\Delta_\commFunc : \R^{\nStates+1}\to \R^{\nFunctions{\tmnuM}}$ be defined as in Lemma \ref{lem:transition-functions-as-cpwl-functions}. Then, there exist a neural network $\nn_M$ such that $\nnWidth{\nn_M} \leq \max\{3\nStates, \nStates + \nFunctions{\tmnuM} + 9\}$, $\nnDepth{\nn_M} = 3$ and $\nnWeights{\nn_M} \subseteq \pm\{0,1,2,3,M\}$ and
    \begin{align*}
        \nn_M(\oneHot{n}{\state},\cantorMap(\tape)) & = \begin{bmatrix}
        \Delta_\state(\oneHot{n}{\state}, \tape)\\
        \Delta_{\tape}(\oneHot{n}{\state}, \tape) - 1_9\\
        M(\Delta_\commFunc(\oneHot{n}{\state}, \tape) - 1_{\nFunctions{\tmnuM}})
    \end{bmatrix},
    \end{align*}
    for every $\state \in \{1, \dots, \nStates\}$ and $\tape \in \workSymbols^\Z$.
\end{lemma}
\begin{proof}
    Let $\tmnuM := (\nStates, \ntapes, \neurDim, \transFunc, \commFunc)$ be a TMNU, and $M>0$. Given $A,B$ two sets and a function $f : A \to B$, we denote by $\funGraph(f) := \{(a,b) \in A \times B : f(a) = b\}$ the graph of $f$. We define the matrices $\transStateMat{\tmnuM} \in \R^{\nStates \times 3\nStates}$, $\transSymbMoveMat{\tmnuM} \in \R^{9 \times 3\nStates }$, 
    % $\transMoveMat{\tmnuM} \in \R^{3\nStates \times 3}$
     and $\transCommMat{\tmnuM} \in \R^{\nFunctions{\tmnuM} \times3\nStates}$ 
     by
    \begin{align*}
        \transStateMat{\tmnuM}_{3\state+\symb-1,\state'}  = \characteristicFunc{\funGraph(\transFuncState)}((\state, \symb), \state'), \quad
        \transSymbMoveMat{\tmnuM}_{3\state+\symb-1,k}  = \characteristicFunc{\funGraph(\transFuncSymbMove)}((\state, \symb), k), \quad
        % \transMoveMat{\tmnuM}_{3\state+\symb-1,\move+2} & = \characteristicFunc{\funGraph(\transFuncMove)}((\state, \symb), \move),\\
        \transCommMat{\tmnuM}_{3\state+\symb-1,\ell} = \characteristicFunc{\funGraph(\tilde \commFunc)}((\state, \symb), \ell),
    \end{align*}
    for all $\state,\state' \in \{1, \dots, \nStates\}$, $\symb \in \{-1,0,1\}$, $k \in \{1,\ldots,9\}$ and $\ell \in \{1,\ldots, \nFunctions{\tmnuM}\}$. Note that, for instance,
    \[
        \transStateMat{\tmnuM} \oneHot{3n}{3\state + \symb - 1} = \oneHot{\nStates}{\transFuncState(\state,\symb)}, \quad \state \in \{1, \dots, \nStates\}, \quad \symb \in \workSymbols,
    \] 
    and the same holds for $\transSymbMoveMat{\tmnuM}$ and $\transCommMat{\tmnuM}$ with respect to $\transFuncSymbMove$ and $\tilde \commFunc$, respectively.
    Define the neural network 
    \[\nn_M := \left(3\nStates, \nStates; 
    \begin{bmatrix}
        \transStateMat{\tmnuM}\\
        \transSymbMoveMat{\tmnuM}\\
        M\transCommMat{\tmnuM}
    \end{bmatrix}, \begin{bmatrix}
        0_{\nStates}\\
        -1_{9}\\
        -M1_{\nFunctions{\tmnuM}}
    \end{bmatrix}\right).\] 
    Note that $\nnDepth{\nn_M} = 1$, $\nnWidth{\nn_M} = \max\{3\nStates, \nStates + \nFunctions{\tmnuM} + 9\}$ and $\nnWeights{\nn_M} \subseteq \pm\{0,1,M\}$. Moreover, for every $\state \in \{1, \dots, \nStates\}$ and $\symb \in \workSymbols$, we have
    \begin{align*}
        \nn_M(\oneHot{3n}{3\state + \symb - 1}) & = \begin{bmatrix}
        \transStateMat{\tmnuM}\\
        \transSymbMoveMat{\tmnuM}\\
        \transCommMat{\tmnuM}
    \end{bmatrix} \oneHot{3n}{3\state + \symb - 1} + \begin{bmatrix}
        0_{\nStates}\\
        -1_{9}\\
        -M1_{\nFunctions{\tmnuM}}
    \end{bmatrix} = 
    \begin{bmatrix}
        \oneHot{\nStates}{\transFuncState(\state,\symb)}\\
        \oneHot{9}{\transFuncSymbMove(\state,\symb)} - 1_9\\
        M\left(\oneHot{\nFunctions{\tmnuM}}{\tilde \commFunc(\state,\symb)}- 1_{\nFunctions{\tmnuM}}\right)
    \end{bmatrix}.
    \end{align*}
    By Lemma \ref{lemma:pattern-making-neural-network}, there exists a right-selector neural network $\nn'$ such that $\nnDepth{\nn'} = 3$, $\nnWidth{\nn'} = 3n$ and $\nnWeights{\nn'} \subseteq \pm\{0,1,2,3\}$, and
    \[
        \nn'(\oneHot{n}{\state},\cantorMap(\tape)) = \oneHot{3n}{3\state + \readOp \tape - 1}.
    \]
    Hence, by Lemma \ref{lemma:composition-of-neural-networks} applied to $\nn'$ and $\nn_M$, there exists a neural network $\tilde \nn_M$ such that $\nnDepth{\tilde \nn_M} = 3$, $\nnWidth{\tilde \nn_M} \leq \max\{3n, n+9+\nFunctions{\tmnuM}\}$ and $\nnWeights{\tilde \nn_M} \subseteq \pm \{0,1,2,3,M\}$, such that
    \begin{align*}
        \tilde \nn_M(\oneHot{n}{\state},\cantorMap(\tape)) & = \nn(\nn'(\oneHot{n}{\state},\cantorMap(\tape))) = \begin{bmatrix}
        \oneHot{\nStates}{\transFuncState(\state,\readOp \tape)}\\
        \oneHot{9}{\transFuncSymbMove(\state,\readOp \tape)} - 1_9\\
        M\left(\oneHot{\nFunctions{\tmnuM}}{\tilde \commFunc(\state,\readOp \tape)}- 1_{\nFunctions{\tmnuM}}\right)
    \end{bmatrix} = \begin{bmatrix}
        \Delta_\state(\oneHot{n}{\state}, \tape)\\
        \Delta_{\tape}(\oneHot{n}{\state}, \tape) - 1_9\\
        M(\Delta_\commFunc(\oneHot{n}{\state}, \tape) - 1_{\nFunctions{\tmnuM}})
    \end{bmatrix}.
    \end{align*}
    This concludes the proof.
\end{proof}

\subsubsection{Neural state management}

This part builds a ReLU neural network that computes all possible outputs of the command function $\commFunc$ of a \TMNU $\tmnuM$ on a given neural state $\neurState$.

The following lemma packages all candidate neural-state updates into one network output.
\begin{lemma}\label{lemma:neurstate-management-neural-network}
    Let $\tmnuM := (\nStates, \neurDim, \transFunc, \commFunc)$ be a \TMNU, and 
    $f_\tmnuM = (f_1, \ldots, f_{\nFunctions{\tmnuM}})$ as in Definition \ref{def:functions-tmnu}. 
    Then, there exists a neural network $\nn$ satisfying $\nnDepth{\nn} = 2$, $\nnWidth{\nn} = 2\nFunctions{\tmnuM}\neurDim$, $\nnWeights{\nn} \subseteq \pm\{0,1\} \cup \pm \nnWeights{\tmnuM}$ and
    \[
        \nn(\neurState) = \ReLU\left(
            f_\tmnuM(\neurState), 
            -f_\tmnuM(\neurState)
        \right),
    \]
    for every $\neurState \in \R^\neurDim$.
\end{lemma}
\begin{proof}
    Let $\tmnuM := (\nStates, \neurDim, \transFunc, \commFunc)$ be a \TMNU, and consider $f_1, \ldots, f_{\nFunctions{\tmnuM}}$ to be the elements of $\Fc_\tmnuM$ ordered as $f_1 <_L f_2 <_L \dots <_L f_{\nFunctions{\tmnuM}}$. Let $i \in \{1, \dots, \nFunctions{\tmnuM}\}$. If $f_i$ is an affine function, then there exist $A_i \in \R^{\neurDim \times \neurDim}$ and $b_i \in \R^\neurDim$ such that $f_i(\neurState) = A_i \neurState + b_i$ for every $\neurState \in \R^\neurDim$. Therefore, we have
    \[
        \ReLU(f_i(\neurState)) = \ReLU(A_i \neurState + b_i) \quad \text{and} \quad \ReLU(-f_i(\neurState)) = \ReLU(-A_i \neurState - b_i) =: \ReLU(A'_i \neurState + b'_i),
    \]
    for every $\neurState \in \R^\neurDim$. If $f_i$ is a function of the form $f_i = \ReLU_{S_i}$ for some $S_i \subseteq \{1, \dots, \neurDim\}$, then we have
    \[
        \ReLU(f_i(\neurState)) = \ReLU(\ReLU_{S_i}(\neurState)) = \ReLU(\neurState) =: \ReLU(A_i\neurState + b_i) \quad \text{and} \quad 
    \]
    and
    \[
        \ReLU(-f_i(\neurState)) = \ReLU(-\ReLU_{S_i}(\neurState)) =: (z_1, \ldots, z_\neurDim),
    \]
    where $z_\ell = 0 = \ReLU(0 \cdot \neurState + 0)$ if $\ell \in S_i$ and $z_\ell = \ReLU(-\neurState_\ell) = \ReLU(-1 \cdot \neurState_\ell + 0)$ if $\ell \notin S_i$, for every $\ell \in \{1, \dots, \neurDim\}$. Accordingly, we define $A'_i \in \{-1,0\}^{\neurDim \times \neurDim}$ and $b'_i := 0_\neurDim$ such that $\ReLU(-f_i(\neurState)) = \ReLU(A'_i \neurState + b'_i)$ for every $\neurState \in \R^\neurDim$. Hence, by defining the matrices
    \[
        A := \begin{bmatrix}
            A_1\\
            \vdots\\
            A_{\nFunctions{\tmnuM}}
        \end{bmatrix}, \quad A' := \begin{bmatrix}
            A'_1\\
            \vdots\\
            A'_{\nFunctions{\tmnuM}}
        \end{bmatrix}, \quad b := \begin{bmatrix}
            b_1\\
            \vdots\\
            b_{\nFunctions{\tmnuM}}
        \end{bmatrix}, \quad \text{and} \quad b' := \begin{bmatrix}
            b'_1\\
            \vdots\\
            b'_{\nFunctions{\tmnuM}}
        \end{bmatrix},
    \]
    and the neural network
    \[
        \nn := \left(2\neurDim, 2\nFunctions{\tmnuM}\neurDim; \begin{bmatrix}
            A\\
            A'
        \end{bmatrix}, \begin{bmatrix}
            b\\
            b'
        \end{bmatrix}\right),
    \]
    we have $\nnDepth{\nn} = 2$, $\nnWidth{\nn} = 2\nFunctions{\tmnuM}\neurDim$, $\nnWeights{\nn} \subseteq \pm\{0,1\} \cup \pm \nnWeights{\tmnuM}$ and
    \[
        \nn(\neurState) = \ReLU\left(
            f_\tmnuM(\neurState), 
            -f_\tmnuM(\neurState)
        \right),
    \]
    for every $\neurState \in \R^\neurDim$. This concludes the proof.
\end{proof}

\subsubsection{Finalization of the proof}

% For $n,d \in \N$, let $g_{n,d} : \R^{n(d+1)} \to \R^d$ and $g_{n,d} : \R^{n(2d+1)} \to \R^d$ be defined by
% \begin{equation}
%     g^+_{n,d}(x,z) := \sum_{i=1}^n \ReLU(z_i + x_i1_d), \quad x \in \R^n, z_1, \dots, z_n \in \R^d,
% \end{equation}
% and
% \begin{equation}
%     g_{n,d}(x,z) := g^+_{n,d}(x,z_1, \dots, z_n) - g^+_{n,d}(x,z_{n+1}, \dots, z_{2n}), \quad x \in \R^n, z_1, \dots, z_{2n} \in \R^d.
% \end{equation}
% The next Lemma follows directly from the definition of $g^+_{n,d}$ and $g_{n,d}$.
The final selector needed in the network construction is itself realized by a small ReLU network.
\begin{lemma}\label{lemma:selection-neural-networks}
    Let $n,d \in \N$, and let $g_{n,d} : \R^{n(d+1)} \to \R^d$ be as in Lemma \ref{lem:selection-function}. Then, there exists two left-separator neural networks $\nn_1$ and $\nn_2$ satisfying $\nnDepth{\nn_1} = \nnDepth{\nn_2} = 2$, $\nnWidth{\nn_1} = n(d+1)$, $\nnWidth{\nn_2} = n(2d+1)$, $\nnWeights{\nn_1},\nnWeights{\nn_2} \subseteq \pm \{0,1\}$, such that $\nn_1(x,z) = g_{n,d}(x,z)$ and $\nn(x,z,z') = g_{n,d}(x,z) - g_{n,d}(x,z')$ for every $x \in \R_+^n$ and $z , z'\in \R^{n d}$.
\end{lemma}
\begin{proof}
    The Lemma follows directly from the definition of $g_{n,d}$ as
    \begin{equation}
    g_{n,d}(x,z_1, \ldots, z_d) := \sum_{i=1}^n \ReLU(z_i + x_i1_d), \quad x \in \R^n, z_1, \dots, z_n \in \R^d,
\end{equation}
as there exists two matrices $A_1 \in \{0,1\}^{nd \times n(d+1)}$ and $A_2 \in \{0,1\}^{d \times nd}$ such that $g_{n,d}(x,z) = A_1 \ReLU(A_2(x,z))$ for every $x \in \R^n$ and $z \in \R^{nd}$, and two matrices $A'_1 \in \{0,1\}^{nd \times n(2d+1)}$ and $A'_2 \in \{-1,0,1\}^{d \times n(2d+1)}$ such that $g_{n,d}(x,z) - g_{n,d}(x,z') = A'_1 \ReLU(A'_2(x,z,z'))$ for every $x \in \R^n$ and $z,z' \in \R^{nd}$.
\end{proof}

We can now combine the tape, command, and selection subnetworks into a neural network for one full \TMNU step.
\begin{theorem}\label{thm:transition-function-as-neural-network:appendix}
    Let $\tmnuM := (\nStates, \neurDim, \transFunc, \commFunc)$ be a \TMNU and $C > 0$. Then, there exists a ReLU neural network $\nn \in \nns{\nStates+2+\neurDim}{\nStates+2+\neurDim}$ such that $\nnDepth{\nn} = 4$, \[\nnWidth{\nn} \leq \max\{3\nStates, \nStates + \nFunctions{\tmnuM} + 9\} + 2\neurDim\nFunctions{\tmnuM} + 18, \qquad \nnWeights{\nn} \subseteq \pm \{0,1/4,1,2,3,4,\|\tmnuM\|_C\} \cup \pm \nnWeights{\tmnuM}\] and 
    \[\nn(\configMap{\tmnuM}(c)) = \configMap{\tmnuM}(M(c)), \quad c \in \Bc_\tmnuM(C).\]
\end{theorem}
\begin{proof}
    Let $M := \|\tmnuM\|_C$, $\nn_M$ be the neural network given by Lemma \ref{lem:transition-functions-as-neural-networks-1}, $\nn_1$ be the neural network given by Lemma \ref{lemma:writing-and-shifting-operations-as-neural-network}, and $\nn_2$ be the neural network given by Lemma \ref{lemma:neurstate-management-neural-network}. Consider also the neural networks $\nnId_{18,2}$ and $\nnId_{2 \neurDim \nFunctions{\tmnuM},2}$ given by Lemma \ref{lemma:identity-neural-network}, and note that they are left- and right-selectors. Therefore, by Lemma \ref{lemma:composition-of-neural-networks} applied to $\nn_1$ and $\nnId_{18,2}$, there exists a neural network $\tilde \nn_1$ such that $\nnDepth{\tilde \nn_1} = 3$, $\nnWidth{\tilde \nn_1} = 18$ and $\nnWeights{\tilde \nn_1} \subseteq \pm\{0,1/4,1,2,3,4\}$, and 
    \[
        \tilde \nn_1(\cantorMap(\tape)) = \nnId_{18,2}(\nn_1(\cantorMap(\tape))) = \ReLU(u(\cantorMap(\tape))) = u(\cantorMap(\tape)), \quad \tape \in \workSymbols^\Z.
    \]
    Moreover, by Lemma \ref{lemma:composition-of-neural-networks} applied to $\nn_2$ and $\nnId_{2\neurDim\nFunctions{\tmnuM},2}$, there exists a neural network $\tilde \nn_2$ such that $\nnDepth{\tilde \nn_2} = 3$, $\nnWidth{\tilde \nn_2} = 2\neurDim\nFunctions{\tmnuM}$ and $\nnWeights{\tilde \nn_2} \subseteq \pm\{0,1\} \cup \pm\nnWeights{\tmnuM}$, and 
    \[
        \tilde \nn_2(x) = \nnId_{2\neurDim\nFunctions{\tmnuM},2}(\nn_2(x)) = \ReLU\left(\ReLU\left(f_\tmnuM(x), -f_\tmnuM(x)\right)\right)  = \ReLU\left(f_\tmnuM(x), -f_\tmnuM(x)\right),
    \]
    for every $x \in \R^{2\neurDim\nFunctions{\tmnuM}}$. Now, by Lemma \ref{lemma:parallelization-of-neural-networks} applied to $d := \nStates + 2 + \neurDim$, $m = 3$, $\nn_1 = \nn_M$, $\nn_2 = \tilde \nn_1$, $\nn_3 = \tilde \nn_2$ and $\iota_1 : \{1, \dots, \nStates+2\} \to \{1, \dots, \nStates + 2 + \neurDim\}, i \mapsto i$, $\iota_2 : \{1, 2\} \to \{1, \dots, \nStates + 2 + \neurDim\}, i \mapsto i + \nStates$ and $\iota_3 : \{1, \dots, \neurDim\} \to \{1, \dots, \nStates + 2 + \neurDim\}, i \mapsto i + \nStates + 2$, there exists a neural network $\tilde\nn$ satisfying $\nnDepth{\tilde\nn} = 3$, 
    \[
        \nnWidth{\tilde\nn} \leq \max\{3\nStates, \nStates + \nFunctions{\tmnuM} + 9\} + 2\neurDim\nFunctions{\tmnuM} + 18,
    \]
    and $\nnWeights{\tilde\nn} \subseteq \pm\{0,1/4,1,2,3,4,M\} \cup \pm\nnWeights{\tmnuM}$, such that
    \[
        \tilde\nn(x) = (\nn_M(\proj_{\iota_1}(x)), \tilde \nn_1(\proj_{\iota_2}(x)), \tilde \nn_2(\proj_{\iota_3}(x))) = (\nn_M(x_{1:\nStates+2}), \tilde \nn_1(x_{\nStates+1:\nStates+2}), \tilde \nn_2(x_{\nStates+3:\nStates+2+\neurDim})),
    \]
    for every $x \in \R^{\nStates + 2 + \neurDim}$. In particular, for $x = \configMap{\tmnuM}(c) \in \R^{\nStates + 2 + \neurDim}$ for some $c \in \Cc_\tmnuM$, we have
    \begin{align}\label{eq:final-neural-network-1}
        \tilde \nn(\configMap{\tmnuM}(c)) & = (\nn_M(\oneHot{\nStates}{\state},\cantorMap(\tape)), \tilde \nn_1(\cantorMap(\tape)), \tilde \nn_2(\neurState)) = \begin{bmatrix}
        \Delta_\state(\oneHot{\nStates}{\state}, \tape)\\
        \Delta_\tape(\oneHot{\nStates}{\state}, \tape) - 1_9\\
        M(\Delta_\commFunc(\oneHot{\nStates}{\state}, \tape) - 1_{\nFunctions{\tmnuM}})\\
        u(\cantorMap(\tape))\\
        \ReLU(f_\tmnuM(\neurState))\\
        \ReLU(-f_\tmnuM(\neurState))
    \end{bmatrix}.
    \end{align}
    Now, let $\nn'_1$ and $\nn'_2$ be the left-separator neural networks given by Lemma \ref{lemma:selection-neural-networks} applied to $n = 18$ and $d = \nStates + 2 + \neurDim$, respectively. Consider also the neural network $\nnId_{\nStates,2}$ given by Lemma \ref{lemma:identity-neural-network}, and note that it is a left-separator. By Lemma \ref{lemma:parallelization-of-neural-networks} applied to $d := \nStates + 9 + \nFunctions{\tmnuM} + 18 + 2\neurDim\nFunctions{\tmnuM}$, $m = 3$, $\nn_1 = \nnId_{\nStates,2}$,  $\nn_1 = \nn^+$, $\nn_2 = \nn$, 
    \[\iota_1 : \{1, \dots, \nStates\} \to \{1, \dots, \nStates + 9 + \nFunctions{\tmnuM} + 18 + 2\neurDim\nFunctions{\tmnuM}\}, i \mapsto i,\]
    \[\iota_2 : \{1, \dots, 27\} \to \{1, \dots, \nStates + 9 + \nFunctions{\tmnuM} + 18 + 2\neurDim\nFunctions{\tmnuM}\}, i \mapsto \begin{cases}
    i + \nStates & \text{if } i \in \{1, \dots, 9\},\\
    i + \nStates + \nFunctions{\tmnuM} & \text{if } i \in \{10, \dots, 27\},
    \end{cases}\] and 
    \[\iota_3 : \{1, \dots, (2\neurDim+1)\nFunctions{\tmnuM}\} \to \{1, \dots, \nStates + 9 + \nFunctions{\tmnuM} + 18 + 2\neurDim\nFunctions{\tmnuM}\}, i \mapsto \begin{cases}
    i + \nStates + 9 & \text{if } i \leq  \nFunctions{\tmnuM},\\
    i + \nStates + 27 + \nFunctions{\tmnuM} & \text{if } i > \nFunctions{\tmnuM},
    \end{cases}\]
    there exists a left-selector neural network $\nn$ satisfying $\nnDepth{\nn'} = 2$, $\nnWidth{\nn'} = \nStates + 27 + (2\neurDim+1)\nFunctions{\tmnuM}$ and $\nnWeights{\nn'} \subseteq \pm\{0,1\} \cup \pm\nnWeights{\tmnuM}$, such that
    \begin{align*}
        \nn'(x) &= (\nnId_{\nStates,2}(\proj_{\iota_1}(x)), \nn'_1(\proj_{\iota_2}(x)), \nn'_2(\proj_{\iota_3}(x)))
    \end{align*}
    for every $x \in \R^{\nStates + 27 + (2\neurDim+1)\nFunctions{\tmnuM}}$. Now, by Lemma \ref{lemma:composition-of-neural-networks} applied to $\tilde \nn$ and $\nn'$, there exists a neural network $\hat \nn$ such that $\nnDepth{\hat \nn} = 4$, $\nnWidth{\hat \nn} = \max\{3\nStates, \nStates + \nFunctions{\tmnuM} + 9\} + 2\neurDim\nFunctions{\tmnuM} + 18$ and $\nnWeights{\hat \nn} \subseteq \pm\{0,1/4,1,2,3,4,M\} \cup \pm\nnWeights{\tmnuM}$, such that
    \begin{align*}
        \hat \nn(x) & = \nn'(\tilde\nn(x)) = (\ReLU(\tilde\nn(x)_{1:\nStates}), \nn'_1(\proj_{\iota_2}(\tilde\nn(x))), \nn'_2(\proj_{\iota_2}(\tilde\nn(x)))),
    \end{align*}
    for every $x \in \R^{\nStates + 2 + \neurDim}$. In particular, for $x = \configMap{\tmnuM}(c)$ for some $c \in \Cc_\tmnuM$, we have
    \begin{align*}
        \ReLU(\tilde\nn(x)_{1:\nStates}) & = \ReLU(\Delta_\state(\oneHot{\nStates}{\state}, \tape)) = F_\state(\configMap{\tmnuM}(c)),
    \end{align*}
    \begin{align*}
        \nn'_1(\proj_{\iota_2}(\tilde\nn(x))) & = g_{9,2}(\Delta_{\tape}(\oneHot{\nStates}{\state}, \tape) - 1_9, u(\cantorMap(\tape))) = F_\tape(\configMap{\tmnuM}(c)),
    \end{align*}
    and
    \begin{align*}
        \nn'_2(\proj_{\iota_3}(\tilde\nn(x))) & = g_{\nFunctions{\tmnuM},\neurDim}(M(\Delta_\commFunc(\oneHot{\nStates}{\state}, \tape) - 1_{\nFunctions{\tmnuM}}), \ReLU(f_\tmnuM(\neurState))) - g_{\nFunctions{\tmnuM},\neurDim}(M(\Delta_\commFunc(\oneHot{\nStates}{\state}, \tape) - 1_{\nFunctions{\tmnuM}}), \ReLU(-f_\tmnuM(\neurState)))\\
        & = F_{\neurState,C}(\configMap{\tmnuM}(c)),
    \end{align*}
    where $F_\state, F_\tape$ and $F_{\neurState,C}$ are as the proof of Theorem \ref{thm:simulation-of-tmnu-by-cpwl-function}. Hence, for every $c \in \Bc_\tmnuM(C)$, we have
    \begin{align*}
        \hat \nn(\configMap{\tmnuM}(c)) & (F_\state(\configMap{\tmnuM}(c)), F_\tape(\configMap{\tmnuM}(c)), F_{\neurState,C}(\configMap{\tmnuM}(c))) = F_\tmnuM(\configMap{\tmnuM}(c)) \overset{(a)}= \configMap{\tmnuM}(M(c)),
    \end{align*}
    where (a) follows from the proof of Theorem \ref{thm:simulation-of-tmnu-by-cpwl-function}. This concludes the proof
\end{proof}

\subsection{Construction of an RNN that simulates a \TMNU}

\label{subsection:rnn-simulates-neural-network-iterations:appendix}

In this appendix, we formally show that we can simulate the iteration of a ReLU neural network by an RNN. 
% In the sequel, we will refer at neural networks as if they were functions. Specifically, we write that $\nn : \R^n \to \R^m$ is a neural network if $\nn$ is a neural network such that $\nnInDim{\nn} =n $ and $\nnOutDim{\nn} = m$.
We introduce the following notation and terminology for RNNs. Let $\Rc := (d,m,d';A_h,b_h,A_x,A_o,b_o)$ be an RNN.
\begin{enumerate}[label=(\alph*)]
    \item We define
    \begin{equation}
        \nnWeights{\Rc} := \nnWeights{A_x} \cup \nnWeights{A_h} \cup \nnWeights{A_o} \cup \nnWeights{b_h} \cup \nnWeights{b_o}.
    \end{equation}
    \item We say that $\Rc$ is a left-selector RNN if $A_x$ is a left-selector matrix and $b_h = 0$, and that $\Rc$ is a right-selector RNN if $A_o$ is a right-selector matrix and $b_o = 0$.
\end{enumerate}

\subsubsection{Technical Lemmata for RNNs}

In this part, we prove two useful lemmata that allow us to construct RNNs that perform linear transformations of the input and output of a given RNN.

We first record that a selector can be appended to the output without changing the hidden dynamics.
\begin{lemma}\label{lem:linear-transform-of-the-output-of-an-rnn}
    Let $n \in \N$, $\Rc$ be an RNN, and $A \in \R^{n \times d'}$ be a left-selector matrix. Then, there exists an RNN $\tilde \Rc$ satisfying $\rnnHidDim{\tilde \Rc} = \rnnHidDim{ \Rc}$ and
    \[
        \nnWeights{\tilde \Rc} \subseteq \nnWeights{\Rc} \cup \{0\}
    \]
    such that for every $x \in \R^n$ and $t \in \N$, we have $\tilde \Rc \Dc x[t] = A (\Rc \Dc x[t])$.
\end{lemma}
\begin{proof}
    Let $\Rc := (d,m,d';A_h,b_h,A_x,A_o,b_o)$. We define $\tilde \Rc := (d,m,n; A_h,b_h,A_x,\tilde A_o, \tilde b_o)$, where $\tilde A_o := A A_o$ and $\tilde b_o := A b_o$. Note that, indeed, we have $\rnnHidDim{\tilde \Rc} = m = \rnnHidDim{\Rc}$ and $\nnWeights{\tilde \Rc} \subseteq \nnWeights{\Rc} \cup \{0\}$. Now, let $x \in \R^n$ and $t \in \N$. Note that
    \[
        \tilde \Rc \Dc x[t] = \tilde A_o (\Hc \Dc x[t]) + \tilde b_o = A A_o (\Hc \Dc x[t]) + A b_o = A (\Rc \Dc x[t]).
    \] This concludes the proof.
\end{proof}

The companion input transformation is slightly more delicate, because it must be absorbed into the recurrent initialization.
\begin{lemma}\label{lem:affine-transform-of-the-input-of-an-rnn}
    Let $n \in \N$, $\Rc$ be a left-selector RNN, $A \in \R^{d \times n}$, and $b \in \R^d$. Then, there exists an RNN $\tilde \Rc$ satisfying $\rnnHidDim{\tilde \Rc} = \rnnHidDim{ \Rc} +1 $ and
    \[
        \nnWeights{\tilde \Rc} = \nnWeights{\Rc} \cup \nnWeights{A} \cup \pm\nnWeights{b},
    \]
    such that for every $x \in \R^n$ and $t \in \N$, we have $\tilde \Rc \Dc x[t] = \Rc \Dc (A x + b)[t]$.
\end{lemma}
\begin{proof}
    Let $(d,m,d';A_h,b_h,A_x,A_o,b_o) := \Rc$. We define $\tilde \Rc := (n, m + 1, d'; \tilde A_h, \tilde b_h, \tilde A_x, \tilde A_o, b_o)$ by
    \[
        \tilde A_h := \begin{bmatrix}
            A_h & -A_x b\\
            0 & 0
        \end{bmatrix} \in \R^{m+1 \times m+1}, \quad \tilde b_h := \begin{bmatrix}
            b_h + A_x b\\
            1
        \end{bmatrix}, \quad \tilde A_x := \begin{bmatrix}
            A_x A\\
            0
        \end{bmatrix}, \quad \text{and} \quad \tilde A_o := (A_o, 0).
    \]
    Note that, indeed, $\rnnHidDim{\tilde \Rc} = m + 1 = \rnnHidDim{\Rc} + 1$, and that since $\Rc$ is a left-selector, we have $\{0,1\} = \nnWeights{A_x}$, $\nnWeights{-A_xb} \subseteq \nnWeights{b}$, $\nnWeights{A_x A} \subseteq \nnWeights{A}$, and $\nnWeights{b_h + A_xb} \subseteq - \nnWeights{b}$, so that $\nnWeights{\tilde \Rc} = \nnWeights{\Rc} \cup \nnWeights{A} \cup \pm\nnWeights{b}$. Now, let $x \in \R^n$ and define $h_t := \tilde \Hc \Dc x[t]_{1:m}$ and $d_t := \tilde \Hc \Dc x[t]_{m+1}$ for every $t \in \No \cup {-1}$, where $\tilde \Hc$ is the hidden state operator of $\tilde \Rc$. In particular, note that $h_{-1} = 0$, $d_{-1} = 0$. Then, note that
    \begin{equation}
        \begin{cases}
        h_{t} &= \ReLU(\tilde A_h (h_{t-1},d_{t-1}) + \tilde b_h + \tilde A_x \Dc x[t]) = \ReLU(A_h h_{t-1} - A_x b d_{t-1} + b_h + A_x b + A_x A \Dc x[t])\\
        d_{t} &= \ReLU(0 \cdot d_{t-1} + 1) = 1
    \end{cases}
    \end{equation}
    for every $t \in \No$. Therefore, for every $t \in \No$, we have
    \begin{align}
        h_t &= \begin{cases}
        \ReLU(A_h h_{t-1} + b_h + A_x b + A_x A \Dc x[t]), & \text{if} \ t = 0\\
        \ReLU(A_h h_{t-1} + b_h + A_x A \Dc x[t]), & \text{if} \ t > 0\\
    \end{cases}\\
    & = \ReLU(A_h h_{t-1} + b_h + A_x \Dc (A x + b)[t]) = \Hc \Dc (A x + b)[t],
    \end{align}
    where (a) follows from the fact that $d_{t-1} = 1$ for every $t \in \N$. Finally, for every $t \in \N$, we have
    \[
        \tilde \Rc \Dc x[t] = \tilde A_o (h_t,d_t) + b_o = A_o h_t + b_o = A_o\Hc \Dc (A x + b)[t] + b_o = \Rc \Dc (A x + b)[t].
    \] 
    This concludes the proof.
\end{proof}

\subsubsection{Simulation of the iterations of a neural network by an RNN}

This part is very technical, and is separated in three steps: given a ReLU neural network $\nn$ of depth $L$, we first show that we can design an RNN that outputs the positive part of the iterates of $\nn$ at time steps that are multiples of $L$, and outputs zero at other time steps. Then, we show how to modify such an RNN to output the positive part of the iterates of $\nn$ at every time step. Finally, we show how to modify such an RNN to output the iterates of $\nn$. The next lemma corresponds to the first step of this construction.

The first construction stores the layers of one network evaluation across $L$ recurrent steps.
\begin{lemma}\label{lem:simulation-of-cpwl-function-iterations-with-an-rnn:app}
    Let $\nn \in \nns{n}{n}$ be a ReLU neural network, and let $L := \nnDepth{\nn}$. Then, there exists $m \in \N$ satisfying $n + L\leq m \leq (\nnWidth{\nn}+1) L$ and $A_\nn \in \R^{m \times m}$ such that $\nnWeights{A_\nn} = \nnWeights{\nn} \cup \{0,1\}$ and for every $x \in \R^n$, the sequence $(h_t \in \R^m)_{t \in \N}$ defined by
    \begin{equation}
        h_0 := (x, 0, \oneHot{L}{1}), \quad h_{t+1} := \ReLU(A_\nn h_t) \text{ for every } t \in \No
    \end{equation}
    satisfies
    \begin{equation}
        \proj_{1:n} h_t = (\ReLU \circ\nn)^{t//L}(x) \kroen{t \bmod L}{0}, \quad t \in \No,
    \end{equation}
    where $//$ denotes the integer division.
\end{lemma}
\begin{proof}
    Let $n \in \N$, $\nn := (N_0, N_1, \ldots, N_L; A^1, b^1, A^2, b^2, \ldots, A^L, b^L)$ be a ReLU neural network, such that $N_0 = N_L =n$ and$\nn \geq 0$. We let $N := n + \sum_{\ell=1}^{L-1} N_\ell$ and $m := N + L$, and we define the matrices 
    \[
        A := \begin{bmatrix}
            0 & 0 & \ldots & 0 & A^L\\
            A^1 & 0 & \ldots & 0 & 0\\
            0 & A_2 & \ldots & 0 & 0\\
            \vdots & \vdots & \ddots & \vdots & \vdots\\
            0 & 0 & \ldots & A^{L-1} & 0
         \end{bmatrix} \in \R^{N \times N},
        \quad 
        B := \begin{bmatrix}
            0 & 0 & \ldots & 0 & b^L\\
            b^1 & 0 & \ldots & 0 & 0\\
            0 & b_2 & \ldots & 0 & 0\\
            \vdots & \vdots & \ddots & \vdots & \vdots\\
            0 & 0 & \ldots & b^{L-1} & 0
         \end{bmatrix} \in \R^{N \times L},
    \]
    \[
         C := \begin{bmatrix}
            0 & 0 & \ldots & 0 & 1\\
            1 & 0 & \ldots & 0 & 0\\
            0 & 1 & \ldots & 0 & 0\\
            \vdots & \vdots & \ddots & \vdots & \vdots\\
            0 & 0 & \ldots & 1 & 0
         \end{bmatrix} \in \R^{L \times L}, \quad \text{and} \quad A_\nn := \begin{bmatrix}
            A & B\\
            0 & C
         \end{bmatrix} \in \R^{m \times m}.
    \]
    Note that, indeed, we have \[n + L \leq m = N + L = n + \sum_{\ell=1}^{L-1} N_\ell + L \leq L \nnWidth{\nn} + L = (\nnWidth{\nn}+1) L,\] and that $\nnWeights{A_\nn} = \nnWeights{\nn} \cup \{0,1\}$. Now, let $x \in \R^n$ and let $(h_{x,t} \in \R^m)_{t \in \N}$ be defined by $h_{x,0} := (x, 0, \oneHot{L}{1})$ and $h_{x,t+1} := \ReLU(A_\nn h_{x,t})$ for every $t \in \No$. We split the sequence $(h_{x,t})_{t \in \N}$ into subsequences $(h_{x,t}^\ell \in \R^{N_\ell})_{t \in \N}$ and $(c_t^\ell \in \R)_{t \in \N}$ for $\ell \in \{0, \ldots, L-1\}$ such that
    \[
        h_{x,t} = (h_{x,t}^0, h_{x,t}^1, \ldots, h_{x,t}^{L-1}, c_t^0, \ldots, c_t^{L-1}) \text{ for every } t \in \N.
    \]
    Note that, in particular,
    \begin{equation}\label{eq:simulation-of-cpwl-function-iterations-with-an-rnn:recursive-definition:base-case}
        h_{x,0}^0 = x, \quad h_{x,0}^1 = 0, \quad \ldots, \quad h_{x,0}^{L-1} = 0, \quad c_0^0 = 1, \quad c_0^1 = 0, \quad \ldots, \quad c_0^{L-1} = 0,
    \end{equation}
    and that
    \begin{equation}\label{eq:simulation-of-cpwl-function-iterations-with-an-rnn:recursive-definition:h}
        h_{x,t+1}^0  = \ReLU(A^L h_{x,t}^{L-1} + b^L c_t^{L-1}), \quad 
        h_{x,t+1}^\ell = \ReLU(A^\ell h_{x,t}^{\ell-1} + b^\ell c_t^{\ell-1}),
    \end{equation}
    and
    \begin{equation}\label{eq:simulation-of-cpwl-function-iterations-with-an-rnn:recursive-definition:c}
        c_{t+1}^0  = \ReLU(c_t^{L-1}), \quad c_{t+1}^\ell  = \ReLU(c_t^{(\ell-1)}),
    \end{equation}
    for every $\ell \in \{1, \ldots, L-1\}$ and $t \in \No$.
    % Note also that for every $t \in \N$, $h_t^0 = \proj_{1:n} h_t$, so that the statement of the Lemma is equivalent to showing that for every $t \in \N$, we have
    % \begin{equation}\label{eq:simulation-of-cpwl-function-iterations-with-an-rnn:induction-hypothesis}
    %     h_t^0 = \begin{cases}
    %         f^{t/L}(x) & \text{if } t \in L \No,\\
    %         0 & \text{if } t \notin L \No.
    %     \end{cases}
    % \end{equation}
    We divide the proof into several claims.
    \begin{claim*}
        For every $t \in \N$, $\ell \in \{0,\ldots, L-1\}$, we have $c_t^\ell = \kroen{\ell}{t \mod L}$.
    \end{claim*}
    \begin{proof}[Proof of the claim]
        We proceed by induction on $t$. For the base case, note that for every $\ell \in \{0,\ldots, L-1\}$, we have $c_0^\ell = \oneHot{L}{1}(\ell) = \kroen{\ell}{0}$. Now, let $t \in \N$ and assume that for every $\ell \in \{0,\ldots, L-1\}$, we have $c_t^\ell = \kroen{\ell}{t \mod L}$. Then, by \eqref{eq:simulation-of-cpwl-function-iterations-with-an-rnn:recursive-definition:c}, we have $c_{t+1}^0 = \ReLU(c_t^{L-1}) = \ReLU(\kroen{L-1}{t \mod L}) = \kroen{0}{(t+1) \mod L}$, and for every $\ell \in \{1, \ldots, L-1\}$, we have $c_{t+1}^\ell = \ReLU(c_t^{(\ell-1)}) = \ReLU(\kroen{\ell-1}{t \mod L}) = \kroen{\ell}{(t+1) \mod L}$. This concludes the proof of the claim.
    \end{proof}
    \begin{claim*}
        For every $t\in \{0, \ldots, L-1\}$, we have $h_{x,t}^0 = x \kroen{t}{0}$ and $h_{x,t}^\ell = \ReLU(\nn_\ell(x))\kroen{t}{\ell}$ for every $\ell \in \{1, \ldots, L-1\}$.
    \end{claim*}
    \begin{proof}[Proof of the claim]
        We proceed by induction on $t$. For the base case, note that by \eqref{eq:simulation-of-cpwl-function-iterations-with-an-rnn:recursive-definition:base-case}, we have $h_{x,0}^\ell = x \kroen{0}{\ell} = \ReLU(\nn_0(x)) \kroen{0}{\ell}$, for every $\ell \in \{0,\ldots, L-1\}$. Now, let $t \in \{0, \ldots, L-2\}$ and assume that for every $\ell \in \{0,\ldots, L-1\}$, we have $h_{x,t}^\ell = \nn_\ell(x)\kroen{t}{\ell}$. Then, by \eqref{eq:simulation-of-cpwl-function-iterations-with-an-rnn:recursive-definition:h}, we have 
        \begin{align*}
            h_{x,t+1}^0 &= \ReLU(A^L h_{x,t}^{L-1} + b^L c_t^{L-1}) = \ReLU(A^L \nn_{L-1}(x)\kroen{t}{L-1} + b^L \kroen{t}{L-1})\\
            & = \ReLU((A^L \nn_{L-1}(x) + b^L)\kroen{t}{L-1}) = \ReLU(A^L \nn_{L-1}(x) + b^L)\kroen{t}{L-1}\\
            & = 0 = x\kroen{t+1}{0},
        \end{align*}
        and for every $\ell \in \{1, \ldots, L-1\}$, we have \begin{align*}
            h_{x,t+1}^\ell &= \ReLU(A^\ell h_{x,t}^{\ell-1} + b^\ell c_t^{\ell-1}) = \ReLU(A^\ell \nn_{\ell-1}(x)\kroen{t}{\ell-1} + b^\ell \kroen{t}{\ell-1})\\
            & = \ReLU((A^\ell \nn_{\ell-1}(x) + b^\ell)\kroen{t}{\ell-1}) = \ReLU(A^\ell \nn_{\ell-1}(x) + b^\ell)\kroen{t}{\ell-1}\\
            & = \ReLU(\nn_\ell(x))\kroen{t+1}{\ell}.
        \end{align*}
        This concludes the proof of the claim.
    \end{proof}
    \begin{claim*}
        For every $k \in \N$, we have $h_{x,kL} = ((\ReLU \circ \nn)^k(x), 0, \oneHot{L}{1})$.
    \end{claim*}
    \begin{proof}[Proof of the claim]
        First note that 
        \begin{equation}
            h_{x,L}^0 = \ReLU(A^L h_{x,L-1}^{L-1} + b^L c_{L-1}^{L-1}) \overset{(a)}= \ReLU(A^L \nn_{L-1}(x) + b^L) = \ReLU(\nn_L(x)) = \ReLU(\nn(x)),
        \end{equation}
        and
        \begin{equation}
            h_{x,L}^\ell = \ReLU(A^\ell h_{x,L-1}^{\ell-1} + b^\ell c_{L-1}^{\ell-1}) \overset{(a)}= \ReLU(A^\ell \nn_{\ell-1}(x) + b^\ell) \kroen{L}{\ell} = 0,
        \end{equation}
         for every $\ell \in \{1, \ldots, L-1\}$, where (a) follows from the preceding two claims. Therefore,
        \begin{equation}
            h_{x,L} = (h_{x,L}^0, h_{x,L}^1, \ldots, h_{x,L}^{L-1}, c_L^0, \ldots, c_L^{L-1}) = (\ReLU \circ \nn(x), 0, \oneHot{L}{1}).
        \end{equation}
        We now proceed by induction on $k$. For the base case, note that $h_0 = (x,0,\oneHot{L}{1}) = ((\ReLU \circ \nn)^0(x),0,\oneHot{L}{1})$. Now, let $k \in \N$ and assume that $h_{x,kL} = ((\ReLU \circ \nn)^k(x),0,\oneHot{L}{1})$. Therefore, we have
        \begin{align*}
            h_{x,(k+1)L} = h_{(\ReLU \circ \nn)^k(x),L} = ((\ReLU \circ \nn)((\ReLU \circ \nn)^k(x)), 0, \oneHot{L}{1}) = ((\ReLU \circ \nn)^{k+1}(x), 0, \oneHot{L}{1}).
        \end{align*}
        This concludes the proof of the claim.
    \end{proof}
    \begin{claim*}
        For every $k \in \N$, $r \in \{1,\ldots,L-1\}$, we have $h_{x,kL+r}^0 = (\ReLU \circ \nn)^{k+1}(x) \kroen{r}{0}$.
    \end{claim*}
    \begin{proof}[Proof of the claim]
        For $r = 0$, the preceding claim implies that for every $k \in \N$, we have $h_{x,kL}^0 = (\ReLU \circ \nn)^k(x) = (\ReLU \circ \nn)^{k+1}(x) \kroen{0}{0}$. Now, suppose $r \in \{1,\ldots,L-1\}$. Then, by the preceding claim,
        \begin{equation}
            h_{x,kL+r}^0 = h_{(\ReLU \circ \nn)^{k+1}(x),r}^0 = (\ReLU \circ \nn)_0((\ReLU \circ \nn)^{k+1}(x)) \kroen{r}{0} = 0 = (\ReLU \circ \nn)^{k+1}(x) \kroen{r}{0}.
        \end{equation}
    \end{proof}
    Now, let $t \in \No$. Note that $t = (t//L) L + (t \bmod L)$, so that by the preceding claim, we have
    \begin{equation}
        \proj_{1:n} h_{x,t} = h_{x,t}^0 = h_{x,(t//L) L + (t \bmod L)}^0 = (\ReLU \circ \nn)^{t//L}(x) \kroen{t \bmod L}{0}.
    \end{equation}
    This concludes the proof.
\end{proof}

We now show how to modify the RNN constructed in Lemma \ref{lem:simulation-of-cpwl-function-iterations-with-an-rnn:app} to output the positive part of the iterates of $\nn$ at every time step, and not only at time steps that are multiples of $L$. The next lemma corresponds to this second step of the construction.

The following stabilization step keeps the last completed iterate available between two multiples of the depth.
\begin{lemma}\label{lem:simulation-of-cpwl-function-iterations-with-an-rnn:stabilisation}
    Let $\nn \in \nns{n}{n}$ be a ReLU neural network, and let $L := \nnDepth{\nn}$. Then, there exists $\tilde m \in \N$ and $\tilde A_\nn \in \R^{\tilde m \times \tilde m}$ satisfying $ (L+2)n+L \leq \tilde m \leq (2L+2)(\nnWidth{\nn}+1)$ and $\nnWeights{\tilde A_\nn} = \nnWeights{\nn} \cup \{-1,0,1\}$, such that for every $x \in \R^n$, the sequence $(h_t \in \R^{\tilde m})_{t \in \N}$ defined by
    \begin{equation}\label{eq:simulation-of-cpwl-function-iterations-with-an-rnn:stabilisation:recursive-definition}
        h_0 := (0_{n(L+1)},x, 0, \oneHot{L}{1}), \quad h_{t+1} := \ReLU(\tilde A_\nn h_t) \text{ for every } t \in \No
    \end{equation}
    satisfies
    \begin{equation}
        \proj_{1:n} h_t = (\ReLU \circ \nn)^{(t-1)//L}(x), \quad t \in \N,
    \end{equation}
    where $//$ denotes the integer division.
\end{lemma}
\begin{proof}
    Let $m \in \N$ and $A_\nn \in \R^{m \times m}$ be as in Lemma \ref{lem:simulation-of-cpwl-function-iterations-with-an-rnn}. We define 
    \[
        \tilde A_\nn := \begin{bmatrix}
            I_n & 0 & -I_n & \begin{array}{cc} I_n & 0 \end{array}\\
            0 & 0 & 0 & \begin{array}{cc} I_n & 0 \end{array}\\
            0 & I_{n(L-1)} & 0 & 0\\
            0 & 0 & 0 & A_\nn
        \end{bmatrix} \in \R^{\tilde m \times \tilde m}.
    \]
    Note that, indeed, we have \[(L+2)n + L \leq (L+1)n + m = \tilde m \leq (L+1)n + (\nnWidth{\nn}+1)L \leq (L+2)(\nnWidth{\nn}+1),\] and that $\nnWeights{\tilde A_\nn} = \nnWeights{A_\nn} \cup \{-1,0,1\} = \nnWeights{\nn} \cup \{-1,0,1\}$. Now, let $x \in \R^n$ and let $(h_t \in \R^m)_{t \in \No}$ be defined as in \eqref{eq:simulation-of-cpwl-function-iterations-with-an-rnn:stabilisation:recursive-definition}. We split the sequence $(h_t)_{t \in \N}$ into subsequences $(h_t^* \in \R^n)_{t \in \No}$, $(h_t^\ell \in \R^n)_{t \in \No}$ for $\ell \in \{0, \ldots, L-1\}$, and $(\tilde h_t \in \R^m)_{t \in \No}$ such that
    \[
        h_t = (h_t^*, h_t^0, \ldots, h_t^{L-1}, \tilde h_t) \text{ for every } t \in \No.
    \] Note that, in particular,
    \begin{equation}\label{eq:simulation-of-cpwl-function-iterations-with-an-rnn:stabilisation:recursive-definition:base-case}
                h_0^* = 0, \quad h_0^0 = 0, \quad h_0^1 = 0, \quad \ldots, \quad h_0^{L-1} = 0, \quad \tilde h_0 = (x, 0, \oneHot{L}{1}),
    \end{equation}
    and that
    \begin{equation}\label{eq:simulation-of-cpwl-function-iterations-with-an-rnn:stabilisation:recursive-definition:iterative-step}
                h_{t+1}^* = \ReLU(h_t^* - h_t^0 + \proj_{1:n}\tilde h_t), \quad h_{t+1}^0 = \ReLU(\proj_{1:n} \tilde h_t), \quad h_{t+1}^\ell = \ReLU(h_t^{\ell-1}), \quad \text{and} \quad \tilde h_{t+1} = \ReLU(A_\nn \tilde h_t),
    \end{equation}
    for every $t \in \No$ and $\ell \in \{1, \ldots, L-1\}$. In particular, by Lemma \ref{lem:simulation-of-cpwl-function-iterations-with-an-rnn:app}, we have 
    \begin{equation}\label{eq:simulation-of-cpwl-function-iterations-with-an-rnn:stabilisation:tilde h t value}
        \proj_{1:n} \tilde h_t = (\ReLU \circ \nn)^{t//L}(x) \kroen{t \bmod L}{0} \geq 0 \text{ for every } t \in \N.
    \end{equation}

    We divide the proof into several claims.
    \begin{claim*}
        For every $\ell \in \{0, \ldots, L-1\}$, we have
        \[
            h_t^\ell = \begin{cases}
                \proj_{1:n} \tilde h_{t-\ell-1} & \text{if } t > \ell,\\
                0 & \text{if } t \leq \ell.
            \end{cases}
        \]
    \end{claim*} 
    \begin{proof}
        We first treat the case $\ell = 0$. Note that by \eqref{eq:simulation-of-cpwl-function-iterations-with-an-rnn:stabilisation:recursive-definition:base-case}, we have $h_0^0 = 0$, and that by \eqref{eq:simulation-of-cpwl-function-iterations-with-an-rnn:stabilisation:recursive-definition:iterative-step}, we have 
        \begin{equation}\label{eq:simulation-of-cpwl-function-iterations-with-an-rnn:stabilisation:tilde h t value-2}
            h_t^0 = \ReLU(\proj_{1:n} \tilde h_{t-1}) = \proj_{1:n} \tilde h_{t-1},
        \end{equation}
        for every $t > 0$, where the second equality follows $\proj_{1:n} \tilde h_{t-1} \leq 0$, which is given by \eqref{eq:simulation-of-cpwl-function-iterations-with-an-rnn:stabilisation:tilde h t value}. 
        Now, $\ell \in \{1, \ldots, L-1\}$. First, let $t \in \{0, \ldots, \ell\}$. Then, by \eqref{eq:simulation-of-cpwl-function-iterations-with-an-rnn:stabilisation:recursive-definition:iterative-step} applied $t$ times, we have
        \[
            h_t^\ell = h_{t-t}^{\ell-t} = h_0^{\ell-t} \overset{(a)}= 0,
        \]
        where (a) follows from \eqref{eq:simulation-of-cpwl-function-iterations-with-an-rnn:stabilisation:recursive-definition:base-case}. Now, let $t > \ell$. Then, by \eqref{eq:simulation-of-cpwl-function-iterations-with-an-rnn:stabilisation:recursive-definition:iterative-step} applied $\ell$ times, we have
        \[
            h_t^\ell = h_{t-\ell}^{\ell - \ell} = h_{t-\ell}^0 \overset{(a)}= \proj_{1:n} \tilde h_{t-\ell-1}
        \]
        where (a) follows from \eqref{eq:simulation-of-cpwl-function-iterations-with-an-rnn:stabilisation:tilde h t value-2}. This concludes the proof of the claim.
    \end{proof}
    \begin{claim*}
        \begin{equation}\label{eq:simulation-of-cpwl-function-iterations-with-an-rnn:stabilisation:h t star value}
            h_t^* = \begin{cases}
                \sum_{j=0}^{t-1} \proj_{1:n} \tilde h_j & \text{if } t \leq L,\\
                \sum_{j=t-L}^{t-1} \proj_{1:n} \tilde h_j & \text{if } t > L.
            \end{cases}
        \end{equation}
    \end{claim*}
    \begin{proof}
        We make the proof by induction on $t$. For the base case, note that by \eqref{eq:simulation-of-cpwl-function-iterations-with-an-rnn:stabilisation:recursive-definition:base-case}, we have $h_0^* = 0$, so that \eqref{eq:simulation-of-cpwl-function-iterations-with-an-rnn:stabilisation:h t star value} holds. Now, let $t \in \No$ and assume that \eqref{eq:simulation-of-cpwl-function-iterations-with-an-rnn:stabilisation:h t star value} holds. Note that in particular, we have $h_t^\ast \geq 0$. First, assume that $t \in \{0, \ldots, L-1\}$. Then, by \eqref{eq:simulation-of-cpwl-function-iterations-with-an-rnn:stabilisation:recursive-definition:iterative-step}, we have
        \begin{equation}\label{eq:simulation-of-cpwl-function-iterations-with-an-rnn:stabilisation:h t star value-2}
            h_{t+1}^* = \ReLU(h_t^* - h_t^0 + \proj_{1:n}\tilde h_t) \overset{(a)}= \ReLU(h_t^* + \proj_{1:n}\tilde h_t) \overset{(b)}= h_t^* + \proj_{1:n}\tilde h_t,
        \end{equation}
        where (a) follows from the preceding claim and (b) is by $h_t^* \geq 0$ and $\proj_{1:n}\tilde h_t \geq 0$, which is given by \eqref{eq:simulation-of-cpwl-function-iterations-with-an-rnn:stabilisation:tilde h t value}. Therefore, be the induction hypothesis, we have
        \[
            h_{t+1}^* = h_t^* + \proj_{1:n}\tilde h_t = \sum_{j=0}^{t-1} \proj_{1:n} \tilde h_j + \proj_{1:n}\tilde h_t = \sum_{j=0}^{t} \proj_{1:n} \tilde h_j,
        \]
        so that \eqref{eq:simulation-of-cpwl-function-iterations-with-an-rnn:stabilisation:h t star value} holds for $t+1$. Now, assume that $t \geq L$. Then, by \eqref{eq:simulation-of-cpwl-function-iterations-with-an-rnn:stabilisation:recursive-definition:iterative-step}, we have
        \begin{equation}\label{eq:simulation-of-cpwl-function-iterations-with-an-rnn:stabilisation:h t star value-3}
            h_{t+1}^* = \ReLU(h_t^* - h_t^0 + \proj_{1:n}\tilde h_t) \overset{(a)}= \ReLU(h_t^* - \proj_{1:n} \tilde h_{t-L} + \proj_{1:n}\tilde h_t) \overset{(b)}= h_t^* - \proj_{1:n} \tilde h_{t-L} + \proj_{1:n}\tilde h_t,
        \end{equation}
        where (a) follows from the preceding claim and (b) is by $h_t^* \geq 0$, $\proj_{1:n}\tilde h_t \geq 0$, and $\proj_{1:n} \tilde h_{t-L} \leq 0$, which is given by \eqref{eq:simulation-of-cpwl-function-iterations-with-an-rnn:stabilisation:tilde h t value}. Therefore, by the induction hypothesis, we have
        \[
            h_{t+1}^* = h_t^* - \proj_{1:n} \tilde h_{t-L} + \proj_{1:n}\tilde h_t = \sum_{j=t-L}^{t-1} \proj_{1:n} \tilde h_j - \proj_{1:n} \tilde h_{t-L} + \proj_{1:n}\tilde h_t = \sum_{j=t+1-L}^{t} \proj_{1:n} \tilde h_j,
        \]
        so that \eqref{eq:simulation-of-cpwl-function-iterations-with-an-rnn:stabilisation:h t star value} holds for $t+1$. This concludes the proof of the claim.
    \end{proof}
    Now, let $t \in \N$. Note that the preceding claim and \eqref{eq:simulation-of-cpwl-function-iterations-with-an-rnn:stabilisation:tilde h t value} can be reformulated as
    \[
        h_t^* = \sum_{j=(t-L)\vee 0}^{t-1} \proj_{1:n} \tilde h_j \overset{\eqref{eq:simulation-of-cpwl-function-iterations-with-an-rnn:stabilisation:tilde h t value}}= \sum_{j=(t-L)\vee 0}^{t-1} (\ReLU \circ \nn)^{j//L}(x) \kroen{j \bmod L}{0}.
    \]
    Note that since the set $S_t := \{(t-L)\vee 0, \ldots, t-1\}$ has at most $L$ elements, there exists at most one element $j \in S_t$ such that $j \bmod L = 0$. Such an element is given by $j = ((t-1)//L) L$. Therefore, we have
    \[
        h_t^* = (\ReLU \circ \nn)^{(((t-1)//L) L)//L}(x) \kroen{(((t-1)//L) L) \bmod L}{0} = (\ReLU \circ \nn)^{(t-1)//L}(x).
    \]
    Finally,
    \[
        \proj_{1:n} h_t = h_t^* = (\ReLU \circ \nn)^{(t-1)//L}(x).
    \]
    This concludes the proof.
\end{proof}

The following technical lemma allows us to split a neural network into its positive and negative parts, which will be useful in the proof of the final simulation result.

This splitting lets the RNN recover signed iterates while still using ReLU states.
\begin{lemma}\label{lem:splitting-nn-into-positive-and-negative-parts}
    Let $\nn \in \nns{n}{n}$ be a ReLU neural network. Then, there exists a neural network $\nn'\in \nns{2n}{2n}$ satisfying $\nnDepth{\nn'} = \nnDepth{\nn}$, $\nnWidth{\nn'} = \max\{\nnWidth{\nn},2n\}$, and $\nnWeights{\nn'} = \pm\nnWeights{\nn}$ such that for every $x \in \R^n$, we have
    \[
        \nn'(x, 0) = \nn'(\ReLU(x), \ReLU(-x)) = (\nn(x), -\nn(x)).
    \]
\end{lemma}
\begin{proof}
    Let $\nn := (N_0, \ldots, N_L, A_1, \ldots, b_L)$, with $N_0 = N_L =: n$. We define $\nn' := (N_0', \ldots, N_L', A_1', \ldots, b_L')$ by $N_0 = N_L := 2n$,
    $N_\ell' := N_\ell$ for every $\ell \in \{1, \ldots, L-1\}$, $A_\ell' = A_\ell$ for every $\ell \in \{2, \ldots, L-1\}$, $b_\ell' := b_\ell$ for every $\ell \in \{1, \ldots, L-1\}$,
    \[
        A_1' := \begin{bmatrix}
        A_1 & -A_1
    \end{bmatrix}, \quad A_L' := \begin{bmatrix}
        A_L\\
        -A_L
    \end{bmatrix}, \quad \text{and} \quad b_L' := \begin{bmatrix}
        b_L\\
        -b_L
    \end{bmatrix}.
    \] 
    Note that, indeed, we have $\nnDepth{\nn'} = \nnDepth{\nn}$, $\nnWidth{\nn'} = \max\{2n, \max_{\ell \in \{1, \ldots, L-1\}} N_\ell\} = \max\{\nnWidth{\nn},2n\}$, and $\nnWeights{\nn'} = \pm\nnWeights{\nn}$. Now, let $x \in \R^n$. Note that for every $\ell \in \{1, \ldots, L-1\}$, we have
    \[
        \nn_1(x,0) = A_1' (x,0) + b_1' = A_1 x + b_1 = \nn_1(x),
    \]
    and
    \[
        \nn_1(\ReLU(x), \ReLU(-x)) = A_1 (\ReLU(x) - \ReLU(-x)) + b_1 = A_1 x + b_1 = \nn_1(x).
    \]
    Therefore, we can show by induction that 
    \[
        \nn_{L-1}(x,0) = \nn_{L-1}(\ReLU(x), \ReLU(-x)) = \nn_{L-1}(x).
    \]
    Finally, we have
    \[
        \nn_L(x,0) = \nn_L(\ReLU(x), \ReLU(-x)) = A_L' (x,0) + b_L' = (A_L x + b_L, -A_L x - b_L) = (\nn_L(x), -\nn_L(x)).
    \]
\end{proof}

We finally show how to modify the RNN constructed in Lemma \ref{lem:simulation-of-cpwl-function-iterations-with-an-rnn:stabilisation} to output the iterates of $\nn$ at every time step, and not only at time steps that are multiples of $L$. The next lemma corresponds to this third and final step of the construction. We deliberately make an RNN that takes as input a vector of the form $(0_{2n(L+1)},x,-x,0,\oneHot{L}{1})$ instead of plain $x$ to avoid having to use negative weights in the output layer of the RNN, which would not be allowed in a left-selector RNN, and will compromise the final simulation result. However, we provide a version of the final simulation result in which the RNN takes as input plain $x$ at the end of this section, see Lemma \ref{thm:simulation-of-cpwl-function-iterations-with-an-rnn:final-simulation-plain-input}.

The next result is the signed-iterate simulation in the left-selector form needed later.
\begin{lemma}\label{lem:simulation-of-cpwl-function-iterations-with-an-rnn:final-simulation}
    Let $\nn \in \nns{n}{n}$ be a ReLU neural network. Then, there exists a left-selector RNN $\Rc$ satisfying $\rnnHidDim{\Rc} \leq (2\nnDepth{\nn}+2)(2n \vee \nnWidth{\nn}+1)$ and $\nnWeights{\Rc} = \pm\nnWeights{\nn} \cup \pm\{0,1\}$ such that for every $x \in \R^n$ and $t \in \N$, we have 
    \[\Rc \Dc y[t] = \nn^{t//L}(x),\]
    where $y := (0_{2n(L+1)},x,-x, 0, \oneHot{L}{1}) \in \R^{\rnnHidDim{\Rc}}$.
\end{lemma}
\begin{proof}
    Let $\nn' : \R^{2n} \to \R^{2n}$ be as in the preceding lemma, and let $\tilde m \in \N$ $\tilde A_{\nn'} \in \R^{\tilde m \times \tilde m}$ be as in Lemma \ref{lem:simulation-of-cpwl-function-iterations-with-an-rnn:stabilisation} applied to $\nn'$. We define $\Rc := (\tilde m, \tilde m, n; \tilde A_{\nn'}, 0, I_{\tilde m}, A_o, 0)$, where
    \[
        A_o = \proj_{1:n} - \proj_{n+1:2n} = \begin{bmatrix}
            I_n & -I_n & 0
        \end{bmatrix} \in \R^{n \times \tilde m}.
    \]
    Note that, indeed, we have $\rnnHidDim{\Rc} = \tilde m \leq (2\nnDepth{\nn'}+2)(\nnWidth{\nn'}+1) = (2\nnDepth{\nn}+2)(2n \vee \nnWidth{\nn}+1)$ and $\nnWeights{\Rc} = \nnWeights{\tilde A_{\nn'}}\cup\{0,1\} = \nnWeights{\nn'} \cup\pm\{0,1\}\subseteq \pm\nnWeights{\nn} \cup\pm\{0,1\}$. Now, let $x \in \R^n$, $y := (0_{2n(L+1)},x, -x, 0, \oneHot{L}{1}) \in \R^{\rnnHidDim{\Rc}}$, and consider the sequence $(h_t \in \R^{\tilde m})_{t \in \No}$ defined by $h_0 := \ReLU(y)$ and $h_{t+1} := \ReLU(\tilde A_{\nn'} h_t)$ for every $t \in \No$. Note that by Lemma \ref{lem:simulation-of-cpwl-function-iterations-with-an-rnn:stabilisation}, we have $\proj_{1:2n} h_t = (\ReLU \circ \nn')^{(t-1)//L}(\ReLU(x,-x))$. Therefore, for every $t \in \N$, we have
    \begin{align*}
        A_o h_t &= \proj_{1:n} h_t - \proj_{n+1:2n} h_t = \proj_{1:n}(\ReLU \circ \nn')^{(t-1)//L}(\ReLU(x,-x)) - \proj_{n+1:2n}(\ReLU \circ \nn')^{(t-1)//L}(\ReLU(x,-x))\\
        & \overset{(a)} = \proj_{1:n}(\ReLU \circ \nn)^{(t-1)//L}(x) - \proj_{n+1:2n}(\ReLU \circ (-\nn))^{(t-1)//L}(x) = \nn^{(t-1)//L}(x),
    \end{align*}
    where (a) follows from Lemma \ref{lem:splitting-nn-into-positive-and-negative-parts}. We now show that for every $t \in \N$, we have $\Hc \Dc y[t] = h_{t+1}$. We make the proof by induction on $t$. For the base case, note that by Definition \ref{def:elman_rnn}, we have
    \[
        \Hc \Dc y[0] = \ReLU(I_{\tilde m}y) = \ReLU(y) = h_0.
    \]
    Now, for the induction step, let $t \in \No$ and assume that $\Hc \Dc y[t] = h_{t+1}$. Then, by Definition \ref{def:elman_rnn}, we have
    \[
        \Hc \Dc y[t+1] = \ReLU(\tilde A_{\nn'} h_{t+1} + I_{\tilde m} \Dc y[t+1]) = \ReLU(\tilde A_{\nn'} h_{t+1}) = h_{t+2}.
    \]
    Therefore, for every $t \in \N$, we have $\Hc \Dc y[t] = h_{t+1}$, so that
    \[\Rc \Dc y[t] = A_o h_{t+1} = \nn^{t//L}(x).\]
    This concludes the proof.
\end{proof}

We close this part with the following Theorem, that modifies the RNN constructed in Lemma \ref{lem:simulation-of-cpwl-function-iterations-with-an-rnn:final-simulation} to take as input plain $x$ instead of a vector of the form $(0_{2n(L+1)},x,-x,0,\oneHot{L}{1})$. This will not be used directly in the proof of the main \TMNU simulation result, because the constructed RNN is not a left-selector, but is informative as a standalone result.

This gives the same iteration simulation in the more familiar plain-input format.
\begin{theorem}\label{thm:simulation-of-cpwl-function-iterations-with-an-rnn:final-simulation-plain-input}
    Let $\nn \in \nns{n}{n}$ be a ReLU neural network. Then, there exists an RNN $\Rc$ satisfying $\rnnHidDim{\Rc} \leq (2\nnDepth{\nn}+2)(2n \vee \nnWidth{\nn}+1) + 2n + L + 1$ and $\nnWeights{\Rc} = \pm\nnWeights{\nn} \cup \pm\{0,1\}$ such that for every $x \in \R^n$ and $t \in \N$, we have 
    \[\Rc \Dc x[t] = \nn^{t//L}(x).\]
\end{theorem}
\begin{proof}
    Let $\Rc' = (\tilde m, \tilde m, n; \tilde A_{\nn'}, 0, I_{\tilde m}, A_o, 0)$ be as in the preceding lemma, such that for every $x \in \R^n$, we have $\Rc' \Dc (0_{2n(L+1)},x,-x,0,\oneHot{L}{1})[t] = \nn^{t//L}(x)$. Note that for every $x \in \R^n$, we have
    \[(0_{2n(L+1)},x,-x,0,\oneHot{L}{1}) = A x + b,\]
    where $A \in \R^{\tilde m \times n}, b \in \R^{\tilde m}$. Therefore, by Lemma \ref{lem:affine-transform-of-the-input-of-an-rnn}, there exists an RNN $\Rc$ satisfying $\rnnHidDim{\Rc} = \tilde m + n + L + 1 \leq (2\nnDepth{\nn}+2)(2n \vee \nnWidth{\nn}+1) + 2n + L + 1$ and $\nnWeights{\Rc} = \nnWeights{\Rc'} \cup \pm\{0,1\} = \pm\nnWeights{\nn} \cup \pm\{0,1\}$ such that for every $x \in \R^n$ and $t \in \N$, we have
    \[\Rc \Dc x[t] = \Rc' \Dc (A x + b)[t] = \nn^{t//L}(x).\]
    This concludes the proof.
\end{proof}

\subsubsection{Construction of an RNN that simulates a \TMNU}

In this part, we apply the preceding results to show that we can simulate the iteration of a \TMNU by an RNN. The next lemma corresponds to the first step of the construction, in which we show that we can simulate the iteration of a \TMNU by the iteration of a neural network, and then apply Lemma \ref{lem:simulation-of-cpwl-function-iterations-with-an-rnn:final-simulation} to show that we can simulate the iteration of a \TMNU by an RNN.

We now obtain the advertised RNN simulation theorem for bounded \TMNU trajectories.
\begin{theorem}\label{thm:simulation-of-a-tmnu-by-an-rnn:app}
    Let $\tmnuM$ be a \TMNU and $u \in \{0,1\}^\N$ such that $\tmnuM$ has uniformly $C$-bounded trajectories at $u$. Then, there exists an RNN $\Rc$ satisfying 
    \[
        \rnnHidDim{\Rc} \leq 10 (\max\{3\nStates, \nStates + \nFunctions{\tmnuM} + 9\} + 2\neurDim\nFunctions{\tmnuM} + 19),
    \]
    and 
    \[\nnWeights{\Rc} = \{0,1/4,1,2,3,4,\|\tmnuM\|_C\} \cup \pm \nnWeights{\tmnuM} \cup \pm\{\cantorMap(u)\},\]
    such that
    \begin{equation}
        \Rc \Dc x[t+1] = \tmnuM^ux[t//4], \quad t \in \No, \quad x \in [-1,1].
    \end{equation}
\end{theorem}
\begin{proof}
    By Theorem \ref{thm:transition-function-as-neural-network:appendix}, there exists a ReLU neural network $\nn\in \nns{\nStates+2+\neurDim}{\nStates+2+\neurDim}$ satisfying $\nnDepth{\nn} = 4$, $\nnWidth{\nn} \leq \max\{3\nStates, \nStates + \nFunctions{\tmnuM} + 9\} + 2\neurDim\nFunctions{\tmnuM} + 18$, $\nnWeights{\nn} \subseteq \pm \{0,1/4,1,2,3,4,\|\tmnuM\|_C\} \cup \pm \nnWeights{\tmnuM}$, such that for every $c \in \Bc_\tmnuM(C)$,
    \begin{equation}
        \nn_{\tmnuM,C}(\configMap{\tmnuM}(c)) = \configMap{\tmnuM}(\tmnuM(c)).
    \end{equation}
    Then, by Lemma \ref{lem:simulation-of-cpwl-function-iterations-with-an-rnn:final-simulation}, there exists a left-selector RNN $\Rc$ satisfying 
    \begin{align*}
        \rnnHidDim{\Rc} &\leq (2\nnDepth{\nn_{\tmnuM,C}}+2)((2(\nStates+2+\neurDim)) \vee \nnWidth{\nn_{\tmnuM,C}}+1)\\
        & \leq 10 (\max\{3\nStates, \nStates + \nFunctions{\tmnuM} + 9\} + 2\neurDim\nFunctions{\tmnuM} + 19)
    \end{align*}
    and $\nnWeights{\Rc} = \pm\nnWeights{\nn} \cup \{0,1\} \subseteq \pm \{0,1/4,1,2,3,4,\|\tmnuM\|_C\} \cup \pm \nnWeights{\tmnuM}$, such that for every $z \in \R^n$,
    \begin{equation}
        \Rc \Dc y_z[t] = \nn^{t//4}(z), \quad t \in \N,
    \end{equation}
    where $y_z := (0_{2N(L+1)},z,-z, 0, \oneHot{L}{1}) \in \R^{\rnnHidDim{\Rc}}$, $N := \nStates+2+\neurDim$, and $L := \nnDepth{\nn} = 4$. Now, fix $u \in \{0,1\}^\N$ and assume that $\tmnuM$ has uniformly $C$-bounded trajectories at $u$. Then, for every $x \in [-1,1]$, we have $c_x := (1;|u;x,0,\ldots,0) \in \Bc_\tmnuM(C)$, so that for every $t \in \N$, we have 
    \begin{equation}
        \Rc \Dc y_{\configMap{\tmnuM}(c_x)}[t] = \nn^{t//4}(\configMap{\tmnuM}(c_x)) = \configMap{\tmnuM}(\tmnuM^{t//4}(c_x)).
    \end{equation}
    Now, note that for every $x \in [-1,1]$,
    \begin{align*}
        y_{\configMap{\tmnuM}(c_x)} &= (0_{2N(L+1)},\configMap{\tmnuM}(c_x),-\configMap{\tmnuM}(c_x), 0, \oneHot{L}{1})\\
        &= (0_{2N(L+1)},\oneHot{\nStates}{1},\cantorMap(u),x, 0_{\neurDim-1}, -\oneHot{\nStates}{1},-\cantorMap(u),-x, 0_{\neurDim-1}, 0, \oneHot{L}{1})\\
        & = (0_{2N(L+1)}, 0_{\nStates + 2}, 1, 0_{\neurDim+\nStates + 1}, -1, 0) x + (0_{2N(L+1)},\oneHot{\nStates}{1},\cantorMap(u),0_d, -\oneHot{\nStates}{1},-\cantorMap(u), 0, \oneHot{L}{1})\\
        & =: A x + b.
    \end{align*}
    Then, by Lemma \ref{lem:affine-transform-of-the-input-of-an-rnn}, there exists an RNN $\tilde \Rc$ satisfying $\rnnHidDim{\tilde \Rc} = \rnnHidDim{ \Rc} +1 $ and
    \[
        \nnWeights{\tilde \Rc} = \nnWeights{\Rc} \cup \nnWeights{A} \cup \pm\nnWeights{b} \subseteq \pm\{\cantorMap(u)\} \subseteq \{0,1/4,1,2,3,4,\|\tmnuM\|_C\} \cup \pm \nnWeights{\tmnuM} \cup \pm\{\cantorMap(u)\},
    \]
    such that for every $x \in [-1,1]$ and $t \in \N$, we have $\tilde \Rc \Dc x[t] = \Rc \Dc y_{\configMap{\tmnuM}(c_x)}[t]$. Hence, for every $x \in [-1,1]$ and $t \in \N$, we have
    \begin{equation}        \tilde \Rc \Dc x[t] = \Rc \Dc y_{\configMap{\tmnuM}(c_x)}[t] = \configMap{\tmnuM}(\tmnuM^{t//4}(c_x)).
    \end{equation}
    Moreover, for every $x \in [-1,1]$ and $t \in \N$, we have
    \begin{equation}
        \tmnuM^u x[t] = \proj_{\neurDim} \proj_\neurState \tmnuM^{t}(c_x) = \proj_{\nStates + 2 + \neurDim} \configMap{\tmnuM}(\tmnuM^{t}(c_x)) =: A' \configMap{\tmnuM}(\tmnuM^{t}(c_x)),
    \end{equation}
    where $A' := \proj_{\nStates + 2 + \neurDim}$ is a left-selector matrix. Hence, by Lemma \ref{lem:linear-transform-of-the-output-of-an-rnn}, there exists an RNN $\Rc'$ satisfying $\rnnHidDim{\Rc'} = \rnnHidDim{\tilde \Rc}$ and $\nnWeights{\Rc'} \subseteq \nnWeights{\tilde \Rc} \cup \{0\} \subseteq \{0,1/4,1,2,3,4,\|\tmnuM\|_C\} \cup \pm \nnWeights{\tmnuM} \cup \pm\{\cantorMap(u)\}$ such that for every $x \in [-1,1]$ and $t \in \N$, we have
    \begin{equation}
        \Rc' \Dc x[t] = A' (\tilde \Rc \Dc x[t]) = A' \configMap{\tmnuM}(\tmnuM^{t//4}(c_x)) = \tmnuM^u x[t//4].
    \end{equation}
    This concludes the proof.
\end{proof}

\section{Detailed TMNU constructions}

\label{sec:tmnu-construction-appendix}

In this section, we build a \TMNU that can approximate any continuous function $f : [-1,1] \to \R$, in the sense of paradigm \eqref{eq:paradigm4}. The construction is made progressively, by first designing simple \TMNUs that can approximate some simple functions, and then by using these simple \TMNUs as subroutines to design more complex \TMNUs that can approximate more complex functions, until we reach the point where we can design a \TMNU that can approximate any continuous function $f : [-1,1] \to \R$. Throughout this section, we will specify the number of states and neural dimension of the \TMNUs explicitly, and the transition and command functions implicitly, by describing what the \TMNU does in each state and for each symbol read by the head. Sometimes, we will not specify the transition and command functions completely, but only partially, by describing what the \TMNU does in some states and for some symbols read by the head, and leaving the rest of the transition and command functions unspecified. For such unspecified state-symbol pairs, we will assume that the \TMNU simply transitions to the halting state and does not update the tape or the neural state, i.e., that $\transFunc(\state, \symb) = (\nStates, \symb, 0)$ and $\commFunc(\state, \symb) = \id$ for every unspecified state-symbol pair $(\state, \symb)$. Moreover, for some state-symbol pairs, we will only specify the actions that modify the tape and the neural state. Specifically, if for some state symbol pair $(\state, \symb)$, the command function applies the identity, i.e., does not update the neural state, then we will simply not specify the command function for this state-symbol pair, and we will assume that $\commFunc(\state, \symb) = \id$. Similarly, if for some state-symbol pair $(\state, \symb)$, the head does not move right or left, we will simply not specify the transition function for this state-symbol pair, and we will assume that $\transFunc(\state, \symb) = (\state', \symb', 0)$ for some $\state' \in \{1, \ldots,\nStates\}$ and $\symb' \in \workSymbols$. This way of describing the \TMNUs is more intuitive, and is sufficient to specify \TMNUs without ambiguity. 

Once a \TMNU is specified, we derive some properties of its evolution. Specifically, we will be interested to show that given some pair of configurations $\config, \config'$ of a \TMNU $\tmnuM$, there exists some $t \in \No$, $t \geq 1$ such that $\tmnuM^t(\config) = \config'$, and whether the computation remains bounded during the computation, i.e., whether
\begin{equation}
    \|\config\|_\tmnuM^t := \max_{0 \leq s \leq t} \|\tmnuM^s(\config)\| \leq C
\end{equation}
for some $C > 0$.

The next Lemma establishes that in order to study whether $\tmnuM^t(\config) = \config'$, and $\|\config\|_\tmnuM^t \leq C$, we can split the evolution from $\config$ to $\config'$ into several sub-evolutions, and study each of these sub-evolutions separately. This is a very useful property, because it allows us to design \TMNUs in a modular way, by designing subroutines that can be used as building blocks to design more complex \TMNUs.
This bookkeeping lemma is the basic concatenation rule used throughout the construction.
\begin{lemma}(Chaining)\label{lem:chaining}
    Let $\tmnuM$ be a \TMNU, $\config_1, \config_2, \config_3 \in \Cc_\tmnuM$, $t_1,t_2 \in \N$, and $C_1,C_2 > 0$ such that
    \begin{equation}        \tmnuM^{t_1}(\config_1) = \config_2, \quad \tmnuM^{t_2}(\config_2) = \config_3, \quad \|\config_1\|_\tmnuM^{t_1} \leq C_1, \quad \|\config_2\|_\tmnuM^{t_2} \leq C_2.
    \end{equation}
    Then,
    \begin{equation}
        \tmnuM^{t_1+t_2}(\config_1) = \config_3, \quad \text{and} \quad \|\config_1\|_\tmnuM^{t_1+t_2} \leq C_1 \vee C_2.
    \end{equation}
\end{lemma}
\begin{proof}
    Since $\tmnuM^{t_1}(\config_1) = \config_2$, we have $\tmnuM^{t_1+t_2}(\config_1) = \tmnuM^{t_2}(\tmnuM^{t_1}(\config_1)) = \tmnuM^{t_2}(\config_2) = \config_3$. Moreover, since $\config_1 \in \Bc_\tmnuM^{t_1}(C_1)$ and $\config_2 \in \Bc_\tmnuM^{t_2}(C_2)$, we have $\|\tmnuM^s(\config_1)\|\leq C_1 \leq C_1 \vee C_2$ for every $s = 0, \ldots, t_1$, and $\|\tmnuM^s(\config_2)\| \leq C_2$ for every $s = 0, \ldots, t_2$. Therefore, for every $s = t_1+1, \ldots, t_1+t_2$, we have
    \begin{equation}
        \|\tmnuM^s(\config_1)\| = \|\tmnuM^{s-t_1}(\tmnuM^{t_1}(\config_1))\| = \|\tmnuM^{s-t_1}(\config_2)\| \leq C_2 \leq C_1 \vee C_2.
    \end{equation}
\end{proof}

\subsection{Subroutines for TMNU constructions}

In this section, we introduce the notion of subroutine for \TMNUs, that will allow us to design more complex \TMNUs by combining simpler \TMNUs as building blocks. 
The idea is that a \TMNU $\tmnuM$ can be designed in such a way that, when it is in some subset of its states, it behaves exactly as some other \TMNU $\tmnuN$, on a subset of its neural dimensions, while leaving the remaining neural dimensions unchanged. For example, assume that the \TMNU $\tmnuM$ has 5 states and 3 neural dimensions, and the \TMNU $\tmnuN$ has 3 states and 2 neural dimensions. Then, we will say that $\tmnuN$ is a subroutine of $\tmnuM$ with state correspondence $(2,3,4)$ and neural dimension correspondence $(1,2)$, or in condensed form, with correspondence $(2,3,4;1,2)$, if in state 2 and 3, the \TMNU $\tmnuM$ updates its tape and neural dimensions 1 and 2 exactly as $\tmnuN$ would in states 1 and 2, respectively, and that when $\tmnuN$ halts, i.e., reaches state 3, $\tmnuM$ transitions to state 4. Such a behavior can be expressed formally by imposing some relation between the transition and command functions of both \TMNUs. 
The following definition makes this relation precise.
\begin{definition}(Subroutine)\label{def:subroutine}
    Let $\tmnuM := (\nStates_\tmnuM, \neurDim_\tmnuM, \transFunc_\tmnuM, \commFunc_\tmnuM)$ and $\tmnuN := (\nStates_\tmnuN, \neurDim_\tmnuN, \transFunc_\tmnuN, \commFunc_\tmnuN)$ be two \TMNUs, such that $\nStates_\tmnuN \leq \nStates_\tmnuM$ and $\neurDim_\tmnuN \leq \neurDim_\tmnuM$, and let $\indexFuncState : \{1,\ldots,\nStates_\tmnuN\} \hookrightarrow \{1,\ldots,\nStates_\tmnuM\}$ and $\indexFuncNeur : \{1,\ldots,\neurDim_\tmnuN\} \hookrightarrow \{1,\ldots,\neurDim_\tmnuM\}$ be two injective functions. We say that $\tmnuN$ is a \newname{subroutine of $\tmnuM$ with correspondance} $(\indexFuncState, \indexFuncNeur)$ if for every $\state \in \{1,\ldots,\nStates_\tmnuN-1\}$, $\symb \in \workSymbols$ and $\neurState \in \R^{\neurDim_\tmnuM}$, we have 
    \begin{equation}\label{eq:subroutine-conditions-1}
        \transFunc_\tmnuM(\indexFuncState(\state), \symb) = (\indexFuncState \circ \transFuncState_\tmnuN(\state, \symb), \transFuncSymb_\tmnuN (\state, \symb), \transFuncMove_\tmnuN (\state, \symb)),
    \end{equation}
    \begin{equation}\label{eq:subroutine-conditions-2}
        \proj_{\indexFuncNeur}\commFunc_\tmnuM(\indexFuncState(\state), \symb)(\neurState) = \commFunc_\tmnuN(\state, \symb)(\proj_{\indexFuncNeur}\neurState), \quad \text{and} \quad \proj^\perp_{\indexFuncNeur}\commFunc_\tmnuM(\indexFuncState(\state), \symb)(\neurState) = \proj^\perp_{\indexFuncNeur}\neurState,
    \end{equation}
    where $\proj_{\indexFuncNeur} : \R^{\neurDim_\tmnuM} \to \R^{\neurDim_\tmnuN}$ is the projection defined by $\proj_{\indexFuncNeur}\neurState = (\neurState_i)_{i \in \indexFuncNeur(\{1,\ldots,\neurDim_\tmnuN\})}$, and $\proj^\perp_{\indexFuncNeur} : \R^{\neurDim_\tmnuM} \to \R^{\neurDim_\tmnuM - \neurDim_\tmnuN}$ is the orthogonal projection defined by $\proj^\perp_{\indexFuncNeur}\neurState = (\neurState_i)_{i \notin \indexFuncNeur(\{1,\ldots,\neurDim_\tmnuN\})}$.
    % where 
    % \begin{align}\label{eq:subroutine-conditions}
    %     \begin{cases}
    %         \transFunc_\tmnuM(\state', \symb) &= (\indexFuncState \circ \transFuncState_\tmnuN(\state, \symb), \transFuncSymb_\tmnuN (\state, \symb), \transFuncMove_\tmnuN (\state, \symb))\\
    %         \\
    %         \commFunc_\tmnuM(\state', \symb)(\neurState)_i &= \neurState_i, \quad \text{for all } i \in \compIndexFuncNeur(\{1,\ldots,\neurDim_\tmnuN\}),
    %     \end{cases}
    % \end{align}
    % where $q' := \indexFuncState(q)$, and $\proj_{\indexFuncNeur} : \R^{\neurDim_\tmnuM} \to \R^{\neurDim_\tmnuN}$ is the projection defined by $\proj_{\indexFuncNeur}\neurState = (\neurState_i)_{i \in \indexFuncNeur(\{1,\ldots,\neurDim_\tmnuN\})}$.
\end{definition}

\newcommand{\Shadow}{\operatorname{Sh}}
\newcommand{\Lift}{\operatorname{Lift}}

We now introduce two pieces of notation that make the bookkeeping in subroutine arguments more transparent. Let $\tmnuN$ be a subroutine of $\tmnuM$ with correspondence $(\indexFuncState,\indexFuncNeur)$. If $\config = (\state;\tape;\neurState)$ is a configuration of $\tmnuM$ with $\state \in \indexFuncState(\{1,\ldots,\nStates_\tmnuN\})$, we define its \newname{subroutine shadow} by
\begin{equation}
    \Shadow_{\indexFuncState,\indexFuncNeur}(\config)
    :=
    \left(\indexFuncState^{-1}(\state);\tape;\proj_{\indexFuncNeur}\neurState\right).
\end{equation}
Conversely, if $\config' = (\state';\tape';\neurState')$ is a configuration of $\tmnuN$ and $h^\perp \in \R^{\neurDim_\tmnuM-\neurDim_\tmnuN}$, we define the \newname{lift} of $\config'$ with frozen complementary neural state $h^\perp$ as the configuration
\begin{equation}
    \Lift_{\indexFuncState,\indexFuncNeur}^{h^\perp}(\config')
    :=
    (\state;\tape;\neurState),
\end{equation}
where
\begin{equation}
    \state := \indexFuncState(\state'), \quad \tape := \tape', \quad
    \proj_{\indexFuncNeur}\neurState := \neurState', \quad \text{and} \quad
    \proj^\perp_{\indexFuncNeur}\neurState := h^\perp.
\end{equation}

The subroutine lemma states that the large machine follows the shadow computation exactly until the subroutine halts.
\begin{lemma}\label{lem:subroutine}
    Let $\tmnuM,\tmnuN$ be two \TMNUs such that $\tmnuN$ is a subroutine of $\tmnuM$ with correspondance $(\indexFuncState,\indexFuncNeur)$. Let $\config := (\state;\tape;\neurState)$ be a configuration of $\tmnuM$ such that $\state \in \indexFuncState(\{1,\ldots,\nStates_\tmnuN-1\})$, and define
    \begin{equation}\label{eq:main:lem:subroutine}
        \config' := \Shadow_{\indexFuncState,\indexFuncNeur}(\config), \quad \text{and} \quad h^\perp := \proj^\perp_{\indexFuncNeur}\neurState.
    \end{equation}
    Then, for all $t \in \{0, \ldots, T_\tmnuN(\config')\}$,
    \begin{equation}\label{eq:trajectory:lem:subroutine}
        \tmnuM^t(\config) = \Lift_{\indexFuncState,\indexFuncNeur}^{h^\perp}\left(\tmnuN^t(\config')\right),
    \end{equation}
    where $T_\tmnuN(\config')$ is the halting time of $\config'$ in $\tmnuN$ defined by $T_\tmnuN(\config') := \inf\{t \in \No : \proj_\state \tmnuN^t(\config') = \nStates_\tmnuN\}$. In particular, for all $t \in \{0, \ldots, T_\tmnuN(\config')\}$, we have
    \begin{equation}\label{eq:main:lem:subroutine-bound}
        \|\config\|_\tmnuM^t \leq \|\config'\|_\tmnuN^t \vee \|\config\|,
    \end{equation}
\end{lemma}
\begin{proof}
    Write $\tmnuM^t(\config) = (\state_t;\tape_t;\neurState_t)$ and $\tmnuN^t(\config') = (\state_t';\tape_t';\neurState_t')$. We prove \eqref{eq:trajectory:lem:subroutine} by induction on $t$. The case $t=0$ follows directly from the definitions of $\Shadow$ and $\Lift$.

    Assume that \eqref{eq:trajectory:lem:subroutine} holds for some $t < T_\tmnuN(\config')$. Equivalently, we have
    \[
        \state_t = \indexFuncState(\state_t'), \quad \tape_t = \tape_t', \quad
        \proj_{\indexFuncNeur}\neurState_t = \neurState_t', \quad \text{and} \quad
        \proj^\perp_{\indexFuncNeur}\neurState_t = h^\perp.
    \]
    Since $t < T_\tmnuN(\config')$, the state $\state_t'$ is not the halting state of $\tmnuN$. Therefore, the subroutine identities \eqref{eq:subroutine-conditions-1} and \eqref{eq:subroutine-conditions-2} apply to $\state_t'$, and yield
    \[
        \state_{t+1} = \indexFuncState(\state_{t+1}'), \quad \tape_{t+1} = \tape_{t+1}', \quad
        \proj_{\indexFuncNeur}\neurState_{t+1} = \neurState_{t+1}', \quad \text{and} \quad
        \proj^\perp_{\indexFuncNeur}\neurState_{t+1} = h^\perp.
    \]
    This is precisely \eqref{eq:trajectory:lem:subroutine} at time $t+1$, and the induction is complete.

    Finally, by \eqref{eq:trajectory:lem:subroutine}, for every $s \in \{0,\ldots,t\}$,
    \[
        \|\tmnuM^s(\config)\| \leq \|\tmnuN^s(\config')\| \vee \|h^\perp\|_\infty \leq \|\config'\|_\tmnuN^t \vee \|\config\|.
    \]
    Taking the maximum over $s=0,\ldots,t$ gives \eqref{eq:main:lem:subroutine-bound}.
\end{proof}

We now initiate the construction of \TMNUs that can approximate any continuous function $f : [-1,1] \to \R$. The general philosophy guiding the incoming constructions is the following. At its roots, our construction exploits the fact that given a continuous function $f : [-1,1] \to \R$, there exists a sequence of polynomials $(P_i)_{i \in \No}$ with coefficients having an arbitrarily long but finite binary representation that converges to $f$. We will show that there exists a \TMNU that, given that its tape is initialized as containing some encoding $u_i \in \{0,1\}^\ast$ of the coefficients of $P_i$ for some $i \in \No$, and that its neural state is initialized as containing some input $x \in [-1,1]$, can approximate $P_i(x)$, and hence $f(x)$. Then, we will use this \TMNU as a subroutine of a larger \TMNU that, given that its tape is initialized as containing an infinite sequence of bits $u := u_0u_1u_2 \ldots$ that is the result of the concatenation of the encodings $u_i$ of the coefficients of $P_i$ for every $i \in \No$, and that its neural state is initialized as containing some input $x \in [-1,1]$, can approximate $f(x)$ by successively approximating the polynomials $P_i(x)$ for every $i \in \No$. 

In other words, the \TMNU we design will be such that, if some infinite binary sequence $u$ is written on its tape, will recognize some finite prefix $u_0$ of $u$ as being some instruction to be executed, will execute this instruction, and delete the prefix $u_0$ from the tape, and then will repeat this process with the remaining infinite binary sequence on the tape. This way, if $u = u_0u_1u_2\ldots$ encodes the coefficients of the polynomials $P_i$ for every $i \in \No$, then the \TMNU will be able to successively approximate $P_i(x)$ for every $i \in \No$, and therefore to approximate $f(x)$. With this philosophy in mind, we will show that every elementary \TMNU that we design will be able to read some finite prefix $p$ of the infinite sequence $u$ written on its tape, recognize $p$ as being some instruction to be executed, execute this instruction, and delete the prefix $p$ from the tape. This will be reflected in all the statements we will establish regarding the properties of the \TMNUs we will design. In the following, we will first give four examples of simple \TMNUs that can be used as building blocks to design more complex \TMNUs, and then we will show how to use these simple \TMNUs as subroutines to design a \TMNU that can approximate any continuous function $f : [-1,1] \to \R$.

\subsection{Elementary \TMNUs}

As a first elementary example of the philosophy described above, we define a \TMNU that reads the first bit of the infinite binary sequence $u$ written on the tape, and depending on this bit, either multiplies the neural state by $-1$ or leaves it unchanged, and then deletes this bit from the tape. This \TMNU can be seen as a \TMNU that reads some finite prefix $u$ of the infinite sequence $u$ written on the tape, recognizes $u$ as being some instruction to be executed, executes this instruction, and deletes the prefix $u$ from the tape, where in this case the instruction is "if the first bit of $u$ is $1$, then multiply the neural state by $-1$, otherwise do nothing".
\begin{definition}\label{def:sign-tmnu}
    We let $\Sign$ be the \TMNU with neural dimension $1$ and 2 states, defined by the following procedure.
    \begin{itemize}
        \item \textbf{State 1:} Let $b \in \{\blanksymb,0,1\}$ under scan. Then, write $\blanksymb$, move right, update the neural state as $\neurState \gets (-1)^b \cdot \neurState$, and go to State $2$.
        \item \textbf{State 2:} Halt.
    \end{itemize}
\end{definition}
% Note that since $\blanksymb = -1$, $(-1)^b$ makes sense even for $b = \blanksymb$.
Its behavior is immediate from the definition and will be used as a one-step subroutine.
\begin{lemma}\label{lem:sign-tmnu}
    Let $b \in \{0,1\}$, $v \in \{0,1\}^\#$ and $x \in \R$. Define the configurations $c := (1;|bv;x)$ and $c' := (2;|v;(-1)^b x)$. Then, 
    \begin{equation}
        \Sign(c) =c', \quad \text{and} \quad \|c\|_\Sign^1 = |x|.
    \end{equation}
\end{lemma}
\begin{proof}
    By definition of $\Sign$, we have $\Sign(c) = c'$. Moreover, we have $\|c\|_\Sign^1 = \|c\| \vee \|c'\| = |x|$, which concludes the proof.
\end{proof}
As a second example of this philosophy, we design a \TMNU that, given that its tape is initialized as containing some finite binary sequence $u = 1^k0$ followed by some infinite binary sequence $u$, and that its neural state is initialized as containing some input $x \in [-1,1]$, will delete $u$ while multiplying $x$ by $2^k$. Essentially, this \TMNU acts as follows. As long as it reads a $1$ on the tape, it multiplies the neural state by $2$ and deletes this $1$ from the tape. When it reads a $0$ on the tape, it deletes this $0$ from the tape and halts. Therefore, if the tape is initialized as containing some finite binary sequence $u = 1^k0$ followed by some infinite binary sequence $u$, and if the neural state is initialized as containing some input $x \in [-1,1]$, then after $k+1$ steps, the tape will contain only the infinite binary sequence $u$, and the neural state will contain $2^k x$. This \TMNU is precisely defined as follows.

% Chaining + subroutines.

% \subsection{Simple constructions of TMNUs}

% % Maybe before the general method --> because they do not use the Chaining and subroutines method. Or maybe Chaining first, then simple constructions that do not use it, then the subroutines method.

% We first design two simple \TMNUs, namely, a \TMNU that multiplies its input by $2^k$ for some $k \in \N$, and a \TMNU that multiplies its input by $a \in [-1,1]$ some dyadic number.

\begin{definition}\label{def:contr-+-tmnu}
    We let $\Scale$ be the \TMNU with neural dimension $1$ and 2 states, defined by the following procedure.
    \begin{itemize}
        \item \textbf{State 1.} Let $b \in \{\blanksymb,0,1\}$ under scan.
        \begin{enumerate}
            \item[i.] If $b \in \{\blanksymb,0\}$, write $\blanksymb$, move right, and go to State $2$.
            \item[ii.] If $b = 1$, write $\blanksymb$, move right, update the neural state as
            $\neurState \gets 2 \cdot \neurState,$
            and go back to State 1.
        \end{enumerate}
        \item \textbf{State 2.} Halt.
    \end{itemize}
\end{definition}
We now establish that $\Scale$ can be used to multiply its input by $2^k$ for some $k \in \N$, by reading some finite binary sequence $u = 1^k0$ on the tape, and deleting it. Note that the proof of this result relies on the Chaining Lemma (Lemma \ref{lem:chaining}), which allows us to split the evolution from some configuration $c$ to some configuration $c'$ into several sub-evolutions, and to study each of these sub-evolutions separately.
The formal statement keeps track of both the final configuration and the trajectory bound.
\begin{lemma}\label{lem:scale-machine}
    Let $k \in \No$, $v \in \{0,1\}^\#$ and $x \in \R$. Define the configurations $c := (1;|1^k0v,|;x)$ and $c' := (2;|v,|;2^k x)$. Then,
    \begin{equation}
        \Scale^{k+1}(c) = c', \quad \|c\|_\Scale^{k+1} \leq 2^k|x|.
    \end{equation}
\end{lemma}
In the above Lemma, the tape in configuration $c$ contains the finite binary sequence $1^k0$ followed by some (potentially) infinite binary sequence $v$, and the neural state in configuration $c$ contains some input $x \in \R$. The Lemma states that after $k+1$ steps, the tape will contain only the infinite binary sequence $v$, and the neural state will contain $2^k x$. In other words, the \TMNU $\Scale$ has read the finite binary sequence $1^k0$ on the tape, recognized it as being an instruction to be executed, executed this instruction by multiplying the input $x$ by $2^k$, and deleted the finite binary sequence $1^k0$ from the tape. This is a very simple example of how a \TMNU can read some finite prefix of an infinite binary sequence written on its tape, recognize it as being some instruction to be executed, execute this instruction, and delete this prefix from the tape. We now give the proof of the above Lemma.
\begin{proof}
    Let $k \in \No$ and $x \in \R$. We define the configurations $c_0, c_1, \ldots, c_{k+1}$ of $\Scale$ by
    \[
        c_t := (1;|1^{k - t}0v,|;2^{t} x), \quad t \in \{0,\ldots,k\},
    \]
    and
    \[
        c_{k+1} := (2;|v,|;2^k x).
    \]
    \begin{claim*}
        For all $t \in \{0,\ldots,k\}$, we have $\Scale(c_t) = c_{t+1}$ and $\|c_t\|_\Scale^1 \leq 2^k |x|$.
    \end{claim*}
    \begin{proof}
        Let $t \in \{0,\ldots,k-1\}$. In configuration $c_t$, the machine is in State 1 and reads $b = 1$ on tape 1, so after one step, the machine is in configuration $c_{t+1}$. Moreover, we have $\|c_t\|_\Scale^1 \leq 2^{t+1}| x| \leq 2^k|x|$.
        Now, let $t = k$. In configuration $c_k$, the machine is in State 1 and reads $b = 0$ on tape 1, so after one step, the machine is in configuration $c_{k+1}$. Moreover, we have $\|c_k\|_\Scale^1 \leq 2^k| x|$. This concludes the proof of the claim.
    \end{proof}
    Now, since $c_0 = c$ and $c_{k+1} = c'$, the claim implies that $\Scale^{k+1}(c) = c',$ and, moreover,
    \begin{equation}
        \|c\|_\Scale^{k+1} = \max_{t = 0, \ldots, k} \|c_t\|_\Scale^1 \leq 2^k|x|.
    \end{equation}
    This concludes the proof.
\end{proof}

We now design a \TMNU that multiplies a real number $x \in \R$ by some dyadic number $a \in [0,1)$, by reading some finite binary sequence $u$ on the tape that encodes $a$ in binary form. We define $\delta_{[0,1)} : \{0,1\}^\ast \to [0,1]$ by
\begin{equation}\label{eq:definition_encoding_0_1}
    \delta_{[0,1)}(u) = \sum_{i=1}^{\len u} u_i 2^{-i}, \quad u \in \{0,1\}^\ast
\end{equation}
We denote by $\Db_1^+$ the output set of the function $\delta_{[0,1)}$, i.e.,
\begin{equation}
    \Db_1^+ := \left\{ \delta_{[0,1)}(u) : u \in \{0,1\}^\ast \right\} \subseteq [0,1].
\end{equation}
Note that $\Db_1^+$ is dense in $[0,1]$, because for every $x \in [0,1]$ and every $n \in \N$, there exists $u \in \{0,1\}^n$ such that $|x - \delta_{[0,1)}(u)| \leq 2^{-n}$. We also define kind of inverse encoding by
\begin{equation}
    \encDya{a} = \arg\min \{ |u| : u \in \{0,1\}^\ast, \delta_{[0,1)}(u) = a \},
\end{equation}
for every $a \in \Db_1^+$.

Continuing with the philosophy introduced above, the \TMNU we design to multiply $x$ by some $a \in \Db_1^+$ needs to be able to identify some prefix of an infinite binary sequence $u$ written on the tape as encoding the number $a$, and to delete this prefix from the tape while multiplying $x$ by $a$. In order to nonambiguously identify some prefix of $u$ as encoding $a$, we will use the following prefix-free encoding of binary sequences, that we denote by $\prfx{\cdot} : \{0,1\}^\ast \to \{0,1\}^\ast$.
We first specify the prefix code used to delimit finite dyadic instructions.
\begin{definition}(Prefix-free encoding)\label{def:prefix-free-encoding}
    For $u \in \{0,1\}^\ast$, we let $\prfx{u} \in \{0,1\}^{2\len u + 1}$ be defined as
    \[
        \prfx{u}_{2k} = u_k, \quad \prfx{u}_{2k-1} = 1, \ \forall k \in \{1,\ldots,\len u\},
    \]
    and $\prfx{u}_{2\len u + 1} =0$.
\end{definition}
The \TMNU we design has a neural state in dimension 2. Initially, the neural state contains $(x,0)$ for some $x \in \R$, and the tape content is of the form $\prfx{u}v$ for some $u \in \{0,1\}^\ast$ encoding a number $a \in \Db_1^+$ and some $v \in \{0,1\}^\#$. Then, the \TMNU will exploit the prefix-free encoding to identify the prefix $\prfx{u}$ of the tape content as encoding the number $a$. Now, remark that $a x$ can be rewritten as
\begin{equation}
    a x = \sum_{k=1}^{\len u} u_k 2^{-k} x = \sum_{k=1}^{\len u} u_k y_k,
\end{equation}
where $y_k := 2^{-k} x$ for every $k \in \{1,\ldots,\len u\}$. Note that, in particular, we have $y_{k+1} = y_k/2$, for every $k \in \{1,\ldots,\len u - 1\}$. Therefore, the \TMNU will compute successively each term $y_k$ on the first coordinate of the neural state, and will add $y_k$ to the second coordinate of the neural state if $u_k = 1$. This way, after reading the whole prefix $\prfx{u}$, the first coordinate of the neural state will contain $y_{\len u} = 2^{-\len u} x$, and the second coordinate of the neural state will contain $\sum_{k=1}^{\len u} u_k y_k = a x$. Then, the \TMNU will delete the prefix $\prfx{u}$ from the tape, and will halt with the first coordinate of the neural state containing $2^{-\len u} x$ and the second coordinate of the neural state containing $a x$. The precise definition of this \TMNU is given as follows.
\begin{definition}\label{def:scale-machine}
    We let $\Contr^+$ be the \TMNU with neural dimension $2$ and 3 states, defined by the following procedure.
    \begin{itemize}
        \item \textbf{State 1.} Let $b \in \{\blanksymb,0,1\}$ be the symbol under scan.
        \begin{enumerate}
            \item[i.] If $b \in \{ 0, \blanksymb\}$, update the neural state as 
            $
                \neurState \gets (\neurState_2,0),
            $
            and go to State 3.
            \item[ii.] If $b =1$, write $\blanksymb$, move right, update the neural state as
            $
                \neurState \gets \left(\frac{1}{2} \neurState_1, \neurState_2\right),
            $
            and go to State 2. 
        \end{enumerate}
        \item \textbf{State 2.} Let $b \in \{\blanksymb,0,1\}$ be the symbol under scan. Then, write $\blanksymb$, move right, update the neural state as
            $
                \neurState \gets \left(\neurState_1, \neurState_2 + b \neurState_1\right),
            $
        and go to State 1.
        \item \textbf{State 3.} Halt.
    \end{itemize}
\end{definition}
We now cast the discussion above into the following formal Lemma, that establishes that $\Contr^+$ can be used to multiply a real number $x \in \R$ by some dyadic number $a \in \Db_1^+$, by reading some finite binary sequence $u$ on the tape that encodes $a$ in binary form, and deleting this prefix from the tape.
The estimate also records that the auxiliary accumulation never exceeds the input magnitude.
\begin{lemma}\label{lem:contr-tmnu}
    Let $a \in \Db_1^+, u := \encDya{a}, v \in \{0,1\}^\#$, and $x \in \R$. Define the configurations $c := (1;|\prfx{u}v;x,0)$ and $c' := (3;|v;a x,0)$. Then,
    \begin{equation}
        (\Contr^+)^{\len{\prfx{u}}}(c) = c', \quad \|c\|_{\Contr^+}^{\len{\prfx{u}}} \leq |x|.
    \end{equation}
\end{lemma}
\begin{proof}
    Let $a \in \Db_1^+$, $u:= \encDya{a}$, $n := \len {u}$, $v \in \{0,1\}^\#$, and $x \in \R$. We define a sequence $(c_t)_{0 \leq t \leq 2n + 1}$ of configurations of ${\Contr^+}$ by
    \begin{equation}\label{eq:1:thm:contr-tmnu}
        \begin{cases}
            c_{2t} &= \left(1;|\prfx{u}_{2t+1:2n+1}v;2^{-t} x,a_t x\right), \quad \text{for all } t = 0, \ldots, n,\\
            c_{2t+1} &= \left(2;|\prfx{u}_{2(t+1):2n+1}v;2^{-(t+1)} x,a_t x\right), \quad \text{for all } t = 0, \ldots, n-1,\\
            c_{2n+1} &= (3;|v; a x,0),
        \end{cases}
    \end{equation}
    where $a_t := \delta_{[0,1)}(u_{1:t})$ for every $t \in \{0,\ldots,n\}$.
    \begin{claim*}
        For all $t = 0, \ldots, 2n$, we have $\Contr^+(c_t) = c_{t+1}$ and $\|c_t\|_{\Contr^+}^1 \leq |x|$.
    \end{claim*}
    \begin{proof}[Proof of the Claim]
        We divide the proof of the claim into three cases. 
    \begin{enumerate}
        \item Let $0 \leq t \leq n-1$, and assume that the machine is in configuration $c_{2t}$. We have 
        \[
            c_{2t} = \left(1;|\prfx{u}_{2t+1:2n+1}v; 2^{-t}x,a_t x\right),
        \]
        therefore the machine is in State 1 and reads $b := \prfx{u}_{2t+1} = 1$ on the tape. By Definition of State 1 of ${\Contr^+}$ (Definition \ref{def:scale-machine}), the machine writes $\blanksymb$ on the tape, moves the head one cell to the right, updates the neural state to $(\neurState_1/2,\neurState_2)$, and goes to State 2. Hence, after one step, the machine is in configuration
        \[
            \left(2;|\prfx{u}_{2t+2:2n+1}v; 2^{-(t+1)} x,a_t x\right) = c_{2t+1}.
        \]
        Moreover, we have $\|c_{2t}\|_{\Contr^+}^1 \leq |x|$.
        \item Let $0 \leq t \leq n-1$. Assume that the machine is in configuration $c_{2t+1}$. We have
        \[
            c_{2t+1} = \left(2;|\prfx{u}_{2(t+1):2n+1}v;2^{-(t+1)} x,a_t x\right),
        \]
        therefore the machine is in State 2 and reads $b := \prfx{u}_{2(t+1)} = u_{t+1}$ on the tape. By Definition of State 2 of ${\Contr^+}$ (Definition \ref{def:scale-machine}), the machine writes $\blanksymb$ on the tape, moves the head one cell to the right, updates the neural state to $(\neurState_1, \neurState_2 + b \cdot \neurState_1)$, where
        \[
            \neurState_2 + b \cdot \neurState_1 = a_t x + u_{t+1} \cdot 2^{-(t+1)} x = a_{t+1} x,
        \]
        and goes to State 1. Hence, after one step, the machine is in configuration
        \[
            \left(1;|\prfx{u}_{2t+3:2n+1}v;2^{-(t+1)} x,a_{t+1}x\right) = c_{2t+2}.
        \]
        Moreover, we have $\|c_{2t+1}\|_{\Contr^+}^1 \leq |x|$.
        \item Case $t = 2n$. Assume that the machine is in configuration \[c_{2n} := \left(1;|\prfx{u}_{2n+1:2n+1}v; 2^{-n} x, a_n x\right) = \left(1;|\prfx{u}_{2n+1}v; 2^{-n} x, a x\right),\] so the machine is in State 1, and reads the symbol $b = \prfx{u}_{2n+1} = 0$. By Definition of State 1 of ${\Contr^+}$ (Definition \ref{def:scale-machine}), the machine updates the neural state to $\left(a x,0\right)$, and goes to State 3. This implies that, after one time step, the machine is in configuration $(3;|v;a x,0) = c_{2n+1}$. Moreover, we have $\|c_{2n}\|_{\Contr^+}^1 \leq |x|$.
    \end{enumerate}
    This concludes the proof of the claim.
    \end{proof}
    Now, since $c = c_0$ and $c' = c_{2n+1}$, the claim implies that
    \begin{equation}
        (\Contr^+)^{2n+1}(c) = c',
    \end{equation}
    and, moreover,
    \begin{equation}
        \|c\|_{\Contr^+}^{2n+1} = \max_{0 \leq t \leq 2n} \|c_t\|_{\Contr^+}^1 \leq |x|.
    \end{equation}
    Since $\len{\prfx{u}} = 2n + 1$, this concludes the proof.
\end{proof}

As a final example of the philosophy of building \TMNUs that identify a finite prefix of an infinite sequence written on their tape as an instruction to be executed, we define a family of \TMNUs that simulate RNNs. Specifically, for every RNN $\Rc$, we define a \TMNU $\tmnuM_\Rc$ such that given that $1^n0u$ is initially written on the tape and the neural state contains some $x \in \R^d$ (where $d$ is the input dimension of $\Rc$), the \TMNU erases $1^n0$ from its tape, and produces $\Rc \Dc x[n]$ on its neural state. This allows us, in particular, to import any results already present in the literature for RNNs to \TMNUs. In particular, this allows to show that \TMNUs can approximate the product of two real numbers, by simulating the RNN defined in \cite[Theorem 11]{hutter2025quantifierRNN}. We define formally the \TMNU that simulates an RNN.
\begin{definition}\label{def:tmnu-simulation-rnn}
    Let $\Rc := (d,m,d'; A_h, b_h, A_x, A_o, b_o)$ be an \RNN. We define the \TMNU $\tmnuM_{\Rc}$ with neural dimension $\neurDim := m$ and 4 states, defined by the following procedure.
    \begin{itemize}
        \item \textbf{State 1.} Update the neural state as
        $
            \neurState \gets A_x \neurState_{1:d} + b_h,
        $
        and go to State 2.
        \item \textbf{State 2.} Update the neural state as
        $
            \neurState \gets \ReLU(\neurState)
        $
        and go to State 3.
        \item \textbf{State 3.} Let $b \in \{\blanksymb,0,1\}$ be the symbol under scan.
        \begin{enumerate}
            \item[i.] If $b \in \{0,\blanksymb\}$, write $\blanksymb$, move right, update the neural state as
            \[
                \neurState \gets \begin{bmatrix}
                    A_o\\
                    0
                \end{bmatrix}  \neurState + \begin{bmatrix}
                    b_o\\
                    0
                \end{bmatrix},
            \]
            and go to State 4.
            \item[ii.] If $b = 1$, write $\blanksymb$, move right, update the neural state as
            $
                \neurState \gets A_h \neurState + b_h
            $
            and go to State 2.
        \end{enumerate}
        \item \textbf{State 4.} Halt.
    \end{itemize}
\end{definition}

The simulation property of this machine is recorded next.
\begin{lemma}\label{lem:tmnu-simulation-rnn}
    Let $\Rc := (d,m,d'; A_h, b_h, A_x, A_o, b_o)$ be an \RNN, and let $\tmnuM_\Rc$ be the \TMNU defined in Definition \ref{def:tmnu-simulation-rnn}. Let $n \in \N$, $v \in \{0,1\}^\#$, and $x \in \R^d$. Define the configurations $c := (1;|1^n0v;x,0,\ldots,0)$ and $c' := (4;|v;\Rc\Dc x[n], 0, \ldots, 0)$. Then,
    \begin{equation}
        \tmnuM_\Rc^{2n+3}(c) = c',
    \end{equation}
    and
    \begin{equation}
        \|c\|_{\tmnuM_\Rc}^{2n+3} \leq \|x\|_\infty \vee \|A_x x + b_h\|_\infty \vee \|\Rc \Dc x[n]\|_\infty \vee \max_{0 \leq t \leq n} \|\Hc \Dc x[t]\|_\infty \vee \max_{0 \leq t \leq n} \|A_h\Hc \Dc x[t] + b_h\|_\infty,
    \end{equation}
    where $\Hc$ is the hidden state operator of $\Rc$ as defined in Definition \ref{def:elman_rnn}.
\end{lemma}
\begin{proof}
    Let $\Rc := (d,m,d'; A_h, b_h, A_x, A_o, b_o)$ be an \RNN, $n \in \No$, $v \in \{0,1\}^\#$, and $x \in \R^d$. We define a sequence of configurations $(c_t)_{0 \leq t \leq 2n + 3}$ of $\tmnuM_\Rc$ by
    \begin{equation}
        \begin{cases}
            c_0 &= (1;|1^n0v;x,0,\ldots,0),\\
            c_1 &= (2;|1^n0v;A_x x + b_h),\\
            c_{2t} &= (3;|1^{n-t+1}0v;\Hc \Dc(x)[t-1]), \quad \text{for all } t = 1, \ldots, n+1,\\
            c_{2t+1} &= (2;|1^{n-t}0v;A_h\Hc \Dc(x)[t-1] + b_h), \quad \text{for all } t = 1, \ldots, n,\\
            c_{2n+3} &= (4;|v;\Rc \Dc(x)[n], 0,\ldots,0).
        \end{cases}
    \end{equation}
    We also let
    \[
        B := \|x\|_\infty \vee \|A_x x + b_h\|_\infty \vee \|\Rc \Dc x[n]\|_\infty \vee \max_{0 \leq t \leq n} \|\Hc \Dc x[t]\|_\infty \vee \max_{0 \leq t \leq n} \|A_h\Hc \Dc x[t] + b_h\|_\infty.
    \]
    \begin{claim*}
        For every $t \in \{0, \ldots, 2n + 2\}$, we have $\tmnuM_\Rc(c_t) = c_{t+1}$.
    \end{claim*}
    \begin{proof}[Proof of the claim]
        First note that $\tmnuM_\Rc(c_0) = c_1$ by State 1 of $\tmnuM_\Rc$ (Definition \ref{def:tmnu-simulation-rnn}), and that $\tmnuM_\Rc(c_1) = c_2$ by State 2 of $\tmnuM_\Rc$. Moreover, in configuration $c_{2n+2} = (3;|0v;\Hc\Dc x[n])$, the machine is in State 3 and reads a symbol $0$ on its tape. Therefore, $\tmnuM_\Rc(c_{2n+2}) = c_{2n+3}$.

        Now, let $t \in \{1, \ldots, n\}$. We have $c_{2t} = (3;|1^{n-t+1}0v;\Hc \Dc x[t-1])$, so the machine is in State 3 and reads the symbol $b = 1$ on the tape. By State 3 of $\tmnuM_\Rc$, we have $\tmnuM_\Rc(c_{2t}) = c_{2t+1}$, since $c_{2t+1} = (2;|1^{n-t}0v;A_h\Hc \Dc x[t-1] + b_h)$. Moreover, we have $c_{2t+1} = (2;|1^{n-t}0v;A_h\Hc \Dc x[t-1]+b_h)$, so the machine is in State 2, and by State 2 of $\tmnuM_\Rc$, we have $\tmnuM_\Rc(c_{2t+1}) = c_{2(t+1)}$, since $c_{2(t+1)} = (3;|1^{n-t}0v;\Hc \Dc x[t])$. This concludes the proof of the claim.
    \end{proof}
    Since $c = c_0$ and $c' = c_{2n+3}$, the claim implies that
    \begin{equation}
        \tmnuM_\Rc^{2n+3}(c) = c'.
    \end{equation}
    Moreover, by denoting $\neurState_t$ the neural state of configuration $c_t$, we have
    \begin{align*}
        &\|c\|_{\tmnuM_\Rc}^{2n+3} = \max_{0 \leq t \leq 2n + 2} \|c_t\|_{\tmnuM_\Rc}^1 \leq \max_{0 \leq t \leq 2n + 3} \|\neurState_t\|_\infty\\
        &\leq \max\{ \|x\|_\infty, \|A_x x + b_h\|_\infty, \|\Rc \Dc x[n]\|_\infty, \max_{0 \leq t \leq n} \|\Hc \Dc x[t]\|_\infty, \max_{0 \leq t \leq n} \|A_h\Hc \Dc x[t] + b_h\|_\infty\}\\
        &= B,
    \end{align*}
    by definition of $B$. This concludes the proof of the Lemma.
\end{proof}

Now, we can use the following result from the literature on RNNs to design a \TMNU that can approximate the multiplication of two inputs $x$ and $y$.
\begin{lemma}\label{lem:multiplication_rnn}\cite[Theorem 11]{hutter2025quantifierRNN}
    There exists an \RNN $\Rc^\times := (2,14,1;A_h,b_h,A_x,A_o,b_o)$, such that for every $\realx \in [-1,1]^2$,
    \[
        \left|\Rc^\times\Dc 
        \realx[n] - (\realx_1\cdot \realx_2)\right| \leq 2^{-2n-1}, \quad \text{and} \quad \left\|\Hc^\times\Dc 
        \realx[n]\right\|_\infty,
        \left\|A_h \Hc^\times\Dc 
        \realx[n] + b_h\right\|_\infty, \left|\Rc^\times\Dc 
        \realx[n] \right| \leq 1,
    \]
    for every $n \in \N$, and $A_x \realx + b_h \in [-1,1]^{14}$.
\end{lemma}

We turn this RNN primitive into a \TMNU by applying the simulation construction above.
\begin{definition}\label{def:tmnu-mult}
    We let $\Times$ be the \TMNU defined as the \TMNU $\tmnuM_{\Rc^\times}$ of Definition \ref{def:tmnu-simulation-rnn}, where $\Rc^\times$ is the \RNN of Lemma \ref{lem:multiplication_rnn}.
\end{definition}

The resulting machine inherits the approximation and boundedness properties of the RNN multiplier.
\begin{lemma}\label{lem:tmnu-mult}
    Let $n \in \No$, $v \in \{0,1\}^\#$, and $x,y \in [-1,1]$. Define the configurations $c := (1;|1^n0v;x,y,0_{12})$ and $c' := (4; |v; \Rc^\times\Dc 
        (x,y)[n], 0_{13})$. Then,
    \begin{equation}
        \Times^{2n+3}(c) = c', \quad \|c\|_\Times^{2n+3} \leq 1.
    \end{equation}
\end{lemma}
\begin{proof}
    Let $\Rc^\times := (2,14,1;A_h,b_h,A_x,A_o,b_o)$ be the \RNN of Lemma \ref{lem:multiplication_rnn}, and let $\Hc^\times$ be its hidden state operator. By Lemma \ref{lem:tmnu-simulation-rnn}, we have $\Times^{2n+3}(c) = c'$ and
    \begin{align*}
        \|c\|_\Times^{2n+3} &\leq |x| \vee |y| \vee \|A_x (x,y) + b_h\|_\infty \vee |\Rc^\times \Dc (x,y)[n]|\\
        &\quad \vee \max_{0 \leq t \leq n} \|\Hc^\times \Dc (x,y)[t]\|_\infty \vee \max_{0 \leq t \leq n} \|A_h\Hc^\times \Dc (x,y)[t] + b_h\|_\infty.
    \end{align*}
    Since $(x,y) \in [-1,1]^2$, we have $|x| \vee |y| \leq 1$. Moreover, Lemma \ref{lem:multiplication_rnn} implies that $\|A_x (x,y) + b_h\|_\infty \leq 1$, $|\Rc^\times \Dc (x,y)[n]| \leq 1$, and, by applying the same lemma for each $t \in \{0,\ldots,n\}$,
    \[
        \max_{0 \leq t \leq n} \|\Hc^\times \Dc (x,y)[t]\|_\infty \vee \max_{0 \leq t \leq n} \|A_h\Hc^\times \Dc (x,y)[t] + b_h\|_\infty \leq 1.
    \]
    This concludes the proof.
\end{proof}

We now have defined four \TMNUs that allows to perform some simple operations: multiplication by $-1$, multiplication by $2^k$ for some $k \in \N_0$, multiplication by some $a \in \Db_1^+$, and approximation of the multiplication of two inputs. These \TMNUs can be combined to design more complex \TMNUs that perform more complex operations. Specifically, by combining these operations we can design a \TMNU that approximate any polynomial, and consequently, and any continuous function. In the next section, we introduce the notion of subroutine for \TMNUs, that will allow us to design more complex \TMNUs by combining simpler \TMNUs as building blocks.

\newcommand\encDyaSign[1]{\ensuremath{v^{(#1)}}}
Now equipped with this tool for combining \TMNUs, we are ready to design more complex \TMNUs that can perform more complex operations, by combining simpler \TMNUs as building blocks. As a first example, we can combine the \TMNU $\Sign$ and the \TMNU $\Contr^+$ to design a \TMNU that can multiply its input by some $a \in \Db_1 := \Db_1^+ \cup \left(- \Db_1^+\right)$. For every $a \in \Db_1$, we let
\[
	s^{(a)} = \begin{cases}
	    0, & \text{if } a \geq 0, \\
	    1, & \text{if } a < 0,
	\end{cases}
\]
and define $\encDyaSign{a} := s^{(a)} \prfx{\encDya{|a|}}$. We now introduce the \TMNU $\Contr$ that can multiply its input by some $a \in \Db_1$ by reading some finite binary sequence $u$ on the tape that encodes $a$ in binary form, and deleting this prefix from the tape.
\begin{definition}\label{def:contr-tmnu}
    We let $\Contr$ be the \TMNU with neural dimension $2$ and 4 states, defined by the following procedure.
    \begin{itemize}
        \item \textbf{State 1.} Use $\Sign$ as a subroutine with correspondence $(1,2;1)$.
        \item \textbf{State 2-3.} Use $\Contr^+$ as a subroutine with correspondence $(2,3,4;1,2)$.
        \item \textbf{State 4.} Halt.
    \end{itemize}
\end{definition}
By using the Subroutine Lemma \ref{lem:subroutine}, we can easily verify that $\Contr$ can be used to multiply an input $x$ by some $a \in \Db_1$, by reading some finite binary sequence $u$ on the tape that encodes $a$ in binary form, and deleting this prefix from the tape, as follows. 
\begin{lemma}\label{lem:contr-tmnu-sign}
    Let $a \in \Db_1, u := \encDyaSign{a}, v \in \{0,1\}^\#$, and $x \in \R$. Define the configurations $c := (1;|uv;x,0)$ and $c' := (4;|v;a x,0)$. Then,
    \begin{equation}
        \Contr^{\len u}(c) = c', \quad \|c\|_{\Contr}^{\len u} \leq |x|.
    \end{equation}
\end{lemma}
\begin{proof}
    Let $a \in \Db_1$, $u := \encDyaSign{a}$, $v \in \{0,1\}^\#$, and $x \in \R$. Define the configurations
    \[
        c_0 := (1;|uv;x,0), \quad c_1 := (2;|u_{2:\len u}v;(-1)^{u_1}x,0), \quad c_2 := (4;|v;a x,0).
    \]
    We first show that $\Contr(c_0) = c_1$ and $\|c_0\|_\Contr^1 \leq |x|$. By definition of State 1 of $\Contr$ (Definition \ref{def:contr-tmnu}), the machine uses the \TMNU $\Sign$ as a subroutine with correspondence $(1,2;1)$. Let
    \[
        c_0' := (1;|uv;x), \quad c_1' := (2;|u_{2:\len u}v;(-1)^{u_1}x)
    \]
    be configurations of $\Sign$. Then,
    \[
        \Shadow_{(1,2),(1)}(c_0) = c_0', \quad \Lift_{(1,2),(1)}^0(c_1') = c_1.
    \]
    By Lemma \ref{lem:sign-tmnu}, we have $\Sign(c_0') = c_1'$ and $\|c_0'\|_\Sign^1 \leq |x|$. Therefore, by Lemma \ref{lem:subroutine}, we obtain
    \[
        \Contr(c_0) = \Lift_{(1,2),(1)}^0(\Sign(c_0')) = c_1
    \]
    and
    \[
        \|c_0\|_\Contr^1 \leq \|c_0'\|_\Sign^1 \vee \|c_0\| \leq |x|.
    \]

    We now show that $\Contr^{\len u - 1}(c_1) = c_2$ and $\|c_1\|_\Contr^{\len u - 1} \leq |x|$. By definition of States 2 and 3 of $\Contr$ (Definition \ref{def:contr-tmnu}), the machine uses the \TMNU $\Contr^+$ as a subroutine with correspondence $(2,3,4;1,2)$. Let
    \[
        \tilde c_1 := (1;|u_{2:\len u}v;(-1)^{u_1}x,0), \quad \tilde c_2 := (3;|v;|a|(-1)^{u_1}x,0) = (3;|v;a x,0)
    \]
    be configurations of $\Contr^+$. Then,
    \[
        \Shadow_{(2,3,4),(1,2)}(c_1) = \tilde c_1, \quad \Lift_{(2,3,4),(1,2)}^{\varnothing}(\tilde c_2) = c_2.
    \]
    Since $u_{2:\len u} = \prfx{\encDya{|a|}}$ and $\len{u_{2:\len u}} = \len u - 1$, Lemma \ref{lem:contr-tmnu} gives
    \[
        (\Contr^+)^{\len u - 1}(\tilde c_1) = \tilde c_2, \quad \|\tilde c_1\|_{\Contr^+}^{\len u - 1} \leq |x|.
    \]
    Therefore, by Lemma \ref{lem:subroutine}, we obtain
    \[
        \Contr^{\len u - 1}(c_1) = \Lift_{(2,3,4),(1,2)}^{\varnothing}\left((\Contr^+)^{\len u - 1}(\tilde c_1)\right) = c_2
    \]
    and
    \[
        \|c_1\|_\Contr^{\len u - 1} \leq \|\tilde c_1\|_{\Contr^+}^{\len u - 1} \vee \|c_1\| \leq |x|.
    \]
    Combining the two stages, and using $c = c_0$ and $c' = c_2$, we obtain
    \[
        \Contr^{\len u}(c) = c',
    \]
    and
    \[
        \|c\|_\Contr^{\len u} = \|c_0\|_\Contr^1 \vee \|c_1\|_\Contr^{\len u - 1} \leq |x|.
    \]
\end{proof}

In the incoming section, we use the Subroutine Lemma \ref{lem:subroutine} in combination with the elementary \TMNUs defined in previous section in order to build a \TMNU that approximate any polynomial with dyadic coefficients.

\subsection{Approximation of polynomials with dyadic coefficients}

In this section, we show how to use the previously defined \TMNUs in order to approximate polynomials with dyadic coefficients. Specifically, we will design a \TMNU that realises the following approximation of a polynomial $P = a_0 + a_1 x + \ldots + a_d x^d$, with $d \in \No$ and $a_0, \ldots, a_d \in \Db := \Z + \Db_1$. For $d,n \in \No$, we define the function $G_{d,n} : [-1,1] \to [-1,1]$ by
    \[G_{0,n}(x) = 1, \quad G_{d+1,n}(x) = \Rc^\times \Dc (G_{d,n}(x),x)[n], \quad d \in \N, n \in \No, x \in [-1,1],\]
and further, for $n \in \No$, the function $F_{P,n} : [-1,1] \to \R$ by
    \[F_{P,n}(x) = \sum_{i=0}^d a_i G_{i,n}(x).\]
The function $F_{P,n}$ is precisely the function that we will realize with a \TMNU, by computing succesively $G_{0,n}(x), G_{1,n}(x), \ldots, G_{d,n}(x)$, and then combining these values with linear transformations to obtain $F_{P,n}(x)$. In Appendix \ref{sec:approximation-polynomials-with-rnns:app} are shown the following bounds on the magnitude and on the approximation rate of $G_{d,n}$ and $F_{P,n}$, for $n,d \in \No$. In this Lemma, we use the quantities
\begin{equation}
    \|P\|_0 := d+1, \quad \text{and} \quad \|P\|_1 := |a_0| + |a_1| + \cdots + |a_d|,
\end{equation}
where $P = a_0 + a_1 x + \ldots + a_d x^d$ is a polynomial with $d \in \No$ and $a_0, \ldots, a_d \in \R$.

% In order to design a \TMNU that can approximate any polynomial function, we essentially need to combine the two following ingredients: a \TMNU that can multiply its input by some $a \in \Db$, and a \TMNU that can approximate the multiplication two inputs $x$ and $y$. By the preceding section, we already have at our disposal three \TMNUs that, put together, can multiply an input $x$ by some $a \in \Db$. To design a \TMNU that can approximate the multiplication of two inputs $x$ and $y$, we will actually use the known fact that the multiplication of two reals can be approximated by an RNN CITE CLEMENS, and we will simply show that any RNN can be simulated by a \TMNU as follows. Given an \RNN $\Rc$ of input dimension $d$, there exists a \TMNU $\tmnuM_\Rc$ such that, if the initial tape of the \TMNU contains an infinite sequence $1^n0 v$, where $n \in \N$ and $v \in \{0,1\}^\#$, and the initial neural state of the \TMNU contains some inputs $x_1, \ldots, x_d \in \R$, then after $2n+1$ steps, the tape of the \TMNU contains $v$, and the neural state of the \TMNU contains $\Rc \Dc(x_1, \ldots, x_d)[n]$. This \TMNU is formally defined as follows.

% We first collect the approximation estimates that the polynomial machine will implement.
\begin{lemma}\label{lem:mono_poly_approx}
    Let $d,n \in \No$ and $P$ be a polynomial. Then, for all $x \in [-1,1]$, we have $|G_{d,n}(x)| \leq 1$, $|F_{P,n}(x)| \leq \|P\|_1$,
    \[
        \left|G_{d,n}(x) - x^d\right| \leq d 2^{-2n-1}, \quad \text{and} \quad |F_{P,n}(x) - P(x)| \leq \|P\|_1 \|P\|_0 2^{-2n-1}.
    \]
    % and 
    % \[
        
    % \]
\end{lemma}
We now explain how we design a \TMNU that computes $F_{P,n}$. The idea is that the \TMNU starts with its tape containing an encoding of the coefficients $a_0, \ldots a_d \in \Db$ of $P$, and an encoding of $n \in \No$, and with its neural state containing $(x,1,0) = (x, G_{0,n}(x),0)$ for some $x \in [-1,1]$, that is transformed successively into $(x, G_{1,n}(x), a_0 G_{0,n}(x))$, $(x, G_{2,n}(x), a_0 G_{0,n}(x) + a_1 G_{1,n}(x))$, \ldots, $(x, G_{d,n}(x), a_0 G_{0,n}(x) + a_1 G_{1,n}(x) + \cdots + a_{d-1} G_{d-1,n}(x))$, and finally into $(x, G_{d,n}(x), F_{P,n}(x))$. We will specifically design a \TMNU $\UpPoly$ that implements one step of this procedure, that is, that transforms $(x, G_{i,n}(x), \sum_{j=0}^{i-1} a_j G_{j,n}(x))$ into $(x, G_{i+1,n}(x), \sum_{j=0}^{i} a_j G_{j,n}(x))$, for any $i \in \{0,\ldots,d-1\}$. Then, we will use $\UpPoly$ as a subroutine of a larger \TMNU that implements the whole procedure.

Before proceeding, we want to add a small correction to the preceding description of the procedure. In view of the simulation of the \TMNU we build by an RNN, we need to make sure that the neural state of the \TMNU remains bounded below some constant, as explained at the beginning of Section \ref{sec:tmnu-constructions}. In the procedure above, the neural state is bounded by $\|P\|_1$, so this \TMNU can indeed be simulated by an RNN. However, ultimately, this \TMNU will be used as a subroutine of a larger \TMNU that approximates continuous functions by means of a sequence $(P_i)_{i \in \No}$ of approximating polynomials, that may satisfy $\|P_i\|_1 \underset{i \to \infty}{\to} \infty$. To cope with this issue, we will make a slight modification of the procedure above, that guarantees that the neural state of the \TMNU remains bounded by $\|P\|_{L^\infty([-1,1])} + \|P\|_0 \|P\|_1 2^{-2n-1}$. This way, when approximating a sequence of polynomials $(P_i)_{i \in \No}$ that converges uniformly to some continuous function, we can always choose $n$ large enough so that $\|P_i\|_0 \|P_i\|_1 2^{-2n-1} \leq 1$, and moreover, the uniform convergence of $(P_i)_{i \in \No}$ implies that there exists some constant $M > 0$ such that $\|P_i\|_{L^\infty([-1,1])} \leq M$ for all $i \in \No$, hence the neural state of the \TMNU remains bounded by $M+1$ when approximating the sequence $(P_i)_{i \in \No}$, and can therefore be simulated by an RNN. We now describe how to modify the above procedure. For every polynomial $P \neq 0$, define $k_P := 1 + \lfloor \log_2 \|P\|_1 \rfloor \vee 0$, and $P^* := 2^{-k_P} P$. Note that $\|P^\ast\|_\infty \leq \|P^*\|_1 < 1$. Then, instead of computing $F_{P,n}(x)$, we will compute $F_{P^*,n}(x)$, and then multiply the result by $2^{k_P}$ to obtain $F_{P,n}(x)$. Therefore, during the process of computing $F_{P^*,n}(x)$, the neural state of the \TMNU remains bounded by $\|P^*\|_1 < 1$, and at the end of the process, the \TMNU multiplies the result by $2^{k_P}$ to obtain $F_{P,n}(x)$, so the neural state remains bounded below $|2^{k_P} F_{P^\ast,n}(x)| = |F_{P,n}(x)| \leq \|P\|_{L^\infty([-1,1])} + \|P\|_0 \|P\|_1 2^{-2n-1}$, by Lemma \ref{lem:mono_poly_approx}.

We now give the formal definition of the \TMNU $\UpPoly$ described above.
This machine performs one coefficient update and one monomial update.
\begin{definition}\label{def:upoly}
     We define the \TMNU $\UpPoly$ with neural dimension $16$ and 9 states, defined by the following procedure.
    \begin{itemize}
        \item \textbf{State 1.} Update the neural state as $
            \neurState \gets (\neurState_1, \neurState_2, \neurState_3, \neurState_2, 0_{12}),
        $
        and go to State 2.
        \item \textbf{State 2-4.} Use $\Contr$ (Def. \ref{def:contr-tmnu}) as subroutine with correspondence $(2,3,4,5;4,5)$.
        \item \textbf{State 5.} Update the neural state as
        $
            \neurState \gets (\neurState_1,\neurState_2,\neurState_3+\neurState_4,\neurState_1,0_{12}),
        $
        and go to State 6.
        \item \textbf{State 6-8.} Use $\Times$ (Def. \ref{def:tmnu-mult}) as subroutine with correspondence $(6,7,8,9;2,4,\ldots,16)$.
        \item \textbf{State 9.} Halt. 
    \end{itemize}
\end{definition}
We now show that $\UpPoly$, in essence, implements one step of the procedure described above, that is, it transforms $(x, G_{i,n}(x), \sum_{j=0}^{i-1} a_j G_{j,n}(x))$ into $(x, G_{i+1,n}(x), \sum_{j=0}^{i} a_j G_{j,n}(x))$, for any $i \in \{0,\ldots,d-1\}$, by reading some finite binary sequence $u$ on the tape that encodes $a_i$ in binary form, and deleting this prefix from the tape, and by reading some finite binary sequence $1^n0$ on the tape that encodes $n$, and deleting this prefix from the tape. In fact, what we prove is slightly more general, as we show that $\UpPoly$ transforms $(x, G_{i,n}(x), z)$ into $(x, G_{i+1,n}(x), z + a_i G_{i,n}(x))$, for any $x \in [-1,1]$, $z \in \R$, $i,n \in \No$.
\begin{lemma}\label{lem:upoly}
    Let $n,i \in \No$, $a \in \Db_1$, $u := \encDyaSign{a}$, and $v \in \{0,1\}^\#$, $x\in [-1,1]$, and $z \in \R$. Define $T := \len{u} + 2n + 5$ and the configurations $c := (1;|u1^n0v;x,G_{i,n}(x),z,0_{13})$ and $c' := (9; |v; x, G_{i+1,n}(x), z + a G_{i,n}(x), 0_{13})$. Then,
    \begin{equation}
        \UpPoly^T(c) = c', \quad \|c\|_{\UpPoly}^T \leq 1 \vee (|z| + |a|).
    \end{equation}
\end{lemma}
\begin{proof}
    Let $n,i \in \No$, $a \in \Db_1$, $u := \encDyaSign{a}$, $v \in \{0,1\}^\#$, $x\in [-1,1]$, $z \in \R$, and define $y := G_{i,n}(x)$. We set
    \[
        T_\Contr := \len u, \quad T_\Times := 2n+3, \quad \text{and} \quad T := T_\Contr + T_\Times + 2 = \len u + 2n + 5.
    \]
    We define the configurations $c_0,c_1,c_2,c_3,c_4$ of $\UpPoly$ by 
    \begin{equation}\label{eq:1:thm:upoly}
        \begin{cases}
            c_0 &= (1;|u1^n0v;x,y,z,0_{13}),\\
            c_1 &= (2;|u1^n0v;x,y,z,y,0_{12}),\\
            c_2 &= (5;|1^n0v;x,y,z, a y,0_{12}),\\
            c_3 &= (6;|1^n0v;x,y,z+ a y,x,0_{12}),\\
            c_4 &= (9;|v;x,G_{i+1,n}(x),z + a y,0_{13}).
        \end{cases}
    \end{equation}
    \begin{claim*}
        The following four statements hold.
        \begin{enumerate}[label=(\alph*)]
            \item $\UpPoly(c_0) = c_1$ and $\|c_0\|_{\UpPoly}^1 \leq 1\vee|z|$.
            \item $\UpPoly^{T_\Contr}(c_1) = c_2$ and $\|c_1\|_{\UpPoly}^{T_\Contr} \leq 1\vee|z|$.
            \item $\UpPoly(c_2) = c_3$ and $\|c_2\|_{\UpPoly}^1 \leq 1\vee(|z|+|a|)$.
            \item $\UpPoly^{T_\Times}(c_3) = c_4$ and $\|c_3\|_{\UpPoly}^{T_\Times} \leq 1\vee(|z|+|a|)$.
        \end{enumerate}
    \end{claim*}
    \begin{proof}
        We prove each statement in turn.
        \begin{enumerate}[label=(\alph*)]
            \item Assume that the machine is in configuration $c_0$. The machine is in State 1, so after one step, the machine is in configuration $c_1$. Moreover, we have 
            \[\|c_0\|_{\UpPoly}^1 \leq \|(x,y,z,y,0_{12})\|_\infty \leq 1 \vee |z|,\]
            where we used $|x| \leq 1$ and $|y| = |G_{i,n}(x)| \leq 1$ by Lemma \ref{lem:mono_poly_approx}.
            \item Let $c_1' := (1;|u1^n0v;y,0)$, and $c'_2 := (4;|1^n0v;ay,0)$. Then,
            \[
                \Shadow_{(2,3,4,5),(4,5)}(c_1) = c_1', \quad \Lift_{(2,3,4,5),(4,5)}^{(x,y,z,0_{11})}(c_2') = c_2.
            \]
            By Lemma \ref{lem:contr-tmnu-sign}, we have $\Contr^{T_\Contr}(c_1') = c'_2$ and $\|c_1'\|_\Contr^{T_\Contr} \leq |y| \leq 1$. Since $\Contr$ is a subroutine of $\UpPoly$ with correspondance $(2,3,4,5;4,5)$ by Definition \ref{def:upoly}, Lemma \ref{lem:subroutine} yields
            \[
                \UpPoly^{T_\Contr}(c_1) = \Lift_{(2,3,4,5),(4,5)}^{(x,y,z,0_{11})}\left(\Contr^{T_\Contr}(c_1')\right) = c_2,
            \]
            and
            \[
                \|c_1\|_{\UpPoly}^{T_\Contr} \leq \|c_1'\|_\Contr^{T_\Contr} \vee \|c_1\| \leq 1 \vee |z|.
            \]
            \item Assume that the machine is in configuration $c_2$. The machine is in State 5, so after one step, the machine is in configuration $c_3$. Moreover, we have
            \[\|c_2\|_{\UpPoly}^1 \leq \|(x,y,z,ay,0_{12})\|_\infty \vee \|(x,y,z+ay,x,0_{12})\|_\infty \leq 1 \vee (|z| + |a|).\]
            \item Let $c_3' := (1;|1^n0v;y,x,0_{12})$, and $c'_4 := (4;|v; \Rc^\times \Dc (y,x)[n],0_{13}) = (4;|v; G_{i+1,n}(x),0_{13})$. Then,
            \[
                \Shadow_{(6,7,8,9),(2,4,\ldots,16)}(c_3) = c_3', \quad \Lift_{(6,7,8,9),(2,4,\ldots,16)}^{(x,z+ay)}(c_4') = c_4.
            \]
            By Lemma \ref{lem:tmnu-mult}, we have $\Times^{T_\Times}(c_3') = c'_4$ and $\|c_3'\|_\Times^{T_\Times} \leq 1$. Since $\Times$ is a subroutine of $\UpPoly$ with correspondence $(6,7,8,9;2,4,\ldots,16)$ by Definition \ref{def:upoly}, Lemma \ref{lem:subroutine} yields
            \[
                \UpPoly^{T_\Times}(c_3) = \Lift_{(6,7,8,9),(2,4,\ldots,16)}^{(x,z+ay)}\left(\Times^{T_\Times}(c_3')\right) = c_4,
            \]
            and
            \[\|c_3\|_{\UpPoly}^{T_\Times} \leq \|c_3'\|_\Times^{T_\Times} \vee \|c_3\| \overset{(a)}\leq 1 \vee (|z| + |a|),\]
            where (a) is by $|x| \leq 1$, $|G_{i+1,n}(x)| \leq 1$ and $|y| = |G_{i,n}(x)| \leq 1$, where $|G_{j,n}(x)| \leq 1$ for $j = i, i+1$ by Lemma \ref{lem:mono_poly_approx}.
        \end{enumerate}
        This concludes the proof of the claim.
    \end{proof}
    Now, by invoking repeatedly the Chaining Lemma (Lemma \ref{lem:chaining}) together with the claim, and using $c = c_0$ and $c' = c_4$, we obtain
    \[
        \UpPoly^T(c) = c',
    \]
    and
    \[
        \|c\|_{\UpPoly}^T \leq 1 \vee (|z| + |a|).
    \]
    This concludes the proof.
\end{proof}

We finally design a \TMNU that uses the \TMNU $\UpPoly$ as a subroutine to compute $F_{P,n}(x)$, for any polynomial $P$ with dyadic coefficients, any $n \in \No$, and any $x \in [-1,1]$. In order to implement this \TMNU in a way that its neural state remains bounded by $\|P\|_{L^\infty([-1,1])} + \|P\|_0 \|P\|_1 2^{-2n-1}$, we will first compute $F_{P^*,n}(x)$, where $P^* := 2^{-k_P} P$ is the rescaled version of $P$ defined above, and then we will multiply the result by $2^{k_P}$ to obtain $F_{P,n}(x)$. This \TMNU will parse an encoding of the coefficients of $P^\ast$, of $n$ and $k_P$ on its tape, that we formally define as follows.
\newcommand\encPolyExact[1]{\ensuremath{u_{#1}}}
% We first specify the exact tape format for a dyadic polynomial instruction.
\begin{definition}
    Let $P : x \mapsto a_0 x^0 + a_1 x^1 + \cdots + a_d x^d$ be a polynomial with coefficients $a_0, \ldots, a_d \in \Db$, and let $n \in \No$. We define 
    \[
        \encPolyExact{P} := \encDyaSign{a_0^\ast}\cdot 1 \cdot  \encDyaSign{a_1^\ast} \cdot 1 \cdot \encDyaSign{a_2^\ast} \cdots 1 \cdot \encDyaSign{a_d^\ast} \cdot 0 \cdot 1^{k_P}0,
    \]
    and
    \[
        \encPoly{P}{n} := \encDyaSign{a_0^\ast}1^n0 \cdot 1 \cdot  \encDyaSign{a_1^\ast}1^n0 \cdot 1 \cdot \encDyaSign{a_2^\ast}1^n0 \cdots 1 \cdot \encDyaSign{a_d^\ast}1^n0 \cdot 0 \cdot 1^{k_P}0,
    \]
    where $a_0^\ast, \ldots, a_d^\ast \in \Db_1$ are the coefficients of the polynomial $P^\ast$.
\end{definition}

We now formally introduce the \TMNU that computes $F_{P,n}(x)$, for any polynomial $P$ with dyadic coefficients, any $n \in \No$, and any $x \in [-1,1]$. This machine iterates $\UpPoly$ across all coefficients and then rescales the result.
\begin{definition}\label{def:poly-machine}
    We let $\Poly$ be the \TMNU with neural dimension $16$ and 11 states, defined by the following procedure.
    \begin{itemize}
        \item \textbf{State 1-8.} Use $\UpPoly$ (Def. \ref{def:upoly}) as subroutine with correspondence $(1,\ldots,9;1,\ldots,16)$.
        \item \textbf{State 9.} Let $b \in \{\blanksymb,0,1\}$ under scan.
        \begin{enumerate}
            \item[i.] If $b \in \{0,\blanksymb\}$, write $\blanksymb$, move right, update the neural state as
            $
                \neurState \gets (\neurState_1,\neurState_3, 0_{14}),
            $
            and go to State 10.
            \item[ii.] If $b = 1$, write $\blanksymb$, move right and go to State 1.
        \end{enumerate}
        \item \textbf{State 10.} Use $\Scale$ (Def. \ref{def:scale-machine}) as subroutine with correspondence $(10,11;2)$.
        \item \textbf{State 11.} Halt. 
    \end{itemize}
\end{definition}

The next lemma verifies the complete polynomial-evaluation cycle and its bound.
\begin{lemma}\label{lem:poly-machine}
    Let $n \in \No$, $P \in \Db[X]$, $v \in \{0,1\}^\#$, and $x \in [-1,1]$. Define the configurations
    $c := \left(1;|\encPoly{P}{n}v;x,1, 0_{14}\right)$,
    and 
    $c' := \left(11;|v; x, F_{P,n}(x),0_{14}\right)$.
    Define $T := \len{\encPolyExact{P}} + \|P\|_0 (2n + 5)$. Then,
    \begin{equation}
        \Poly^T(c) = c', \quad \|c\|_{\Poly}^T \leq 1 \vee |F_{P,n}(x)|.
    \end{equation}
\end{lemma}
\begin{proof}
    Let $n \in \No$, $P$ a polynomial of degree $d \in \N$ with coefficients $a_0, \ldots, a_d \in \Db$, $v \in \{0,1\}^\#$, and $x \in [-1,1]$. Let $b_0, \ldots, b_d \in \Db_1$ be the coefficients of $P^\ast$, and define the polynomials $P^\ast_0 = 0$, and $P^\ast_1, \ldots, P^\ast_{d+1}$ by 
    \[P^\ast_i : x \mapsto b_0 x^0 + b_1 x^1 + \cdots + b_{i-1} x^{i-1}, \ \quad i \in \{1,\ldots,d+1\}.\] 
    Define also \[u^{(i)} := \encDyaSign{b_i} 1^n 0 \cdot 1 \cdot \encDyaSign{b_{i+1}} 1^n 0\cdots 1 \cdot \encDyaSign{b_{d}} 1^n 0, \ \quad i \in \{0,\ldots,d\},\] and $u^{(d+1)} = \epsilon$. Note that $u^{(0)} \cdot 0 \cdot 1^{k_P}0 = \encPoly{P}{n}$.
    For $t \in \{0,\ldots,d\}$, set
    \[
        T_t := \len{\encDyaSign{b_t}} + 2n + 5,
    \]
    and set $T_\Scale := k_P+1$ and $T := \len{\encPolyExact{P}} + \|P\|_0(2n+5)$.
    We define the configurations $c_t$ of $\Poly$ for $t \in \{0,\ldots,2d+3\}$ by
    \begin{equation}
        \begin{cases}
            c_{2t} &:= \left(1;| u^{(t)}01^{k_P}0v,|;x, G_{t,n}(x), F_{P^\ast_{t},n}(x), 0_{13}\right), \quad t \in \{0,\ldots,d\},\\
            c_{2t+1} &:= \left(9;| 1 u^{(t+1)}01^{k_P}0v,|;x, G_{t+1,n}(x), F_{P^\ast_{t+1},n}(x), 0_{13}\right), \quad t \in \{0,\ldots,d-1\},\\
            c_{2d+1} &:= \left(9;|01^{k_P}0v,|;x, G_{d+1,n}(x), F_{P^\ast,n}(x), 0_{13}\right),\\
            c_{2d+2} &:= \left(10; |1^{k_P}0v; x, F_{P^\ast,n}(x),0_{14}\right),\\
            c_{2d+3} &:= \left(11; |v; x, F_{P,n}(x),0_{14}\right).
        \end{cases}
    \end{equation}
    \begin{claim*}
        The following four statements hold.
        \begin{enumerate}[label=(\alph*)]
            \item For all $t \in \{0,\ldots,d\}$, $\Poly^{T_t}(c_{2t}) = c_{2t+1}$ and $\|c_{2t}\|_{\Poly}^{T_t} \leq 1$.
            \item For all $t \in \{0,\ldots,d-1\}$, $\Poly(c_{2t+1}) = c_{2t+2}$ and $\|c_{2t+1}\|_{\Poly}^1 \leq 1$.
            \item $\Poly(c_{2d+1}) = c_{2d+2}$ and $\|c_{2d+1}\|_{\Poly}^1 \leq 1$.
            \item $\Poly^{T_\Scale}(c_{2d+2}) = c_{2d+3}$ and $\|c_{2d+2}\|_{\Poly}^{T_\Scale} \leq 1 \vee |F_{P,n}(x)|$.
        \end{enumerate}
    \end{claim*}
    \begin{proof}[Proof of the Claim.]
        We prove each statement in turn.
        \begin{enumerate}[label=(\alph*)]
            \item Let $t \in \{0,\ldots,d\}$. Note that we can write $c_{2t}$ and $c_{2t+1}$ as 
            \[
                c_{2t} = (1;|\encDyaSign{b_t}1^n0\tilde v; x, G_{t,n}(x),z,0_{13}),
            \]
            and
            \[
                c_{2t+1} = (9;|\tilde v;x, G_{t+1,n}(x), z + b_t G_{t,n}(x), 0_{13}),
            \]
            where $z := F_{P^\ast_{t},n}(x)$, and $\tilde v := 1u^{(t+1)}01^{k_P}0v$ if $t < d$, and $\tilde v := 01^{k_P}0v$ if $t = d$. Therefore, by Lemma \ref{lem:upoly}, we have $\UpPoly^{T_t}(c_{2t}) = c_{2t+1}$ and
            \begin{align}
                \|c_{2t}\|_{\UpPoly}^{T_t} &\leq 1 \vee (|z| + |b_t|) \overset{(a)}\leq 1 \vee (\|P^\ast_{t}\|_1 + |b_t|) \leq 1 \vee \|P^\ast_{t+1}\|_1 \leq 1 \vee \|P^\ast\|_1 \leq 1,
            \end{align}
            where (a) follows by $|z| = |F_{P^\ast_{t},n}(x)| \leq \|P^\ast_{t}\|_1$ (Lemma \ref{lem:mono_poly_approx}).
            Since $\UpPoly$ is a subroutine of $\Poly$ with correspondence $(1,\ldots,9;1,\ldots,16)$ by Definition \ref{def:poly-machine}, and since $\Shadow_{(1,\ldots,9),(1,\ldots,16)}(c_{2t}) = c_{2t}$ and $\Lift_{(1,\ldots,9),(1,\ldots,16)}^{\varnothing}(c_{2t+1})=c_{2t+1}$, Lemma \ref{lem:subroutine} gives
            \[
                \Poly^{T_t}(c_{2t}) = c_{2t+1}, \quad \|c_{2t}\|_{\Poly}^{T_t} \leq \|c_{2t}\|_{\UpPoly}^{T_t} \vee \|c_{2t}\| \leq 1.
            \]
            \item Let $t \in \{0,\ldots,d-1\}$. In configuration $c_{2t+1}$, the machine is in State 9 and reads the symbol $1$, so after one step, the machine is in configuration $c_{2(t+1)}$. Moreover, we have
            \[\|c_{2t+1}\|_{\Poly}^1 \leq \|c_{2t+1}\| \vee \|c_{2t+2}\| \leq 1.\]
            \item In configuration $c_{2d+1}$, the machine is in State 9 and reads the symbol $0$, so after one step, the machine is in configuration $c_{2d+2}$. Moreover, we have
            \[\|c_{2d+1}\|_{\Poly}^1 \leq \|c_{2d+1}\| \vee \|c_{2d+2}\| \leq 1.\]
            \item Define the configurations $c'_{2d+2} := (1;|1^{k_P}0v;F_{P^\ast,n}(x))$, and $c'_{2d+3} := (2;|v;2^{k_P} F_{P^\ast,n}(x))$. By Lemma \ref{lem:scale-machine}, we have
            \[
                \Scale^{T_\Scale}(c'_{2d+2}) = c'_{2d+3}, \quad \|c'_{2d+2}\|_\Scale^{T_\Scale} \leq |2^{k_P} F_{P^\ast,n}(x)| = |F_{P,n}(x)|.
            \]
            Since $\Scale$ is a subroutine of $\Poly$ with correspondence $(10,11;2)$ by Definition \ref{def:poly-machine}, and since
            \[
                \Shadow_{(10,11),(2)}(c_{2d+2}) = c'_{2d+2}, \quad \Lift_{(10,11),(2)}^{(x,0_{14})}(c'_{2d+3}) = c_{2d+3},
            \]
            Lemma \ref{lem:subroutine} gives
            \[
                \Poly^{T_\Scale}(c_{2d+2}) = c_{2d+3}, \quad \|c_{2d+2}\|_{\Poly}^{T_\Scale} \leq \|c'_{2d+2}\|_\Scale^{T_\Scale} \vee \|c_{2d+2}\| \leq 1 \vee |F_{P,n}(x)|.
            \]
        \end{enumerate}
        This concludes the proof of the claim.
    \end{proof}
    By invoking repeatedly the Chaining Lemma \ref{lem:chaining} together with the claim, and using $c=c_0$ and $c'=c_{2d+3}$, we obtain
    \[
        \Poly^T(c) = c',
    \]
    where
    \begin{align*}
        T &= \sum_{t=0}^d T_t + d + 1 + T_\Scale\\
        &= \sum_{t=0}^d \len{\encDyaSign{b_t}} + (d+1)(2n+5) + d + 1 + k_P + 1\\
        &= \len{\encPolyExact{P}} + \|P\|_0(2n+5).
    \end{align*}
    Moreover,
    \[
        \|c\|_{\Poly}^T \leq 1 \vee |F_{P,n}(x)|.
    \]
\end{proof}

\subsection{Approximation of continuous functions by \TMNUs}
The final piece is to construct a \TMNU that, given the encoding of a sequence $\mathbf{P} := (P_i)_{i\in\No} \in \Db[X]$ of polynomials with dyadic coefficients, a sequence of natural numbers $\mathbf{n} := (n_i)_{i \in \No}$ initially written on its tape, and its neural state initially containing some $x \in [-1,1]$, computes successively $F_{P_0,n_0}, F_{P_1,n_1}, F_{P_2,n_2}, \ldots$. More precisely, we encode the sequence $\mathbf{P}$ and $\mathbf{n}$ on the tape of the machine as follows.
\newcommand\encSeq[2]{\ensuremath{\langle #1, #2 \rangle}}
% The sequence encoding is just the concatenation of the individual polynomial instructions.
\begin{definition}
    Let $\mathbf{P} := (P_i)_{i\in\No} \in \Db[X]$ be a sequence of polynomials with dyadic coefficients, and $\mathbf{n} := (n_i)_{i \in \No}$ be a sequence of natural numbers. We define the encoding of $\mathbf{P}$ and $\mathbf{n}$ by
    \[
        \encSeq{\mathbf{P}}{\mathbf{n}} := \encPoly{P_0}{n_0}\encPoly{P_1}{n_1}\encPoly{P_2}{n_2}\cdots.
    \]
    For every $k \in \No$, we let $\mathbf{P}_{\geq k} := (P_{i+k})_{i \in \No}$ and $\mathbf{n}_{\geq k} := (n_{i+k})_{i \in \No}$. Note that
    \[
        \encSeq{\mathbf{P}_{\geq k}}{\mathbf{n}_{\geq k}} = \encPoly{P_k}{n_k}\encPoly{P_{k+1}}{n_{k+1}}\cdots.
    \]
\end{definition}

This \TMNU has neural dimension of 17, and uses the 16 first dimensions to simulate the \TMNU $\Poly$, and the last dimension is designed to store the value of the last computation of $\Poly$. The formal definition of such a machine is given as follows.
% The machine below repeatedly calls $\Poly$ and keeps the last completed output available.
\begin{definition}\label{def:continuous-machine}
    We let $\Continuous$ be the \TMNU with neural dimension $17$ and 12 states defined by the following procedure.
    \begin{itemize}
        \item \textbf{State 1.} Update the neural state as
        $
                \neurState \gets (\neurState_1,1, 0_{14}, \neurState_2),
            $
            and go to State 2.
        \item \textbf{State 2-11.} Use $\Poly$ (Def. \ref{def:poly-machine}) as subroutine with correspondence $(2,\ldots,11,1;1,\ldots,16)$.
        \item \textbf{State 12.} Halt.
    \end{itemize}
\end{definition}

The following lemma describes one full cycle of this infinite polynomial-evaluation machine.
\begin{lemma}\label{lem:continuous-machine}
    Let $\mathbf{P} := (P_i)_{i\in \No}$ be a sequence of polynomials, $\mathbf{n} := (n_i)_{i\in\No}$ be a sequence of natural numbers, and $x \in [-1,1]$. For all $i \in \No$, define the configuration
    \[
        c_i := \left(1;|\encSeq{\mathbf{P}_{\geq i}}{\mathbf{n}_{\geq i}};x, F_{P_{i-1},n_{i-1}}(x), 0_{14}, F_{P_{i-2},n_{i-2}}(x)\right),
    \]
    with the convention that $F_{P_{-1},n_{-1}}(x) := F_{P_{-2},n_{-2}}(x) := 0$. Note that, in particular, $c_0 := (1;|\encSeq{\mathbf{P}}{\mathbf{n}};x, 0_{16})$.
    Define
    \begin{equation}
        T_i := \len{\encPolyExact{P_i}} + \|P_i\|_0(2 n_i + 5) + 1.
    \end{equation}
    Then,
    \begin{equation}
        \Continuous^{T_i}(c_i) = c_{i+1}, \quad \|c_i\|_{\Continuous}^{T_i} \leq 1 \vee \sup_{j \in \No} |F_{P_j,n_j}(x)| .
    \end{equation}
    In particular, by denoting $N_i := \sum_{j=0}^{i-1} T_j$, we have $\Continuous^{N_i}(c_0) = c_i$, and
    \begin{equation}\label{eq:output_tmnu:lem:continuous-machine}
       {\Continuous}^{\encSeq{\mathbf{P}}{\mathbf{n}}}x[t] = F_{P_{i-1},n_{i-1}}(x), \quad \text{for all } t \in \{N_i+1, \ldots, N_{i+1}\}, \ i \in \No.
    \end{equation}
\end{lemma}
\begin{proof}
    Let $\mathbf{P} := (P_i)_{i\in \No}$ be a sequence of polynomials, $\mathbf{n} := (n_i)_{i\in\No}$ be a sequence of natural numbers, and $x \in [-1,1]$. For all $i \in \No$, define the configuration
    \[
        c_i := \left(1;|\encSeq{\mathbf{P}_{\geq i}}{\mathbf{n}_{\geq i}};x,F_{P_{i-1},n_{i-1}}(x), 0_{14},  F_{P_{i-2},n_{i-2}}(x)\right).
    \]
    We also define the configuration $\tilde c_i$ by 
    \[\tilde c_i := \left(2; |\encSeq{\mathbf{P}_{\geq i}}{\mathbf{n}_{\geq i}};x,1,0_{14},F_{P_{i-1},n_{i-1}}(x)\right).\]
    Fix $i \in \No$, and set
    \[
        S_i := \len{\encPolyExact{P_i}} + \|P_i\|_0(2n_i+5), \quad \text{and} \quad T_i := S_i+1.
    \]
    If $\Continuous$ is in configuration $c_i$, then by definition it goes to configuration $\tilde c_i$ in one transition. Therefore,
    \[
        \Continuous(c_i) = \tilde c_i,
    \]
    and
    \begin{align*}
        \|c_i\|_{\Continuous}^1 &\leq \|(x,F_{P_{i-1},n_{i-1}}(x),0_{14},  F_{P_{i-2},n_{i-2}}(x))\|_\infty \vee \|(x,1,0_{14},  F_{P_{i-1},n_{i-1}}(x))\|_\infty\\
        & \overset{(a)}\leq 1 \vee |F_{P_{i-2},n_{i-2}}(x)| \vee |F_{P_{i-1},n_{i-1}}(x)|,
    \end{align*}
    where (a) follows from $|x| \leq 1$.

    We now analyze the $\Poly$ subroutine started from $\tilde c_i$. Define the configurations $\tilde c'_i$ and $c'_{i+1}$ by 
    \[\tilde c'_i := \left(1;|\encSeq{\mathbf{P}_{\geq i}}{\mathbf{n}_{\geq i}};x,1, 0_{14}\right),\]
    and 
    \[
        c'_{i+1} := \left(11; |\encSeq{\mathbf{P}_{\geq i+1}}{\mathbf{n}_{\geq i+1}};x,F_{P_i,n_i}(x),0_{14}\right).
    \]
    Note that $\tilde c'_i$ can be rewritten as $\tilde c'_i = (1;|\encPoly{P_i}{n_i}\encSeq{\mathbf{P}_{\geq i+1}}{\mathbf{n}_{\geq i+1}};x,1,0_{14})$. Therefore, by Lemma \ref{lem:poly-machine}, we have 
    \[
        \Poly^{S_i}(\tilde c'_i) = c'_{i+1}, \quad \|\tilde c'_i\|_{\Poly}^{S_i} \leq 1 \vee |F_{P_i,n_i}(x)|.
    \]
    Since $\Poly$ is a subroutine of $\Continuous$ with correspondence $(2,\ldots,11,1;1,\ldots,16)$ by Definition \ref{def:continuous-machine}, and since
    \[
        \Shadow_{(2,\ldots,11,1),(1,\ldots,16)}(\tilde c_i) = \tilde c'_i, \quad \Lift_{(2,\ldots,11,1),(1,\ldots,16)}^{F_{P_{i-1},n_{i-1}}(x)}(c'_{i+1}) = c_{i+1},
    \]
    Lemma \ref{lem:subroutine} yields
    \[
        \Continuous^{S_i}(\tilde c_i) = c_{i+1},
    \]
    and, for all $t \in \{0,\ldots,S_i\}$,
    \begin{equation}\label{eq:traj_neur_1:proof:lem:continuous-machine}
        \proj_{17} \proj_\neurState {\Continuous}^t(\tilde c_i) = F_{P_{i-1},n_{i-1}}(x).
    \end{equation}
    Moreover, by Lemma \ref{lem:subroutine}, we have
    \[
        \|\tilde c_i\|_{\Continuous}^{S_i} \leq \|\tilde c'_i\|_{\Poly}^{S_i} \vee \|\tilde c_i\| \leq 1 \vee |F_{P_{i-1},n_{i-1}}(x)| \vee |F_{P_i,n_i}(x)|.
    \]
    By the Chaining Lemma \ref{lem:chaining}, we obtain
    \[
        \Continuous^{T_i}(c_i) = c_{i+1},
    \]
    and
    \[
        \|c_i\|_{\Continuous}^{T_i} \leq 1 \vee |F_{P_{i-2},n_{i-2}}(x)| \vee |F_{P_{i-1},n_{i-1}}(x)| \vee |F_{P_i,n_i}(x)| \leq 1 \vee \sup_{j \in \No} |F_{P_j,n_j}(x)|.
    \]

    We now move to the proof of the last statement. For $i \in \No$, define $N_i := \sum_{j=0}^{i-1} T_j$. By induction,
    \begin{equation}\label{eq:traj_config_0:proof:lem:continuous-machine}
        {\Continuous}^{N_i}(c_0) = c_i, \quad \text{for all } i \in \No.
    \end{equation}
    Now, fix $i \in \No$ and $t \in \{N_i+1, \ldots, N_{i+1}\}$. Then $s := t-N_i-1$ belongs to $\{0,\ldots,S_i\}$, and
    \begin{align*}
        \Continuous^t(c_0) &= \Continuous^s(\Continuous(\Continuous^{N_i}(c_0)))
         \overset{\eqref{eq:traj_config_0:proof:lem:continuous-machine}}= \Continuous^s(\Continuous(c_i)) 
         \overset{(a)}= \Continuous^s(\tilde c_i),
    \end{align*}
    where (a) is by $\Continuous(c_i)=\tilde c_i$. Therefore, by Definition \ref{def:tmnu-as-operator},
    \begin{align*}
        {\Continuous}^{\encSeq{\mathbf{P}}{\mathbf{n}}}x[t] &= \proj_{17} \proj_\neurState {\Continuous}^t(c_0) = \proj_{17} \proj_\neurState {\Continuous}^s(\tilde c_i) \overset{(a)}= F_{P_{i-1},n_{i-1}}(x),
    \end{align*}
    where (a) is by \eqref{eq:traj_neur_1:proof:lem:continuous-machine}. This concludes the proof of the last statement.
\end{proof}

We finally close this section by showing that the \TMNU $\Continuous$ can approximate uniformly any continuous function $f : [-1,1] \to \R$, and is uniformly bounded over its computations. This gives the qualitative approximation theorem at the \TMNU level.

\begin{theorem}\label{thm:continuous-approximation-tmnu}
    For every continuous function $f : [-1,1] \to \R$, there exists $u \in \{0,1\}^\N$ such that
    \[
        \lim_{t \to \infty} \|\Continuous^u(x)[t] - f(x)\|_{L^{\infty}([-1,1])} = 0,
    \]
    and 
    \[
        (1;|u;x,0_{16}) \in \Bc_\Continuous^\infty \left(1 + \|f\|_{L^\infty([-1,1])}\right).
    \]
\end{theorem}
\begin{proof}
    Let $f: [-1,1] \to \R$ be a continuous function. By the Weierstrass approximation theorem and density of dyadic numbers in $\R$, there exists a sequence of polynomials with dyadic coefficients $(P_i)_{i\in\No} \in \Db[X]$ such that $\|f - P_i\|_{L^\infty([-1,1])} \leq 2^{-i-1}$ for all $i \in \No$. Further, we define $\mathbf{n} := (n_i)_{i \in \No} := (i + \lceil (k_{P_i} + \log_2 \|P_i\|_0)/2 \rceil)_{i \in \No}$. Note that
    \begin{equation}\label{eq:approx_poly:proof:thm:continuous-approximation}
        |F_{P_i,n_i} - P_i(x)| \overset{(a)}\leq \|P_i\|_1 \|P_i\|_0 2^{-2n_i-1} \leq \|P_i\|_1 \|P_i\|_0 2^{-2i - k_{P_i} - \log_2\|P_i\|_0 -1} \leq 2^{-2i-1},
    \end{equation}
    where (a) follows by Lemma \ref{lem:mono_poly_approx}. By Lemma \ref{lem:continuous-machine}, we have that for every $i \in \No$ and $t \in \{N_i+1, \ldots, N_{i+1}\}$,
    \begin{align*}
        |\Continuous^{\encSeq{\mathbf{P}}{\mathbf{n}}}x[t] - f(x)| &\leq |\Continuous^{\encSeq{\mathbf{P}}{\mathbf{n}}} x[t] - P_i(x)| + |P_i(x) - f(x)| \leq |F_{P_i,n_i} - P_i(x)| + 2^{-i-1} \overset{\eqref{eq:approx_poly:proof:thm:continuous-approximation}}\leq 2^{-i}.
    \end{align*} 
    Upon noting that $t \geq N_i + 1 \iff t \in \cup_{j \geq i} \{N_j + 1, \ldots, N_{j+1}\}$, we get that for every $i \in \No$, $t \geq N_i + 1$,
    \[
        |\Continuous^{\encSeq{\mathbf{P}}{\mathbf{n}}}x[t] - f(x)| \leq \sup_{j \geq i} 2^{-j} = 2^{-i}, \ \text{for all} \ x \in [-1,1].
    \]
For the second statement, set $C_f := 1 + \|f\|_{L^\infty([-1,1])}$. First note that, for every $x \in [-1,1]$,
\begin{align*}
    1 \vee \sup_{i \in \No} |F_{P_i,n_i}(x)| &\leq 1 \vee \sup_{i \in \No} \left(|F_{P_i,n_i}(x) - P_i(x)| + |P_i(x) - f(x)| + |f(x)|\right)\\
    &\leq 1 \vee \sup_{i \in \No} (2^{-2i-1} + 2^{-i-1} + |f(x)|) \leq C_f.
\end{align*}
Let $c_0 := (1;|\encSeq{\mathbf{P}}{\mathbf{n}};x,0_{16})$. With the notation of Lemma \ref{lem:continuous-machine}, we have $\Continuous^{N_i}(c_0)=c_i$ and $\|c_i\|_{\Continuous}^{T_i}\leq C_f$ for every $i\in\No$. Hence, for every $t\in\No$, either $t=0$, in which case $\|\Continuous^t(c_0)\|=\|c_0\|\leq 1\leq C_f$, or there exist $i\in\No$ and $s\in\{1,\ldots,T_i\}$ such that $t=N_i+s$, and then
\[
    \|\Continuous^t(c_0)\|=\|\Continuous^s(c_i)\|\leq \|c_i\|_{\Continuous}^{T_i}\leq C_f.
\]
Therefore, $c_0 \in \Bc_\Continuous^\infty(C_f)$.
This concludes the proof.
\end{proof}

Note that in the above proof, we do not know a priori the properties sequence of approximating polynomials given by the Weierstrass approximation theorem, therefore, the speed of convergence of the \TMNU $\Continuous$ to $f$ is not known. In the remainder of this section, we analyze the convergence rate of $\Continuous$ to $f$ when $f$ is a polynomial with arbitrary real coefficients. 
We first  extend the definitions given for dyadic polynomials to polynomials with real coefficients. Specifically, given $P := a_0 + a_1 X + a_2 X^2 + \ldots + a_d X^d \in \R[X]$, we define $||P\|_0 := d + 1$, $\|P\|_1 := \sum_{i=0}^d |a_i|$, $k_P := \lfloor \log_2 \|P\|_1\rfloor + 1$ and $P^\ast := 2^{-k_P}P$. Note that $\|P\|_1 < 2^{k_P}$, and hence $\|P^\ast\|_1 < 1$.

For $a \in [0,1]$ and $n \in \No$, we define the dyadic approximation of $a$ at precision $n$ by
\[\approxRea{a}{n} := s(a) \frac{\lfloor 2^n |a| \rfloor}{2^n} \in 2^{-n} \Z \subseteq \Db.\]
Note that 
\begin{equation}\label{eq:approx-rea}
    |a - 2^{-n}| < |\approxRea{a}{n}| \leq |a|.
\end{equation}
Given $P = a_0 + a_1 X + a_2 X^2 + \ldots + a_d X^d \in \R[X]$ and $n \in \No$, we define the dyadic approximation of $P$ at precision $n$ by $\approxPoly{P}{n} := \sum_{j=0}^d \approxRea{(a_j)}{n} X^j \in 2^{-n} \Z[X]$. Note that
\begin{equation}\label{eq:approx-poly-dyadic-inequalities}
    \|\approxPoly{P}{n} - P\|_{L^\infty([-1,1])} \leq \|P\|_1 (d+1) 2^{-n},  \|\approxPoly{P}{n}\|_1 \leq \|P\|_1 \quad \text{and} \quad \|\approxPoly{P}{n}\|_0 \leq \|P\|_0.
\end{equation}

Before proving polynomial rates, we need a simple bound on the length of the polynomial encoding.
\begin{lemma}\label{lem:len-enc-poly-exact-dyadic-length}
    Let $n \in \No$, and $P \in 2^{-n}\Z[X]$ be a polynomial of degree $d$. Then, we have
    \[
        \len{\encPolyExact{P}} \leq (d+1)(2(n + k_P) + 2) + 1 + k_P.
    \]
\end{lemma}
\begin{proof}
    Let $n \in \No$, and $P \in 2^{-n}\Z[X]$ be a polynomial of degree $d$. Then, $P^\ast = 2^{-k_P} P \in 2^{-n-k_P}\Z[X] \cap \Db_1[X]$. Now, for every $a \in 2^{-n-k_P}\Z \cap \Db_1$, we have $|a| \in 2^{-n-k_P}\Z \cap \Db^+_1$, and hence $\len{\encDya{|a|}} \leq n + k_P$, which implies that 
    \begin{equation}\label{eq:len-enc-dya-sign}
        \len{\encDyaSign{a}} = 2 \len{\encDya{|a|}} + 2 \leq 2(n + k_P) + 2.
    \end{equation}
    Therefore, by definition of $\encPolyExact{P}$, we have
    \[
        \len{\encPolyExact{P}} = \sum_{j=0}^d \len{\encDyaSign{b_j}} + 2(d+1) + 1 + k_P \leq (d+1)(2(n + k_P) + 2) + 1 + k_P.
    \]
\end{proof}

We now are ready to prove the main result of this section, which gives an upper bound on the convergence rate of $\Continuous$ to a polynomial $P$ with real coefficients.
% The theorem turns the dyadic polynomial evaluator into a quantitative rate for arbitrary real coefficients.
\begin{theorem}\label{thm:convergence-rate-polynomial}
    Let $P \in \R[X]$ be a polynomial of degree $d$. Then, there exists $u \in \{0,1\}^\N$ such that for all $x \in [-1,1]$ and $t \geq \tau_P$,
    \[
        \left|\Continuous^u x[t] - P(x)\right| \leq 2 \cdot 2^{-\frac{t}{t_P}},
    \]
    where $t_P := 18(d+1)$ and $\tau_P := t_P(6k_P + d + 11) + 1$, and
    \[
        (1;|u;x,0_{16}) \in \Bc_\Continuous^\infty\left(1 + \|P\|_{L^\infty([-1,1])}\right), \quad \text{for all } x \in [-1,1].
    \]
\end{theorem}
\begin{proof}
    Let $P \in \R[X]$ be a polynomial of degree $d$. Define $\mathbf{n} := (n_i)_{i\in\No}$ by $n_i := 2^{i+1} +  k_{P} + \lceil \log_2 (d+1) \rceil + 1$ and $\mathbf{P} := (P_i)_{i\in\No}$ by $P_i := \approxPoly{P}{n_i}$. Set $u := \encSeq{\mathbf{P}}{\mathbf{n}}$. Note that
    \begin{align}
        |F_{P_{i-1},n_{i-1}}(x) - P(x)| &\leq |F_{P_{i-1},n_{i-1}}(x) - P_{i-1}(x)| + |P_{i-1}(x) - P(x)| \nonumber\\
        &\overset{(a)}\leq \|P_{i-1}\|_1 \|P_{i-1}\|_0 2^{-2n_{i-1}-1} + \|P\|_1 (d+1) 2^{-n_{i-1}} \nonumber\\
        &\leq 2\|P\|_1 (d+1) 2^{-n_{i-1}} \leq  2^{- 2^i} \label{eq:approx_poly:proof:thm:convergence-rate-polynomial}
    \end{align}
    for every $i \in \No$ and $x \in [-1,1]$, where (a) follows by Lemma \ref{lem:mono_poly_approx}. 
    By Lemma \ref{lem:continuous-machine}, there exists a nondecreasing sequence $(N_i)_{i \in \No} \in \N$ such that 
    \begin{equation}
       |\Continuous^{u}x[t] - P(x)| = |F_{P_{i-1},n_{i-1}}(x) - P(x)| \leq 2^{-2^i}, \quad \text{for all } t \in \{N_i+1, \ldots, N_{i+1}\}, \ i \in \No,
    \end{equation}
    which upon noting that $t \geq N_i + 1 \iff t \in \bigcup_{j \geq i} \{N_j + 1, \ldots, N_{j+1}\}$, delivers that for all $t \geq N_i + 1$,
    \[
        |\Continuous^{u}x[t] - P(x)| \leq \sup_{j \geq i} 2^{-2^j} =  2^{-2^i}.
    \]
    Moreover, Lemma \ref{lem:continuous-machine} also delivers that for all $i \in \No$,
    \begin{align*}
        N_i &\leq \sum_{j=0}^{i-1} \left( \len{\encPolyExact{P_j}} + \|P_j\|_0(2n_j + 5) + 1\right)
        % &\overset{(a)}\leq \sum_{j=0}^{i-1} \left( (d+1)(2(2^{j+1} + k_{P_j}) + 1) + 1 + k_{P_j} + \|P_j\|_0(2\cdot 2^j + k_{P} + \lceil \log_2 (d+1) \rceil + 1 + 5) + 1\right)\\
        % &\overset{(b)}\leq \sum_{j=0}^{i-1} \left( (d+1)(2(2^j + k_{P}) + 1) + 1 + k_{P} + (d+1)(2\cdot 2^j + k_P + d + 8) + 1\right)\\
        % & = \sum_{j=0}^j \left( (d+1)2^{j+2} + (d+1)(2k_P + 6) + k_P + 2\right)\\
        % &= 4(d+1)(2^i -1) + i((d+1)(2k_P + 6) + k_P + 2)\\
        % &< 4(d+1)2^i + i(d+1)(2k_P + 7) =: \beta 2^i + \gamma i,
    \end{align*}
    where by noting that $P_j \in 2^j \Z$ and applying Lemma \ref{lem:len-enc-poly-exact-dyadic-length}, we get
    \[
    \len{\encPolyExact{P_j}} \leq (d+1)(2(n_j + k_{P_j}) + 2) + 1 + k_{P_j},
    \]
    and hence, by injecting $n_j \leq 2^{j+1} + k_P + d + 2$, and noting that $k_{P_j} \leq k_P$ and $\|P_j\|_0 \leq d+1$, we get
    \begin{align*}
        N_i &\leq \sum_{j=0}^{i-1} \left( (d+1)\left(4\left(2^{j+1} + k_P + d + 2\right) + k_P + 7\right) + 2 + k_P\right)\\
        & \leq \sum_{j=0}^{i-1} (d+1)\left(8 \cdot 2^j + 6k_P + d + 11\right) \leq 9(d+1)2^i,
    \end{align*}
    for $i \geq \alpha_P := \log_2(6k_P + d + 11)$.
    Now, we let $\tilde t_P := 9(d+1)$, and note that for every 
    \[
        t \geq \tau_P := \tilde t_P 2^{\alpha_P+1} + 1 = 18(d+1)(6k_P + d + 11) + 1,
    \]
    we have $\lfloor \log_2((t-1)/\tilde t_P) \rfloor \geq \lfloor \alpha_P+1 \rfloor \geq \alpha_P$, and hence
    \[
        N_{\lfloor \log_2((t-1)/\tilde t_P) \rfloor} + 1 \leq \tilde t_P 2^{\lfloor \log_2((t-1)/\tilde t_P) \rfloor} + 1 \leq \tilde t_P 2^{\log_2((t-1)/\tilde t_P)} + 1= t,
    \]
    so that, by \eqref{eq:approx_poly:proof:thm:convergence-rate-polynomial},
    \[
        |\Continuous^u x[t] - P(x)| \leq 2^{-2^{\lfloor \log_2((t-1)/\tilde t_P) \rfloor}} \leq 2^{-2^{ \log_2((t-1)/\tilde t_P) - 1}} = 2^{-\frac{t-1}{2\tilde t_P}} = 2^{-\frac{t}{t_P}} \cdot 2^{\frac{1}{t_P}} \leq 2 \cdot 2^{-\frac{t}{t_P}}.
    \]
   Finally, set $C_P := 1 + \|P\|_{L^\infty([-1,1])}$ and fix $x \in [-1,1]$. By \eqref{eq:approx_poly:proof:thm:convergence-rate-polynomial}, applied with $i+1$ in place of $i$, we have
   \[
       \sup_{i \in \No} |F_{P_i,n_i}(x) - P(x)| \leq 1.
   \]
   Hence,
   \[
       1 \vee \sup_{i \in \No} |F_{P_i,n_i}(x)| \leq 1 \vee \left(1 + |P(x)|\right) \leq C_P.
   \]
   Let $c_0 := (1;|u;x,0_{16})$. With the notation of Lemma \ref{lem:continuous-machine}, we have $\Continuous^{N_i}(c_0)=c_i$ and $\|c_i\|_{\Continuous}^{T_i}\leq C_P$ for every $i\in\No$. Therefore, for every $t\in\No$, either $t=0$, in which case $\|\Continuous^t(c_0)\|=\|c_0\|\leq 1\leq C_P$, or there exist $i\in\No$ and $s\in\{1,\ldots,T_i\}$ such that $t=N_i+s$, and then
   \[
       \|\Continuous^t(c_0)\|=\|\Continuous^s(c_i)\|\leq \|c_i\|_{\Continuous}^{T_i}\leq C_P.
   \]
   This proves that $c_0 \in \Bc_\Continuous^\infty(C_P)$.
\end{proof}

\subsection{Native evaluation of Chebyshev expansions}

% Let $(T_n)_{n \in \No}$ be the sequence of Chebyshev polynomials defined by $T_0 := 1$, $T_1 := X$, and for all $n \geq 1$, $T_{n+1} := 2X T_n - T_{n-1}$. It is known that Chebyshev polynomials can be used to approximate Dini-Lipschitz continuous functions.
% We recall the classical convergence statement in the form used below.
% \begin{theorem}(Chebyshev series)\label{thm:chebychev-series}\cite[Theorem 3.1]{trefethen2019approximation}
%     Let $f : [-1,1] \to \R$ be a Lipschitz continuous function. Then,
%     \begin{equation}\label{eq:chebychev-series:thm:chebychev-series}
%         \limi{n} \|f - S^{(f)}_n\|_{L^\infty([-1,1])} = 0
%     \end{equation}
%     where
%     \begin{equation}\label{eq:main:def:chebychev-series-truncation}
%         S^{(f)}_n(x) := \sum_{j=0}^n c^{(f)}_j T_j(x), \quad x \in [-1,1], n \in \No,
%     \end{equation}
%     such that
%     \begin{equation}\label{eq:chebychev-coefficients:thm:chebychev-series}
%         c^{(f)}_0 := \frac{1}{\pi} \int_{-1}^1 \frac{f(x)T_0(x)}{\sqrt{1-x^2}} dx, \quad \text{and} \quad c^{(f)}_j := \frac{2}{\pi} \int_{-1}^1 \frac{f(x) T_j(x)}{\sqrt{1-x^2}} dx, \quad j \in \N.
%     \end{equation}
% \end{theorem}

In this section, we show how to implement a native evaluation of Chebyshev expansions using the \TMNU $\Times$ as a subroutine. The main idea is to use the recurrence relation of Chebyshev polynomials to evaluate the expansion without explicitly computing the polynomials themselves. Namely, we use the recurrence relation
\begin{equation}\label{eq:recurrence-chebychev-polynomials-appendix}
    T_0(x) = 1, \quad T_1(x) = x, \quad T_{n+1}(x) = 2xT_n(x) - T_{n-1}(x), \quad n \in \No,
\end{equation}
to compute the Chebyshev polynomials iteratively. This allows us to evaluate the Chebyshev partial sums $S^{(f)}_n(x) = \sum_{j=0}^n c^{(f)}_j T_j(x)$ efficiently, using the coefficients $c^{(f)}_j$ obtained from the Chebyshev series representation of the function $f$. 

Before proceeding, we introduce some notation to simplify the presentation.For $n\in\No$, write
\[
    \operatorname{Mult}_n(x,y):=\Rc^\times\Dc(x,y)[n],
    \qquad (x,y)\in[-1,1]^2,
\]
and define the approximate normalized Chebyshev sequence by
\begin{equation}\label{eq:def-native-chebychev-sequence}
    Q_{0,n}(x):=\frac12,\qquad
    Q_{1,n}(x):=\frac{x}{2},\qquad
    Q_{k+1,n}(x):=2\operatorname{Mult}_n(x,Q_{k,n}(x))-Q_{k-1,n}(x).
\end{equation}
The normalization by $1/2$ leaves room for the numerical error while calling $\Times$. The next machine implements one step of this normalized recurrence.
\begin{definition}\label{def:cheb-step-machine}
    We let $\ChebStep$ be the \TMNU with neural dimension $17$ and $6$ states defined by the following procedure.
    \begin{itemize}
        \item \textbf{State 1.} Update the neural state as
        \[
            \neurState\gets
            (\neurState_1,\neurState_2,\neurState_3,
            \neurState_1,\neurState_3,0_{12}),
        \]
        and go to State $2$.
        \item \textbf{States 2--5.} Use $\Times$ as a subroutine with correspondence $(2,3,4,5;4,\ldots,17)$.
        \item \textbf{State 5.} Update the neural state as
        \[
            \neurState\gets
            (\neurState_1,\neurState_3,2\neurState_4-\neurState_2,0_{14}),
        \]
        and go to State $6$.
        \item \textbf{State 6.} Halt.
\end{itemize}
\end{definition}

We first verify the effect and boundedness of a single Chebyshev step.
\begin{lemma}\label{lem:cheb-step-machine}
    Let $n\in\No$, $v\in\{0,1\}^{\#}$, $x,a,b\in[-1,1]$, and define
    \[
        \widetilde b:=2\operatorname{Mult}_n(x,b)-a.
    \]
    For the configurations
    \[
        c:=(1;|1^n0v;x,a,b,0_{14}),
        \qquad
        c':=(6;|v;x,b,\widetilde b,0_{14}),
    \]
    and $T:=2n+5$, we have
    \[
        \ChebStep^T(c)=c',
        \qquad
        \|c\|_{\ChebStep}^T\leq 3.
    \]
\end{lemma}
\begin{proof}
    Let
    \[
        T_\Times:=2n+3,
        \qquad
        T:=T_\Times+2=2n+5.
    \]
    We define the following four configurations of $\ChebStep$:
    \begin{equation}\label{eq:configurations:proof:lem:cheb-step-machine}
        \begin{cases}
            c_0&:=(1;|1^n0v;x,a,b,0_{14}),\\
            c_1&:=(2;|1^n0v;x,a,b,x,b,0_{12}),\\
            c_2&:=(5;|v;x,a,b,\operatorname{Mult}_n(x,b),0_{13}),\\
            c_3&:=(6;|v;x,b,2\operatorname{Mult}_n(x,b)-a,0_{14}).
        \end{cases}
    \end{equation}
    Note that $c_0=c$ and $c_3=c'$. We prove separately the three stages of the computation.

    \begin{claim*}
        The following statements hold:
        \begin{enumerate}[label=(\alph*)]
            \item $\ChebStep(c_0)=c_1$ and $\|c_0\|_{\ChebStep}^1\leq1$.
            \item $\ChebStep^{T_\Times}(c_1)=c_2$ and
            $\|c_1\|_{\ChebStep}^{T_\Times}\leq1$.
            \item $\ChebStep(c_2)=c_3$ and
            $\|c_2\|_{\ChebStep}^1\leq3$.
        \end{enumerate}
    \end{claim*}
    \begin{proof}[Proof of the Claim.]
        We prove each statement in turn.
        \begin{enumerate}[label=(\alph*)]
            \item In configuration $c_0$, the machine is in State $1$. Therefore, by Definition \ref{def:cheb-step-machine}, it preserves the first three neural coordinates, copies $x$ and $b$ into coordinates $4$ and $5$, sets all remaining coordinates to zero, and goes to State $2$. Hence,
            \[
                \ChebStep(c_0)=c_1.
            \]
            Moreover, since $x,a,b\in[-1,1]$,
            \[
                \|c_0\|_{\ChebStep}^1
                \leq
                \|(x,a,b,0_{14})\|_\infty
                \vee
                \|(x,a,b,x,b,0_{12})\|_\infty
                \leq1.
            \]

            \item Define the configurations of $\Times$
            \[
                c_1':=(1;|1^n0v;x,b,0_{12}),
                \qquad
                c_2':=(4;|v;\operatorname{Mult}_n(x,b),0_{13}).
            \]
            By Lemma \ref{lem:tmnu-mult},
            \[
                \Times^{T_\Times}(c_1')=c_2',
                \qquad
                \|c_1'\|_\Times^{T_\Times}\leq1.
            \]
            By Definition \ref{def:cheb-step-machine}, $\Times$ is a subroutine of $\ChebStep$ with correspondence
            $(2,3,4,5;4,\ldots,17)$. Furthermore,
            \[
                \Shadow_{(2,3,4,5),(4,\ldots,17)}(c_1)=c_1',
            \]
            and
            \[
                \Lift_{(2,3,4,5),(4,\ldots,17)}^{(x,a,b)}(c_2')=c_2.
            \]
            Therefore, Lemma \ref{lem:subroutine} yields
            \[
                \ChebStep^{T_\Times}(c_1)
                =
                \Lift_{(2,3,4,5),(4,\ldots,17)}^{(x,a,b)}
                \left(\Times^{T_\Times}(c_1')\right)
                =
                c_2,
            \]
            and
            \[
                \|c_1\|_{\ChebStep}^{T_\Times}
                \leq
                \|c_1'\|_\Times^{T_\Times}\vee\|c_1\|
                \leq1.
            \]

            \item In configuration $c_2$, the machine is in State $5$. By Definition \ref{def:cheb-step-machine}, it preserves $x$, moves $b$ into the second coordinate, places
            $2\operatorname{Mult}_n(x,b)-a$ in the third coordinate, sets the remaining coordinates to zero, and goes to State $6$. Thus,
            \[
                \ChebStep(c_2)=c_3.
            \]
            Lemma \ref{lem:multiplication_rnn} gives
            $|\operatorname{Mult}_n(x,b)|\leq1$. Consequently,
            \begin{align*}
                \|c_2\|_{\ChebStep}^1
                &\leq
                \|c_2\|\vee\|c_3\|\\
                &\leq
                1\vee
                \left|2\operatorname{Mult}_n(x,b)-a\right|\\
                &\leq3.
            \end{align*}
        \end{enumerate}
        This concludes the proof of the claim.
    \end{proof}

    By repeatedly applying the Chaining Lemma \ref{lem:chaining} to the three stages of the claim, we obtain
    \[
        \ChebStep^T(c)=c',
    \]
    where $T=1+T_\Times+1=2n+5$. Moreover,
    \[
        \|c\|_{\ChebStep}^T
        \leq
        \|c_0\|_{\ChebStep}^1
        \vee
        \|c_1\|_{\ChebStep}^{T_\Times}
        \vee
        \|c_2\|_{\ChebStep}^1
        \leq3.
    \]
\end{proof}

The next estimate translates the approximate multiplication error into a recurrence error.
\begin{lemma}\label{lem:approximate-chebychev-recurrence}
    Let $n\in\No$, $k\in\N$, $x,a,b\in[-1,1]$, and define
    \[
        \widetilde b:=2\operatorname{Mult}_n(x,b)-a.
    \]
    Define also the incoming errors
    \[
        E_-:=a-\frac{T_{k-1}(x)}2,
        \qquad
        E:=b-\frac{T_k(x)}2,
    \]
    and the local multiplication error
    \[
        \varepsilon_n(x,b):=\operatorname{Mult}_n(x,b)-xb.
    \]
    Then,
    \begin{equation}\label{eq:cheb-step-error-recurrence}
        \widetilde b-\frac{T_{k+1}(x)}2
        =
        2xE-E_-+2\varepsilon_n(x,b),
        \qquad
        |\varepsilon_n(x,b)|\leq2^{-2n-1}.
    \end{equation}
\end{lemma}
\begin{proof}
    By definition,
    \[
        \operatorname{Mult}_n(x,b)=xb+\varepsilon_n(x,b).
    \]
    Since $x,b\in[-1,1]$, Lemma \ref{lem:multiplication_rnn} gives
    \[
        |\varepsilon_n(x,b)|
        =
        |\operatorname{Mult}_n(x,b)-xb|
        \leq2^{-2n-1}.
    \]
    Using
    \[
        a=\frac{T_{k-1}(x)}2+E_-,
        \qquad
        b=\frac{T_k(x)}2+E,
    \]
    we obtain
    \begin{align*}
        \widetilde b
        &=
        2\operatorname{Mult}_n(x,b)-a\\
        &= 2(xb+\varepsilon_n(x,b))-\left(\frac{T_{k-1}(x)}2+E_-\right)\\
        &=
        2x\left(\frac{T_k(x)}2+E\right)
        +2\varepsilon_n(x,b)
        -\left(\frac{T_{k-1}(x)}2+E_-\right)\\
        &=
        \frac{2xT_k(x)-T_{k-1}(x)}2
        +2xE-E_-+2\varepsilon_n(x,b)\\
        &=
        \frac{T_{k+1}(x)}2
        +2xE-E_-+2\varepsilon_n(x,b).
    \end{align*}
    Rearranging proves \eqref{eq:cheb-step-error-recurrence}.
\end{proof}

Iterating the previous recurrence gives a uniform stability bound for the normalized Chebyshev values. The key is to remark that the error propagation follows the same recurrence as the Chebyshev polynomials of the second kind, defined by the recurrence
\begin{equation}\label{eq:chebychev-second-kind}
    U_0(x)=1,\qquad U_1(x)=2x,\qquad
    U_{r+1}(x)=2xU_r(x)-U_{r-1}(x), \quad r\in\N,
\end{equation}
for $x \in [-1,1]$. We show the following classical boundedness property of the Chebyshev polynomials of the second kind.
\begin{lemma}\label{lem:chebychev-second-kind-bound}
    For $r \in \N$, 
\[
        \sup_{x \in [-1,1]} |U_r(x)| = r+1.
\]
\end{lemma}
\begin{proof}
    The proof follows by first showing a useful formula for $U_r(x)$, and then using it to find the maximum value of $|U_r(x)|$ on the interval $[-1,1]$. We show directly from the recurrence that, for every $\theta\in[0,\pi]$,
    \[
        U_r(\cos(\theta))
        =
        \sum_{\ell=0}^r e^{i(r-2\ell)\theta}.
    \]
    For $r=0$, this says $U_0(\cos(\theta))=1$, which is true. For
    $r=1$, this says
    \[
        U_1(\cos(\theta))
        =
        e^{i\theta}+e^{-i\theta}
        =
        2\cos(\theta),
    \]
    which is also true. For the inductive step, assume the formula holds at
    the two consecutive indices $r$ and $r-1$. Then the recurrence
    \eqref{eq:chebychev-second-kind} gives
    \begin{align*}
        U_{r+1}(\cos(\theta))
        &=
        2\cos(\theta)U_r(\cos(\theta))-U_{r-1}(\cos(\theta))\\
        &=
        (e^{i\theta}+e^{-i\theta})
        \sum_{\ell=0}^r e^{i(r-2\ell)\theta}
        -
        \sum_{\ell=0}^{r-1} e^{i(r-1-2\ell)\theta}\\
        &=
        \sum_{\ell=0}^r e^{i(r+1-2\ell)\theta}
        +
        \sum_{\ell=0}^r e^{i(r-1-2\ell)\theta}
        -
        \sum_{\ell=0}^{r-1} e^{i(r-1-2\ell)\theta}\\
        &=
        \sum_{\ell=0}^r e^{i(r+1-2\ell)\theta}
        +
        e^{-i(r+1)\theta}\\
        &=
        \sum_{\ell=0}^{r+1} e^{i(r+1-2\ell)\theta}.
    \end{align*}
    This completes the induction. Now, now that
    \[
        \sup_{x \in [-1,1]} |U_r(x)| = \sup_{\theta \in [0,\pi]} |U_r(\cos(\theta))| = \sup_{\theta \in [0,\pi]} \left|\sum_{\ell=0}^r e^{i(r-2\ell)\theta}\right| \leq \sum_{\ell=0}^r |e^{i(r-2\ell)\theta}| = r+1.
    \]
\end{proof}

We are now ready to prove the stability of the normalized Chebyshev sequence.
\begin{lemma}\label{lem:native-chebychev-stability}
    Let $k,n\in\No$ satisfy
    \begin{equation}\label{eq:native-chebychev-validity}
        k\leq 2^n.
    \end{equation}
    Then,
    \[
        |Q_{k,n}(x)|\leq1,
        \qquad
        \left|Q_{k,n}(x)-\frac{T_k(x)}2\right|
        \leq k^2 2^{-2n-1}, \quad x \in [-1,1].
    \]
\end{lemma}
\begin{proof}
    Fix $n \in \No$.
    For every $k\in\No$, define
    \[
        E_{k,n}(x):=Q_{k,n}(x)-\frac{T_k(x)}2,
    \]
    and the local multiplication error
    \[
        \varepsilon_{k,n}(x)
        :=
        \operatorname{Mult}_n(x,Q_{k,n}(x))
        -
        xQ_{k,n}(x).
    \]
    We claim that, for every $k \in \{0, \ldots, 2^n\}$,
    \begin{equation}\label{eq:chebychev-error-formula}
        |Q_{k,n}(x)|\leq1,
        \qquad
        E_{k,n}(x)
        =
        2\sum_{j=1}^{k-1}U_{k-j-1}(x)\varepsilon_{j,n}(x),
    \end{equation}
    for $x \in [-1,1]$, where $U_{r}, r \in \No$ are the Chebyshev polynomials of the second kind defined by \eqref{eq:chebychev-second-kind}. We prove the claim by induction on $k$. For $k=0,1$, the sum is empty, and \eqref{eq:recurrence-chebychev-polynomials-appendix} and \eqref{eq:def-native-chebychev-sequence} yield $E_{0,n}(x) = E_{1,n}(x) = 0$ and $|Q_{0,n}(x)| = |Q_{1,n}(x)| \leq 1/2$. This verifies the claim at the base cases.
    Now assume that \eqref{eq:chebychev-error-formula} holds at the two consecutive indices $k-1$ and
    $k$ such that $k < 2^n$. Then, by Lemma \ref{lem:approximate-chebychev-recurrence},
    \begin{align*}
        E_{k+1,n}(x)
        &=
        2xE_{k,n}(x)-E_{k-1,n}(x)+2\varepsilon_{k,n}(x)\\
        &=
        4x\sum_{j=1}^{k-1}U_{k-j-1}(x)\varepsilon_{j,n}(x)
        -
        2\sum_{j=1}^{k-2}U_{k-j-2}(x)\varepsilon_{j,n}(x)
        +
        2\varepsilon_{k,n}(x)\\
        &=
        2\sum_{j=1}^{k-2}
        \left(2xU_{k-j-1}(x)-U_{k-j-2}(x)\right)
        \varepsilon_{j,n}(x)
        +
        4xU_0(x)\varepsilon_{k-1,n}(x)
        +
        2\varepsilon_{k,n}(x)\\
        &=
        2\sum_{j=1}^{k-2}
        U_{k-j}(x)\varepsilon_{j,n}(x)
        +
        2U_1(x)\varepsilon_{k-1,n}(x)
        +
        2U_0(x)\varepsilon_{k,n}(x)\\
        &=
        2\sum_{j=1}^{k}U_{k-j}(x)\varepsilon_{j,n}(x).
    \end{align*}
    Now, since by Lemma \ref{lem:chebychev-second-kind-bound}, $|U_r(x)|\leq r+1$ on $[-1,1]$, we have
    \[
        |E_{k,n}(x)|
        \leq
        2^{-2n}\sum_{j=1}^{k-1}(k-j)
        \leq k^2 2^{-2n-1}.
    \]
    Moreover, since $|T_k(x)|\leq1$ on $[-1,1]$,
    \[
        |Q_{k,n}(x)|
        \leq
        \left|\frac{T_k(x)}2\right|+|E_{k,n}(x)|
        \leq
        \frac12 + k^2 2^{-2n-1} \overset{(a)}\leq \frac12 + \frac12 = 1,
    \]
    where (a) is by the assumption $k \leq 2^n$. This closes the induction and proves the claim. Finally, the error bound follows from the claim and the bound on $|U_r(x)|$.
\end{proof}

For a dyadic vector $\mathbf a=(a_0,\ldots,a_d)\in\Db^{d+1}$, define
\begin{equation}\label{eq:def-native-chebychev-sum-approximation}
    H_{\mathbf a,n}(x):=
    2\sum_{k=0}^d a_kQ_{k,n}(x).
\end{equation}
If $\mathbf a\neq0$, let
\begin{equation}\label{eq:normalization-chebychev-sum}
    k_{\mathbf a}
    :=
    1+\left(\left\lfloor\log_2(2\|\mathbf a\|_1)\right\rfloor\vee0\right),
    \qquad
    \mathbf a^\ast:=2^{-k_{\mathbf a}}\mathbf a,
\end{equation}
and set $k_{\mathbf 0}:=0$ and $\mathbf 0^\ast:=\mathbf0$. Thus,
$2\|\mathbf a^\ast\|_1<1$. Define the encoding
\begin{align*}
    \encCheb{\mathbf a}{n}
    &:=
    \encDyaSign{a_0^\ast}1^n0\,1\,
    \encDyaSign{a_1^\ast}1^n0\,1\cdots
    1\,\encDyaSign{a_d^\ast}1^n0\,0\,1^{k_{\mathbf a}}0.
\end{align*}

The summation machine now loops over the encoded coefficients and calls $\ChebStep$ after each accumulation.
\begin{definition}\label{def:cheb-sum-machine}
    We let $\ChebSum$ be the \TMNU with neural dimension $18$ and $13$ states defined by the following procedure.
    \begin{itemize}
        \item \textbf{State 1.} Update the neural state as
        \[
            \neurState\gets
            (\neurState_1,\neurState_2,\neurState_3,\neurState_4,
            2\neurState_2,0_{13}),
        \]
        and go to State $2$.
        \item \textbf{States 2--4.} Use $\Contr$ as a subroutine with correspondence $(2,3,4,5;5,6)$.
        \item \textbf{State 5.} Update the neural state as
        \[
            \neurState\gets
            (\neurState_1,\neurState_2,\neurState_3,
            \neurState_4+\neurState_5,0_{14}),
        \]
        and go to State $6$.
        \item \textbf{States 6--10.} Use $\ChebStep$ as a subroutine with correspondence
        $(6,\ldots,11;1,2,3,5,\ldots,18)$.
        \item \textbf{State 11.} If the scanned symbol is $1$, erase it, move right, and go to State $1$. If it is $0$, erase it, move right, update the neural state as
        \[
            \neurState\gets(\neurState_1,\neurState_4,0_{16}),
        \]
        and go to State $12$.
        \item \textbf{State 12.} Use $\Scale$ as a subroutine with correspondence $(12,13;2)$.
        \item \textbf{State 13.} Halt.
\end{itemize}
\end{definition}

The next lemma verifies one complete encoded Chebyshev-sum evaluation.
\begin{lemma}\label{lem:cheb-sum-machine}
    Let $\mathbf a=(a_0,\ldots,a_d)\in\Db^{d+1}$, $n\in\No$ such that $d+1 \leq 2^n$, $v\in\{0,1\}^{\#}$, and $x\in[-1,1]$. Define
    \begin{equation}\label{eq:cheb-sum-time-bound}
        T_{\mathbf a,n}
        :=
        \sum_{k=0}^d\left(\len{\encDyaSign{a_k^\ast}}+2n+8\right)
        +k_{\mathbf a}+1.
    \end{equation}
    For
    \[
        c:=(1;|\encCheb{\mathbf a}{n}v;x,\tfrac12,\tfrac{x}{2},0,0_{14}),
        \qquad
        c':=(13;|v;x,H_{\mathbf a,n}(x),0_{16}),
    \]
    we have
    \[
        \ChebSum^{T_{\mathbf a,n}}(c)=c',
        \qquad
        \|c\|_{\ChebSum}^{T_{\mathbf a,n}}
        \leq 3\vee|H_{\mathbf a,n}(x)|.
    \]
    % Furthermore,
    % \begin{equation}\label{eq:cheb-sum-error}
    %     \left|
    %     H_{\mathbf a,n}(x)-\sum_{k=0}^da_kT_k(x)
    %     \right|
    %     \leq
    %     2\|\mathbf a\|_1d^2 2^{-2n-1}.
    % \end{equation}
\end{lemma}
\begin{proof}
    For every $k\in\{0,\ldots,d+1\}$, define
    \[
        z_k
        :=
        2\sum_{j=0}^{k-1}a_j^\ast Q_{j,n}(x),
    \]
    with the convention $z_0:=0$. Since by
    Lemma \ref{lem:native-chebychev-stability}, $|Q_{k,n}(x)|\leq1$ for every $k \in \{0,\ldots, d+1\}$, we have
    \begin{equation}\label{eq:bound-zk:proof:lem:cheb-sum-machine}
        |z_k|
        \leq
        2\|\mathbf a^\ast\|_1
        <1,
        \qquad
        k\in\{0,\ldots,d+1\}.
    \end{equation}

    For every $k\in\{0,\ldots,d\}$, let $w_k$ denote the suffix of the
    encoding beginning with the coefficient $a_k^\ast$, namely
    \[
        w_k
        :=
        \encDyaSign{a_k^\ast}1^n0\,1\cdots
        1\,\encDyaSign{a_d^\ast}1^n0\,0\,1^{k_{\mathbf a}}0v,
    \]
    and set $w_{d+1}:=1^{k_{\mathbf a}}0v$. We also define
    \[
        s_k
        :=
        \begin{cases}
            1w_{k+1}, & k\in\{0,\ldots,d-1\},\\
            0w_{d+1}, & k=d.
        \end{cases}
    \]
    Thus,
    \[
        w_k=\encDyaSign{a_k^\ast}1^n0s_k,
        \qquad
        w_0=\encCheb{\mathbf a}{n}v.
    \]

    For every $k\in\{0,\ldots,d\}$, define
    \[
        T_{\Contr,k}:=\len{\encDyaSign{a_k^\ast}},
        \qquad
        T_{\ChebStep}:=2n+5,
    \]
    and the following configurations of $\ChebSum$:
    \begin{equation}\label{eq:configurations:proof:lem:cheb-sum-machine}
        \begin{cases}
            c_{k,0}
            &:=
            (1;|w_k;x,Q_{k,n}(x),Q_{k+1,n}(x),z_k,0_{14}),\\
            c_{k,1}
            &:=
            (2;|w_k;x,Q_{k,n}(x),Q_{k+1,n}(x),z_k,
            2Q_{k,n}(x),0_{13}),\\
            c_{k,2}
            &:=
            (5;|1^n0s_k;x,Q_{k,n}(x),Q_{k+1,n}(x),z_k,
            2a_k^\ast Q_{k,n}(x),0_{13}),\\
            c_{k,3}
            &:=
            (6;|1^n0s_k;x,Q_{k,n}(x),Q_{k+1,n}(x),z_{k+1},0_{14}),\\
            c_{k,4}
            &:=
            (11;|s_k;x,Q_{k+1,n}(x),Q_{k+2,n}(x),z_{k+1},0_{14}).
        \end{cases}
    \end{equation}
    Finally, define
    \[
        c_{\Scale}
        :=
        (12;|w_{d+1};x,z_{d+1},0_{16}),
        \qquad
        c_{\mathrm{out}}
        :=
        (13;|v;x,H_{\mathbf a,n}(x),0_{16}), \qquad T_\Scale:=k_{\mathbf a}+1.
    \]
    Note that $c_{0,0}=c$. Moreover,
    \[
        2^{k_{\mathbf a}}z_{d+1}
        =
        2\sum_{j=0}^d a_jQ_{j,n}(x)
        =
        H_{\mathbf a,n}(x),
    \]
    and therefore $c_{\mathrm{out}}=c'$.
    \begin{claim*}
        For every $k\in\{0,\ldots,d\}$, the following statements hold:
        \begin{enumerate}[label=(\alph*)]
            \item $\ChebSum(c_{k,0})=c_{k,1}$ and
            $\|c_{k,0}\|_{\ChebSum}^1\leq2$.
            \item $\ChebSum^{T_{\Contr,k}}(c_{k,1})=c_{k,2}$ and
            $\|c_{k,1}\|_{\ChebSum}^{T_{\Contr,k}}\leq2$.
            \item $\ChebSum(c_{k,2})=c_{k,3}$ and
            $\|c_{k,2}\|_{\ChebSum}^1\leq2$.
            \item $\ChebSum^{T_{\ChebStep}}(c_{k,3})=c_{k,4}$ and
            $\|c_{k,3}\|_{\ChebSum}^{T_{\ChebStep}}\leq3$.
            \item If $k<d$, then $\ChebSum(c_{k,4})=c_{k+1,0}$ and
            $\|c_{k,4}\|_{\ChebSum}^1\leq1$.
        \end{enumerate}
        Moreover,
        \[
            \ChebSum(c_{d,4})=c_{\Scale},
            \qquad
            \|c_{d,4}\|_{\ChebSum}^1\leq1.
        \]
        Finally,
        \[
            \ChebSum^{T_\Scale}(c_{\Scale})=c_{\mathrm{out}},
            \qquad
            \|c_{\Scale}\|_{\ChebSum}^{T_\Scale}
            \leq1\vee|H_{\mathbf a,n}(x)|.
        \]
    \end{claim*}
    \begin{proof}[Proof of the Claim.]
        Fix $k\in\{0,\ldots,d\}$.
        \begin{enumerate}[label=(\alph*)]
            \item In configuration $c_{k,0}$, the machine is in State $1$.
            Therefore, Definition \ref{def:cheb-sum-machine} gives
            $\ChebSum(c_{k,0})=c_{k,1}$. Moreover,
            \[
                \|c_{k,0}\|_{\ChebSum}^1
                \leq
                2,
            \]
            since $|x|,|Q_{k,n}(x)|,|Q_{k+1,n}(x)|\leq1$,
            $|z_k|<1$, and $|2Q_{k,n}(x)|\leq2$.

            \item Define the configurations of $\Contr$
            \[
                c_{k,1}'
                :=
                (1;|w_k;2Q_{k,n}(x),0),
                \qquad
                c_{k,2}'
                :=
                (4;|1^n0s_k;2a_k^\ast Q_{k,n}(x),0).
            \]
            Lemma \ref{lem:contr-tmnu-sign} gives
            \[
                \Contr^{T_{\Contr,k}}(c_{k,1}')=c_{k,2}',
                \qquad
                \|c_{k,1}'\|_\Contr^{T_{\Contr,k}}
                \leq2|Q_{k,n}(x)|
                \leq2.
            \]
            Since $\Contr$ is a subroutine of $\ChebSum$ with correspondence
            $(2,3,4,5;5,6)$, and since
            \[
                \Shadow_{(2,3,4,5),(5,6)}(c_{k,1})=c_{k,1}',
                \qquad
                \Lift_{(2,3,4,5),(5,6)}^{(x,Q_{k,n}(x),Q_{k+1,n}(x),z_k,0_{12})}
                (c_{k,2}')=c_{k,2},
            \]
            Lemma \ref{lem:subroutine} yields
            \[
                \ChebSum^{T_{\Contr,k}}(c_{k,1})=c_{k,2},
                \qquad
                \|c_{k,1}\|_{\ChebSum}^{T_{\Contr,k}}\leq2.
            \]

            \item In configuration $c_{k,2}$, the machine is in State $5$.
            Definition \ref{def:cheb-sum-machine} therefore gives
            $\ChebSum(c_{k,2})=c_{k,3}$, since
            \[
                z_k+2a_k^\ast Q_{k,n}(x)=z_{k+1}.
            \]
            Furthermore, \eqref{eq:bound-zk:proof:lem:cheb-sum-machine} and
            $|2a_k^\ast Q_{k,n}(x)|\leq2|a_k^\ast|\leq2$ imply
            $\|c_{k,2}\|_{\ChebSum}^1\leq2$.

            \item Define the configurations of $\ChebStep$
            \[
                c_{k,3}'
                :=
                (1;|1^n0s_k;x,Q_{k,n}(x),Q_{k+1,n}(x),0_{14}),
            \]
            and
            \[
                c_{k,4}'
                :=
                (6;|s_k;x,Q_{k+1,n}(x),Q_{k+2,n}(x),0_{14}).
            \]
            Lemma \ref{lem:cheb-step-machine} gives
            \[
                \ChebStep^{T_{\ChebStep}}(c_{k,3}')=c_{k,4}',
                \qquad
                \|c_{k,3}'\|_{\ChebStep}^{T_{\ChebStep}}\leq3.
            \]
            Since $\ChebStep$ is a subroutine of $\ChebSum$ with
            correspondence $(6,\ldots,11;1,2,3,5,\ldots,18)$, and since
            \[
                \Shadow_{(6,\ldots,11),(1,2,3,5,\ldots,18)}(c_{k,3})
                =
                c_{k,3}',
            \]
            \[
                \Lift_{(6,\ldots,11),(1,2,3,5,\ldots,18)}^{z_{k+1}}
                (c_{k,4}')
                =
                c_{k,4},
            \]
            Lemma \ref{lem:subroutine} yields
            \[
                \ChebSum^{T_{\ChebStep}}(c_{k,3})=c_{k,4},
                \qquad
                \|c_{k,3}\|_{\ChebSum}^{T_{\ChebStep}}\leq3.
            \]

            \item If $k<d$, then $s_k=1w_{k+1}$. Thus, in configuration
            $c_{k,4}$, the machine is in State $11$ and reads the symbol $1$.
            It erases this symbol, moves right, and goes to State $1$, so
            $\ChebSum(c_{k,4})=c_{k+1,0}$. The neural state is unchanged and
            bounded by one, hence $\|c_{k,4}\|_{\ChebSum}^1\leq1$.
        \end{enumerate}

        Finally, $s_d=0w_{d+1}$. Therefore, in configuration $c_{d,4}$,
        the machine is in State $11$ and reads the symbol $0$. It erases this
        symbol, moves right, updates its neural state to
        $(x,z_{d+1},0_{16})$, and goes to State $12$. Hence,
        \[
            \ChebSum(c_{d,4})=c_{\Scale},
            \qquad
            \|c_{d,4}\|_{\ChebSum}^1\leq1.
        \]
        Now, define the configurations of $\Scale$
        \[
            c_{\Scale}'
            :=
            (1;|w_{d+1};x,z_{d+1},0_{16}),
            \qquad
            c_{\mathrm{out}}'
            :=
            (2;|v;x,H_{\mathbf a,n}(x),0_{16}).
        \]
        By Lemma \ref{lem:scale-machine},
        \[
            \Scale^{T_\Scale}(c_{\Scale}')=c_{\mathrm{out}}',
            \qquad
            \|c_{\Scale}'\|_\Scale^{T_\Scale}
            \leq|H_{\mathbf a,n}(x)|.
        \]
        Since $\Scale$ is a subroutine of $\ChebSum$ with correspondence
        $(12,13;2)$, and since
        \[
            \Shadow_{(12,13),(2)}(c_{\Scale})=c_{\Scale}',
            \qquad
            \Lift_{(12,13),(2)}^{(x,0_{16})}(c_{\mathrm{out}}')
            =
            c_{\mathrm{out}},
        \]
        Lemma \ref{lem:subroutine} gives
        \[
            \ChebSum^{T_\Scale}(c_{\Scale})=c_{\mathrm{out}},
            \qquad
            \|c_{\Scale}\|_{\ChebSum}^{T_\Scale}
            \leq1\vee|H_{\mathbf a,n}(x)|.
        \]
        This concludes the proof of the claim. 
    \end{proof}

    Repeated applications of the Chaining Lemma \ref{lem:chaining}, together
    with the claim now give
    \[
        \ChebSum^{T_{\mathbf a,n}}(c)=c',
        \qquad
        \|c\|_{\ChebSum}^{T_{\mathbf a,n}}
        \leq3\vee|H_{\mathbf a,n}(x)|.
    \]
    Indeed,
    \begin{align*}
        T_{\mathbf a,n}
        &=
        \sum_{k=0}^d
        \left(
        1+T_{\Contr,k}+1+T_{\ChebStep}+1
        \right)
        +T_\Scale\\
        &=
        \sum_{k=0}^d
        \left(
        \len{\encDyaSign{a_k^\ast}}+2n+8
        \right)
        +k_{\mathbf a}+1.
    \end{align*}

    % Finally, Lemma \ref{lem:native-chebychev-stability} gives
    % \[
    %     \left|
    %     H_{\mathbf a,n}(x)-\sum_{k=0}^da_kT_k(x)
    %     \right|
    %     \leq
    %     2\sum_{k=0}^d|a_k|k^2 2^{-2n-1}
    %     \leq
    %     2\|\mathbf a\|_1d^2 2^{-2n-1}.
    % \]
\end{proof}

For sequences $\mathbf A=(\mathbf a_i)_{i\in\No}$ of finite dyadic vectors and
$\mathbf n=(n_i)_{i\in\No}$, define
\[
    \encChebSeq{\mathbf A}{\mathbf n}
    :=
    \encCheb{\mathbf a_0}{n_0}
    \encCheb{\mathbf a_1}{n_1}
    \encCheb{\mathbf a_2}{n_2}\cdots.
\]

The continuous Chebyshev machine repeatedly executes these finite sum instructions.
\begin{definition}\label{def:cheb-continuous-machine}
    We let $\ChebContinuous$ be the \TMNU with neural dimension $19$ and $14$ states defined by the following procedure.
    \begin{itemize}
        \item \textbf{State 1.} Update the neural state as
        \[
            \neurState\gets
            (\neurState_1,\tfrac12,\tfrac{\neurState_1}{2},0_{15},\neurState_2),
        \]
        and go to State $2$.
        \item \textbf{States 2--13.} Use $\ChebSum$ as a subroutine with correspondence
        $(2,\ldots,13,1;1,\ldots,18)$.
        \item \textbf{State 14.} Halt.
\end{itemize}
\end{definition}

The following lemma records one cycle and the resulting output convention.
\begin{lemma}\label{lem:cheb-continuous-machine}
    Let $\mathbf A=(\mathbf a_i)_{i\in\No}$ be a sequence of finite dyadic
    vectors, let $\mathbf n=(n_i)_{i\in\No}$, and let $x\in[-1,1]$.
    For every $i\in\No$, write
    $\mathbf a_i=(a_{i,0},\ldots,a_{i,d_i})$, and assume that
    \[
        d_i+1\leq 2^{n_i}.
    \]
    Define the tails
    \[
        \mathbf A_{\geq i}:=(\mathbf a_{i+j})_{j\in\No},
        \qquad
        \mathbf n_{\geq i}:=(n_{i+j})_{j\in\No},
    \]
    and use the conventions
    \[
        H_{\mathbf a_{-1},n_{-1}}(x)
        :=
        H_{\mathbf a_{-2},n_{-2}}(x)
        :=
        0.
    \]
    For every $i\in\No$, define the configuration
    \[
        c_i
        :=
        \left(
        1;
        |\encChebSeq{\mathbf A_{\geq i}}{\mathbf n_{\geq i}};
        x,H_{\mathbf a_{i-1},n_{i-1}}(x),0_{16},
        H_{\mathbf a_{i-2},n_{i-2}}(x)
        \right),
    \]
    and set
    \[
        T_i:=T_{\mathbf a_i,n_i}+1,
    \]
    where $T_{\mathbf a_i,n_i}$ is as in \eqref{eq:cheb-sum-time-bound}. Then, for every $i\in\No$,
    \begin{equation}\label{eq:cycle:lem:cheb-continuous-machine}
        \ChebContinuous^{T_i}(c_i)=c_{i+1},
    \end{equation}
    and
    \begin{equation}\label{eq:bound:lem:cheb-continuous-machine}
        \|c_i\|_{\ChebContinuous}^{T_i}
        \leq
        3
        \vee
        |H_{\mathbf a_{i-2},n_{i-2}}(x)|
        \vee
        |H_{\mathbf a_{i-1},n_{i-1}}(x)|
        \vee
        |H_{\mathbf a_i,n_i}(x)|.
    \end{equation}
    In particular, $c_0=(1;|\encChebSeq{\mathbf A}{\mathbf n};x,0_{18})$.
    By denoting
    \[
        N_i:=\sum_{j=0}^{i-1}T_j,
    \]
    we have $\ChebContinuous^{N_i}(c_0)=c_i$ for every $i\in\No$, and
    \begin{equation}\label{eq:output:lem:cheb-continuous-machine}
        \ChebContinuous^{\encChebSeq{\mathbf A}{\mathbf n}}x[t]
        =
        H_{\mathbf a_{i-1},n_{i-1}}(x),
        \qquad
        t\in\{N_i+1,\ldots,N_{i+1}\}.
    \end{equation}
\end{lemma}
\begin{proof}
    Fix $i\in\No$, and define
    \[
        h_j:=H_{\mathbf a_j,n_j}(x),
        \qquad j\in\No,
    \]
    together with the conventions $h_{-1}:=h_{-2}:=0$. We also define the
    intermediate configuration
    \[
        \widetilde c_i
        :=
        \left(
        2;
        |\encChebSeq{\mathbf A_{\geq i}}{\mathbf n_{\geq i}};
        x,\frac12,\frac{x}{2},0_{15},h_{i-1}
        \right).
    \]
    We prove separately the initialization transition and the
    $\ChebSum$ subroutine stage.

    \begin{claim*}
        The following statements hold:
        \begin{enumerate}[label=(\alph*)]
            \item $\ChebContinuous(c_i)=\widetilde c_i$ and
            \[
                \|c_i\|_{\ChebContinuous}^1
                \leq
                1\vee|h_{i-2}|\vee|h_{i-1}|.
            \]
            \item
            \[
                \ChebContinuous^{T_{\mathbf a_i,n_i}}(\widetilde c_i)
                =
                c_{i+1},
            \]
            and
            \[
                \|\widetilde c_i\|_{\ChebContinuous}^{T_{\mathbf a_i,n_i}}
                \leq
                3\vee|h_{i-1}|\vee|h_i|,
            \]
            where $T_{\mathbf a_i,n_i}$ is as in \eqref{eq:cheb-sum-time-bound}. Moreover, for every
            $s\in\{0,\ldots,T_{\mathbf a_i,n_i}\}$,
            \begin{equation}\label{eq:preserved-output:proof:lem:cheb-continuous-machine}
                \proj_{19}\proj_\neurState
                \ChebContinuous^s(\widetilde c_i)
                =
                h_{i-1}.
            \end{equation}
        \end{enumerate}
    \end{claim*}
    \begin{proof}[Proof of the Claim.]
        We prove both statements in turn.
        \begin{enumerate}[label=(\alph*)]
            \item In configuration $c_i$, the machine is in State $1$.
            Therefore, by Definition \ref{def:cheb-continuous-machine}, it
            preserves $x$, initializes the first two normalized Chebyshev
            values as $1/2$ and $x/2$, sets the accumulator and the remaining
            working coordinates to zero, copies $h_{i-1}$ into coordinate
            $19$, and goes to State $2$. Hence,
            \[
                \ChebContinuous(c_i)=\widetilde c_i.
            \]
            Since $|x|\leq1$, we also have
            \begin{align*}
                \|c_i\|_{\ChebContinuous}^1
                &\leq
                \|(x,h_{i-1},0_{16},h_{i-2})\|_\infty\\
                &\quad\vee
                \left\|
                \left(x,\frac12,\frac{x}{2},0_{15},h_{i-1}\right)
                \right\|_\infty\\
                &\leq
                1\vee|h_{i-2}|\vee|h_{i-1}|.
            \end{align*}

            \item Define the configurations of $\ChebSum$
            \[
                \widetilde c_i'
                :=
                \left(
                1;
                |\encChebSeq{\mathbf A_{\geq i}}{\mathbf n_{\geq i}};
                x,\frac12,\frac{x}{2},0_{15}
                \right),
            \]
            and
            \[
                c_{i+1}'
                :=
                \left(
                13;
                |\encChebSeq{\mathbf A_{\geq i+1}}{\mathbf n_{\geq i+1}};
                x,h_i,0_{16}
                \right).
            \]
            Since
            \[
                \encChebSeq{\mathbf A_{\geq i}}{\mathbf n_{\geq i}}
                =
                \encCheb{\mathbf a_i}{n_i}
                \encChebSeq{\mathbf A_{\geq i+1}}{\mathbf n_{\geq i+1}},
            \]
            Lemma \ref{lem:cheb-sum-machine} gives
            \[
                \ChebSum^{T_{\mathbf a_i,n_i}}(\widetilde c_i')
                =
                c_{i+1}',
            \]
            and
            \[
                \|\widetilde c_i'\|_{\ChebSum}^{T_{\mathbf a_i,n_i}}
                \leq
                3\vee|h_i|.
            \]

            By Definition \ref{def:cheb-continuous-machine}, $\ChebSum$ is a
            subroutine of $\ChebContinuous$ with correspondence
            $(2,\ldots,13,1;1,\ldots,18)$. Furthermore,
            \[
                \Shadow_{(2,\ldots,13,1),(1,\ldots,18)}(\widetilde c_i)
                =
                \widetilde c_i',
            \]
            and
            \[
                \Lift_{(2,\ldots,13,1),(1,\ldots,18)}^{h_{i-1}}
                (c_{i+1}')
                =
                c_{i+1}.
            \]
            Therefore, Lemma \ref{lem:subroutine} yields
            \[
                \ChebContinuous^{T_{\mathbf a_i,n_i}}(\widetilde c_i)
                =
                c_{i+1},
            \]
            and
            \[
                \|\widetilde c_i\|_{\ChebContinuous}^{T_{\mathbf a_i,n_i}}
                \leq
                \|\widetilde c_i'\|_{\ChebSum}^{T_{\mathbf a_i,n_i}}
                \vee
                \|\widetilde c_i\|
                \leq
                3\vee|h_{i-1}|\vee|h_i|.
            \]
            Since coordinate $19$ is not used by the $\ChebSum$ subroutine,
            the same application of Lemma \ref{lem:subroutine} gives
            \eqref{eq:preserved-output:proof:lem:cheb-continuous-machine}.
        \end{enumerate}
        This concludes the proof of the claim.
    \end{proof}

    Since $T_i=1+T_{\mathbf a_i,n_i}$, the Chaining Lemma
    \ref{lem:chaining} and the claim give
    \[
        \ChebContinuous^{T_i}(c_i)=c_{i+1},
    \]
    and
    \[
        \|c_i\|_{\ChebContinuous}^{T_i}
        \leq
        3\vee|h_{i-2}|\vee|h_{i-1}|\vee|h_i|.
    \]
    This proves \eqref{eq:cycle:lem:cheb-continuous-machine} and
    \eqref{eq:bound:lem:cheb-continuous-machine}. We now prove the last statements. By induction on $i$,
    \[
        \ChebContinuous^{N_i}(c_0)=c_i,
        \qquad i\in\No.
    \]
    Fix $i\in\No$ and
    $t\in\{N_i+1,\ldots,N_{i+1}\}$. Then
    \[
        s:=t-N_i-1
        \in
        \{0,\ldots,T_{\mathbf a_i,n_i}\}.
    \]
    Using $\ChebContinuous^{N_i}(c_0)=c_i$ and
    $\ChebContinuous(c_i)=\widetilde c_i$, we obtain
    \[
        \ChebContinuous^t(c_0)
        =
        \ChebContinuous^s(\widetilde c_i).
    \]
    Therefore, by Definition \ref{def:tmnu-as-operator} and
    \eqref{eq:preserved-output:proof:lem:cheb-continuous-machine},
    \[
        \ChebContinuous^{\encChebSeq{\mathbf A}{\mathbf n}}x[t]
        =
        \proj_{19}\proj_\neurState
        \ChebContinuous^s(\widetilde c_i)
        =
        h_{i-1}.
    \]
    This proves \eqref{eq:output:lem:cheb-continuous-machine}.
\end{proof}

We can now balance degree and precision to obtain the native Chebyshev rate.
\begin{theorem}[Native Chebyshev tradeoff]\label{thm:native-chebychev-tradeoff}
    Let $f:[-1,1]\to\R$ be a Dini-continuous continuous function. Let
    $\eta:[1,\infty)\to\N$ be nondecreasing, with
    $\eta(t)\to\infty$ and $t/\eta(t)$ nondecreasing. Suppose that
    \begin{equation}\label{eq:native-eta-condition}
        \eta(t)\log_2(\eta(t)+1)\leq A t,
        \qquad t\geq1,
    \end{equation}
    for some $A>0$. Define 
    \[
        B_f:=2\|f\|_{L^\infty([-1,1])}+1,
        \qquad
        \gamma_f:=\left\lceil\log_2(B_f)\right\rceil,
    \]
    and
    \begin{equation}\label{eq:def-native-chebychev-time-constant}
        \tau := \tau_{f,A}
        :=
        4\bigl(10+\gamma_f+A(80+6\gamma_f)\bigr).
    \end{equation}
    Then, there exists $u\in\{0,1\}^{\N}$ such that
    \[
        (1;|u;x,0_{18})
        \in
        \Bc_{\ChebContinuous}^\infty
        \left(
        3 + S(f,0) + \|f\|_{L^\infty([-1,1])}
        \right),
        \qquad x\in[-1,1],
    \]
    and, for every $x\in[-1,1]$ and $t\geq \tau$,
    \begin{equation}\label{eq:native-chebychev-rate}
        \left|\ChebContinuous^u x[t]-f(x)\right|
        \leq
        2\cdot
        2^{
        -\frac{t/\tau}{\eta(t/\tau)}
        }
        +
        S\left(f,\eta(t/\tau)\right),
    \end{equation}
    where $S(f,d)$ is as in \eqref{eq:rate-of-approximation-by-tchebychev-series}.
\end{theorem}
\begin{proof}
    % Write $\tau:=\tau_{f,A}$. 
    % Let $(c_k^{(f)})_{k\in\No}$ be the Chebyshev coefficients of $f$. 
    First note that since $f$ is DiniLipschitz continuous, $S(f,d)\to0$ as $d\to\infty$ and $S(f,d) < \infty$ for every $d\in\No$. For every $i\in\No$, set
    \begin{equation}\label{eq:native-chebychev-parameters}
        d_i:=\eta(2^i),
        \quad
        q_i:=\left\lceil\frac{2^i}{d_i}\right\rceil,
        \quad
        \ell_i:=\left\lceil\log_2(d_i+1)\right\rceil, \quad
        p_i:=q_i+\ell_i+2,
        \quad
        n_i:=
        \left\lceil
        \frac{q_i+3\ell_i+\gamma_f}{2}
        \right\rceil,
    \end{equation}
    and for $k\in\{0,\ldots,d_i\}$, choose
    $a_{i,k}\in2^{-p_i}\Z$ such that
    \begin{equation}\label{eq:approx-cheb-coeffs:proof:native-tradeoff}
        |a_{i,k}-c_k^{(f)}|\leq2^{-p_i},
    \end{equation}
    where $c_k^{(f)}$ is the $k$-th Chebyshev coefficient of $f$,
    and set
    \[
        \mathbf a_i:=(a_{i,0},\ldots,a_{i,d_i}),
        \qquad
        \mathbf A:=(\mathbf a_i)_{i\in\No},
        \qquad
        \mathbf n:=(n_i)_{i\in\No},
        \qquad
        u:=\encChebSeq{\mathbf A}{\mathbf n}.
    \]
    The proof is effected by estimating the error and duration of each cycle of the computation.
    \begin{enumerate}[label=(\alph*)]
        \item  We first estimate the error of cycle $i \in \No$. 
    Note that $d_i+1\leq2^{\ell_i} \leq 2^{n_i}$, so Lemma \ref{lem:native-chebychev-stability} applies. Therefore, we get
    \begin{align*}
        \hspace{-0.5cm}\left|
        H_{\mathbf a_i,n_i}(x)-S^{(f)}_{d_i}(x)
        \right| &= \left| 2\sum_{k=0}^{d_i}a_{i,k}Q_{n_i,k}(x)-\sum_{k=0}^{d_i}c_k^{(f)}T_k(x)\right|\\
        &\leq \left| 2\sum_{k=0}^{d_i}a_{i,k}Q_{n_i,k}(x)-2\sum_{k=0}^{d_i}c_k^{(f)}Q_{n_i,k}(x)\right| + \left| 2\sum_{k=0}^{d_i}c_k^{(f)}Q_{n_i,k}(x)-\sum_{k=0}^{d_i}c_k^{(f)}T_k(x)\right|\\
        &\leq 2 \sum_{k=0}^{d_i}|a_{i,k}-c_k^{(f)}|+2\sum_{k=0}^{d_i}| c_k^{(f)}| \left| Q_{n_i,k}(x)-\frac{T_k(x)}2\right|\\
        &\overset{(a)}\leq 2 d_i 2^{-p_i}+B_f d_i^2 2^{-2n_i-1} \overset{\eqref{eq:native-chebychev-parameters}}\leq 2^{-q_i-1}+2^{-q_i-1}=2^{-q_i},
    \end{align*}
    where (a) follows from Lemma \ref{lem:native-chebychev-stability} and \eqref{eq:approx-cheb-coeffs:proof:native-tradeoff}. Hence,
    \begin{equation}\label{eq:native-cycle-error}
        |H_{\mathbf a_i,n_i}(x)-f(x)|
        \leq
        2^{-q_i}+S(f,d_i),
        \qquad x\in[-1,1].
    \end{equation}
    \item We now derive an explicit bound on the duration of cycle $i$. First note that
        \begin{align}\label{eq:bound-ai-one-norm:proof:native-tradeoff}
            \|\mathbf a_i\|_1
            &\leq
            \sum_{k=0}^{d_i}(|a_{i,k}-c_k^{(f)}|+|c_k^{(f)}|)
            \overset{\eqref{eq:approx-cheb-coeffs:proof:native-tradeoff}}\leq (d_i+1) \left(2^{-p_i} + \max_{k\in\{0,\ldots,d_i\}}|c_k^{(f)}|\right)
            \nonumber\\
            &\overset{\eqref{eq:chebyshev-coefficients}}\leq (d_i+1) \left(1 + 2\|f\|_{L^\infty([-1,1])}\right) = (d_i+1)B_f.
        \end{align} From
    \eqref{eq:bound-ai-one-norm:proof:native-tradeoff} and the definition \eqref{eq:normalization-chebychev-sum} of
    $k_{\mathbf a_i}$, we have
    \begin{equation}\label{eq:bound-k-ai:proof:native-tradeoff}
        k_{\mathbf a_i}
        \leq
        \ell_i+\gamma_f+2.
    \end{equation}
    Since $a_{i,k}\in2^{-p_i}\Z$, we have
    $a_{i,k}^\ast\in2^{-p_i-k_{\mathbf a_i}}\Z\cap\Db_1$. Therefore,
    \begin{equation}\label{eq:bound-encoding-ai:proof:native-tradeoff}
        \len{\encDyaSign{a_{i,k}^\ast}}
        \overset{\eqref{eq:len-enc-dya-sign}}\leq
        2(p_i+k_{\mathbf a_i})+2.
    \end{equation}
    Let $T_i:=T_{\mathbf a_i,n_i}+1$ be the duration of cycle $i$ in
    $\ChebContinuous$, where $T_{\mathbf a_i,n_i}$ is as in \eqref{eq:cheb-sum-time-bound}. By
    \eqref{eq:bound-k-ai:proof:native-tradeoff}, and
    \eqref{eq:bound-encoding-ai:proof:native-tradeoff},
    \begin{align}
        T_i
        &\overset{\eqref{eq:cheb-sum-time-bound}}\leq
        1 + \sum_{k=0}^{d_i}\left(\len{\encDyaSign{a_{i,k}^\ast}}+2n_i+8\right)
        +k_{\mathbf a_i}+1\nonumber\\
        &\overset{\eqref{eq:bound-encoding-ai:proof:native-tradeoff}}\leq (d_i+1)(2p_i+2k_{\mathbf a_i}+2n_i+10)+k_{\mathbf a_i}+2\nonumber\\
        &\overset{\eqref{eq:bound-k-ai:proof:native-tradeoff}}\leq (d_i+1)(2p_i+2(\ell_i+\gamma_f+2)+2n_i+10)+\ell_i+\gamma_f+2+2\nonumber\\
        &\overset{\eqref{eq:native-chebychev-parameters}}\leq
        (d_i+1)
        \left(
        3q_i+7\ell_i+3\gamma_f+22
        \right)
        +\ell_i+\gamma_f+4.
        \label{eq:cycle-time-explicit:proof:native-tradeoff}
    \end{align}

    We bound each term in \eqref{eq:cycle-time-explicit:proof:native-tradeoff}.
    First, since $d_i=\eta(2^i) \geq 1$, \eqref{eq:native-eta-condition} gives
    \begin{equation}\label{eq:cycle-degree-geometric:proof:native-tradeoff}
        d_i\log_2(d_i+1)\leq A2^i,
        \qquad
        d_i\leq \frac{A2^i}{\log_2(d_i+1)} \leq \frac{A2^i}{\log_2(2)} = A2^i, \qquad d_i + 1\leq 2d_i \leq 2A2^i.
    \end{equation}
    Moreover, by \eqref{eq:native-chebychev-parameters},
    \begin{equation}\label{eq:cycle-qi-li-geometric:proof:native-tradeoff}
        q_i\leq\frac{2^i}{d_i}+1,
        \qquad
        \ell_i\leq\log_2(d_i+1)+1\leq2\log_2(d_i+1) \leq 2A2^i.
    \end{equation}
    which yields the two product bounds
    \begin{align}
        (d_i+1)q_i
        &\leq
        2d_i\left(\frac{2^i}{d_i}+1\right)
        =
        2\cdot2^i+2d_i
        \leq
        (2+2A)2^i,\label{eq:cycle-di-qi:proof:native-tradeoff}\\
        (d_i+1)\ell_i
        &\leq
        4d_i\log_2(d_i+1)
        \leq
        4A2^i. \label{eq:cycle-di-li:proof:native-tradeoff}
    \end{align}
    Substituting \eqref{eq:cycle-degree-geometric:proof:native-tradeoff}, \eqref{eq:cycle-qi-li-geometric:proof:native-tradeoff}, \eqref{eq:cycle-di-qi:proof:native-tradeoff} and \eqref{eq:cycle-di-li:proof:native-tradeoff} into
    \eqref{eq:cycle-time-explicit:proof:native-tradeoff}, and using
    $\gamma_f+4\leq(\gamma_f+4)2^i$, gives
    \begin{align}
        T_i
        &\leq
        \bigl(
        3(2+2A)+7(4A)
        +(3\gamma_f+22)(2A)
        +2A+\gamma_f+4
        \bigr)2^i\nonumber\\
        &=
        \bigl(
        10+\gamma_f+A(80+6\gamma_f)
        \bigr)2^i
        =:
        \frac{\tau}{4}2^i.
        \label{eq:cycle-time-geometric:proof:native-tradeoff}
    \end{align}
    Hence, with the notation of Lemma \ref{lem:cheb-continuous-machine},
    \begin{equation}\label{eq:cumulative-time-geometric:proof:native-tradeoff}
        N_i + 1
        =
        \sum_{j=0}^{i-1}T_j + 1
        \leq
        \frac{\tau}{4}(2^i-1) + 1
        \leq
        \frac{\tau}{4}2^i,
    \end{equation}
    where the last inequality uses $\tau/4\geq 10/4 \geq 1$.
    \end{enumerate}
    We now express the error in terms of machine time. Fix $t\geq \tau$,
    and set
    \[
        i:=\left\lfloor\log_2(4t/\tau)\right\rfloor.
    \]
    Then
    \begin{equation}\label{eq:cycle-index-time-ineq:proof:native-tradeoff}
        i \geq 2, \qquad
        N_i + 1 \leq \frac{\tau}{4}2^i\leq t, \qquad \text{and} \qquad
        2^{i-1}
        \geq
        \frac{t}{\tau}.
    \end{equation}
    In particular, $t \in \{N_j+1,\ldots,N_{j+1}\}$ for some $j\geq i$, so by Lemma \ref{lem:cheb-continuous-machine}, 
    \begin{align}\label{eq:native-cycle-error:proof:native-tradeoff}
        |\ChebContinuous^u x[t]-f(x)|
        &=|H_{\mathbf a_{j-1},n_{j-1}}(x)-f(x)| \overset{\eqref{eq:native-cycle-error}}\leq
        2^{-q_{j-1}}+S(f,d_{j-1}) \leq \sup_{k\geq i-1} \left(2^{-q_k}+S(f,d_k)\right)\\
        & = \sup_{k\geq i-1} \left(2^{-\frac{2^k}{\eta(2^k)}}+S(f,\eta(2^k))\right)
        \overset{(a)}\leq 2^{-\frac{2^{i-1}}{\eta(2^{i-1)}}}+S(f,\eta(2^{i-1})),
    \end{align}
    where (a) follows from the fact that $\eta(t)$ and $t/\eta(t)$ are nondecreasing, and $t \mapsto 2^{-t}$ and $t \mapsto S(f,t)$ are nonincreasing. By \eqref{eq:cycle-index-time-ineq:proof:native-tradeoff},
    \[
        d_{i-1}
        =
        \eta(2^{i-1})
        \geq
        \eta(t/\tau), \qquad \text{and} \qquad q_{i-1}
        \geq
        \left\lceil
        \frac{2^{i-1}}{\eta(2^{i-1})}
        \right\rceil
        \geq
        \frac{t/\tau}{\eta(t/\tau)}.
    \]
    Substituting these inequalities into
    \eqref{eq:native-cycle-error:proof:native-tradeoff} proves
    \eqref{eq:native-chebychev-rate}. Finally, by \eqref{eq:native-cycle-error}, for every $i\in\No$ and $x\in[-1,1]$,
    \[
        |H_{\mathbf a_i,n_i}(x)|
        \leq |H_{\mathbf a_i,n_i}(x) - f(x)| + |f(x)| \leq 1 + S(f,d_i) + \|f\|_{L^\infty([-1,1])} \leq 1 + S(f,0) + \|f\|_{L^\infty([-1,1])}
    \]
    Lemma \ref{lem:cheb-continuous-machine} therefore gives
    \[
        (1;|u;x,0_{18})
        \in
        \Bc_{\ChebContinuous}^\infty
        \left(
        3 + S(f,0) + \|f\|_{L^\infty([-1,1])}
        \right).
    \]
\end{proof}

\subsection{Magnitude estimates for the \TMNUs used in the constructions}
\label{subsec:tmnu-construction-appendix-magnitude-estimates}

We close the appendix by recording the estimates on the quantity
$\|\tmnuM\|_C$ that enters Theorem
\ref{thm:simulation-of-a-tmnu-by-an-rnn}. Recall from Section
\ref{sec:tmnu-rnn-simulation} that, if
$\Fc_\tmnuM=\{f_1,\ldots,f_{\nFunctions{\tmnuM}}\}$ denotes the set of
command maps of a \TMNU $\tmnuM$ with neural dimension $d$, then
\[
    \|\tmnuM\|_C
    :=
    \max_{1\leq j\leq\nFunctions{\tmnuM}}
    \sup_{\omega\in[-C,C]^d}\|f_j(\omega)\|_\infty .
\]

We first introduce a global magnitude for command maps. If
$W:\R^d\to\R^d$ is affine, say $W(\omega)=A\omega+b$, set
\[
    \mathfrak m(W)
    :=
    \max_{1\leq i\leq d}\left(\sum_{j=1}^d |A_{ij}|+|b_i|\right).
\]
If $W=\ReLU_S\in\reluSet d$, set $\mathfrak m(W):=1$. Finally, for a
\TMNU $\tmnuM$, define
\[
    \mathfrak m(\tmnuM)
    :=
    1\vee\max_{f\in\Fc_\tmnuM}\mathfrak m(f).
\]

This global command magnitude controls the bounded-region quantity used in the simulation theorem.
\begin{lemma}\label{lem:tmnu-C-magnitude-from-command-magnitude}
    Let $\tmnuM$ be a \TMNU. Then, for every $C>0$,
    \[
        \|\tmnuM\|_C\leq \mathfrak m(\tmnuM)(C\vee1).
    \]
\end{lemma}
\begin{proof}
    Let $f\in\Fc_\tmnuM$ and $\omega\in[-C,C]^d$. If $f=\ReLU_S$ for some
    $S\subseteq\{1,\ldots,d\}$, then
    $\|f(\omega)\|_\infty\leq\|\omega\|_\infty\leq C\leq C\vee1$.
    If $f(\omega)=A\omega+b$ is affine, then
    \[
        \|f(\omega)\|_\infty
        \leq
        \max_{1\leq i\leq d}
        \left(
        \sum_{j=1}^d |A_{ij}|\,C+|b_i|
        \right)
        \leq
        \mathfrak m(f)(C\vee1).
    \]
    Taking the maximum over $f\in\Fc_\tmnuM$ proves the claim.
\end{proof}

The next estimate explains how this magnitude behaves when a machine is built from a subroutine.
\begin{lemma}\label{lem:tmnu-magnitude-under-subroutines}
    Let $\tmnuN$ be a subroutine of $\tmnuM$ with correspondence
    $(\indexFuncState,\indexFuncNeur)$, and define the set of states of
    $\tmnuM$ that do not simulate non-halting states of $\tmnuN$ by
    \[
        Q_{\mathrm{rem}}
        :=
        \{1,\ldots,\nStates_\tmnuM\}
        \setminus
        \indexFuncState(\{1,\ldots,\nStates_\tmnuN-1\}).
    \]
    Then, with the convention that the maximum over an empty set is $0$,
    \[
        \mathfrak m(\tmnuM)
        \leq
        \mathfrak m(\tmnuN)
        \vee 1
        \vee
        \max_{\state\in Q_{\mathrm{rem}},\,\symb\in\workSymbols}
        \mathfrak m\!\left(\commFunc_\tmnuM(\state,\symb)\right).
    \]
\end{lemma}
\begin{proof}
    For a command map $f:\R^{\neurDim_\tmnuN}\to\R^{\neurDim_\tmnuN}$, let
    $\widetilde f:\R^{\neurDim_\tmnuM}\to\R^{\neurDim_\tmnuM}$ be the map
    obtained by applying $f$ to the coordinates indexed by
    $\indexFuncNeur$ and by leaving all complementary coordinates unchanged,
    i.e.,
    \begin{equation}\label{eq:definition-extended-command-map}
        \proj_{\indexFuncNeur}\widetilde f(\omega)
        =
        f(\proj_{\indexFuncNeur}\omega),
        \qquad
        \proj_{\indexFuncNeur}^\perp \widetilde f(\omega)
        =
        \proj_{\indexFuncNeur}^\perp\omega.
    \end{equation}
    Note that
    \begin{equation}\label{eq:extended-command-map-magnitude}
        \mathfrak m(\widetilde f)\leq \mathfrak m(f)\vee1.
    \end{equation}
    Now fix $\state\in\{1,\ldots,\nStates_\tmnuN-1\}$ and
    $\symb\in\workSymbols$, and set $f:=\commFunc_\tmnuN(\state,\symb)$.
    Let $\widetilde f:\R^{\neurDim_\tmnuM}\to\R^{\neurDim_\tmnuM}$ be the extension of
    $f$ defined by \eqref{eq:definition-extended-command-map}. By the
    subroutine identity \eqref{eq:subroutine-conditions-2},
    \[
        \commFunc_\tmnuM(\indexFuncState(\state),\symb)
        =
        \widetilde f.
    \]
    Hence, by \eqref{eq:extended-command-map-magnitude},
    \[
        \mathfrak m\!\left(
        \commFunc_\tmnuM(\indexFuncState(\state),\symb)
        \right)
        \leq
        \mathfrak m\!\left(\commFunc_\tmnuN(\state,\symb)\right)\vee1
        \leq
        \mathfrak m(\tmnuN)\vee1.
    \]
    The remaining command maps of $\tmnuM$ are exactly those indexed by
    $\state\in Q_{\mathrm{rem}}$ and $\symb\in\workSymbols$. Taking the
    maximum over all command maps of $\tmnuM$, and recalling the outer
    maximum with $1$ in the definition of $\mathfrak m(\tmnuM)$, gives the
    desired estimate.
\end{proof}

We finish by tabulating the command magnitudes for every machine used above.
\begin{lemma}\label{lem:constructed-tmnus-command-magnitudes}
    The \TMNUs used in the monomial and native Chebyshev constructions
    satisfy
    \begin{center}
    \resizebox{\textwidth}{!}{$
    \begin{array}{c|ccccccccccc}
    \tmnuM
    & \Sign & \Scale & \Contr^+ & \Times & \Contr & \UpPoly
    & \Poly & \Continuous & \ChebStep & \ChebSum & \ChebContinuous\\
    \hline 
    \nStates_\tmnuM
    & 2 & 2 & 3 & 4 & 4 & 9
    & 11 & 12 & 6 & 13 & 14\\
    \neurDim_\tmnuM
    & 1 & 1 & 2 & 14 & 2 & 16
    & 16 & 17 & 17 & 18 & 19\\
    \nFunctions{\tmnuM}
    & 2 & 2 & 4 & 5 & 5 & 11
    & 13 & 14 & 7 & 15 & 16\\
    \mathfrak m(\tmnuM)
    & 1 & 2 & 2 & \leq5 & \leq2 & \leq5
    & \leq5 & \leq5 & \leq5 & \leq5 & \leq5 .
    \end{array}
    $}
    \end{center}
    In particular, for every $C>0$,
    \[
        \|\Continuous\|_C\leq5(C\vee1),
        \qquad
        \|\ChebContinuous\|_C\leq5(C\vee1).
    \]
\end{lemma}
\begin{proof}
    The rows for $\nStates_\tmnuM$ and $\neurDim_\tmnuM$ are read directly
    from Definitions \ref{def:sign-tmnu}, \ref{def:contr-+-tmnu},
    \ref{def:scale-machine}, \ref{def:tmnu-mult}, \ref{def:contr-tmnu},
    \ref{def:upoly}, \ref{def:poly-machine},
    \ref{def:continuous-machine}, \ref{def:cheb-step-machine},
    \ref{def:cheb-sum-machine}, and
    \ref{def:cheb-continuous-machine}. For $\Times$, Definition
    \ref{def:tmnu-mult} identifies it with the \TMNU associated to the
    multiplication \RNN of Lemma \ref{lem:multiplication_rnn}, which has hidden
    dimension $14$; the \TMNU construction of Definition
    \ref{def:tmnu-simulation-rnn} has $4$ states.

    We next count the distinct command maps. By inspection of Definitions
    \ref{def:sign-tmnu}, \ref{def:contr-+-tmnu}, and
    \ref{def:scale-machine}, the elementary machines have
    \[
        \nFunctions{\Sign}=2,\qquad
        \nFunctions{\Scale}=2,\qquad
        \nFunctions{\Contr^+}=4.
    \]
    Moreover, Definition \ref{def:tmnu-simulation-rnn} gives five command maps
    for a simulated \RNN, and therefore
    $\nFunctions{\Times}=5$. When a machine is built by inserting subroutines,
    we count the union of the extended command maps of the subroutines and the
    additional explicit command maps; the identity command is counted only
    once. Inspection of the coordinates on which the non-identity parts act
    shows that there are no further coincidences between the command maps
    listed below.
    Thus
    \[
        \nFunctions{\Contr}
        =
        \nFunctions{\Sign}+\nFunctions{\Contr^+}-1
        =
        5.
    \]
    The machine $\UpPoly$ adds two explicit commands to the $\Contr$ and
    $\Times$ subroutines, whence
    \[
        \nFunctions{\UpPoly}
        =
        \nFunctions{\Contr}+\nFunctions{\Times}-1+2
        =
        11.
    \]
    Similarly, $\Poly$ adds one explicit command to the $\UpPoly$ and $\Scale$
    subroutines, and $\Continuous$ adds one explicit command to the $\Poly$
    subroutine. Hence
    \[
        \nFunctions{\Poly}=13,
        \qquad
        \nFunctions{\Continuous}=14.
    \]
    On the native Chebyshev side, $\ChebStep$ adds two explicit commands to
    the $\Times$ subroutine, so
    \[
        \nFunctions{\ChebStep}=5+2=7.
    \]
    The machine $\ChebSum$ combines the $\Contr$, $\ChebStep$, and $\Scale$
    subroutines and adds three explicit commands: the initialization command,
    the accumulation command, and the exit command before the final scaling
    subroutine. Therefore
    \[
        \nFunctions{\ChebSum}
        =
        \nFunctions{\Contr}
        +(\nFunctions{\ChebStep}-1)
        +(\nFunctions{\Scale}-1)
        +3
        =
        15.
    \]
    Finally, $\ChebContinuous$ adds one explicit initialization command to the
    $\ChebSum$ subroutine, and hence
    \[
        \nFunctions{\ChebContinuous}=16.
    \]

    It remains to prove the command-magnitude row. We inspect the command maps
    in the order in which the machines are constructed. Throughout, the command
    maps that are not explicitly listed are identities, or frozen complementary
    coordinates coming from subroutines, and therefore have magnitude $1$.
    When a machine contains several disjoint subroutine blocks, we apply Lemma
    \ref{lem:tmnu-magnitude-under-subroutines} successively to these blocks and
    then take the maximum with the magnitudes of the remaining explicit
    commands.
    \begin{itemize}
        \item For $\Sign$, the only nontrivial command is
        $\omega\mapsto\pm\omega$, hence $\mathfrak m(\Sign)=1$.

        \item For $\Scale$, the nontrivial command is
        $\omega\mapsto2\omega$, while all other commands are the identity.
        Hence $\mathfrak m(\Scale)=2$.

        \item For $\Contr^+$, the possible affine commands are
        \[
            (\omega_1,\omega_2)\mapsto(\omega_2,0),\qquad
            (\omega_1,\omega_2)\mapsto(\tfrac12\omega_1,\omega_2),
        \]
        and
        \[
            (\omega_1,\omega_2)\mapsto(\omega_1,\omega_2+b\omega_1),
            \qquad b\in\{\blanksymb,0,1\}.
        \]
        Since $\blanksymb=-1$, these maps have magnitudes at most $1$,
        $1$, and $2$, respectively. Thus
        $\mathfrak m(\Contr^+)=2$.

        \item For $\Times$, we inspect of the RNN multiplication
        construction used in Lemma \ref{lem:multiplication_rnn}. This RNN was introduced  in \cite[Theorem 11]{hutter2025quantifierRNN}. This yields
        $\mathfrak m(\Times)\leq5$.

        \item The machine $\Contr$ uses $\Sign$ and $\Contr^+$ as subroutines
        and has no additional nontrivial command map. Lemma
        \ref{lem:tmnu-magnitude-under-subroutines} therefore gives
        \[
            \mathfrak m(\Contr)
            \leq
            \mathfrak m(\Sign)\vee\mathfrak m(\Contr^+)\vee1
            =
            2.
        \]

        \item The explicit commands of $\UpPoly$ are
        \[
            \omega\mapsto(\omega_1,\omega_2,\omega_3,\omega_2,0_{12})
        \]
        and
        \[
            \omega\mapsto
            (\omega_1,\omega_2,\omega_3+\omega_4,\omega_1,0_{12}),
        \]
        with magnitudes $1$ and $2$. Its subroutines are $\Contr$ and
        $\Times$, hence
        \[
            \mathfrak m(\UpPoly)
            \leq
            2\vee\mathfrak m(\Contr)\vee\mathfrak m(\Times)
            \leq5.
        \]

        \item The only explicit nontrivial command of $\Poly$ outside its
        subroutines is
        \[
            \omega\mapsto(\omega_1,\omega_3,0_{14}),
        \]
        which has magnitude $1$. Its subroutines are $\UpPoly$ and $\Scale$,
        so
        \[
            \mathfrak m(\Poly)
            \leq
            1\vee\mathfrak m(\UpPoly)\vee\mathfrak m(\Scale)
            \leq5.
        \]

        \item The explicit initialization command of $\Continuous$ is
        \[
            \omega\mapsto(\omega_1,1,0_{14},\omega_2),
        \]
        which has magnitude $1$. Since the only subroutine is $\Poly$,
        \[
            \mathfrak m(\Continuous)\leq
            1\vee\mathfrak m(\Poly)
            \leq5.
        \]

        \item For $\ChebStep$, the explicit commands are
        \[
            \omega\mapsto
            (\omega_1,\omega_2,\omega_3,\omega_1,\omega_3,0_{12})
        \]
        and
        \[
            \omega\mapsto
            (\omega_1,\omega_3,2\omega_4-\omega_2,0_{14}),
        \]
        with magnitudes $1$ and $3$. The only subroutine is $\Times$, hence
        $\mathfrak m(\ChebStep)\leq5$.

        \item For $\ChebSum$, the explicit commands outside subroutines have
        magnitudes at most $2$: the initialization command has the row
        $2\omega_2$, the accumulation command has the row
        $\omega_4+\omega_5$, and the exit command
        $\omega\mapsto(\omega_1,\omega_4,0_{16})$ has magnitude $1$. Its
        subroutines are $\Contr$, $\ChebStep$, and $\Scale$, so
        \[
            \mathfrak m(\ChebSum)
            \leq
            2\vee\mathfrak m(\Contr)\vee\mathfrak m(\ChebStep)
            \vee\mathfrak m(\Scale)
            \leq5.
        \]

        \item Finally, the explicit initialization command of
        $\ChebContinuous$ is
        \[
            \omega\mapsto
            (\omega_1,\tfrac12,\tfrac{\omega_1}{2},0_{15},\omega_2),
        \]
        which has magnitude at most $1$. The only subroutine is $\ChebSum$,
        and therefore $\mathfrak m(\ChebContinuous)\leq5$.
    \end{itemize}

    The two displayed bounds for $\|\cdot\|_C$ are now immediate from Lemma
    \ref{lem:tmnu-C-magnitude-from-command-magnitude}.
\end{proof}

% Given a continuous function $f : [-1,1] \to \R$, the $n$-order Chebyshev interpolation polynomial of $f$ is defined by
% \[P^{(f)}_{n}(x) := \sum_{j=0}^n c^{(f)}_{n,j} T_j(x),\]
% where
% \[c^{(f)}_{n,0} := \frac{1}{n} \sum_{k=0}^{n-1} f\left(x_k\right),\]
% and for all $j \in \{1,\ldots,n-1\}$,
% \[c^{(f)}_{n,j} := \frac{2}{n} \sum_{k=0}^{n-1} f\left(x_k\right) T_j\left(x_k\right),\]
% with $x_k := \cos\left(\frac{\pi (k + 0.5)}{n}\right)$ for $k \in \{0,\ldots,n-1\}$. Note that, for all $n \in \No$, and $j \in \{0,\ldots,n-1\}$, we have
% \[
%     |c^{(f)}_{n,j}| \leq \frac{2}{n} \sum_{k=0}^{n-1} |f(x_k)| \leq 2 \|f\|_{L^\infty([-1,1])}.
% \]
% A classical result states that 
% \[
%     \lim_{n \to \infty} \|f - P^{(f)}_{n}\|_{L^\infty([-1,1])} = 0.
% \]
% We denote
% \[
%     S(f,n) := \sup_{k \geq n} \|f - P^{(f)}_{k}\|_{L^\infty([-1,1])} \to_{n \to \infty} 0.
% \]

% \subfile{subfiles/tmnu/tmnu-subroutines-appendix.tex}

\section{Technical results on approximation of polynomials with RNNs}

\label{sec:approximation-polynomials-with-rnns:app}

We begin with the monomial estimate produced by repeated approximate multiplication.
\begin{lemma}\label{lem:mono_approx}
    For all $d,n \in \No$ and $x \in [-1,1]$, we have
    \[
        \left|G_{d,n}(x) - x^d\right| \leq d 2^{-2n-1}.
    \]
    and 
    \[
        |G_{d,n}(x)| \leq 1.
    \]
\end{lemma}
\begin{proof}
    We prove the first statement by induction on $d$.
    \begin{enumerate}[label=(\alph*)]
        \item Base case $d = 0$. For all $n \in \No$ and $x \in [-1,1]$, we have
        \[
            |G_{0,n}(x) - x^0| = |1 - 1| = 0 \leq 0 \cdot 2^{-2n-1}.
        \]
        \item Inductive step. Assume that the statement holds for some $d \in \No$. We prove that it also holds for $d+1$. Let $n \in \No$ and $x \in [-1,1]$. We have
        \begin{align*}
            |G_{d+1,n}(x) - x^{d+1}| &= |\Rc^\times \Dc (G_{d,n}(x),x)[n] - G_{d,n}(x) x + G_{d,n}(x) x - x^{d+1}|\\
            &\leq |\Rc^\times \Dc (G_{d,n}(x),x)[n] - G_{d,n}(x) x| + |G_{d,n}(x) x - x^{d+1}|\\
            &\leq 2^{-2n-1} + |G_{d,n}(x) - x^d| |x|\\
            &\leq 2^{-2n-1} + d 2^{-2n-1} \cdot 1\\
            &= (d+1) 2^{-2n-1},
        \end{align*}
        where we used the induction hypothesis in the last inequality.
    \end{enumerate}
    This concludes the proof of the first statement. The second statement is a direct consequence of Lemma \ref{lem:multiplication_rnn}.
\end{proof}

Summing the monomial estimates gives the corresponding bound for dyadic polynomials.
\begin{lemma}\label{lem:poly_approx}
    Let $P : x \mapsto a_0 x^0 + a_1 x^1 + \cdots + a_d x^d$ be a polynomial with coefficients $a_0, \ldots, a_d \in \Db$. For all $n \in \No$ and $x \in [-1,1]$, we have
    \[
        |F_{P,n}(x)| \leq \|P\|_1.
    \]
    and
    \[
        |F_{P,n}(x) - P(x)| \leq \|P\|_1 \|P\|_0 2^{-2n-1}.
    \]
\end{lemma}
\begin{proof}
    Let $P : x \mapsto a_0 x^0 + a_1 x^1 + \cdots + a_d x^d$ be a polynomial with coefficients $a_0, \ldots, a_d \in \Db$. Let also $n \in \No$ and $x \in [-1,1]$. We have
    \[
        |F_{P,n}(x)| = \left|\sum_{i=0}^d a_i G_{i,n}(x)\right| \leq \sum_{i=0}^d |a_i| |G_{i,n}(x)| \leq \sum_{i=0}^d |a_i| = \|P\|_1,
    \]
    where we used Lemma \ref{lem:mono_approx} in the second inequality. Moreover, we have
    \begin{align*}
        |F_{P,n}(x) - P(x)| &= \left|\sum_{i=0}^d a_i G_{i,n}(x) - \sum_{i=0}^d a_i x^i\right|\\
        &\leq \sum_{i=0}^d |a_i| |G_{i,n}(x) - x^i|\\
        &\leq \sum_{i=0}^d |a_i| i 2^{-2n-1}\\
        &\leq \|P\|_1 \|P\|_0 2^{-2n-1},
    \end{align*}
    where we used Lemma \ref{lem:mono_approx} in the second inequality, and the fact that $i \leq d \leq \|P\|_0$ for all $i \in \{0,\ldots,d\}$ in the last inequality. This concludes the proof.
\end{proof}

% \subfile{subfiles/tmnu/chebychev-approximation-appendix.tex}

\section{Lower bounds on convergence speed}
\label{sec:lower-bound-convergence-speed-appendix}

This appendix gathers lower-bound estimates showing that the runtime dependence
in the RNN approximation paradigm cannot be removed in general.

\subsection{Fixed-time lower bound for the squaring function}
\label{subsec:fixed-time-lower-bound}

We shall use the following elementary terminology. A continuous piecewise-affine
function $g:[-1,1]\to\R$ has at most $N$ breakpoints if there exist
\[
    -1=x_0<x_1<\cdots<x_{N}<x_{N+1}=1
\]
such that $g$ is affine on each interval $[x_j,x_{j+1}]$.

We begin by recording that a fixed-time RNN realization is piecewise affine with controlled breakpoints.
\begin{lemma}\label{lem:rnn-fixed-time-breakpoints}
    Let $\Rc=(1,m,1;A_h,b_h,A_x,A_o,b_o)$ be a scalar-input scalar-output RNN,
    and let $t_0\in\No$. Then the map
    \[
        x\in[-1,1]\longmapsto (\Rc\Dc x)[t_0]
    \]
    is continuous piecewise-affine and has at most $(m+1)^{t_0+1}-1$
    breakpoints.
\end{lemma}
\begin{proof}
    For $t\in\No$, write
    \[
        h_t(x):=(\Hc\Dc x)[t].
    \]
    We prove by induction that each coordinate of $h_t$ is affine on every
    interval of a partition of $[-1,1]$ with at most $(m+1)^{t+1}$ intervals.

    For $t=0$, we have
    \[
        h_0(x)=\ReLU(A_xx+b_h).
    \]
    The $m$ scalar affine functions appearing before the ReLU have at most
    $m$ zeros in total. These zeros partition $[-1,1]$ into at most $m+1$
    intervals, and on each such interval every coordinate of $h_0$ is affine.

    Assume that the claim holds at time $t-1$. On each interval of the
    corresponding partition, each coordinate of $h_{t-1}$ is affine. Therefore
    each coordinate of
    \[
        A_hh_{t-1}(x)+b_h
    \]
    is affine on that interval and has at most one zero there. Adding the zeros
    of these $m$ affine functions refines the interval into at most $m+1$
    subintervals. Thus the number of intervals is multiplied by at most $m+1$,
    and on every resulting interval
    \[
        h_t(x)=\ReLU(A_hh_{t-1}(x)+b_h)
    \]
    is affine. This proves that $h_{t_0}$ is affine on at most
    $(m+1)^{t_0+1}$ intervals. Since $(\Rc\Dc x)[t_0]=A_oh_{t_0}(x)+b_o$, the
    same partition works for the output map, which therefore has at most
    $(m+1)^{t_0+1} - 1$ breakpoints.
\end{proof}

The lower-bound argument needs a simple counting estimate for approximating a quadratic by piecewise-affine functions.
\begin{lemma}\label{lem:quadratic-piecewise-linear-breakpoints}
    Let $\varepsilon>0$, and let $g:[-1,1]\to\R$ be continuous
    piecewise-affine. If
    \[
        \sup_{x\in[-1,1]}|g(x)-x^2|\leq\varepsilon,
    \]
    then $g$ has at least $\frac1{\sqrt{2\varepsilon}} - 1$ breakpoints.
\end{lemma}
\begin{proof}
    We first record a local obstruction. Let $I=[a,b]\subset[-1,1]$ have length
    $\ell=b-a$, and let $\lambda$ be affine on $I$. Denote by $m=(a+b)/2$ the
    midpoint of $I$ and put $r=\ell/2$. Define
    \[
        e(y):=(m+y)^2-\lambda(m+y),
        \qquad y\in[-r,r].
    \]
    Since $\lambda$ is affine, there exist $\alpha,\beta\in\R$ such that
    \[
        e(y)=y^2+\alpha y+\beta.
    \]
    In particular,
    \[
        e(0)=\beta,\qquad
        e(r)=r^2+\alpha r+\beta,
        \qquad
        e(-r)=r^2-\alpha r+\beta,
    \]
    yielding
    \[
        \frac{e(r)+e(-r)}{2}-e(0)=r^2.
    \]
    If $E:=\sup_{x\in I}|x^2-\lambda(x)|$, then
    $|e(-r)|,|e(0)|,|e(r)|\leq E$. Therefore
    \[
        r^2
        =
        \left|\frac{e(r)+e(-r)}{2}-e(0)\right|
        \leq
        \frac{|e(r)|+|e(-r)|}{2}+|e(0)|
        \leq
        2E.
    \]
    Since $r=\ell/2$, we obtain $E\geq r^2/2=\ell^2/8$. Hence every affine
    $\lambda$ satisfies
    \[
        \sup_{x\in I}|x^2-\lambda(x)|\geq\frac{\ell^2}{8}.
    \]

    Suppose now that $g$ has $N$ breakpoints. Then $[-1,1]$ is decomposed into
    $N+1$ intervals $I_0,\ldots,I_N$ on which $g$ is affine. Fix one of these
    intervals, say $I_j$, and denote its length by $\ell_j$. Since $g$ is affine
    on $I_j$, the local estimate above can be applied with
    $\lambda=g_{\mid I_j}$. Hence
    \[
        \frac{\ell_j^2}{8}
        \leq
        \sup_{x\in I_j}|x^2-g(x)|
        \leq
        \sup_{x\in[-1,1]}|x^2-g(x)|
        \leq
        \varepsilon.
    \]
    Therefore $\ell_j\leq\sqrt{8\varepsilon}$. Since this holds for every
    $j\in\{0,\ldots,N\}$ and since the lengths of the intervals sum to $2$, we
    get
    \[
        N+1\geq\frac{2}{\sqrt{8\varepsilon}}
        =
        \frac{1}{\sqrt{2\varepsilon}},
    \]
    as claimed.
\end{proof}

Combining Lemmas \ref{lem:rnn-fixed-time-breakpoints} and
\ref{lem:quadratic-piecewise-linear-breakpoints} shows the following. If an RNN
$\Rc$ with scalar input and scalar output satisfies
\[
    \sup_{x\in[-1,1]}|(\Rc\Dc x)[t_0]-x^2|\leq\varepsilon
\]
for some $\varepsilon > 0$, then
\[
    (\rnnHidDim{\Rc}+1)^{t_0+1} - 1
    \geq
    \frac{1}{\sqrt{2\varepsilon}} - 1,
\]
and hence
\[
    \rnnHidDim{\Rc}
    \geq \left(\frac1{2\varepsilon}\right)^{\frac1{2(t_0+1)}} - 1.
\]

\subsection{Minimax errors over compact classes}
\label{subsec:minimax-lower-bound-notation}

We now introduce the notation used for minimax lower bounds. For
$m\in\N$ and $B\geq0$, let
\[
    \mathfrak R_{m,B}
    :=
    \left\{
        \Rc=(1,m',1;A_h,b_h,A_x,A_o,b_o)
        \,:\,
        m'\leq m,\ \nnMag{\Rc}\leq B
    \right\}.
\]
Thus $\mathfrak R_{m,B}$ is the class of scalar-input scalar-output \RNNs whose
hidden state size is at most $m$ and whose weights have magnitude at most $B$.
For $t\in\No$ and $\Rc\in\mathfrak R_{m,B}$, we define
\[
    \Phi_t(\Rc):[-1,1]\to\R,
    \qquad
    \Phi_t(\Rc)(x):=(\Rc\Dc x)[t].
\]
Equivalently, the set of functions realized at time $t$ by RNNs in
$\mathfrak R_{m,B}$ is
\[
    \mathcal A_t(m,B)
    :=
    \left\{
        \Phi_t(\Rc)
        \,:\,
        \Rc\in\mathfrak R_{m,B}
    \right\}
    \subset C([-1,1]).
\]
Let $\Xc\subset C([-1,1])$ be compact for the uniform norm
\[
    \|f\|_\infty:=\sup_{x\in[-1,1]}|f(x)|.
\]
The fixed-time minimax error of the class $\mathfrak R_{m,B}$ over $\Xc$ is
defined by
\[
    \mathcal E_t(\Xc;m,B)
    :=
    \sup_{f\in\Xc}
    \inf_{\Rc\in\mathfrak R_{m,B}}
    \sup_{x\in[-1,1]}
    \left|(\Rc\Dc x)[t]-f(x)\right|.
\]
In terms of the realized function class $\mathcal A_t(m,B)$, this is simply
\[
    \mathcal E_t(\Xc;m,B)
    =
    \sup_{f\in\Xc}
    \inf_{g\in\mathcal A_t(m,B)}
    \|g-f\|_\infty.
\]
The lower bounds below will estimate this quantity from below as a function of
the time $t$, the hidden state size $m$, and the weight magnitude $B$. We next recall the metric-entropy language used to formulate these lower bounds.
\begin{definition}[Covering number and metric entropy]
\label{def:covering-number-metric-entropy}
    Let $\Yc\subset C([-1,1])$ be a compact set and $\varepsilon>0$. We define the $\varepsilon$-covering
    number of $\Yc$ by
    \[
        \mathcal N(\varepsilon,\Yc)
        :=
        \inf\left\{
            N\in\N
            \,:\,
            \exists g_1,\ldots,g_N\in C([-1,1]),
            \ \Yc\subset\bigcup_{j=1}^{N}
            \overline B_\infty(g_j,\varepsilon)
        \right\},
    \]
    where
    \[
        \overline B_\infty(g,\varepsilon)
        :=
        \left\{h\in C([-1,1])\,:\,\|h-g\|_\infty\leq\varepsilon\right\}.
    \]
    The metric entropy of $\Yc$ at scale $\varepsilon$ is
    \[
        \mathcal H(\varepsilon,\Yc)
        :=
        \log\mathcal N(\varepsilon,\Yc).
    \]
\end{definition}
We next define the parameter-to-realization map.
\begin{definition}[Parameter-to-realization map]
\label{def:parameter-to-realization-map}
    Let $m\in\N$ and $t\in\No$, and set
    \[
        p_m:=m^2+3m+1.
    \]
    We identify $\R^{p_m}$ with
    \[
        \R^{m\times m}\times\R^m\times\R^{m\times1}
        \times\R^{1\times m}\times\R.
    \]
    For
    \[
        \theta=(A_h,b_h,A_x,A_o,b_o)
        \in\R^{p_m},
    \]
    let
    \[
        \Rc_\theta:=(1,m,1;A_h,b_h,A_x,A_o,b_o),
    \]
    and define
    \[
        F_{t,m}:\R^{p_m}\to C([-1,1]),
        \qquad
        F_{t,m}(\theta)(x):=\Phi_t(\Rc_\theta)(x).
    \]
\end{definition}
Note that Definition \ref{def:parameter-to-realization-map} gives, for
every $B\geq0$,
    \[
        \mathcal A_t(m,B)
        =
        F_{t,m}\left([-B,B]^{p_m}\right).
    \]
The following lemma gives Lipschitz control of this realization map with respect to the parameters.
\begin{lemma}\label{lem:rnn-realization-lipschitz}
    Let $m\in\N$, $B\geq1$, and $t\in\No$. Define
    \[
        \Lambda_{t,m,B}:=8B^2m^2(2mB)^{2t}.
    \]
    Then $F_{t,m}$ is $\Lambda_{t,m,B}$-Lipschitz from
    $([-B,B]^{m^2+3m+1},\|\cdot\|_\infty)$ to
    $(C([-1,1]),\|\cdot\|_\infty)$.
\end{lemma}
\begin{proof}
    Let $\theta,\theta'\in[-B,B]^{m^2+3m+1}$ be two parameter vectors, and let
    \[
        \delta:=\|\theta-\theta'\|_\infty.
    \]
    We write $\Rc,\Rc'$ for the corresponding RNNs and
    \[
        h_s(x):=(\Hc\Dc x)[s],
        \qquad
        h'_s(x):=(\Hc'\Dc x)[s],
        \qquad s\in\No.
    \]
    Set
    \[
        S_{t,m,B}:=\sum_{j=0}^{t}(mB)^j
        \qquad \text{and}\qquad
        H_{t,m,B}:=2B S_{t,m,B}.
    \]
    First we bound the size of the hidden states. First note that
    \[
        \|h_0(x)\|_\infty = \|\ReLU(A_h h_{-1}(x) + A_x(\Dc x)[0]+b_h)\|_\infty = \|\ReLU(A_x x+b_h)\|_\infty \leq \|A_x x\|_\infty + \|b_h\|_\infty \leq B|x| + B \leq 2B,
    \]
    for every $x\in[-1,1]$. Similarly, $\|h'_0(x)\|_\infty \leq 2B$.
    Moreover, for every $s\in\{1,\ldots,t\}$ and $x\in[-1,1]$ we have
    \begin{align*}
        \|h_s(x)\|_\infty &= \|\ReLU(A_hh_{s-1}(x)+A_x(\Dc x)[s]+b_h)\|_\infty = \|\ReLU(A_hh_{s-1}(x)+b_h)\|_\infty \leq \|A_hh_{s-1}(x)\|_\infty + \|b_h\|_\infty\\
        & \leq mB\|h_{s-1}(x)\|_\infty + B \leq mB\|h_{s-1}(x)\|_\infty + 2B.
    \end{align*}
    It follows by induction that
    \begin{equation}\label{eq:hidden-bound:minimax}
        \|h_s(x)\|_\infty 
        \leq
        2B\sum_{j=0}^{s}(mB)^j
        \leq
        H_{t,m,B},
        \qquad
        s\in\{0,\ldots,t\},
    \end{equation}
    and we have the same bound for $h'_s(x)$.
    We now estimate the sensitivity of the hidden state to the parameters. Put
    \[
        D_s:=\sup_{x\in[-1,1]}\|h_s(x)-h'_s(x)\|_\infty.
    \]
    Using the $1$-Lipschitz property of $\ReLU$, for every
    $s\in\{0,\ldots,t\}$ and every $x\in[-1,1]$, we have
    \begin{align*}
        \|h_s(x)-h'_s(x)\|_\infty
        &\leq
        \|A_hh_{s-1}(x)+A_x(\Dc x)[s]+b_h
        -A'_hh'_{s-1}(x)-A'_x(\Dc x)[s]-b'_h\|_\infty\\
        &\leq
        \|A'_h(h_{s-1}(x)-h'_{s-1}(x))\|_\infty
        +
        \|(A_h-A'_h)h_{s-1}(x)\|_\infty\\
        &\quad
        +
        \|(A_x-A'_x)(\Dc x)[s]\|_\infty
        +
        \|b_h-b'_h\|_\infty\\
        &\leq
        mB D_{s-1}
        +
        mH_{t,m,B}\delta
        +
        \delta
        +
        \delta.
    \end{align*}
    Taking the supremum over $x\in[-1,1]$ gives
    \[
        D_s
        \leq
        mB D_{s-1}
        +
        (mH_{t,m,B}+2)\delta,
    \]
    with $D_{-1}=0$. Therefore
    \begin{equation}\label{eq:hidden-lipschitz:minimax}
        D_t
        \leq
        (mH_{t,m,B}+2)\delta\sum_{j=0}^{t}(mB)^j.
    \end{equation}

    Finally, for the output layer,
    \[
        |(\Rc\Dc x)[t]-(\Rc'\Dc x)[t]|
        \leq
        Bm\|h_t(x)-h'_t(x)\|_\infty
        +
        mH_{t,m,B}\delta
        +
        \delta.
    \]
    Combining this with \eqref{eq:hidden-lipschitz:minimax} gives
    \begin{align*}
        \|F_{t,m}(\theta)-F_{t,m}(\theta')\|_\infty
        &\leq
        \left[
        1
        +
        Bm(mH_{t,m,B}+2)\sum_{j=0}^{t}(mB)^j
        +
        mH_{t,m,B}
        \right]\delta \leq \Lambda_{t,m,B}\delta.
    \end{align*}
\end{proof}

This Lipschitz control converts parameter dimension into a covering-number bound.
\begin{lemma}\label{lem:metric-entropy-realized-class}
    Let $m\in\N$, $B\geq1$, $t\in\No$, and $\varepsilon>0$. Then
    \[
        \mathcal N\left(\varepsilon,\mathcal A_t(m,B)\right)
        \leq
        \left(
            1+\frac{2B \Lambda_{t,m,B}}{\varepsilon}
        \right)^{m^2+3m+1}.
    \]
\end{lemma}
\begin{proof}
    Let $p:=m^2+3m+1$. The cube $[-B,B]^p$ can be covered, for the
    $\ell^\infty$ norm, by at most
    \[
        \left(1+\frac{2B \Lambda_{t,m,B}}{\varepsilon}\right)^p
    \]
    balls of radius $\varepsilon/\Lambda_{t,m,B}$. By Lemma
    \ref{lem:rnn-realization-lipschitz}, the image under $F_{t,m}$ of each such
    ball is contained in a ball of radius $\varepsilon$ in $C([-1,1])$. Since
    $\mathcal A_t(m,B)\subset F_{t,m}([-B,B]^p)$ by Definition
    \ref{def:parameter-to-realization-map}, this yields the claimed covering
    bound for $\mathcal A_t(m,B)$.
\end{proof}

The next comparison turns small uniform approximation error into an entropy inequality.
\begin{lemma}\label{lem:minimax-entropy-comparison}
    Let $\Xc\subset C([-1,1])$ be compact, let $m\in\N$, $B\geq1$,
    and $t\in\No$. Then
    \[
        \mathcal E_t(\Xc;m,B)
        \geq
        \inf\left\{
            \varepsilon>0
            \,:\,
            \mathcal N(2\varepsilon,\Xc)
            \leq
            \mathcal N\left(\varepsilon,\mathcal A_t(m,B)\right)
        \right\}.
    \]
\end{lemma}
\begin{proof}
    Let $\eta>\mathcal E_t(\Xc;m,B)$. By definition of
    $\mathcal E_t(\Xc;m,B)$, for every $f\in\Xc$ there exists
    $g_f\in\mathcal A_t(m,B)$ such that
    \[
        \|f-g_f\|_\infty<\eta.
    \]
    Let $g_1,\ldots,g_M$ be an $\eta$-covering of $\mathcal A_t(m,B)$, with
    $M=\mathcal N(\eta,\mathcal A_t(m,B))$. For each $f\in\Xc$, choose
    $j\in\{1,\ldots,M\}$ such that $\|g_f-g_j\|_\infty\leq\eta$. Then
    \[
        \|f-g_j\|_\infty
        \leq
        \|f-g_f\|_\infty+\|g_f-g_j\|_\infty
        <
        2\eta.
    \]
    Hence $\{g_1,\ldots,g_M\}$ is a $2\eta$-covering of $\Xc$, and so
    \[
        \mathcal N(2\eta,\Xc)
        \leq
        \mathcal N\left(\eta,\mathcal A_t(m,B)\right).
    \]
    Therefore for every $\eta>\mathcal E_t(\Xc;m,B)$, we also have
    \[
        \eta\in 
        \left\{ 
            \varepsilon>0
            \,:\,
            \mathcal N(2\varepsilon,\Xc)
            \leq
            \mathcal N\left(\varepsilon,\mathcal A_t(m,B)\right)
        \right\}.
    \]
    Taking the infimum over $\eta>\mathcal E_t(\Xc;m,B)$ gives the claim.
\end{proof}

Combining the two entropy estimates gives a necessary condition for minimax approximation.
\begin{corollary}\label{cor:necessary-entropy-condition-minimax}
    Let $\Xc\subset C([-1,1])$ be compact, let $m\in\N$, $B\geq1$,
    and $t\in\No$. Then
    \[
        \mathcal E_t(\Xc;m,B)
        \geq
        \frac12
        \inf\left\{
            \varepsilon\in(0,1]
            \,:\,
            \mathcal N(\varepsilon,\Xc)
            \leq
            \left(
                \frac{(2mB)^{2t+7}}{\varepsilon}
            \right)^{5m^2}
        \right\}
    \]
\end{corollary}
\begin{proof}
    Let
    \[
        p:=m^2+3m+1
    \]
    and
    \[
        \mathcal S
        :=
        \left\{
            \varepsilon\in(0,1]
            \,:\,
            \mathcal N(\varepsilon,\Xc)
            \leq
            \left(
                \frac{(2mB)^{2t+7}}{\varepsilon}
            \right)^{5m^2}
        \right\}.
    \]
    By Lemma \ref{lem:metric-entropy-realized-class}, for every
    $\varepsilon>0$,
        \[
            \mathcal N\left(\varepsilon,\mathcal A_t(m,B)\right)
            \leq
            \left(
                1+\frac{2B \Lambda_{t,m,B}}{\varepsilon}
            \right)^p.
        \]
    Since $m\geq1$ and $B\geq1$, we have
    \[
        p=m^2+3m+1\leq5m^2
    \]
    and
    \[
        2B\Lambda_{t,m,B}
        =
        16B^3m^2(2mB)^{2t}
        \leq
        (2mB)^{2t+5}.
    \]
    Therefore
    \[
        \mathcal N\left(\varepsilon,\mathcal A_t(m,B)\right)
        \leq
        \left(
            1+\frac{(2mB)^{2t+5}}{\varepsilon}
        \right)^{5m^2}.
    \]
    Hence, if $\rho\in(0,1]$ and
    \[
        \mathcal N(\rho,\Xc)
        \leq
        \mathcal N\left(\frac{\rho}{2},\mathcal A_t(m,B)\right),
    \]
    then
    \[
        \mathcal N(\rho,\Xc)
        \leq
        \left(
            \frac{(2mB)^{2t+6}}{\rho/2}
        \right)^{5m^2}
        \leq
        \left(
            \frac{(2mB)^{2t+7}}{\rho}
        \right)^{5m^2}.
    \]
    Hence
    \[
        \left\{
            \rho\in(0,1]
            \,:\,
            \mathcal N(\rho,\Xc)
            \leq
            \mathcal N\left(\frac{\rho}{2},\mathcal A_t(m,B)\right)
        \right\}
        \subset
        \mathcal S.
    \]
    If $\mathcal E_t(\Xc;m,B)\geq1/2$, then the claim is immediate. Assume
    $\mathcal E_t(\Xc;m,B)<1/2$. The proof of Lemma
    \ref{lem:minimax-entropy-comparison} shows that every
    $\eta>\mathcal E_t(\Xc;m,B)$ satisfies
    \[
        \mathcal N(2\eta,\Xc)
        \leq
        \mathcal N\left(\eta,\mathcal A_t(m,B)\right).
    \]
    Thus every $2\eta$ with
    $\eta\in(2\mathcal E_t(\Xc;m,B),1/2]$ belongs to $\mathcal S$, and so
    \[
        \inf\mathcal S
        \leq
        2\mathcal E_t(\Xc;m,B).
    \]
    This proves the claim.
\end{proof}

% \subsection{Explicit minimax rates from entropy}
% \label{subsec:explicit-minimax-rates-from-entropy}

We now spell out what Corollary
\ref{cor:necessary-entropy-condition-minimax} gives for the function classes
used in the main text. Throughout this subsection, we fix $m\in\N$, $B\geq1$,
and put
\[
    A_{t,m,B}:=5m^2(2t+7)\log(2mB),
    \qquad
    C_m:=5m^2.
\]
With this notation, the condition appearing in Corollary
\ref{cor:necessary-entropy-condition-minimax} can be written as
\begin{equation}\label{eq:entropy-condition:explicit-rates}
    \log\mathcal N(\varepsilon,\Xc)
    \leq
    A_{t,m,B}+C_m\log(\varepsilon^{-1}).
\end{equation}
Thus the minimax lower bound is obtained by inverting the metric entropy of
$\Xc$. The following theorem gives a convenient rate-level inversion, avoiding
the exact transcendental formulas.
\begin{theorem}[Entropy inversion]\label{thm:entropy-inversion}
    Let $\Xc\subset C([-1,1])$ be compact. Let
    $h:(0,1]\to(0,\infty)$ be nonincreasing and assume that
    \begin{equation}\label{eq:thm:entropy-inversion}
        \log(\varepsilon^{-1})=o(h(\varepsilon))
        \qquad
        \text{and} \qquad \log\mathcal N(\varepsilon,\Xc)
        \geq
        h(\varepsilon), \quad \varepsilon \in (0,1].
    \end{equation}
    Then there exist constants $K>0$ and $t_0\in\N$, depending only on
    $m,B$ and $h$, such that, for every $t\geq t_0$,
    \[
        \mathcal E_t(\Xc;m,B)
        \geq
        \frac12 h^{-1}(K(t+1)),
    \]
    where $h^{-1}(T)
        :=
        \inf\left\{
            \varepsilon\in(0,1]
            \,:\,
            h(\varepsilon)\leq T
        \right\}$.
\end{theorem}
\begin{proof}
    Since $\log(\varepsilon^{-1})=o(h(\varepsilon))$, there exists
    $\varepsilon_1\in(0,1]$ such that
    \begin{equation}\label{eq:log-negligible:proof:entropy-inversion}
        C_m\log(\varepsilon^{-1})
        \leq
        \frac 12 h(\varepsilon),
        \qquad
        \varepsilon\in(0,\varepsilon_1].
    \end{equation}
    Let $\varepsilon\in(0,\varepsilon_1]$ satisfy
    \eqref{eq:entropy-condition:explicit-rates}. Combining
    \eqref{eq:entropy-condition:explicit-rates}, \eqref{eq:thm:entropy-inversion} and
    \eqref{eq:log-negligible:proof:entropy-inversion}, we obtain
    \[
        h(\varepsilon)
        \leq
        A_{t,m,B}+C_m\log(\varepsilon^{-1})
        \leq
        A_{t,m,B}+\frac 12 h(\varepsilon), 
    \]
    hence
    \[
        h(\varepsilon)
        \leq
        2 A_{t,m,B}.
    \]
    Choose $K>0$ such that
    \[
        2 A_{t,m,B}\leq K(t+1),
        \qquad
        t\in\No.
    \]
    Then every $\varepsilon\in(0,\varepsilon_1]$ satisfying
    \eqref{eq:entropy-condition:explicit-rates} also satisfies
    \[
        \varepsilon\geq h^{-1}(K(t+1)).
    \]
    Since $h^{-1}(K(t+1))\to0$ as $t\to\infty$, there exists $t_0\in\N$ such
    that $h^{-1}(K(t+1))\leq\varepsilon_1$ for every $t\geq t_0$. For such
    $t$, the same lower bound is automatic for admissible scales
    $\varepsilon>\varepsilon_1$. Therefore
    \[
        \inf\left\{
            \varepsilon\in(0,1]
            \,:\,
            \mathcal N(\varepsilon,\Xc)
            \leq
            \left(
                \frac{(2mB)^{2t+7}}{\varepsilon}
            \right)^{5m^2}
        \right\}
        \geq
        h^{-1}(K(t+1)).
    \]
    Corollary \ref{cor:necessary-entropy-condition-minimax} proves the claim.
\end{proof}

\end{document}

\subsection{Construction of an RNN that simulates the iterations of a neural network}

\label{subsection:simulation-of-a-tmnu-by-an-rnn:appendix}

In this appendix, we formally show that we can simulate the iteration of a neural network by an RNN.

In this section, we will refer at neural networks as if they were functions. Specifically, we write that $\nn : \R^n \to \R^m$ is a neural network if $\nn$ is a neural network such that $\nnInDim{\nn} =n $ and $\nnOutDim{\nn} = m$. We say that a function $f : \R^n \to \R^m$ is non-negative, and we write $f \geq 0$ if $f_i(x) \geq 0, i \in \{1, \ldots, m\}$ for every $x \in \R^n$. Accordingly, we say that a neural network $\nn$ is non-negative, and we write $\nn \geq 0$ if $\nn(x) \geq 0$ for every $x \in \R^n$, where $n = \nnInDim{\nn}$.

We introduce two notations for RNNs. $\rnnHidDim{\Rc}$ denotes the dimension of the hidden state of $\Rc$, and $\nnWeights{\Rc}$ denotes the set of weights of $\Rc$, that is, the set of entries of the matrices and vectors defining $\Rc$.

\begin{lemma}\label{lem:simulation-of-cpwl-function-iterations-with-an-rnn:app}
    Let $\nn : \R^n \to \R^n$ be a neural network such that $\nn \geq 0$, and let $L := \nnDepth{\nn}$. Then, there exists $m \in \N$ satisfying $n + L\leq m \leq (\nnWidth{\nn}+1) L$ and $A_\nn \in \R^{m \times m}$ such that $\nnWeights{A_\nn} = \nnWeights{\nn} \cup \{0,1\}$ and for every $x \in \R^n$, the sequence $(h_t \in \R^m)_{t \in \N}$ defined by
    \begin{equation}
        h_0 := (x, 0, \oneHot{L}{1}), \quad h_{t+1} := \ReLU(A_\nn h_t) \text{ for every } t \in \No
    \end{equation}
    satisfies
    \begin{equation}
        \proj_{1:n} h_t = \nn^{t//L}(x) \kroen{t \bmod L}{0}, \quad t \in \No,
    \end{equation}
    where $//$ denotes the integer division.
\end{lemma}
\begin{proof}
    Let $n \in \N$, $\nn := (N_0, N_1, \ldots, N_L; A^1, b^1, A^2, b^2, \ldots, A^L, b^L)$ be a neural network, such that $N_0 = N_L =n$ and$\nn \geq 0$. We let $N := n + \sum_{\ell=1}^{L-1} N_\ell$ and $m := N + L$, and we define the matrices 
    \[
        A := \begin{bmatrix}
            0 & 0 & \ldots & 0 & A^L\\
            A^1 & 0 & \ldots & 0 & 0\\
            0 & A_2 & \ldots & 0 & 0\\
            \vdots & \vdots & \ddots & \vdots & \vdots\\
            0 & 0 & \ldots & A^{L-1} & 0
         \end{bmatrix} \in \R^{N \times N},
        \quad 
        B := \begin{bmatrix}
            0 & 0 & \ldots & 0 & b^L\\
            b^1 & 0 & \ldots & 0 & 0\\
            0 & b_2 & \ldots & 0 & 0\\
            \vdots & \vdots & \ddots & \vdots & \vdots\\
            0 & 0 & \ldots & b^{L-1} & 0
         \end{bmatrix} \in \R^{N \times L},
    \]
    \[
         C := \begin{bmatrix}
            0 & 0 & \ldots & 0 & 1\\
            1 & 0 & \ldots & 0 & 0\\
            0 & 1 & \ldots & 0 & 0\\
            \vdots & \vdots & \ddots & \vdots & \vdots\\
            0 & 0 & \ldots & 1 & 0
         \end{bmatrix} \in \R^{L \times L}, \quad \text{and} \quad A_\nn := \begin{bmatrix}
            A & B\\
            0 & C
         \end{bmatrix} \in \R^{m \times m}.
    \]
    Note that, indeed, we have \[n + L \leq m = N + L = n + \sum_{\ell=1}^{L-1} N_\ell + L \leq L \nnWidth{\nn} + L = (\nnWidth{\nn}+1) L,\] and that $\nnWeights{A_\nn} = \nnWeights{\nn} \cup \{0,1\}$. Now, let $x \in \R^n$ and let $(h_{x,t} \in \R^m)_{t \in \N}$ be defined by $h_{x,0} := (x, 0, \oneHot{L}{1})$ and $h_{x,t+1} := \ReLU(A_\nn h_{x,t})$ for every $t \in \No$. We split the sequence $(h_{x,t})_{t \in \N}$ into subsequences $(h_{x,t}^\ell \in \R^{N_\ell})_{t \in \N}$ and $(c_t^\ell \in \R)_{t \in \N}$ for $\ell \in \{0, \ldots, L-1\}$ such that
    \[
        h_{x,t} = (h_{x,t}^0, h_{x,t}^1, \ldots, h_{x,t}^{L-1}, c_t^0, \ldots, c_t^{L-1}) \text{ for every } t \in \N.
    \]
    Note that, in particular,
    \begin{equation}\label{eq:simulation-of-cpwl-function-iterations-with-an-rnn:recursive-definition:base-case}
        h_{x,0}^0 = x, \quad h_{x,0}^1 = 0, \quad \ldots, \quad h_{x,0}^{L-1} = 0, \quad c_0^0 = 1, \quad c_0^1 = 0, \quad \ldots, \quad c_0^{L-1} = 0,
    \end{equation}
    and that
    \begin{equation}\label{eq:simulation-of-cpwl-function-iterations-with-an-rnn:recursive-definition:h}
        h_{x,t+1}^0  = \ReLU(A^L h_{x,t}^{L-1} + b^L c_t^{L-1}), \quad 
        h_{x,t+1}^\ell = \ReLU(A^\ell h_{x,t}^{\ell-1} + b^\ell c_t^{\ell-1}),
    \end{equation}
    and
    \begin{equation}\label{eq:simulation-of-cpwl-function-iterations-with-an-rnn:recursive-definition:c}
        c_{t+1}^0  = \ReLU(c_t^{L-1}), \quad c_{t+1}^\ell  = \ReLU(c_t^{(\ell-1)}),
    \end{equation}
    for every $\ell \in \{1, \ldots, L-1\}$ and $t \in \No$.
    % Note also that for every $t \in \N$, $h_t^0 = \proj_{1:n} h_t$, so that the statement of the Lemma is equivalent to showing that for every $t \in \N$, we have
    % \begin{equation}\label{eq:simulation-of-cpwl-function-iterations-with-an-rnn:induction-hypothesis}
    %     h_t^0 = \begin{cases}
    %         f^{t/L}(x) & \text{if } t \in L \No,\\
    %         0 & \text{if } t \notin L \No.
    %     \end{cases}
    % \end{equation}
    We divide the proof into several claims.
    \begin{claim*}
        For every $t \in \N$, $\ell \in \{0,\ldots, L-1\}$, we have $c_t^\ell = \kroen{\ell}{t \mod L}$.
    \end{claim*}
    \begin{proof}[Proof of the claim]
        We proceed by induction on $t$. For the base case, note that for every $\ell \in \{0,\ldots, L-1\}$, we have $c_0^\ell = \oneHot{L}{1}(\ell) = \kroen{\ell}{0}$. Now, let $t \in \N$ and assume that for every $\ell \in \{0,\ldots, L-1\}$, we have $c_t^\ell = \kroen{\ell}{t \mod L}$. Then, by \eqref{eq:simulation-of-cpwl-function-iterations-with-an-rnn:recursive-definition:c}, we have $c_{t+1}^0 = \ReLU(c_t^{L-1}) = \ReLU(\kroen{L-1}{t \mod L}) = \kroen{0}{(t+1) \mod L}$, and for every $\ell \in \{1, \ldots, L-1\}$, we have $c_{t+1}^\ell = \ReLU(c_t^{(\ell-1)}) = \ReLU(\kroen{\ell-1}{t \mod L}) = \kroen{\ell}{(t+1) \mod L}$. This concludes the proof of the claim.
    \end{proof}
    \begin{claim*}
        For every $t\in \{0, \ldots, L-1\}$, we have $h_{x,t}^0 = x \kroen{t}{0}$ and $h_{x,t}^\ell = \ReLU(\nn_\ell(x))\kroen{t}{\ell}$ for every $\ell \in \{1, \ldots, L-1\}$.
    \end{claim*}
    \begin{proof}[Proof of the claim]
        We proceed by induction on $t$. For the base case, note that by \eqref{eq:simulation-of-cpwl-function-iterations-with-an-rnn:recursive-definition:base-case}, we have $h_{x,0}^\ell = x \kroen{0}{\ell} = \ReLU(\nn_0(x)) \kroen{0}{\ell}$, for every $\ell \in \{0,\ldots, L-1\}$. Now, let $t \in \{0, \ldots, L-2\}$ and assume that for every $\ell \in \{0,\ldots, L-1\}$, we have $h_{x,t}^\ell = \nn_\ell(x)\kroen{t}{\ell}$. Then, by \eqref{eq:simulation-of-cpwl-function-iterations-with-an-rnn:recursive-definition:h}, we have 
        \begin{align*}
            h_{x,t+1}^0 &= \ReLU(A^L h_{x,t}^{L-1} + b^L c_t^{L-1}) = \ReLU(A^L \nn_{L-1}(x)\kroen{t}{L-1} + b^L \kroen{t}{L-1})\\
            & = \ReLU((A^L \nn_{L-1}(x) + b^L)\kroen{t}{L-1}) = \ReLU(A^L \nn_{L-1}(x) + b^L)\kroen{t}{L-1}\\
            & = 0 = x\kroen{t+1}{0},
        \end{align*}
        and for every $\ell \in \{1, \ldots, L-1\}$, we have \begin{align*}
            h_{x,t+1}^\ell &= \ReLU(A^\ell h_{x,t}^{\ell-1} + b^\ell c_t^{\ell-1}) = \ReLU(A^\ell \nn_{\ell-1}(x)\kroen{t}{\ell-1} + b^\ell \kroen{t}{\ell-1})\\
            & = \ReLU((A^\ell \nn_{\ell-1}(x) + b^\ell)\kroen{t}{\ell-1}) = \ReLU(A^\ell \nn_{\ell-1}(x) + b^\ell)\kroen{t}{\ell-1}\\
            & = \ReLU(\nn_\ell(x))\kroen{t+1}{\ell}.
        \end{align*}
        This concludes the proof of the claim.
    \end{proof}
    \begin{claim*}
        For every $k \in \N$, we have $h_{x,kL} = (\nn^k(x), 0, \oneHot{L}{1})$.
    \end{claim*}
    \begin{proof}[Proof of the claim]
        First note that 
        \begin{equation}
            h_{x,L}^0 = \ReLU(A^L h_{x,L-1}^{L-1} + b^L c_{L-1}^{L-1}) \overset{(a)}= \ReLU(A^L \nn_{L-1}(x) + b^L) = \ReLU(\nn_L(x)) = \ReLU(\nn(x)) \overset{(b)}= \nn(x),
        \end{equation}
        and
        \begin{equation}
            h_{x,L}^\ell = \ReLU(A^\ell h_{x,L-1}^{\ell-1} + b^\ell c_{L-1}^{\ell-1}) \overset{(a)}= \ReLU(A^\ell \nn_{\ell-1}(x) + b^\ell) \kroen{L}{\ell} = 0,
        \end{equation}
         for every $\ell \in \{1, \ldots, L-1\}$, where (a) follows from the preceding two claims and (b) is by $\nn \geq 0$. Therefore,
        \begin{equation}
            h_{x,L} = (h_{x,L}^0, h_{x,L}^1, \ldots, h_{x,L}^{L-1}, c_L^0, \ldots, c_L^{L-1}) = (\nn(x), 0, \oneHot{L}{1}).
        \end{equation}
        We now proceed by induction on $k$. For the base case, note that $h_0 = (x,0,\oneHot{L}{1}) = (\nn^0(x),0,\oneHot{L}{1})$. Now, let $k \in \N$ and assume that $h_{x,kL} = (\nn^k(x),0,\oneHot{L}{1})$. Therefore, we have
        \begin{align*}
            h_{x,(k+1)L} = h_{\nn^k(x),L} = (\nn(\nn^k(x)), 0, \oneHot{L}{1}) = (\nn^{k+1}(x), 0, \oneHot{L}{1}).
        \end{align*}
        This concludes the proof of the claim.
    \end{proof}
    \begin{claim*}
        For every $k \in \N$, $r \in \{1,\ldots,L-1\}$, we have $h_{x,kL+r}^0 = \nn^{k+1}(x) \kroen{r}{0}$.
    \end{claim*}
    \begin{proof}[Proof of the claim]
        For $r = 0$, the preceding claim implies that for every $k \in \N$, we have $h_{x,kL}^0 = \nn^k(x) = \nn^{k+1}(x) \kroen{0}{0}$. Now, suppose $r \in \{1,\ldots,L-1\}$. Then, by the preceding claim,
        \begin{equation}
            h_{x,kL+r}^0 = h_{\nn^{k+1}(x),r}^0 = \nn_0(\nn^{k+1}(x)) \kroen{r}{0} = 0 = \nn^{k+1}(x) \kroen{r}{0}.
        \end{equation}
    \end{proof}
    Now, let $t \in \No$. Note that $t = (t//L) L + (t \bmod L)$, so that by the preceding claim, we have
    \begin{equation}
        \proj_{1:n} h_{x,t} = h_{x,t}^0 = h_{x,(t//L) L + (t \bmod L)}^0 = \nn^{t//L}(x) \kroen{t \bmod L}{0}.
    \end{equation}
    This concludes the proof.
\end{proof}

\begin{lemma}\label{lem:simulation-of-cpwl-function-iterations-with-an-rnn:stabilisation}
    Let $\nn : \R^n \to \R^n$ be a neural network such that $\nn \geq 0$, and let $L := \nnDepth{\nn}$. Then, there exists $\tilde m \in \N$ and $\tilde A_\nn \in \R^{\tilde m \times \tilde m}$ satisfying $ (L+2)n+L \leq \tilde m \leq (2L+2)(\nnWidth{\nn}+1)$ and $\nnWeights{\tilde A_\nn} = \nnWeights{\nn} \cup \{-1,0,1\}$, such that for every $x \in \R^n$, the sequence $(h_t \in \R^{\tilde m})_{t \in \N}$ defined by
    \begin{equation}\label{eq:simulation-of-cpwl-function-iterations-with-an-rnn:stabilisation:recursive-definition}
        h_0 := (0_{n(L+1)},x, 0, \oneHot{L}{1}), \quad h_{t+1} := \ReLU(\tilde A_\nn h_t) \text{ for every } t \in \No
    \end{equation}
    satisfies
    \begin{equation}
        \proj_{1:n} h_t = \nn^{(t-1)//L}(x), \quad t \in \N,
    \end{equation}
    where $//$ denotes the integer division.
\end{lemma}
\begin{proof}
    Let $m \in \N$ and $A_\nn \in \R^{m \times m}$ be as in Lemma \ref{lem:simulation-of-cpwl-function-iterations-with-an-rnn}. We define 
    \[
        \tilde A_\nn := \begin{bmatrix}
            I_n & 0 & -I_n & \begin{array}{cc} I_n & 0 \end{array}\\
            0 & 0 & 0 & \begin{array}{cc} I_n & 0 \end{array}\\
            0 & I_{n(L-1)} & 0 & 0\\
            0 & 0 & 0 & A_\nn
        \end{bmatrix} \in \R^{\tilde m \times \tilde m}.
    \]
    Note that, indeed, we have \[(L+2)n + L \leq (L+1)n + m = \tilde m \leq (L+1)n + (\nnWidth{\nn}+1)L \leq (L+2)(\nnWidth{\nn}+1),\] and that $\nnWeights{\tilde A_\nn} = \nnWeights{A_\nn} \cup \{-1,0,1\} = \nnWeights{\nn} \cup \{-1,0,1\}$. Now, let $x \in \R^n$ and let $(h_t \in \R^m)_{t \in \No}$ be defined as in \eqref{eq:simulation-of-cpwl-function-iterations-with-an-rnn:stabilisation:recursive-definition}. We split the sequence $(h_t)_{t \in \N}$ into subsequences $(h_t^* \in \R^n)_{t \in \No}$, $(h_t^\ell \in \R^n)_{t \in \No}$ for $\ell \in \{0, \ldots, L-1\}$, and $(\tilde h_t \in \R^m)_{t \in \No}$ such that
    \[
        h_t = (h_t^*, h_t^0, \ldots, h_t^{L-1}, \tilde h_t) \text{ for every } t \in \No.
    \] Note that, in particular,
    \begin{equation}\label{eq:simulation-of-cpwl-function-iterations-with-an-rnn:stabilisation:recursive-definition:base-case}
                h_0^* = 0, \quad h_0^0 = 0, \quad h_0^1 = 0, \quad \ldots, \quad h_0^{L-1} = 0, \quad \tilde h_0 = (x, 0, \oneHot{L}{1}),
    \end{equation}
    and that
    \begin{equation}\label{eq:simulation-of-cpwl-function-iterations-with-an-rnn:stabilisation:recursive-definition:iterative-step}
                h_{t+1}^* = \ReLU(h_t^* - h_t^0 + \proj_{1:n}\tilde h_t), \quad h_{t+1}^0 = \ReLU(\proj_{1:n} \tilde h_t), \quad h_{t+1}^\ell = \ReLU(h_t^{\ell-1}), \quad \text{and} \quad \tilde h_{t+1} = \ReLU(A_\nn \tilde h_t),
    \end{equation}
    for every $t \in \No$ and $\ell \in \{1, \ldots, L-1\}$. In particular, by Lemma \ref{lem:simulation-of-cpwl-function-iterations-with-an-rnn:app}, we have 
    \begin{equation}\label{eq:simulation-of-cpwl-function-iterations-with-an-rnn:stabilisation:tilde h t value}
        \proj_{1:n} \tilde h_t = \nn^{t//L}(x) \kroen{t \bmod L}{0} \geq 0 \text{ for every } t \in \N.
    \end{equation}

    We divide the proof into several claims.
    \begin{claim*}
        For every $\ell \in \{0, \ldots, L-1\}$, we have
        \[
            h_t^\ell = \begin{cases}
                \proj_{1:n} \tilde h_{t-\ell-1} & \text{if } t > \ell,\\
                0 & \text{if } t \leq \ell.
            \end{cases}
        \]
    \end{claim*} 
    \begin{proof}
        We first treat the case $\ell = 0$. Note that by \eqref{eq:simulation-of-cpwl-function-iterations-with-an-rnn:stabilisation:recursive-definition:base-case}, we have $h_0^0 = 0$, and that by \eqref{eq:simulation-of-cpwl-function-iterations-with-an-rnn:stabilisation:recursive-definition:iterative-step}, we have 
        \begin{equation}\label{eq:simulation-of-cpwl-function-iterations-with-an-rnn:stabilisation:tilde h t value-2}
            h_t^0 = \ReLU(\proj_{1:n} \tilde h_{t-1}) = \proj_{1:n} \tilde h_{t-1},
        \end{equation}
        for every $t > 0$, where the second equality follows $\proj_{1:n} \tilde h_{t-1} \leq 0$, which is given by \eqref{eq:simulation-of-cpwl-function-iterations-with-an-rnn:stabilisation:tilde h t value}. 
        Now, $\ell \in \{1, \ldots, L-1\}$. First, let $t \in \{0, \ldots, \ell\}$. Then, by \eqref{eq:simulation-of-cpwl-function-iterations-with-an-rnn:stabilisation:recursive-definition:iterative-step} applied $t$ times, we have
        \[
            h_t^\ell = h_{t-t}^{\ell-t} = h_0^{\ell-t} \overset{(a)}= 0,
        \]
        where (a) follows from \eqref{eq:simulation-of-cpwl-function-iterations-with-an-rnn:stabilisation:recursive-definition:base-case}. Now, let $t > \ell$. Then, by \eqref{eq:simulation-of-cpwl-function-iterations-with-an-rnn:stabilisation:recursive-definition:iterative-step} applied $\ell$ times, we have
        \[
            h_t^\ell = h_{t-\ell}^{\ell - \ell} = h_{t-\ell}^0 \overset{(a)}= \proj_{1:n} \tilde h_{t-\ell-1}
        \]
        where (a) follows from \eqref{eq:simulation-of-cpwl-function-iterations-with-an-rnn:stabilisation:tilde h t value-2}. This concludes the proof of the claim.
    \end{proof}
    \begin{claim*}
        \begin{equation}\label{eq:simulation-of-cpwl-function-iterations-with-an-rnn:stabilisation:h t star value}
            h_t^* = \begin{cases}
                \sum_{j=0}^{t-1} \proj_{1:n} \tilde h_j & \text{if } t \leq L,\\
                \sum_{j=t-L}^{t-1} \proj_{1:n} \tilde h_j & \text{if } t > L.
            \end{cases}
        \end{equation}
    \end{claim*}
    \begin{proof}
        We make the proof by induction on $t$. For the base case, note that by \eqref{eq:simulation-of-cpwl-function-iterations-with-an-rnn:stabilisation:recursive-definition:base-case}, we have $h_0^* = 0$, so that \eqref{eq:simulation-of-cpwl-function-iterations-with-an-rnn:stabilisation:h t star value} holds. Now, let $t \in \No$ and assume that \eqref{eq:simulation-of-cpwl-function-iterations-with-an-rnn:stabilisation:h t star value} holds. Note that in particular, we have $h_t^\ast \geq 0$. First, assume that $t \in \{0, \ldots, L-1\}$. Then, by \eqref{eq:simulation-of-cpwl-function-iterations-with-an-rnn:stabilisation:recursive-definition:iterative-step}, we have
        \begin{equation}\label{eq:simulation-of-cpwl-function-iterations-with-an-rnn:stabilisation:h t star value-2}
            h_{t+1}^* = \ReLU(h_t^* - h_t^0 + \proj_{1:n}\tilde h_t) \overset{(a)}= \ReLU(h_t^* + \proj_{1:n}\tilde h_t) \overset{(b)}= h_t^* + \proj_{1:n}\tilde h_t,
        \end{equation}
        where (a) follows from the preceding claim and (b) is by $h_t^* \geq 0$ and $\proj_{1:n}\tilde h_t \geq 0$, which is given by \eqref{eq:simulation-of-cpwl-function-iterations-with-an-rnn:stabilisation:tilde h t value}. Therefore, be the induction hypothesis, we have
        \[
            h_{t+1}^* = h_t^* + \proj_{1:n}\tilde h_t = \sum_{j=0}^{t-1} \proj_{1:n} \tilde h_j + \proj_{1:n}\tilde h_t = \sum_{j=0}^{t} \proj_{1:n} \tilde h_j,
        \]
        so that \eqref{eq:simulation-of-cpwl-function-iterations-with-an-rnn:stabilisation:h t star value} holds for $t+1$. Now, assume that $t \geq L$. Then, by \eqref{eq:simulation-of-cpwl-function-iterations-with-an-rnn:stabilisation:recursive-definition:iterative-step}, we have
        \begin{equation}\label{eq:simulation-of-cpwl-function-iterations-with-an-rnn:stabilisation:h t star value-3}
            h_{t+1}^* = \ReLU(h_t^* - h_t^0 + \proj_{1:n}\tilde h_t) \overset{(a)}= \ReLU(h_t^* - \proj_{1:n} \tilde h_{t-L} + \proj_{1:n}\tilde h_t) \overset{(b)}= h_t^* - \proj_{1:n} \tilde h_{t-L} + \proj_{1:n}\tilde h_t,
        \end{equation}
        where (a) follows from the preceding claim and (b) is by $h_t^* \geq 0$, $\proj_{1:n}\tilde h_t \geq 0$, and $\proj_{1:n} \tilde h_{t-L} \leq 0$, which is given by \eqref{eq:simulation-of-cpwl-function-iterations-with-an-rnn:stabilisation:tilde h t value}. Therefore, by the induction hypothesis, we have
        \[
            h_{t+1}^* = h_t^* - \proj_{1:n} \tilde h_{t-L} + \proj_{1:n}\tilde h_t = \sum_{j=t-L}^{t-1} \proj_{1:n} \tilde h_j - \proj_{1:n} \tilde h_{t-L} + \proj_{1:n}\tilde h_t = \sum_{j=t+1-L}^{t} \proj_{1:n} \tilde h_j,
        \]
        so that \eqref{eq:simulation-of-cpwl-function-iterations-with-an-rnn:stabilisation:h t star value} holds for $t+1$. This concludes the proof of the claim.
    \end{proof}
    Now, let $t \in \N$. Note that the preceding claim and \eqref{eq:simulation-of-cpwl-function-iterations-with-an-rnn:stabilisation:tilde h t value} can be reformulated as
    \[
        h_t^* = \sum_{j=(t-L)\vee 0}^{t-1} \proj_{1:n} \tilde h_j \overset{\eqref{eq:simulation-of-cpwl-function-iterations-with-an-rnn:stabilisation:tilde h t value}}= \sum_{j=(t-L)\vee 0}^{t-1} \nn^{j//L}(x) \kroen{j \bmod L}{0}.
    \]
    Note that since the set $S_t := \{(t-L)\vee 0, \ldots, t-1\}$ has at most $L$ elements, there exists at most one element $j \in S_t$ such that $j \bmod L = 0$. Such an element is given by $j = ((t-1)//L) L$. Therefore, we have
    \[
        h_t^* = \nn^{(((t-1)//L) L)//L}(x) \kroen{(((t-1)//L) L) \bmod L}{0} = \nn^{(t-1)//L}(x).
    \]
    Finally,
    \[
        \proj_{1:n} h_t = h_t^* = \nn^{(t-1)//L}(x).
    \]
    This concludes the proof.
\end{proof}

We now generalize the preceding construction to show that we can simulate the iteration of a neural network that is not necessarily non-negative, by an RNN. We first need the following technical lemma, that asserts that we can apply an affine transform to the input of an RNN.
\begin{lemma}\label{lem:affine-transform-of-the-input-of-an-rnn}
    Let $n \in \N$, $\Rc = (d,m,d';A_h,b_h,A_x,A_o,b_o)$ be an RNN, $A \in \R^{d \times n}$, and $b \in \R^d$. Then, there exists an RNN $\tilde \Rc$ satisfying $\rnnHidDim{\tilde \Rc} = \rnnHidDim{ \Rc} +1 $ and
    \[
        \nnWeights{\tilde \Rc} = \nnWeights{A_h} \cup \nnWeights{-A_x b} \cup \nnWeights{b_h + A_x b} \cup \nnWeights{A_x A} \cup \nnWeights{A_o} \cup \nnWeights{b_o} \cup \{0,1\},
    \]
    such that for every $x \in \R^n$ and $t \in \N$, we have $\tilde \Rc \Dc x[t] = \Rc \Dc (A x + b)[t]$.
\end{lemma}
\begin{proof}
    Let $(d,m,d';A_h,b_h,A_x,A_o,b_o) := \Rc$. We define $\tilde \Rc := (n, m + 1, d'; \tilde A_h, \tilde b_h, \tilde A_x, \tilde A_o, b_o)$ by
    \[
        \tilde A_h := \begin{bmatrix}
            A_h & -A_x b\\
            0 & 0
        \end{bmatrix} \in \R^{m+1 \times m+1}, \quad \tilde b_h := \begin{bmatrix}
            b_h + A_x b\\
            1
        \end{bmatrix}, \quad \tilde A_x := \begin{bmatrix}
            A_x A\\
            0
        \end{bmatrix}, \quad \text{and} \quad \tilde A_o := (A_o, 0).
    \]
    Note that $\nnWeights{\tilde \Rc} = \nnWeights{\Rc} \cup \{0,1\}$ and that $\rnnHidDim{\tilde \Rc} = m + 1 = \rnnHidDim{\Rc} + 1$. Now, let $x \in \R^n$ and define $h_t := \tilde \Hc \Dc x[t]_{1:m}$ and $d_t := \tilde \Hc \Dc x[t]_{m+1}$ for every $t \in \No \cup {-1}$. In particular, note that $h_{-1} = 0$, $d_{-1} = 0$. Then, note that
    \begin{equation}
        \begin{cases}
        h_{t} &= \ReLU(\tilde A_h (h_{t-1},d_{t-1}) + \tilde b_h + \tilde A_x \Dc x[t]) = \ReLU(A_h h_{t-1} - A_x b d_{t-1} + b_h + A_x b + A_x A \Dc x[t])\\
        d_{t} &= \ReLU(0 \cdot d_{t-1} + 1) = 1
    \end{cases}
    \end{equation}
    for every $t \in \No$. Therefore, for every $t \in \No$, we have
    \begin{align}
        h_t &= \begin{cases}
        \ReLU(A_h h_{t-1} + b_h + A_x b + A_x A \Dc x[t]), & \text{if} \ t = 0\\
        \ReLU(A_h h_{t-1} + b_h + A_x A \Dc x[t]), & \text{if} \ t > 0\\
    \end{cases}\\
    & = \ReLU(A_h h_{t-1} + b_h + A_x \Dc (A x + b)[t]) = \Hc \Dc (A x + b)[t],
    \end{align}
    where (a) follows from the fact that $d_{t-1} = 1$ for every $t \in \N$. Finally, for every $t \in \N$, we have
    \[
        \tilde \Rc \Dc x[t] = \tilde A_o (h_t,d_t) + b_o = A_o h_t + b_o = A_o\Hc \Dc (A x + b)[t] + b_o = \Rc \Dc (A x + b)[t].
    \] 
    This concludes the proof.
\end{proof}

\begin{lemma}\label{lem:simulation-of-cpwl-function-iterations-with-an-rnn:RNN-construction}
    Let $\nn : \R^n \to \R^n$ be a neural network and $L := \nnDepth{\nn}$. Then, there exists an RNN $\Rc$ satisfying $\rnnHidDim{\Rc} \leq (2L+6)(2\nnWidth{\nn}+1) + 1$ and $\nnWeights{\Rc} = \pm\nnWeights{\nn} \cup \nnWeights{0,1}$, 
    such that for every $x \in \R^n$,
    \begin{equation}
        \Rc \Dc x[t] = \nn^{(t-1)//(L+1)}(x), \quad t \in \N.
    \end{equation}
\end{lemma}
\begin{proof}
    Let $\nn = (N_0, \ldots, N_{L}; A^1, \ldots, b^L)$ be a neural network with $N_0 = N_l = n$. We define the ReLU neural network $\nn^\pm := (2n, N_1, \ldots, N_{L-1}, 2n, 2n; A^{1,\pm}, \ldots, b^{L+1,\pm})$ by 
    \[
        A^{1,\pm} := \begin{bmatrix}
            A^1& -A^1
        \end{bmatrix} \in \R^{N_1 \times 2n}, A^{L,\pm} := \begin{bmatrix}
            A^L\\
            -A^L
        \end{bmatrix} \in \R^{2n \times N_{L-1}}, A^{L+1,\pm} := \begin{bmatrix}
            I_n\\
            I_n
        \end{bmatrix} \in \R^{2n \times 2n}, \quad \text{and} \quad A^{\ell,\pm} := A_\ell, 
    \]
    for $\ell \in \{2, \ldots, L-1\}$, and
    \[
        b^{L,\pm} := \begin{bmatrix}
            b^L\\
            -b^L
        \end{bmatrix} \in \R^{2n}, \quad b^{L+1,\pm} := 0_{2n} \in \R^{2n} \quad \text{and} \quad b^{\ell,\pm} := b_\ell, 
    \]
    for $\ell \in \{1, \ldots, L-1\}$. First note that $\nnWeights{\nn^\pm} \subseteq \pm \nnWeights{\nn} \cup \nnWeights{0,1}$, $\nnDepth{\nn^\pm} = \nnDepth{\nn} + 1$ and $\nnWidth{\nn^\pm} \leq 2 \nnWidth{\nn}$. Moreover, note that for every $x \in \R^n$, we have
    \begin{equation}
        \nn^\pm(x,0) = \nn^\pm(x^+,x^-) = (\nn^+(x), \nn^-(x)),
    \end{equation}
    and 
    \begin{equation}
        \nn^\pm(x,0) = (\nn^+(x), \nn^-(x)).
    \end{equation}
    By induction, we can show that for every $k \in \N$, we have
    \begin{equation}        \nn_\pm^k(x^+,x^-) = (\nn^k(x)^+,\nn^k(x)^-),
    \end{equation}
    where $\nn^k$ denotes the $k$-th iterate of $\nn$. Therefore, for every $k \in \N$, we have
    \begin{equation}        \nn_\pm^k(x,0) = \nn_\pm^{k-1}(\nn_\pm(x,0)) = \nn_\pm^{k-1}(\nn_\pm(x^+,x^-)) =\nn_\pm^{k}(x^+,x^-) =  (\nn^k(x)^+,\nn^k(x)^-).
    \end{equation}
    Moreover, $\nn_\pm \geq 0$. Therefore, by Lemma \ref{lem:simulation-of-cpwl-function-iterations-with-an-rnn:stabilisation}, $m \in \N$ and $\tilde A_{\nn_\pm} \in \R^{m \times m}$ satisfying $ 2(L+1+2)n+L+1 \leq \tilde m \leq (2(L+2)+2)(2\nnWidth{\nn}+1)$ and $\nnWeights{\tilde A_{\nn^\pm}} = \pm\nnWeights{\nn} \cup \{-1,0,1\}$, such that for every $x \in \R^n$, the sequence $(h_t \in \R^m)_{t \in \N}$ defined by
    \begin{equation}
        h_0 := (0_{2n(L+2)},y, 0, \oneHot{L+1}{1}), \quad h_{t+1} := \ReLU(\tilde A_{\nn_\pm} h_t) \text{ for every } t \in \No
    \end{equation}
    where $y := (x,0) \in \R^{2n}$, satisfies
    \begin{equation}        \proj_{1:2n} h_t = \nn_\pm^{(t-1)//(L+1)}(y) = \nn_\pm^{(t-1)//(L+1)}(x,0) = (\nn^{(t-1)//(L+1)}(x)^+,\nn^{(t-1)//L}(x)^-), \quad t \in \N.
    \end{equation}
    We then define the RNN $\Rc := (2n, m, n; \tilde A_{\nn_\pm}, 0, I_m, A_o, 0)$, where $A_o := \begin{bmatrix}
        I_n & -I_n & 0
    \end{bmatrix} \in \R^{n \times m}$. Note that for every $x \in \R^n$ and $t \in \N$, we have
    \begin{equation}
        \Rc \Dc h_0[t] = A_o h_t = \proj_{1:n} h_t - \proj_{n+1:2n} h_t = \nn^{(t-1)//(L+1)}_+(x) -\nn^{(t-1)//(L+1)}_-(x) = \nn^{(t-1)//(L+1)}(x).
    \end{equation}
    Finally, note that $h_0 = (0_{2n(L+2)},x, 0, \oneHot{L+1}{1})$ is an affine transform of $x \in \R^n$, specifically, we have $h_0 = A x + b$, where 
    \[
        A := \begin{bmatrix}
            0_{2n(L+2) \times n}\\
            I_n\\
            0
        \end{bmatrix} \in \R^{m \times n}, \quad b := \begin{pmatrix}
            0\\
            \oneHot{L+1}{1}
        \end{pmatrix} \in \R^m.
    \]
    Hence, by Lemma \ref{lem:affine-transform-of-the-input-of-an-rnn}, there exists an RNN $\tilde \Rc$ satisfying $\rnnHidDim{\tilde \Rc} = \rnnHidDim{ \Rc} +1 $ and 
    \[
        \nnWeights{\tilde \Rc} = \nnWeights{\tilde A_{\nn_\pm}} \cup \nnWeights{-I_m b} \cup \nnWeights{0 + I_m b} \cup \nnWeights{I_m A} \cup \nnWeights{A_o} \cup \nnWeights{0} \cup \{0,1\} = \pm \nnWeights{\nn} \cup \{-1,0,1\},
    \]
    such that for every $x \in \R^n$ and $t \in \N$, we have $\tilde \Rc \Dc x[t] = \Rc \Dc h_0[t]$. We then define $\Rc_\nn := \tilde \Rc$. Note that for every $x \in \R^n$ and $t \in \N$, we have
    \begin{equation}
        \Rc_\nn \Dc x[t] = \Rc \Dc h_0[t] = \nn^{(t-1)//(L+1)}(x).
    \end{equation} This concludes the proof.
\end{proof}

\subsection{Application: an RNN that simulates a \TMNU}

\begin{lemma}\label{lem:simulation-of-a-tmnu-by-an-rnn:app}
    Let $\tmnuM$ be a \TMNU and $u \in \{0,1\}^\N$ such that $\tmnuM$ is uniformly bounded by $C > 0$ at $u$. Then, there exists an RNN $\Rc$ satisfying 
    \[
        \rnnHidDim{\Rc} \leq 42(2\max\{5\nStates, \nStates + 2\nFunctions{\tmnuM}\neurDim\} + 4\nFunctions{\tmnuM}\neurDim + 75) + 2,
    \]
    and 
    \[\nnWeights{\Rc} = \{0,1/4,1,2,3,4,C\} \cup \pm \nnWeights{\tmnuM} \cup \pm\{\cantorMap(u)\},\]
    such that
    \begin{equation}
        \Rc \Dc x[t+1] = \tmnuM^ux[t//19], \quad t \in \No, \quad x \in [-1,1].
    \end{equation}
\end{lemma}
\begin{proof}
    By Lemma \ref{lem:update-function-as-neural-network} and Theorem \ref{thm:simulation-of-tmnu-by-cpwl-function}, there exists a neural network $\nn : \R^{\nStates+2+\neurDim} \to \R^{\nStates+2+\neurDim}$ satisfying $\nnDepth{\nn} = 18$, $\nnWidth{\nn} \leq \max\{5\nStates, \nStates + 2\nFunctions{\tmnuM}\neurDim\} + 2\nFunctions{\tmnuM}\neurDim + 37$, $\nnWeights{\nn} \subseteq \pm \{0,1/4,1,2,3,4,C\} \cup \pm \nnWeights{\tmnuM}$,
    such that for every $c \in \Bc_\tmnuM(C)$,
    \begin{equation}
        \nn_{\tmnuM,C}(\configMap{\tmnuM}(c)) = \configMap{\tmnuM}(\tmnuM(c)).
    \end{equation}
    Then, by Lemma, there exists $\Rc = (\nStates+2+\neurDim, m, \nStates+2+\neurDim; A_h, b_h, A_x, A_o,b_o)$ satisfying 
    \begin{align*}
        \rnnHidDim{\Rc} &\leq (2\nnDepth{\nn}+6)(2\nnWidth{\nn}+1) + 1\\
        & \leq (2\cdot 18+6)(2(\max\{5\nStates, \nStates + 2\nFunctions{\tmnuM}\neurDim\} + 2\nFunctions{\tmnuM}\neurDim + 37)+1) + 1\\
        & = 42(2\max\{5\nStates, \nStates + 2\nFunctions{\tmnuM}\neurDim\} + 4\nFunctions{\tmnuM}\neurDim + 75) + 1
    \end{align*}
    and $\nnWeights{\Rc} = \pm\nnWeights{\nn} \cup \{0,1\} \subseteq \pm \{0,1/4,1,2,3,4,C\} \cup \pm \nnWeights{\tmnuM}$, such that for every $x \in \R^n$,
    \begin{equation}
        \Rc \Dc x[t] = \nn^{(t-1)//(18+1)}(x) = \nn^{(t-1)//19}(x), \quad t \in \N.
    \end{equation}
    In particular, for every $c \in \Bc_\tmnuM(C)$ and $t \in \N$, we have
    \begin{equation}
        \Rc \Dc \configMap{\tmnuM}(c)[t] = \nn^{(t-1)//19}(\configMap{\tmnuM}(c)) = \configMap{\tmnuM}(\tmnuM^{(t-1)//19}(c)).
    \end{equation}
    Now let $ u \in \{0,1\}^\N$ and assume that $\tmnuM$ is uniformly bounded by $C > 0$ at $u$. Then, for every $x \in [-1,1]$, $c_x := (1;|u;x,0,\ldots,0) \in \Bc_\tmnuM(C)$, so that for every $t \in \N$, we have
    \begin{equation}        \Rc \Dc \configMap{\tmnuM}(c_x)[t] = \configMap{\tmnuM}(\tmnuM^{(t-1)//19}(c_x)).
    \end{equation}
    Consider the RNN $\Rc' := (\nStates+2+\neurDim, m, 1; A_h, b_h, A_x, A_o' := \proj_{\nStates + 2 + \neurDim} A_o, b_o' := \proj_{\nStates + 2 + \neurDim}b_o)$. Then, $\rnnHidDim{\Rc'} = \rnnHidDim{\Rc}$ and $\nnWeights{\Rc'} \subseteq \nnWeights{\Rc}$. Moreover, for every $x \in [-1,1]$ and $t \in \N$, we have
    \begin{equation}        
        \Rc' \Dc \configMap{\tmnuM}(c_x)[t] = \proj_{\nStates + 2 + \neurDim} \Rc \Dc \configMap{\tmnuM}(c_x)[t] = \proj_{\nStates + 2 + \neurDim}  \configMap{\tmnuM}(\tmnuM^{(t-1)//19}(c_x)) = \tmnuM^u x[(t-1)//19].
    \end{equation}
    Finally, note that for every $x \in [-1,1]$, 
    \[
        \configMap{\tmnuM}(c_x) = (\oneHot{\nStates}{1},\cantorMap(u),x,0,\ldots,0) = A x + b,
    \]
    where $A = (0_{\nStates + 2}, 1, 0_{\neurDim-1})$ and $b = (\oneHot{\nStates}{1},\cantorMap(u),0,\ldots,0)$. Hence, by Lemma \ref{lem:affine-transform-of-the-input-of-an-rnn}, there exists an RNN $\tilde \Rc$ satisfying $\rnnHidDim{\tilde \Rc} = \rnnHidDim{ \Rc'} +1 $ and 
    \begin{align*}
        \nnWeights{\tilde \Rc} &= \nnWeights{A_h} \cup \nnWeights{- A_x b} \cup \nnWeights{b_h + A_x b} \cup \nnWeights{A_x A} \cup \nnWeights{A_o'} \cup \nnWeights{b_o'} \cup \{0,1\}\\
        &\overset{(a)} \subseteq \nnWeights{A_h} \cup \nnWeights{-b} \cup \nnWeights{b} \cup \nnWeights{A} \cup \nnWeights{A_o} \cup \nnWeights{b_o} \\
        & \pm \{0,1/4,1,2,3,4,C\} \cup \pm \nnWeights{\tmnuM} \cup \pm\{\cantorMap(u)\}
    \end{align*}
    where (a) follows from the form of $A_x$ and $b_h$ by inspection of the proof of Lemma \ref{lem:simulation-of-cpwl-function-iterations-with-an-rnn:RNN-construction},
    such that for every $x \in [-1,1]$ and $t \in \N$, we have $\tilde \Rc \Dc (Ax+b)[t] = \Rc' \Dc x[t]$. We then define $\Rc_\tmnuM := \tilde\Rc$. Note that for every $x \in [-1,1]$ and $t \in \N$, we have
    \begin{equation}
        \Rc_\tmnuM \Dc x[t] = \Rc' \Dc (Ax+b)[t] = \Rc' \Dc \configMap{\tmnuM}(c_x)[t] = \tmnuM^u x[(t-1)//19].
    \end{equation}
    This concludes the proof.
\end{proof}

We now establish $\nFunctions{\Continuous}$ and $\nnWeights{\Continuous}$, where $\Continuous$ is the \TMNU defined in Definition \ref{def:continuous-machine}. For that, we establish these quantities for all the \TMNUs that lead to its definition.
\begin{enumerate}
    \item By inspection of Definition \ref{def:sign-tmnu}, we have $\nFunctions{\Sign} = 2$, and $\nnWeights{\Sign} = \{-1,1\}$.
    \item By inspection of Definition \ref{def:scale-machine}, we have $\nFunctions{\Scale} = 2$, and $\nnWeights{\Scale} = \{1,2\}$.
    \item By inspection of Definition \ref{def:contr-+-tmnu}, we have $\nFunctions{\Contr^+} = 4$, and $\nnWeights{\Contr^+} = \{0,1/2,1\}$.
    \item Let $\Rc$ be an RNN. By inspection of Definition \ref{lem:tmnu-simulation-rnn}, we have $\nFunctions{\tmnuM_\Rc} = 5$, and $\nnWeights{\tmnuM_\Rc} = \{0,1\} \cup \nnWeights{\Rc}$. In particular, by inspection of PAPER CLEMENS, we have $\nFunctions{\Times} = 5$, and $\nnWeights{\Times} = \{-1,0,1/4,1/2,1\}$.
    \item By inspection of Definition \ref{def:contr-tmnu}, we have $\nFunctions{\Contr} = 5$, and $\nnWeights{\Contr} = \{-1,0,1/2,1\}$.
    \item By inspection of Definition \ref{def:upoly}, we have $\nFunctions{\UpPoly} = 11$, and $\nnWeights{\UpPoly} = \{-1,0,1/4,1/2,1\}$. 
    \item By inspection of Definition \ref{def:poly-machine}, we have $\nFunctions{\Poly} = 13$, and $\nnWeights{\Poly} = \{-1,0,1/4,1/2,1,2\}$.
    \item By inspection of Definition \ref{def:continuous-machine}, we have $\nFunctions{\Continuous} = 14$, and $\nnWeights{\Continuous} = \{-1,0,1/4,1/2,1,2\}$.
\end{enumerate}

Let $(\nStates,\neurDim,\transFunc,\commFunc) := \Continuous$ and note that by Definition \ref{def:continuous-machine}, we have $\nStates = 12$ and $\neurDim = 17$.

\begin{corollary}\label{cor:simulation-of-a-tmnu-by-an-rnn:app}
    Let $u \in \{0,1\}^\N$ such that $\Continuous$ is uniformly bounded by $C > 0$ at $u$. Then, there exists an RNN $\Rc$ satisfying 
    \[
        \rnnHidDim{\Rc} \leq 84128, \quad \text{and} \quad \nnMag{\Rc} = 4 \vee C\]
    such that
    \begin{equation}
        \Rc \Dc x[t+1] = \Continuous^ux[t//19], \quad t \in \No, \quad x \in [-1,1].
    \end{equation}
\end{corollary}
\begin{proof}
    By Lemma \ref{lem:simulation-of-a-tmnu-by-an-rnn:app}, there exists an RNN $\Rc$ satisfying 
    \[
        \rnnHidDim{\Rc} \leq 42(2\max\{5\nStates, \nStates + 2\nFunctions{\Continuous}\neurDim\} + 4\nFunctions{\Continuous}\neurDim + 75) + 2 = 84128,
    \]
    and 
    \[\nnMag{\Rc} = \max\left\{|\{0,1/4,1,2,3,4,C\}| \cup |\nnWeights{\Continuous}| \cup \{|\cantorMap(u)|\}\right\} = 4\vee C,\]
    such that
    \begin{equation}
        \Rc \Dc x[t+1] = \tmnuM^ux[t//19], \quad t \in \No, \quad x \in [-1,1].
    \end{equation}
\end{proof}

\begin{corollary}\label{cor:continuous-function-approximation-by-rnns}
    Let $f : [-1,1] \to \R$ be a continuous function. Then, there exists an RNN $\Rc$ satisfying 
    \[
        \rnnHidDim{\Rc} \leq 84128, \quad \text{and} \quad\nnMag{\Rc} \leq 36 \left(1+ \|f\|_{L^\infty([-1,1])}\right)\]
    such that
    \begin{equation}
        \limi{t} \sup_{x \in [-1,1]} |\Rc \Dc x[t] - f(x)| = 0.
    \end{equation}
\end{corollary}
\begin{proof}
    Let $f : [-1,1] \to \R$ be a continuous function. By Theorem \ref{thm:continuous-approximation-tmnu}, there exists $u \in \{0,1\}^\N$ such that
    \[
        \lim_{t \to \infty} \|\Continuous^u(x)[t] - f(x)\|_{L^{\infty}([-1,1])} = 0,
    \]
    and 
    \[
        \satComp{\Continuous}(c_x) \leq 1 + \|f\|_{L^\infty([-1,1])},
    \]
    where $c_x := (1;|u;x,0_{16}) \in \Cc_\Continuous$ for every $x \in [-1,1]$. By Lemma \ref{lem:magnitude-configuration},
    \begin{equation}
        \nnMag{\Continuous} \leq (17+1) (\nnMag{\Continuous})\satComp{\Continuous}(c_x) \leq  36(1 + \|f\|_{L^\infty([-1,1])}).
    \end{equation}
    Therefore, $\Continuous$ is uniformly bounded by $36 (1 + \|f\|_{L^\infty([-1,1])})$ at $u$. The result then follows from Corollary \ref{cor:simulation-of-a-tmnu-by-an-rnn:app}.
\end{proof}

Note that Corollary \ref{cor:continuous-function-approximation-by-rnns} is exactly Theorem \ref{thm:main-theorem-continuous}. Theorems \ref{thm:main-theorem-poly} and \ref{thm:main-theorem-tchebychev} can be proven with exactly the same proof technique.

\subsection{Notation and definitions}

\label{sec:notation-appendix}

Let $n,m \in \N$, $A = (A_{jk}) \in \R^{m \times n}$ and $b = (b_j) \in \R^m$. We define
\begin{equation}
    \nnWeights{A} := \{ A_{jk} : j \in \{1, \ldots, m\}, k \in \{1, \ldots, n\} \}, \quad \nnWeights{b} := \{ b_j : j \in \{1, \ldots, m\} \}.
\end{equation}
Given an affine map $W : \R^n \to \R^m$ defined by $W(x) = Ax + b$ for all $x \in \R^n$, we define $\nnWeights{W} := \nnWeights{A} \cup \nnWeights{b}$. Note that this notion is well-defined since $A$ and $b$ are uniquely defined by $W$. We also define $\nnWeights{\ReLU} := \{0,1\}$, where $\ReLU : \R \to \R$ is the ReLU activation function. Given a \TMNU $\tmnuM := (\nStates, \neurDim, \transFunc, \commFunc)$, we define
\begin{equation}
    \nnWeights{\tmnuM} := \bigcup_{\state \in \states, \symb \in \workSymbols} \nnWeights{\commFunc(\state,\symb)}.
\end{equation}
Given $A \subseteq \R$, we let $\pm A := A \cup \{-x : x \in A\}$. 
For $n \in \N$, we define the lexicographical ordering on $\R^m$ by letting $x <_L y$ if there exists $i \in \{1, \dots, m\}$ such that $x_i <_L y_i$ and $x_j = y_j$ for all $j \in \{1, \dots, i-1\}$. For $n,m \in \N$, we define the lexicographical ordering on $\R^{n \times m}$ by letting $A <_L B$ if there exists $i \in \{1, \dots, n\}$ such that $A_i <_L B_i$ and $A_j = B_j$ for all $j \in \{1, \dots, i-1\}$, where $A_i$ and $B_i$ denote the $i$-th rows of $A$ and $B$, respectively. Given two affine maps $f,g : \R^n \to \R^m$, given by $f(x) = A x + b$ and $g(x) = A' x + b'$, we write $f <_L g$ if $A <_L A'$ or $A = A'$ and $b <_L b'$. Given two sets $A,B \subseteq \{1,\ldots,n\}$, we write $A <_L B$ if $\#A < \#B$ or if $k := \#A = \#B$ and there exists $i \in \{1, \ldots, k\}$ such that $A_i <_L B_i$ and $A_j = B_j$ for all $j \in \{1, \ldots, i-1\}$, where $A_i$ and $B_i$ denote the $i$-th smallest elements of $A$ and $B$, respectively. For any two $\ReLU_A, \ReLU_B \in \reluSet{n}$, we write $\ReLU_A <_L \ReLU_B$ if $A <_L B$. 
Finally, for two functions $f,g \in \setAffine{n}{n} \cup \reluSet{n}$, we write $f <_L g$ if $f,g \in \setAffine{n}{n}$ and $f <_L g$, or if $f,g \in \reluSet{n}$ and $f <_L g$, or if $f \in \setAffine{n}{n}$ and $g \in \reluSet{n}$.

\subsection{Construction of a neural network that simulates a \TMNU}

In Section \ref{sec:tmnu-rnn-simulation}, we have established that given a \TMNU $\tmnuM$, there exists a continuous piecewise-linear function $F_\tmnuM$ such that 
\[
    F_\tmnuM(\configMap{\tmnuM}(c)) = \configMap{\tmnuM}(\tmnuM(c)), \quad \config \in \Cc_\tmnuM.
\]
Here, we construct explicitly a neural network that realizes this function $F_\tmnuM$, where a neural network is defined as follows.

\begin{definition}(Neural network)\label{def:neural network}
    We call neural network an ordered sequence
    \begin{equation}\label{eq:definition-of-neural-network}
        \nn := (N_0, N_1, \ldots, N_L; A^1, b^1, A^2, b^2, \ldots, A^L, b^L),
    \end{equation}
    where $L \in \N$ and $N_0, N_1, \ldots, N_L \in \N$, $A^\ell = (A^\ell_{jk}) \in \R^{N_\ell \times N_{\ell-1}}$ and $b^\ell = (b^\ell_j) \in \R^{N_\ell}$ for $\ell \in \{1, \ldots, L\}$. We consider the neural network $\nn$ as a function $\nn : \R^{N_0} \to \R^{N_L}$ defined by
    \begin{equation}\label{eq:function-represented-by-a-neural-network}
        \nn(\realx) := A^L \ReLU\left( A^{L-1} \ReLU\left( \ldots \ReLU\left( A^1 \realx + b^1 \right) + b^{L-1} \right) + b^L \right) \text{ for all } \realx \in \R^{N_0}.
    \end{equation}
    For $\ell \in \{1, \ldots, L\}$, we define the $\ell$-th ancestor subnetwork of $\nn$ as the neural network 
    \begin{equation}\label{eq:definition-of-ancestor-subnetwork}
        \nn_\ell := (N_0, N_1, \ldots, N_\ell; A^1, b^1, A^2, b^2, \ldots, A^\ell, b^\ell),
    \end{equation}
    and by convention we let $\nn_0(x) = x$ for every $x \in \R^{N_0}$.
    We let 
    \[
        \nnWidth{\nn} := \max_{\ell \in \{0, \ldots, L\}} N_\ell, \quad \nnDepth{\nn} := L, \quad \nnWeights{\nn} := \bigcup_{\ell=1}^L \nnWeights{A^\ell} \cup \nnWeights{b^\ell}, \nnMag{\nn} := \max \{ |x| : x \in \nnWeights{\nn}\},
    \]
    and also
    \[
        \nnInDim{\nn} := N_0, \quad \nnOutDim{\nn} := N_L.
    \]
\end{definition}
In the following, we construct explicitly a neural network that realizes the function $F_\tmnuM$ defined in Section \ref{sec:tmnu-rnn-simulation}.

\newcommand\nnId{\bfI}
\begin{lemma}\label{lemma:identity-neural-network}
    Let $n,d\in\N$. Then, there exists a neural network $\nnId_{n,d}$ with $\nnWidth{\nnId_{n,d}} \leq 2n$, $\nnDepth{\nnId_n} = d$ and $\nnWeights{\nnId_n} \subseteq \{-1,0,1\}$ such that $\nnId_n(x) = x$ for all $x \in \R^n$.
\end{lemma}
\begin{proof}
    We let $\nnId_{n,1} := (n,n; I_n,0)$, where $I_n$ is the $n \times n$ identity matrix. Note that indeed, $\nnDepth{\nnId_{n,1}} = 1$, $\nnWidth{\nnId_{n,1}} = n$ and $\nnWeights{\nnId_{n,1}} = \{0,1\}$. Moreover, for every $x \in \R^n$, we have $\nnId_{n,1}(x) = I_n x = x$. Now, for $d \in \N$ such that $d > 1$, we define 
    \[\nnId_{n,d} := (n,\underbrace{2n, \ldots, 2n}_{d-1 \ \text{times}}, n; \begin{pmatrix}
        I_n\\
        -I_n
    \end{pmatrix}, 0, \underbrace{I_{2n},0, \ldots, I_{2n},0}_{d-2 \ \text{times}}, \begin{pmatrix}
        I_n&
        -I_n
    \end{pmatrix}, 0).\] 
    Note that indeed, $\nnDepth{\nnId_{n,d}} = d$, $\nnWidth{\nnId_{n,d}} = 2n$ and $\nnWeights{\nnId_{n,d}} = \{-1,0,1\}$. Now, let $x \in \R^n$, and note that by definition of $\nnId_n$, we have
    \[
        \nnId_{n,d}(x) = \underbrace{\ReLU \circ \ldots \circ \ReLU}_{d - 1 \ \text{times}}(x) - \underbrace{\ReLU \circ \ldots \circ \ReLU}_{d - 1 \ \text{times}}(-x)= x.
    \]
    This concludes the proof.
\end{proof}

The next lemma asserts that neural networks are closed under composition.
\begin{lemma}\label{lemma:composition-of-neural-networks}
    Let $\nn_1,\nn_2$ be two neural networks such that $\nnOutDim{\nn_1} = \nnInDim{\nn_2}$. Then, there exists a neural network $\nn$ such that $\nnDepth{\nn} = \nnDepth{\nn_1} + \nnDepth{\nn_2}$, $\nnWidth{\nn} \leq \max\{2\nnOutDim{\nn_1}, \nnWidth{\nn_1}, \nnWidth{\nn_2}\}$, $\nnWeights{\nn} \subseteq \pm\nnWeights{\nn_1} \cup \pm \nnWeights{\nn_2}$ and $\nn(x) = \nn_2(\nn_1(x))$ for all $x \in \R^{\nnInDim{\nn_1}}$.
\end{lemma}
\begin{proof}
    Suppose that \[\nn_1 = (N_0^{(1)}, N_1^{(1)}, \ldots, N_{L_1}^{(1)}; A^1_{(1)}, b^1_{(1)}, A^2_{(1)}, b^2_{(1)}, \ldots, A^{L_1}_{(1)}, b^{L_1}_{(1)})\] and \[\nn_2 = (N_0^{(2)}, N_1^{(2)}, \ldots, N_{L_2}^{(2)}; A^1_{(2)}, b^1_{(2)}, A^2_{(2)}, b^2_{(2)}, \ldots, A^{L_2}_{(2)}, b^{L_2}_{(2)}).\] Note that by assumption, we have $N_{L_1}^{(1)} = N_0^{(2)}$. We define \[\nn := (N_0, N_1, \ldots, N_{L_1+L_2}; A^1, b^1, \ldots, A^{L_1+L_1}, b^{L_1+L_2}),\] where
    \begin{enumerate}
        \item $N_\ell = N_\ell^{(1)}$ for $\ell \in \{0, \ldots, L_1-1\}$, $N_{L_1} = 2 N_{L_1}^{(1)}$ and $N_{L_1+\ell} = N_{\ell}^{(2)}$ for $\ell \in \{1, \ldots, L_2\}$,
        \item $A^\ell = A^\ell_{(1)}$ and $b^\ell = b^\ell_{(1)}$ for $\ell \in \{1, \ldots, L_1-1\}$, $A^{L_1} = \begin{pmatrix}
            A^{L_1}_{(1)}\\
            -A^{L_1}_{(1)}
        \end{pmatrix}$ and $b^{L_1} = \begin{pmatrix}
            b^{L_1}_{(1)}\\
            -b^{L_1}_{(1)}
        \end{pmatrix}$, $A^{L_1+1} = \begin{pmatrix}
            A^1_{(2)} & -A^1_{(2)}
        \end{pmatrix}$, $b^{L_1+1} = b^1_{(2)}$ and $A^{L_1+\ell} = A^{\ell}_{(2)}$ and $b^{L_1+\ell} = b^{\ell}_{(2)}$ for $\ell \in \{2, \ldots, L_2\}$.
    \end{enumerate}
    Note that indeed, $\nnDepth{\nn} = \nnDepth{\nn_1} + \nnDepth{\nn_2}$, $\nnWidth{\nn} \leq 2\max\{\nnWidth{\nn_1}, \nnWidth{\nn_2}\}$ and $\nnWeights{\nn} \subseteq \pm\nnWeights{\nn_1} \cup \pm \nnWeights{\nn_2}$. Now, let $x \in \R^{\nnInDim{\nn_1}}$, and note that by definition of $\nn$, we have
    \[
        \nn_{L_1}(x) = \begin{pmatrix}
            \nn_1(x)\\
            -\nn_1(x)
        \end{pmatrix},
    \]
    and therefore
    \[
    \nn_{L_1+1}(x) = \begin{pmatrix}
        A^1_{(2)} & -A^1_{(2)}
    \end{pmatrix} \ReLU\begin{pmatrix}
        \nn_1(x)\\
        -\nn_1(x)
    \end{pmatrix} + b^1_{(2)} = A^1_{(2)} \nn_1(x) + b^1_{(2)}.
    \]
    It follows that
    \[
        \nn(x) = \nn_2(\nn_1(x)).
    \] This concludes the proof.
\end{proof}

\begin{lemma}\label{lemma:parallelization-of-neural-networks}
    Let $m,d \in \N$, $\nn_1,\nn_2, \ldots, \nn_m$ be $m$ neural networks such that $\nnDepth{\nn_1} = \ldots = \nnDepth{\nn_m}$ and let $\iota_i : \{1, \ldots, \nnInDim{\nn_i}\} \to \{1,\ldots,d\}$, $i \in \{1,\ldots,m\}$. Then, there exists a neural network $\nn$ such that $\nnDepth{\nn} = \nnDepth{\nn_1}$, $\nnWidth{\nn} = d \vee \sum_{i=1}^m \nnWidth{\nn_i}$, $\nnWeights{\nn} \subseteq \bigcup_{i=1}^m \nnWeights{\nn_i} \cup \{0\}$ and \[\nn(x) = (\nn_1(\proj_{\iota_1}x),\nn_2(\proj_{\iota_2}x), \ldots, \nn_m(\proj_{\iota_m}x))\] for all $x \in \R^d$.
\end{lemma}
\begin{proof}
    Suppose that \[\nn_i = (N_0^{(i)}, N_1^{(i)}, \ldots, N_{L}^{(i)}; A^1_{(i)}, b^1_{(i)}, A^2_{(i)}, b^2_{(i)}, \ldots, A^{L}_{(i)}, b^{L}_{(i)}),\] for every $i \in \{1,\ldots, m\}$.
    We define \[\nn := (d, N_1, \ldots, N_L; A^1, b^1, \ldots, A^L, b^L),\] where $N_\ell = \sum_{i=1}^m N_\ell^{(i)}$ for $\ell \in \{1, \ldots, L\}$ and $A^\ell = \begin{pmatrix}
    A^\ell_{(1)} & 0 & \ldots & 0\\
    0 & A^\ell_{(2)} & \ldots & 0\\
    \vdots & \vdots & \ddots & \vdots\\
    0 & 0 & \ldots & A^\ell_{(m)}
\end{pmatrix}$ and $b^\ell = \begin{pmatrix}
    b^\ell_{(1)}\\
    b^\ell_{(2)}\\
    \vdots\\
    b^\ell_{(m)}
\end{pmatrix}$ for $\ell \in \{2, \ldots, L\}$, and 
\[
    A^1 = \begin{pmatrix}
        A^1_{(1)} \proj_{\iota_1}\\
        A^1_{(2)} \proj_{\iota_2}\\
        \vdots\\
        A^1_{(m)} \proj_{\iota_m}
    \end{pmatrix}, \quad b^1 = \begin{pmatrix}
        b^1_{(1)}\\
        b^1_{(2)}\\
        \vdots\\
        b^1_{(m)}
    \end{pmatrix}.
\]
    Note that indeed, $\nnDepth{\nn} = L$, $\nnWidth{\nn} = d \vee \sum_{i=1}^m \nnWidth{\nn_i}$ and $\nnWeights{\nn} \subseteq \bigcup_{i=1}^m \nnWeights{\nn_i} \cup \{0\}$. Now, let $x \in \R^d$, and note that by definition of $\nn$, we have
    \[       \nn(x) = (\nn_1(\proj_{\iota_1}x),\nn_2(\proj_{\iota_2}x), \ldots, \nn_m(\proj_{\iota_m}x)).
    \] This concludes the proof.
\end{proof}
\begin{corollary}\label{cor:parallelization-of-neural-networks}
    Let $m,d \in \N$, $\nn_1,\nn_2, \ldots, \nn_m$ be $m$ neural networks such that $\nnInDim{\nn_i} \leq d$, for all $i \in \{1,\ldots,m\}$, and let $\iota_i : \{1, \ldots, \nnInDim{\nn_i}\} \to \{1,\ldots,d\}$ be an injection for every $i \in \{1,\ldots,m\}$. Then, there exists a neural network $\nn$ such that $\nnDepth{\nn} = \max_{i = 1, \ldots, m} \nnDepth{\nn_i}$, $\nnWidth{\nn} \leq d \vee \sum_{i=1}^m (\nnWidth{\nn_i} \vee 2 \nnOutDim{\nn_i})$, $\nnWeights{\nn} \subseteq \bigcup_{i=1}^m \nnWeights{\nn_i} \cup \{-1,0,1\}$ and $\nn(x,y) = (\nn_1(\proj_{\iota_1}x),\nn_2(\proj_{\iota_2}x), \ldots, \nn_m(\proj_{\iota_m}x))$ for all $x \in \R^d$.
\end{corollary}
\begin{proof}
    Let $L := \max_{i = 1, \ldots, m} \nnDepth{\nn_i}$, and for every $i \in \{1,\ldots,m\}$. For every $i \in \{1,\ldots,m\}$, we define $\nn_i' := \nn_i$ if $\nnDepth{\nn_i} = L$, and $\nn_i'$ as the neural network obtained by applying Lemma \ref{lemma:composition-of-neural-networks} to $\nn_i$ and $\nnId_{\nnOutDim{\nn_i}, L - \nnDepth{\nn_i}}$ otherwise. Note that indeed, $\nnDepth{\nn_i'} = L$, $\nnWidth{\nn_i'} \leq \nnWidth{\nn_i} \vee 2 \nnOutDim{\nn_i}$ and $\nnWeights{\nn_i'} \subseteq \pm\nnWeights{\nn_i} \cup \{-1,0,1\}$. Now, we apply Lemma \ref{lemma:parallelization-of-neural-networks} to the neural networks $\nn_1',\ldots,\nn_m'$, and we let $\nn$ be the resulting neural network. Note that indeed, $\nnDepth{\nn} = L$, $\nnWidth{\nn} \leq d \vee \sum_{i=1}^m (\nnWidth{\nn_i} \vee 2 \nnOutDim{\nn_i})$ and $\nnWeights{\nn} \subseteq \bigcup_{i=1}^m \pm\nnWeights{\nn_i} \cup \{0\}$. Moreover, for every $x \in \R^d$, we have
\[    \nn(x) = (\nn_1'(\proj_{\iota_1}x),\nn_2'(\proj_{\iota_2}x), \ldots, \nn_m'(\proj_{\iota_m}x)) = (\nn_1(\proj_{\iota_1}x),\nn_2(\proj_{\iota_2}x), \ldots, \nn_m(\proj_{\iota_m}x)).
\] This concludes the proof.
\end{proof}

% \begin{lemma}\label{lemma:parallelization-of-neural-networks}
%     Let $\nn_1,\nn_2$ be two neural networks such that $\nnDepth{\nn_1} = \nnDepth{\nn_2}$. Then, there exists a neural network $\nn$ such that $\nnDepth{\nn} = \nnDepth{\nn_1}$, $\nnWidth{\nn} = \nnWidth{\nn_1} + \nnWidth{\nn_2}$, $\nnWeights{\nn} \subseteq \nnWeights{\nn_1} \cup \nnWeights{\nn_2} \cup \{0\}$ and $\nn(x,y) = (\nn_1(x),\nn_2(y))$ for all $x \in \R^{\nnInDim{\nn_1}}, y \in \R^{\nnInDim{\nn_2}}$.
% \end{lemma}
% \begin{proof}
%     Suppose that \[\nn_1 = (N_0^{(1)}, N_1^{(1)}, \ldots, N_{L}^{(1)}; A^1_{(1)}, b^1_{(1)}, A^2_{(1)}, b^2_{(1)}, \ldots, A^{L}_{(1)}, b^{L}_{(1)})\] and \[\nn_2 = (N_0^{(2)}, N_1^{(2)}, \ldots, N_{L}^{(2)}; A^1_{(2)}, b^1_{(2)}, A^2_{(2)}, b^2_{(2)}, \ldots, A^{L}_{(2)}, b^{L}_{(2)}).\] We define \[\nn := (N_0, N_1, \ldots, N_L; A^1, b^1, \ldots, A^L, b^L),\] where $N_\ell = N_\ell^{(1)} + N_\ell^{(2)}$ for $\ell \in \{0, \ldots, L\}$ and $A^\ell = \begin{pmatrix}
%     A^\ell_{(1)} & 0\\
%     0 & A^\ell_{(2)}
% \end{pmatrix}$ and $b^\ell = \begin{pmatrix}
%     b^\ell_{(1)}\\
%     b^\ell_{(2)}
% \end{pmatrix}$ for $\ell \in \{1, \ldots, L\}$. Note that indeed, $\nnDepth{\nn} = \nnDepth{\nn_1} = \nnDepth{\nn_2}$, $\nnWidth{\nn} = \nnWidth{\nn_1} + \nnWidth{\nn_2}$ and $\nnWeights{\nn} \subseteq \nnWeights{\nn_1} \cup \nnWeights{\nn_2} \cup \{0\}$. Now, let $x \in \R^{\nnInDim{\nn_1}}$ and $y \in \R^{\nnInDim{\nn_2}}$, and note that by definition of $\nn$, we have
%  \[       \nn(x,y) = (\nn_1(x),\nn_2(y)).
%     \] This concludes the proof.
% \end{proof}

\begin{lemma}\label{lemma:first-element-reading}
    Let $f : \R \to \R$ be the function defined by \eqref{eq:piecewise-linear-function-read-first-symbol:explicit}. Then, there exists a neural network $\firstNN$ with $\nnWidth{\firstNN} = 4$, $\nnDepth{\firstNN} = 2$ and $\nnWeights{\firstNN} = \{-2,-1,0,1,2,3\}$,
     such that $\firstNN(x) = f(x)$ for all $x \in \R$.
\end{lemma}
\begin{proof}
    We define $\firstNN : \R \to \R$ by $\firstNN := (1, 4, 1; \firstNNmat 1, \firstNNvec 1, \firstNNmat 2, \firstNNvec 2)$, where
    \[
        \firstNNmat 1 := \begin{pmatrix}
            3\\
            3\\
            3\\
            3
        \end{pmatrix}, \quad \firstNNvec 1 := \begin{pmatrix}
            2\\
            1\\
            -1\\
            -2
        \end{pmatrix}, \quad \firstNNmat 2 := \begin{pmatrix}
            1 & -1 & 1 & -1
        \end{pmatrix}, \quad \firstNNvec 2 := -1.
    \]
Note that indeed, $\nnDepth{\firstNN} = 2$, $\nnWidth{\firstNN} = 4$ and $\nnWeights{\firstNN} = \{0,1,2,3,-1,-2\}$. Now, let $x \in \R$, and note that by definition of $\firstNN$, we have
\[
    \firstNN(x) = \ReLU(3x+2) - \ReLU(3x+1) + \ReLU(3x-1) - \ReLU(3x-2) - 1.
\]
\begin{enumerate}[label=(\alph*)]
    \item If $x \leq -2/3$, then
    \[
        \firstNN(x) = 0 - 0 + 0 - 0 - 1 = -1 = f(x).
    \]
    \item If $-2/3 < x < -1/3$, then
    \[
        \firstNN(x) = 3x+2 - 0 + 0 - 0 - 1 = 3x + 1 = f(x).
    \]
    \item If $-1/3 \leq x < 1/3$, then
    \[
        \firstNN(x) = 3x+2 - (3x+1) + 0 - 0 - 1 = 0 = f(x).
    \]
    \item If $1/3 \leq x < 2/3$, then
    \[
        \firstNN(x) = 3x+2 - (3x+1) + (3x-1) - 0 - 1 = 3x - 1 = f(x).
    \]
    \item If $x \geq 2/3$, then
    \[       \firstNN(x) = 3x+2 - (3x+1) + (3x-1) - (3x-2) - 1 = 1 = f(x).
    \]
\end{enumerate}
Therefore, $\firstNN(x) = f(x)$ for all $x \in \R$. This concludes the proof.
\end{proof}

\begin{lemma}\label{lemma:reading-operation-as-neural-network}
    Let $r : \R^2 \to \R$ be the function defined by \eqref{eq:reading-writing-shifting-operations:reading}. Then, there exists a neural network $\readNN$ with $\nnWidth{\readNN} = 4$, $\nnDepth{\readNN} = 2$ and $\nnWeights{\readNN} = \{-2,-1,0,1,2,3\}$, such that $\readNN(x,y) = r(x,y)$ for all $x,y \in \R$.
\end{lemma}
\begin{proof}
By Lemma \ref{lemma:first-element-reading}, there exists a neural network $\firstNN$ with $\nnWidth{\firstNN} = 4$, $\nnDepth{\firstNN} = 2$ and $\nnWeights{\firstNN} = \{-2,-1,0,1,2,3\}$ such that $\firstNN(x) = f(x)$ for all $x \in \R$. Consider the injection $\iota : \{1\} \to \{1\}, i \mapsto i$. By Lemma \ref{lemma:parallelization-of-neural-networks} applied to $m =1$, $d=2$, $\nn_1 = \firstNN$ and $\iota_1 = \iota$, there exists a neural network $\readNN$ with $\nnWidth{\readNN} = 2 \vee \nnWidth{\firstNN} = 4$, $\nnDepth{\readNN} = \nnDepth{\firstNN} = 2$ and $\nnWeights{\readNN} = \nnWeights{\firstNN} \cup \{0\} = \{-2,-1,0,1,2,3\}$ such that \[\readNN(x,y) = \firstNN(\proj_{\iota}(x,y)) = \firstNN(x) = f(x) \overset{(a)}= r(x,y),\] for all $x,y \in \R$, where (a) follows from the definition of $r$ given by \eqref{eq:reading-writing-shifting-operations:reading}. This concludes the proof.
\end{proof}

\newcommand\configReadNN[1]{A_{\text{cread}}^{#1}}
\newcommand\configReadNNmat[2]{A_{\text{cread}}^{#1,#2}}
\newcommand\configReadNNvec[2]{b_{\text{cread}}^{#1,#2}}
% \begin{definition}(Configuration reading neural network)
%     Let $n \in \N$. We define the neural network $\configReadNN n : \R^{n + 2} \to \R$ by $\configReadNN n := (n+2,n+4,n+1; \configReadNNmat n 1, \configReadNNvec n 1, \configReadNNmat n 2, \configReadNNvec n 2)$, where \[\configReadNNmat n 1 = \begin{pmatrix}
%         \idMat_n & 0\\
%         0 & \readNNmat 1
%     \end{pmatrix}, \quad \configReadNNvec n 1 = \begin{pmatrix}
%         0\\
%         \readNNvec 1
%     \end{pmatrix}, \quad \configReadNNmat n 2 = \begin{pmatrix}
%         \idMat_n & 0\\
%         0 & \readNNmat 2
%     \end{pmatrix}, \quad \configReadNNvec n 2 = \begin{pmatrix}
%         0\\
%         \readNNvec 2
%     \end{pmatrix}.
%     \]
% \end{definition}

We now show that the writing operation can be computed by a neural network, in the following sense.

\begin{lemma}\label{lemma:writing-operation-as-neural-network}
    Let $s \in \{-1,0,1\}$, and $w_s : \R^2 \to \R^2$ be the function defined by \eqref{eq:reading-writing-shifting-operations:writing}. Then, there exists a neural network $\writeNN s$ with $\nnWidth{\writeNN s} = 6$, $\nnDepth{\writeNN s} = 2$ and $\nnWeights{\writeNN s} \subseteq \{-2,-1,0,1,2,3\}$, such that $\writeNN s(x,y) = w_s(x,y)$ for all $x,y \in \R_+$.
\end{lemma}
\begin{proof}
    Let $s \in \{-1,0,1\}$. We define the neural network $\writeNN{s} := (2,6,2; \writeNNmat 1 s, \writeNNvec 1 s, \writeNNmat 2 s, \writeNNvec 2 s)$, where
    \[
        \writeNNmat 1 s := \begin{bmatrix}
            \firstNNmat 1 & 0\\
            1 & 0\\
            0 & 1
        \end{bmatrix}, \quad \writeNNvec 1 s := \begin{bmatrix}
            \firstNNvec 1\\
            0
        \end{bmatrix}, \quad \writeNNmat 2 s := \begin{bmatrix}
                -\firstNNmat 2 & 1 & 0\\
                0 & 0 & 1
        \end{bmatrix}, \quad \writeNNvec 2 s := \begin{pmatrix}
                1 + s\\
                0
            \end{pmatrix}.
    \]
Note that we have $\nnDepth{\writeNN{s}} = 2$, $\nnWidth{\writeNN{s}} = 6$ and 
\begin{align*}
    \nnWeights{\writeNN{s}}& = \nnWeights{\firstNNmat 1} \cup \nnWeights{\firstNNvec 1} \cup \nnWeights{-\firstNNmat 2} \cup \{0,1,1+s\}\\
    & = \{0,3\} \cup \{-1,1,-2,2\} \cup \{-1,1\} \cup \{0,1,1+s\} = \{0,1,2,3,-1,-2\}.
\end{align*}
    Note that for every $x,y \in \R_+$, we have
\begin{align*}
    \writeNNmat 1 s (x,y) + \writeNNvec 1 s & = \begin{pmatrix}
            \firstNNmat 1 & 0\\
            1 & 0\\
            0 & 1
        \end{pmatrix} (x,y) + \begin{pmatrix}
            \firstNNvec 1\\
            0
        \end{pmatrix} = \begin{pmatrix}
            \firstNNmat 1 x + \firstNNvec 1\\
            y
        \end{pmatrix},
\end{align*}
and hence
\begin{align*}
    \writeNN s (x,y) &= \writeNNmat 2 s \ReLU(\writeNNmat 1 s (x,y) + \writeNNvec 1 s) + \writeNNvec 2 s\\
     & = \begin{pmatrix}
            -\firstNNmat 2 & 1 & 0\\
            0 & 0 & 1
        \end{pmatrix} \begin{pmatrix}
            \ReLU(\firstNNmat 1 x + \firstNNvec 1)\\
            y
        \end{pmatrix} + \begin{pmatrix}
            1 + s\\
            0
        \end{pmatrix}\\
        & = \begin{pmatrix}
            -\firstNNmat 2 \ReLU(\firstNNmat 1 x_1 + \firstNNvec 1) + x + 1+s\\
            y
        \end{pmatrix}\\
        & = \begin{pmatrix}
            -\firstNN(x) + x_1 + s\\
            y
        \end{pmatrix} \overset{\eqref{eq:reading-writing-shifting-operations:writing}}= w_s(x,y).
\end{align*}
\end{proof}

\begin{lemma}\label{lemma:shifting-operation-as-neural-network}
    Let $m \in \{-1,0,1\}$, and $s_m : \R^2 \to \R^2$ be the function defined by \eqref{eq:reading-writing-shifting-operations:shifting}. Then, there exists a neural network $\shiftNN m$ with $\nnWidth{\shiftNN m} \leq 6$, $\nnDepth{\shiftNN m} \leq 2$ and $\nnWeights{\shiftNN m} \subseteq \{0,1,2,3,1/4,4,-1,-2,-4\}$, such that $\shiftNN m (x,y) = s_m(x,y)$ for all $x,y \in \R_+$.
\end{lemma}
\begin{proof}
    We define the neural networks $\shiftNN{-1}, \shiftNN{0}, \shiftNN{1} : \R^2 \to \R^2$ by 
    \begin{enumerate}[label=(\alph*)]
        \item $\shiftNN{0} := (2,2; \idMat_2, 0)$;
        \item $\shiftNN{1} := (2,6,2; \shiftNNmat 1 1, \shiftNNvec 1 1, \shiftNNmat 2 1, \shiftNNvec 2 1)$, where
        \[
            \shiftNNmat 1 1 := \begin{pmatrix}
                \firstNNmat 1 & 0\\
                1 & 0\\
                0 & 1
            \end{pmatrix}, \quad \shiftNNvec 1 1 := \begin{pmatrix}
                \firstNNvec 1\\
                0
            \end{pmatrix}, \quad \shiftNNmat 2 1 := \begin{bmatrix}
                -4 \firstNNmat 2 & 4 & 0\\
                \firstNNmat 2 & 0 & 1/4
            \end{bmatrix}, \quad \shiftNNvec 2 1 := \begin{pmatrix}
                4\\
                -1
            \end{pmatrix};
        \]
        \item $\shiftNN{-1} := (2,6,2; \shiftNNmat 1 {-1}, \shiftNNvec 1 {-1}, \shiftNNmat 2 {-1}, \shiftNNvec 2 {-1})$, where
        \[
            \shiftNNmat 1 {-1}:= \begin{pmatrix}
                0 & \firstNNmat 1\\
                1 & 0\\
                0 & 1
            \end{pmatrix}, \quad \shiftNNvec 1 {-1} := \begin{pmatrix}
                \firstNNvec 1\\
                0
            \end{pmatrix}, \quad \shiftNNmat 2 {-1} := \begin{bmatrix}
                \firstNNmat 2 & 1/4 & 0\\
                -4\firstNNmat 2 & 0 & 4
            \end{bmatrix}, \quad \shiftNNvec 2 {-1} := \begin{pmatrix}
                -1\\
                4
            \end{pmatrix}.
        \]
    \end{enumerate}
Note that $\nnDepth{\shiftNN{0}} = 1 \leq 2$, $\nnWidth{\shiftNN{0}} = 4 \leq 6$ and $\nnWeights{\shiftNN{0}} = \{-1,0,1\}$, and for $m \in \{-1,1\}$, we have $\nnDepth{\shiftNN{m}} = 2$, $\nnWidth{\shiftNN{m}} = 6$ and
\begin{align*}
    \nnWeights{\shiftNN{m}} & = \nnWeights{\firstNNmat 1} \cup \nnWeights{\firstNNvec 1} \cup \nnWeights{-4 \firstNNmat 2} \cup \nnWeights{\firstNNmat 2} \cup \{0,1/4,4\} \\
    & = \{0,3\} \cup \{-1,1,-2,2\} \cup \{-4, 4\} \cup \{-1,1\} \cup \{0,1/4,4\} = \{0,1,2,3,1/4,4,-1,-2,-4\}.
\end{align*}
    We split the rest of the analysis into three cases, depending on the value of $\move$.
    \begin{enumerate}
        \item Suppose that $\move = 0$. Then, for every $x,y \in \R_+$, we have
        \[
            \shiftNN 0 (x,y) = (x,y) = s_0(x,y).
        \]
        \item Suppose that $\move = 1$. Let $x,y \in \R_+$. We have
        \begin{align*}
            \shiftNNmat 1 1 (x,y) + \shiftNNvec 1 1 & = \begin{pmatrix}
                \firstNNmat 1 & 0\\
                1 & 0\\
                0 & 1
            \end{pmatrix} (x,y) + \begin{pmatrix}
                \firstNNvec 1\\
                0
            \end{pmatrix} = \begin{pmatrix}
                \firstNNmat 1 x + \firstNNvec 1\\
                y
            \end{pmatrix},
        \end{align*}
        and hence
        \begin{align*}
            \shiftNN 1 (x) & = \shiftNNmat 2 1 \ReLU(\shiftNNmat 1 1 (x,y) + \shiftNNvec 1 1) + \shiftNNvec 2 1\\
            & = \begin{pmatrix}
                -4 \firstNNmat 2 & 4 & 0\\
                \firstNNmat 2 & 0 & 1/4
            \end{pmatrix} \begin{pmatrix}
                \ReLU(\firstNNmat 1 x_1 + \firstNNvec 1)\\
                (x,y)
            \end{pmatrix} + \begin{pmatrix}
                4\\
                -1
            \end{pmatrix}\\
            & = \begin{pmatrix}
                -4 \firstNNmat 2 \ReLU(\firstNNmat 1 x + \firstNNvec 1) + 4 x + 4\\
                \firstNNmat 2 \ReLU(\firstNNmat 1 x + \firstNNvec 1) + \frac{y}{4} - 1
            \end{pmatrix}\\
            & = \begin{pmatrix}
                -4 \firstNN(x) + 4 x\\
                \firstNN(x) + \frac{y}{4} 
            \end{pmatrix} = \begin{pmatrix}
                4(x - f(x))\\
                f(x) + \frac{y}{4}
            \end{pmatrix} \overset{\eqref{eq:reading-writing-shifting-operations:shifting}}= s_1(x,y),
        \end{align*}
        where $f(x) = \firstNN(x)$ is the function defined by \eqref{eq:piecewise-linear-function-read-first-symbol:explicit}.
        \item Suppose that $\move = -1$. Inspecting the definition of $\shiftNN{-1}$ and $\shiftNN{1}$, we see that
        $\shiftNN{-1}(x,y) = \begin{pmatrix}
        \shiftNN{1}(y,x)_2\\
        \shiftNN{1}(y,x)_1
        \end{pmatrix} =  \begin{pmatrix}
                f(y) + \frac{x}{4}\\
                4(y - f(y))
            \end{pmatrix} \overset{\eqref{eq:reading-writing-shifting-operations:shifting}}= s_{-1}(x,y),
        $ for every $x,y \in \R_+$. 
    \end{enumerate}
\end{proof}

\newcommand{\writeMoveOp}{SW}
% We defined the operation of exhaustive write-moves, which is the operation that, given a tape, computes the result of applying all possible write-moves to the tape. We define this operation formally as $\writeMoveOp : \workSymbols^\Z \to (\workSymbols^\Z)^9$ by
% \begin{align}\label{eq:write-move-operation}
%     \writeMoveOp(\tape) &:= (\shiftOp(\writeOp(\tape,-1),-1), \shiftOp(\writeOp(\tape,-1),0), \shiftOp(\writeOp(\tape,-1),1),\\
%     & \quad \quad \shiftOp(\writeOp(\tape,0),-1), \shiftOp(\writeOp(\tape,0),0), \shiftOp(\writeOp(\tape,0),1),\nonumber\\
%     & \quad \quad \shiftOp(\writeOp(\tape,1),-1), \shiftOp(\writeOp(\tape,1),0), \shiftOp(\writeOp(\tape,1),1)).\nonumber
% \end{align}

% \subsection{Simulation of the transition and command functions}

\newcommand\symbEncNN{\mathcal{N}_{\mathrm{symb}}}
\newcommand\symbEncNNmat[1]{A^{#1}_{\mathrm{symb}}}
\newcommand\symbEncNNvec[1]{b^{#1}_{\mathrm{symb}}}

\begin{lemma}
    There exists a neural network $\symbEncNN$ with $\nnWidth{\symbEncNN} = 5$, $\nnDepth{\symbEncNN} = 2$ and $\nnWeights{\symbEncNN} \subseteq \{-2,-1,0,1,2\}$, such that for every $\symb \in \workSymbols$, we have $\symbEncNN(\symb) = \oneHot{3}{\symb+2}$.
\end{lemma}

\begin{proof}
    We define the neural network $\symbEncNN := (1,5,3; \symbEncNNmat 1, \symbEncNNvec 1, \symbEncNNmat 2, \zeroVec_3)$, where
    \[
        \symbEncNNmat 1 := (-1,1,1,1,1), \quad \symbEncNNvec 1 := (0,1,0,-1,0)
        , \quad \symbEncNNmat 2 := \begin{pmatrix}
            1 & 0 & 0 & 0 & 0\\
            0 & 1 & -2 & 1 & 0\\
            0 & 0 & 0 & 0 & 1
        \end{pmatrix}.
    \]
Note that $\nnDepth{\symbEncNN} = 2$, $\nnWidth{\symbEncNN} = 5$ and $\nnWeights{\symbEncNN} = \{0,1,-2,-1\}$. Now, let $\symb \in \workSymbols$. We split the analysis into three cases, depending on the value of $\symb$.
    \begin{enumerate}
        \item Suppose that $\symb = -1$. Then, we have
        \begin{align*}
            \symbEncNNmat 1 \symb + \symbEncNNvec 1 & = (-1,1,1,1,1) \symb + (0,1,0,-1,0) = (1,0,-1,-2,-1),
        \end{align*}
        so 
        \[
            \ReLU(\symbEncNNmat 1 \symb + \symbEncNNvec 1) = (1,0,0,0,0),
        \]
        and hence
        \[
            \symbEncNN(\symb) = \symbEncNNmat 2 \ReLU(\symbEncNNmat 1 \symb + \symbEncNNvec 1) = (1,0,0) = \oneHot{3}{1}.
        \]
        \item Suppose that $\symb = 0$. Then, we have
        \begin{align*}
            \symbEncNNmat 1 \symb + \symbEncNNvec 1 & = (-1,1,1,1,1) \symb + (0,1,0,-1,0) = (0,1,0,-1,0),
        \end{align*}
        so 
        \[
            \ReLU(\symbEncNNmat 1 \symb + \symbEncNNvec 1) = (0,1,0,0,0),
        \]
        and hence
        \[
            \symbEncNN(\symb) = \symbEncNNmat 2 \ReLU(\symbEncNNmat 1 \symb + \symbEncNNvec 1) = (0,1,0) = \oneHot{3}{2}.
        \]
        \item Suppose that $\symb = 1$. Then, we have
        \begin{align*}
            \symbEncNNmat 1 \symb + \symbEncNNvec 1 & = (-1,1,1,1,1) \symb + (0,1,0,-1,0) = (-1,2,1,0,1),
        \end{align*}
        so 
        \[
            \ReLU(\symbEncNNmat 1 \symb + \symbEncNNvec 1) = (0,2,1,0,1),
        \]
        and hence
        \[
            \symbEncNN(\symb) = \symbEncNNmat 2 \ReLU(\symbEncNNmat 1 \symb + \symbEncNNvec 1) = (0,0,1) = \oneHot{3}{3}.
        \]
    \end{enumerate}
    This concludes the proof.
\end{proof}

\newcommand\patternNN[1]{\mathcal{N}^{#1}_{\mathrm{pat}}}
\newcommand\patternNNmat[2]{A^{#1,#2}_{\mathrm{pat}}}
\newcommand\patternNNvec[2]{b^{#1,#2}_{\mathrm{pat}}}
\newcommand\diracNotation[1]{\delta_{#1}}
\begin{lemma}
    Let $n \in \N$. There exists a neural network $\patternNN n$ with $\nnWidth{\patternNN n} \leq 3n + 5$, $\nnDepth{\patternNN n} = 3$ and $\nnWeights{\patternNN n} \subseteq \{-2,-1,0,1\}$, such that for every $\state \in \{1, \dots, n\}$ and $\symb \in \workSymbols$, we have $\patternNN n (\oneHot{n}{\state}, \symb) = \oneHot{3n}{3\state + \symb - 1}$.
\end{lemma}
\begin{proof}
    Let $n \in \N$. We define the neural network \[\patternNN n := (n+1,n+5,3n,3n; \patternNNmat n 1, \patternNNvec n 1, \patternNNmat n 1, \patternNNvec n 2,  -1_{3n}, \idMat_{3n}, \zeroVec_{3n}),\] where
    \[
        \patternNNmat n 1 := \begin{bmatrix}
            \idMat_n & 0\\
            0 & \symbEncNNmat 1
        \end{bmatrix}, \quad \patternNNvec n 1 := \begin{bmatrix}
            \zeroVec_n\\
            \symbEncNNvec 1
        \end{bmatrix}, \quad \patternNNmat n 2 := \begin{bmatrix}
            W_1\\
            W_2\\
            \vdots\\
            W_n
        \end{bmatrix}, \quad
    \text{with} \quad W_i := \begin{bmatrix}
            \begin{array}{c}
                \oneHot{n}{\ell}^T\\
                \oneHot{n}{\ell}^T\\
                \oneHot{n}{\ell}^T
            \end{array} & \symbEncNNmat 2
        \end{bmatrix} \in \R^{3 \times (n+5)},
    \]
    for all $\ell \in \{1, \dots, n\}$.
Note that $\nnDepth{\patternNN n} = 3$, $\nnWidth{\patternNN n} = \max\{n+5, 3n\} \leq 3n+5$ and $\nnWeights{\patternNN n} = \{0,1,-1\} \cup \nnWeights{\symbEncNNmat 1} \cup \nnWeights{\symbEncNNvec 1} \cup \nnWeights{\symbEncNNmat 2} = \{0,1,-1,-2\}$. Now, let $\state \in \{1, \dots, n\}$ and $\symb \in \workSymbols$. We have
    \begin{align*}
        y := \ReLU\left(\patternNNmat n 1 \begin{pmatrix}
            \oneHot{n}{\state}\\
            \symb
        \end{pmatrix} + \patternNNvec n 1 \right) & = \ReLU\left(\begin{bmatrix}
            \idMat_n & 0\\
            0 & \symbEncNNmat 1
        \end{bmatrix} \begin{pmatrix}
            \oneHot{n}{\state}\\
            \symb
        \end{pmatrix} + \begin{bmatrix}
            \zeroVec_n\\
            \symbEncNNvec 1
        \end{bmatrix}\right) = \begin{pmatrix}
            \oneHot{n}{\state}\\
            \ReLU\left(\symbEncNNmat 1 \symb + \symbEncNNvec 1\right)
        \end{pmatrix}.
    \end{align*}
    Hence, we have
    \begin{align*}
        \patternNN n (\oneHot{n}{\state},\symb) & = \ReLU\left(\patternNNmat n 2  (\ReLU(\patternNNmat n 1 (\oneHot{n}{\state},\symb) + \patternNNvec n 1)) + (-1_{3n})\right)
        = \begin{pmatrix}
            \ReLU\left( W_1 y - 1_{3}\right)\\
            \ReLU\left( W_2 y - 1_{3}\right)\\
            \vdots\\
            \ReLU\left( W_n y - 1_{3}\right)
        \end{pmatrix}
    \end{align*}
    Note that, for every $\ell \in \{1, \dots, n\}$, we have
    \begin{align*}
        \ReLU\left( W_\ell y - 1_3\right) & = \ReLU\left(\begin{bmatrix}
            \begin{array}{c}
                \oneHot{n}{\ell}^T\\
                \oneHot{n}{\ell}^T\\
                \oneHot{n}{\ell}^T
            \end{array} & \symbEncNNmat 2
        \end{bmatrix} \begin{pmatrix}
            \oneHot{n}{\state}\\
            \ReLU\left(\symbEncNNmat 1 \symb + \symbEncNNvec 1\right)
        \end{pmatrix} - 1_3 \right)\\
        & = \ReLU\left(\begin{bmatrix}
            \oneHot{n}{\ell}^T \oneHot{n}{\state}\\
            \oneHot{n}{\ell}^T \oneHot{n}{\state}\\
            \oneHot{n}{\ell}^T \oneHot{n}{\state}
        \end{bmatrix} - 1_3 + \symbEncNNmat 2 \ReLU\left(\symbEncNNmat 1 \symb + \symbEncNNvec 1\right)\right)\\
        & = \ReLU\left(\begin{bmatrix}
            \diracNotation{\ell,\state}\\
            \diracNotation{\ell,\state}\\
            \diracNotation{\ell,\state}
        \end{bmatrix} - 1_3 + \symbEncNN(\symb)\right)
        = \ReLU\left(\begin{bmatrix}
            \diracNotation{\ell,\state} - 1\\
            \diracNotation{\ell,\state} - 1\\
            \diracNotation{\ell,\state} - 1
        \end{bmatrix} + \oneHot{3}{\symb+2} \right)
         = \diracNotation{\ell,\state} \oneHot{3}{\symb+2}.
    \end{align*}
    Hence, we have
    \begin{align*}
        \patternNN n (\oneHot{n}{\state},\symb) & = \begin{pmatrix}
            \diracNotation{1,\state} \oneHot{3}{\symb+2}\\
            \diracNotation{2,\state} \oneHot{3}{\symb+2}\\
            \vdots\\
            \diracNotation{n,\state} \oneHot{3}{\symb+2}
        \end{pmatrix} = \begin{pmatrix}
            0_{3(q-1)}\\
            \oneHot{3}{\symb+2}\\
            0_{3(n-q)}
        \end{pmatrix} = \oneHot{3n}{3\state + \symb - 1}.
    \end{align*}
    This concludes the proof.
\end{proof}

\newcommand\transStateMat[1]{W^{#1,\state}}
\newcommand\transSymbMat[1]{W^{#1,\symb}}
\newcommand\transMoveMat[1]{W^{#1,\move}}
\newcommand\transSymbMoveMat[1]{W^{#1,\mathrm{symb-move}}}
\newcommand\transCommMat[1]{W^{#1,\commFunc}}
\newcommand\patternMatchingNNMat[1]{W^{#1,\mathrm{match}}}
\newcommand\patternMatchingNNvec[1]{b^{#1,\mathrm{match}}}
\newcommand\patternMatchingNN[1]{\nn^{#1,\mathrm{match}}}
\newcommand\transNN[1]{\mathcal{N}^{#1}_{\mathrm{trans}}}
\newcommand\funGraph{\operatorname{graph}}
\newcommand\characteristicFunc[1]{\operatorname{\chi}_{#1}}

Given a \TMNU $\tmnuM := (\nStates, \ntapes, \neurDim, \transFunc, \commFunc)$, we let $\Fc_\tmnuM := \{ \commFunc(\state,\symb) : \state \in \{1, \dots, \nStates\}, \symb \in \workSymbols\}$ be the set of all possible outputs of the command function $\commFunc$ of $\tmnuM$, and we let $\nFunctions{\tmnuM} := \# \Fc_\tmnuM$. We order the elements of $\Fc_\tmnuM$ as $f_1 <_L f_2 <_L \dots <_L f_{\nFunctions{\tmnuM}}$. We define $\tilde \commFunc : \{1, \dots, \nStates\} \times \workSymbols \to \{1, \ldots, \nFunctions{\tmnuM}\}$ to be such that 
\[
    f_{\tilde \commFunc(\state,\symb)} = \commFunc(\state,\symb), \quad \state \in \{1, \dots, \nStates\}, \symb \in \workSymbols.
\]

Let $A,B$ be two sets. Given a function $f : A \to B$, we denote by $\funGraph(f) := \{(a,b) \in A \times B : f(a) = b\}$ the graph of $f$. Given a set $S \subseteq A$, we denote by $\characteristicFunc{S} : A \to \{0,1\}$ the characteristic function of $S$, defined by $\characteristicFunc{S}(a) := 1$ if $a \in S$ and $\characteristicFunc{S}(a) := 0$ otherwise. 

\newcommand\funcTMNUindex{\ell}
\newcommand\symbMove{j}
\begin{lemma}\label{lem:transition-functions-as-neural-networks-1}
    Let $\tmnuM := (\nStates, \ntapes, \neurDim, \transFunc, \commFunc)$ be a \TMNU, and let the functions $f_\state : \R^{\nStates + 1} \to \R^{\nStates}$, $f_\symb : \R^{\nStates + 1} \to \R^3$, $f_\move : \R^{\nStates+1} \to \R^3 $ and $f_\commFunc : \R^{\nStates+1}\to \R^{\nFunctions{\tmnuM}}$ be defined as in Lemma \ref{lem:transition-functions-as-cpwl-functions}. Then, for $\gamma \in \{\state, \symb, \move, \commFunc\}$, there exist a neural network $\nn_\gamma$ such that $\nnWidth{\nn_\gamma} \leq \max\{3\nStates+5, \nFunctions{\tmnuM}\}$, $\nnDepth{\nn_\gamma} = 3$ and $\nnWeights{\nn_\gamma} \subseteq \{-2,-1,0,1\}$ and
    \begin{align*}
        \nn_\gamma(\oneHot{n}{\state},\symb) & = f_\gamma(\oneHot{n}{\state},\symb),
    \end{align*}
    for every $\state \in \{1, \dots, \nStates\}$ and $\symb \in \workSymbols$.
\end{lemma}
\begin{proof}
    Let $\tmnuM := (\nStates, \ntapes, \neurDim, \transFunc, \commFunc)$ be a TMNU. We define the matrices $\transStateMat{\tmnuM} \in \R^{3\nStates \times \nStates}$, $\transSymbMat{\tmnuM} \in \R^{3\nStates \times 3}$, $\transMoveMat{\tmnuM} \in \R^{3\nStates \times 3}$
     and $\transCommMat{\tmnuM} \in \R^{3\nStates \times \nFunctions{\tmnuM}}$ by
    \begin{align*}
        \transStateMat{\tmnuM}_{3\state+\symb-1,\state'} & = \characteristicFunc{\funGraph(\transFuncState)}((\state, \symb), \state'),\\
        \transSymbMat{\tmnuM}_{3\state+\symb-1,\symb'} & = \characteristicFunc{\funGraph(\transFuncSymb)}((\state, \symb), \symb'),\\
        \transMoveMat{\tmnuM}_{3\state+\symb-1,\move+2} & = \characteristicFunc{\funGraph(\transFuncMove)}((\state, \symb), \move),\\
        \transCommMat{\tmnuM}_{3\state+\symb-1,\funcTMNUindex} & = \characteristicFunc{\funGraph(\tilde \commFunc)}((\state, \symb), \funcTMNUindex),
    \end{align*}
    for all $\state,\state' \in \{1, \dots, \nStates\}$, $\symb,\symb' \in \{-1,0,1\}$, $\move \in \{-1,0,1\}$, and $\funcTMNUindex \in \{1,\ldots, \nFunctions{\tmnuM}\}$. We define the neural networks \[\nn_\gamma := (n+1,n+5,3n,3n; \patternNNmat n 1, \patternNNvec n 1, \patternNNmat n 1, \patternNNvec n 2,  -1_{3n}, W^{\tmnuM,\gamma}, 0),\] where $\gamma \in \{\state, \symb, \move, \commFunc\}$. Note that $\nn_\gamma$ has $\nnDepth{\nn_\gamma} = 3$, $\nnWidth{\nn_\gamma} \leq \max\{3n+5, \nFunctions{\tmnuM}\}$ and $\nnWeights{\nn_\gamma} = \{0,1,-1,-2\}$, for every $\gamma \in \{\state, \symb, \move, \commFunc\}$. Now, let $\state \in \{1, \dots, \nStates\}$ and $\symb \in \workSymbols$. We have
    \begin{align*}
        \nn_\gamma(\oneHot{n}{\state},\symb) & = W^{\tmnuM,\gamma} (\patternNN n (\oneHot{n}{\state},\symb)) = W^{\tmnuM,\gamma}(\oneHot{3n}{3\state + \symb - 1}).
    \end{align*}
    For example, if $\gamma = \state$, we have
    \[
    W^{\tmnuM,\mathrm{state}}(\oneHot{3n}{3\state + \symb - 1}) = \oneHot{\nStates}{\transFuncState(\state,\symb)} = f_\state(\oneHot{n}{\state},\symb).
    \]
    The analysis for the other cases is analogous. This concludes the proof.
\end{proof}
\begin{lemma}\label{lem:transition-functions-as-neural-networks-2}
     Let $\tmnuM := (\nStates, \neurDim, \transFunc, \commFunc)$ be a \TMNU and the functions $\Delta_\state : \R^{\nStates + 2} \to \R^{\nStates}$, $\Delta_\symb : \R^{\nStates + 2} \to \R^3$, $\Delta_\move : \R^{\nStates+2} \to \R^3 $ and $\Delta_\commFunc : \R^{\nStates+2}\to \R^{\nFunctions{\tmnuM}}$ be defined as in Corollary \ref{cor:transition-functions-as-cpwl-functions}. Then, for $\gamma \in \{\state, \symb, \move, \commFunc\}$, there exist a neural network $\tilde \nn_\gamma$ such that $\nnWidth{\tilde \nn_\gamma} \leq 2\max\{3\nStates+5, \nFunctions{\tmnuM}\}$, $\nnDepth{\tilde \nn_\gamma} = 5$, $\nnWeights{\tilde \nn_\gamma} \subseteq \{-3,-2,-1,0,1,2,3\}$, and
    \begin{align*}
        \tilde \nn_\gamma(\oneHot{n}{\state},\cantorMap(\tape)) & = \Delta_\gamma(\oneHot{n}{\state},\cantorMap(\tape)),
    \end{align*}
    for every $\state \in \{1, \dots, \nStates\}$ and $\tape \in \workSymbols^\Z$.
\end{lemma}
\begin{proof}
    We make the proof for $\gamma = \state$. The other cases are analogous. Note that $\Delta_\state(x,y) = f_\state(x, r(y))$, for $x \in \R^n$ and $y \in \R^2$ where $r : \R^2 \to \R$ is defined by \eqref{eq:reading-writing-shifting-operations:reading} and $f_q$ is as in Lemma \ref{lem:transition-functions-as-cpwl-functions}. By Lemma \ref{lemma:identity-neural-network}, there exists a neural network $\nnId_n$ such that $\nnDepth{\nnId_n} = 2$, $\nnWidth{\nnId_n} = 2n$ and $\nnWeights{\nnId_n} = \{-1,0,1\}$, and $\nnId_n(x) = x$ for every $x \in \R^n$. Moreover, by Lemma \ref{lemma:reading-operation-as-neural-network}, there exists a neural network $\readNN$ such that $\nnDepth{\readNN} = 2$, $\nnWidth{\readNN} = 4$ and $\nnWeights{\readNN} \subseteq \{-2,-1,0,1,2,3\}$, and $\readNN(y) = r(y)$ for every $y \in \R^2$.
    Furthermore, consider the injections $\iota_1 : \{1, \ldots, n\} \to \{1,\ldots,n\}, i \mapsto x$ and $\iota_2 : \{1,2\} \to \{n+1,n+2\}, i \mapsto i + n$. By application of Lemma \ref{lemma:parallelization-of-neural-networks} with $d = n+2$, $m =2$, $\nn_1 = \nnId_n$, $\nn_2 = \readNN$, $\iota_1 = \iota_1$ and $\iota_2 = \iota_2$, there exists a neural network $\nn$ such that $\nnDepth{\nn} = 2$, $\nnWidth{\nn} = 2n + 4$ and $\nnWeights{\nn} \subseteq \{-2,-1,0,1,2,3\}$, and 
    \[
        \nn(x,y) = (\nnId_n(\proj_{\iota_1}(x,y)), \readNN(\proj_{\iota_2}(x,y))) = (\nnId_n(x), \readNN(y)) =  (x,r(y)), \quad x \in \R^n, y \in \R^2.
    \]
    Now, by Lemma \ref{lem:transition-functions-as-neural-networks-1}, there exists a neural network $\nn_\state$ such that $\nnDepth{\nn_\state} = 3$, $\nnWidth{\nn_\state} \leq \max\{3n+5, \nFunctions{\tmnuM}\}$ and $\nnWeights{\nn_\state} \subseteq \{-2,-1,0,1\}$, and $\nn_\state(\oneHot{n}{\state},\symb) = f_\state(\oneHot{n}{\state},\symb)$ for every $\state \in \{1,\ldots,n\}$ and $\symb \in \workSymbols$. Therefore, by Lemma \ref{lemma:composition-of-neural-networks} applied to $\nn$ and $\nn_\state$, there exists a neural network $\tilde \nn_\state$ such that $\nnDepth{\tilde \nn_\state} = 5$, $\nnWidth{\tilde \nn_\state} \leq \max\{3n+5, \nFunctions{\tmnuM}\}$ and $\nnWeights{\tilde \nn_\state} \subseteq \{-3,-2,-1,0,1,2,3\}$, and 
    \[
        \tilde \nn_\state(\oneHot{n}{\state},\cantorMap(\tape)) = \nn_\state(\nn(\oneHot{n}{\state},\cantorMap(\tape))) = \nn_\state(\oneHot{n}{\state}, r(\cantorMap(\tape))) \overset{(a)}= f_\state(\oneHot{n}{\state}, r(\cantorMap(\tape))),
    \]
    for every $\state \in \{1, \dots, n\}$ and $\tape \in \workSymbols^\Z$, where (a) follows from $r(\cantorMap(\tape)) = \tape_0 \in \workSymbols$. This concludes the proof.
\end{proof}

\begin{lemma}
    Let $n,d \in \N$, $M>0$, and $g_{M,n,d} : \R^{n(d+1)}\to \R^d$ be defined as in Lemma \ref{lem:selection-function}. Then, there exists a neural network $\nn$ such that $\nnDepth{\nn} = 4$, $\nnWidth{\nn} = 4n(d+1)$, $\nnWeights{\nn} \subseteq \{-M,-1,0,1,M\}$, and $\nn(x) = g_{M,n,d}(x,z)$ for every $x \in \R_+^n$ and $z \in \R^{n d}$.
\end{lemma}
\begin{proof}
    First consider the function $g^+_{M,n,d} : \R^{n(d+1)}\to \R^d$ as in the proof Lemma \ref{lem:selection-function}. For every $x^1,x^2 \in \R^n$, $z^1_1, \dots, z^1_n, z^2_1, \dots, z^2_n  \in \R^d$ we have
    \[
        g^+_{M,n,d}(x^1,z^1_1, \dots, z^1_n) - g^+_{M,n,d}(x^2,z^2_1, \dots, z^2_n) = \sum_{i=1}^n \ReLU(z^1_i + M (x^2_i - 1) 1_d) - \sum_{i=1}^n \ReLU(z^2_i + M (x^2_i - 1) 1_d).
    \]
    Note that the map
    \[
        W :(x,z) = z + M (x - 1) 1_d \in \R^d, \quad x \in \R, z \in \R^d,
    \]
    is an affine map such that $\nnWeights{W} \subseteq \{-M,0,1,M\}$, so the map
    \[
        W_1 : (x^1,z^2_1, \dots, z^2_n,x^2,z^2_1, \dots, z^2_n) \mapsto (W(x^1_1,z^1_1), \dots, W(x^1_n,z^1_n),W(x^2_1,z^2_1), \dots, W(x^2_n,z^2_n)) \in \R^{2nd},
    \]
    is an affine map such that $\nnWeights{W_1} \subseteq \{-M,0,1,M\}$, and the map
    \[
        W_2 : (z^1_1, \dots, z^1_n, z^2_1, \dots, z^2_n) \mapsto \sum_{i=1}^n z^1_i - \sum_{i=1}^n z^2_i \in \R^d
    \]
    is an affine map such that $\nnWeights{W_2} \subseteq \{-1,0,1\}$. Hence, there
    i.e., there exists $A_1 \in \R^{2n(d+1) \times 2nd}, A_2 \in \R^{2nd \times d},b_1 \in \R^{2nd}$, and $b_2 \in \R^d$, such that $\nnWeights{A_1}, \nnWeights{A_2}, \nnWeights{b_1}, \nnWeights{b_2} \subseteq \{-M,-1,0,1,M\}$, and the neural network $\nn_g := (2n(d+1), 2n d, d; A_1, b_1, A_2, b_2)$ is such that $\nnDepth{\nn_g} = 2$, $\nnWidth{\nn_g} = 2n(d+1)$, $\nnWeights{\nn_g} \subseteq \{-M,-1,0,1,M\}$ and
    \[\nn_g(x^1,z^1_1, \dots, z^1_n,x^2,z^2_1, \dots, z^2_n) = g^+_{M,n,d}(x^1,z^1_1, \dots, z^1_n) - g^+_{M,n,d}(x^2,z^2_1, \dots, z^2_n),\] 
    for every $x^1,x^2 \in \R^n$, $z^1_1, \dots, z^1_n, z^1_2, \dots, z^2_n \in \R^d$. 
    Now, for $\ell \in \N$, consider the neural networks 
    \[\nn^+_\ell := (\ell, \ell, \ell; \idMat_\ell, \zeroVec_\ell, \idMat_\ell, \zeroVec_\ell)\]
    and
    \[\nn^-_\ell := (\ell, \ell, \ell; -\idMat_\ell, \zeroVec_\ell, \idMat_\ell, \zeroVec_\ell).\]
    Note that $\nnDepth{\nn^+_\ell} = \nnDepth{\nn^-_\ell} = 2$, $\nnWidth{\nn^+_\ell} = \nnWidth{\nn^-_\ell} = \ell$, $\nnWeights{\nn^+_\ell}, \nnWeights{\nn^-_\ell} \subseteq \{-1,0,1\}$, $\nn^+_\ell(z) = \ReLU(z)$ and $\nn^-_\ell(z) = \ReLU(-z)$ for every $z \in \R^\ell$. In particular, if $z \in \R_+^\ell$, then $\nn_\ell(z) = z$. Furthermore, consider the injections $\iota_1 : \{1, \dots, n\} \to \{1, \dots, n (d+1)\}, i \mapsto i$, $\iota_{k+1} : \{1, \dots, d\} \to \{n + d(k-1) + 1, \dots, n + dk\}, i \mapsto i + d(k-1) + n$, for $k \in \{1,\ldots,n\}$, $\iota_{n+2} : \{1, \dots, n\} \to \{n(d+1)+1, \dots, n(d+2)\}, i \mapsto i + n(d+1)$, and $\iota_{n+2+k} : \{1, \dots, d\} \to \{n(d+1) + n + d(k-1) + 1, \dots, n(d+1) + n + dk\}, i \mapsto i + n(d+1) + n + d(k-1)$ for $k \in \{1,\ldots,n\}$. By Lemma \ref{lemma:parallelization-of-neural-networks} with $d = 2n (d+1)$, $m = 2(n+1)$, $\nn_1 = \nn_{n+2} = \nn^+_n$, $\nn_{k+1} = \nn^+_{d}$ and $\nn_{k+n+2} = \nn^-_d$ for $k \in \{1,\ldots ,n\}$, and $\iota_k = \iota_k$ for $k \in \{1,\ldots ,2(n+1)\}$, there exists a neural network $\nn$ such that $\nnDepth{\nn} = 2$, $\nnWidth{\nn} = 2n(d+1)$ and $\nnWeights{\nn} \subseteq \{-1,0,1\}$, and
    \[
        \nn(x,z_1, \dots, z_d) = (x, \ReLU(z_1), \dots, \ReLU(z_d), x, \ReLU(-z_1), \dots, \ReLU(-z_d)), \quad x \in \R_+^n, z_1, \dots, z_d \in \R^{n}.
    \]
    Now, by application of Lemma \ref{lemma:composition-of-neural-networks} to $\nn_g$ and $\nn$, there exists a neural network $\tilde \nn_g$ such that $\nnDepth{\tilde \nn_g} = 4$, $\nnWidth{\tilde \nn_g} = 4n(d+1)$, $\nnWeights{\tilde \nn_g} \subseteq \{-M,-1,0,1,M\}$, and
    \begin{align*}
        \tilde \nn_g(x,z_1, \dots, z_d) & = \nn_g(\nn(x,z_1, \dots, z_d)) = \nn_g(x, \ReLU(z_1), \dots, \ReLU(z_d), x, \ReLU(-z_1), \dots, \ReLU(-z_d))\\
        & = g^+_{M,n,d}(x,\ReLU(z_1), \dots, \ReLU(z_d)) - g^+_{M,n,d}(x,\ReLU(-z_1), \dots, \ReLU(-z_d))\\
        & = g_{M,n,d}(x,z_1, \dots, z_d),
    \end{align*}
    for every $x \in \R_+^n$ and $z_1, \dots, z_d \in \R^n$. This concludes the proof.
\end{proof}
\begin{lemma}\label{lem:update-function-as-neural-network}
    Let $\tmnuM := (\nStates, \neurDim, \transFunc, \commFunc)$ be a \TMNU, and $C > 0$, and $F_{\tmnuM,C} : \R^{\nStates + 2 + \neurDim} \to \R^{\nStates + 2 + \neurDim}$ be defined as in Theorem \ref{thm:simulation-of-tmnu-by-cpwl-function}. Then, there exists a neural network $\nn$ such that $\nnDepth{\nn} = 18$, $\nnWidth{\nn} \leq \max\{5\nStates, \nStates + 2\nFunctions{\tmnuM}\neurDim\} + 2\nFunctions{\tmnuM}\neurDim + 37$, $\nnWeights{\nn} \subseteq \pm \{0,1/4,1,2,3,4,C\} \cup \pm \nnWeights{\tmnuM}$,
    and $\nn(\configMap{\tmnuM}(c)) = F_{\tmnuM,C}(\configMap{\tmnuM}(c))$ for every $c \in \Bc_\tmnuM(C)$.
\end{lemma}
\begin{proof}
    Following the proof of Theorem \ref{thm:simulation-of-tmnu-by-cpwl-function}, we split the function $F_{\tmnuM,C}$ into $F_q$, $F_\tape$ and $F_{\neurState,C}$.
    \begin{enumerate}
        \item Note that $F_q(x,y,z) := \Delta_\state(x,y)$, where $\Delta_\state$ is as in Corollary \ref{cor:transition-functions-as-cpwl-functions}. By Lemma \ref{lem:transition-functions-as-neural-networks-2} there exists a neural network $\tilde \nn_\state$ such that $\nnDepth{\tilde \nn_\state} = 5$, $\nnWidth{\tilde \nn_\state} \leq \max\{3\nStates+5, \nFunctions{\tmnuM}\}$, $\nnWeights{\tilde \nn_\state} \subseteq \{-3,-2,-1,0,1,2,3\}$ and $\tilde \nn_\state(\oneHot{\nStates},\cantorMap{\tape}) = \Delta_\state(\oneHot{\nStates},\cantorMap(\tape))$ for every $\state \in \{1,\ldots,\nStates\}$, $\tape \in \workSymbols^\Z$ and $z \in \R^\neurDim$. Now, by Lemma \ref{lemma:parallelization-of-neural-networks} with $d = \nStates + 2 + \neurDim$, $m = 1$, $\nn_1 = \tilde \nn_\state$ and $\iota_1 : \{1, \dots, \nStates+2\} \to \{1, \dots, \nStates + 2 + \neurDim\}, i \mapsto i$, there exists a neural network $\nn_q$ such that $\nnDepth{\nn_q} = 5$, $\nnWidth{\nn_q} \leq \max\{3\nStates, \nFunctions{\tmnuM},\nStates + \neurDim\} + 5$, $\nnWeights{\nn_q} \subseteq \{-3,-2,-1,0,1,2,3\}$ and 
        \[
            \nn_q(x) = \tilde \nn_\state(\proj_{\iota_1}(x)) = \tilde \nn_\state(x_{1:\nStates+2})
        \]
        for every $x \in \R^{\nStates + 2 + \neurDim}$. In particular, for $x = \configMap{\tmnuM}(c) \in \R^{\nStates + 2 + \neurDim}$, we have $\nn_q(\configMap{\tmnuM}(c)) = \tilde \nn_\state(\oneHot{\nStates},\cantorMap(\tape)) = \Delta_\state(\oneHot{\nStates},\cantorMap(\tape)) = F_q(\configMap{\tmnuM}(c))$.
        \item $F_\tape$ is decomposed into two functions 
        \begin{equation}
            F_{\text{write}}(x,y,z) := g_{1,3,2}(\Delta_\symb(x,y), w_{-1}(y), w_0(y), w_1(y)),
        \end{equation}
        and 
        \begin{equation}
            F_{\text{shift}}(x,y,z) := g_{1,3,2}(\Delta_\move(x,y), s_{-1}(y), s_0(y), s_1(y)),
        \end{equation}
        for every $x \in \R^{\nStates}$, $y \in \R^2$ and $z \in \R^\neurDim$, where $\Delta_\symb$ and $\Delta_\move$ are as in Corollary \ref{cor:transition-functions-as-cpwl-functions}, $w_{-1}$, $w_0$ and $w_1$ are as in Lemma \ref{lem:reading-writing-shifting-operations} $s_{-1}$, $s_0$ and $s_1$ are as in Lemma \ref{lem:reading-writing-shifting-operations}, and $g_{1,3,2}$ is as in Lemma \ref{lem:selection-function}. 
        \begin{enumerate}
            \item By Lemma \ref{lem:transition-functions-as-neural-networks-2} there exist neural networks $\tilde \nn_\symb$ and $\tilde \nn_\move$ such that $\nnDepth{\tilde \nn_\symb} = \nnDepth{\tilde \nn_\move} = 5$, $\nnWidth{\tilde \nn_\symb}, \nnWidth{\tilde \nn_\move} \leq \max\{3\nStates+5, \nFunctions{\tmnuM}\}$, $\nnWeights{\tilde \nn_\symb}, \nnWeights{\tilde \nn_\move} \subseteq \{-3,-2,-1,0,1,2,3\}$ and $\tilde \nn_\symb(\oneHot{\nStates},\cantorMap(\tape)) = \Delta_\symb(\oneHot{\nStates},\cantorMap(\tape))$ and $\tilde \nn_\move(\oneHot{\nStates},\cantorMap(\tape)) = \Delta_\move(\oneHot{\nStates},\cantorMap(\tape))$ for every $\state \in \{1,\ldots,\nStates\}$, and $\tape \in \workSymbols^\Z$.
            \item By Lemma \ref{lemma:writing-operation-as-neural-network} there exists neural networks $\nn^\symb_{-1},\nn^\symb_0,\nn^\symb_1$ such that $\nnDepth{\nn^\symb_{-1}} = \nnDepth{\nn^\symb_0} = \nnDepth{\nn^\symb_1} = 2$, $\nnWidth{\nn^\symb_{-1}} = \nnWidth{\nn^\symb_0} = \nnWidth{\nn^\symb_1} = 6$, $\nnWeights{\nn^\symb_{-1}}, \nnWeights{\nn^\symb_0}, \nnWeights{\nn^\symb_1} \subseteq \{-2,-1,0,1,2,3\}$ and $\nn^\symb_{-1}(x,y) = w_{-1}(y)$, $\nn^\symb_0(x,y) = w_0(y)$ and $\nn^\symb_1(x,y) = w_1(y)$ for every $x \in \R^{\nStates}$, $y \in \R^2$ and $z \in \R^\neurDim$.
            \item By Lemma \ref{lemma:shifting-operation-as-neural-network} there exists neural networks $\nn^\move_{-1},\nn^\move_0,\nn^\move_1$ such that $\nnDepth{\nn^\move_{-1}} = \nnDepth{\nn^\move_0} = \nnDepth{\nn^\move_1} = 2$, $\nnWidth{\nn^\move_{-1}} = \nnWidth{\nn^\move_0} = \nnWidth{\nn^\move_1} = 6$, $\nnWeights{\nn^\move_{-1}}, \nnWeights{\nn^\move_0}, \nnWeights{\nn^\move_1} \subseteq \{-4,-2,-1,0,1/4,1,2,3,4\}$ and $\nn^\move_{-1}(x,y) = s_{-1}(y)$, $\nn^\move_0(x,y) = s_0(y)$ and $\nn^\move_1(x,y) = s_1(y)$ for every $x \in \R^{\nStates}$, $y \in \R^2$ and $z \in \R^\neurDim$.
        \end{enumerate}
        
        Now, consider $\iota_\Delta: \{1, \dots, \nStates+2\} \to \{1, \dots, \nStates + 2\}, i \mapsto i$, and $\iota: \{1,2\} \to \{1, \dots, \nStates + 2\}, i \mapsto i + \nStates$. By Corollary \ref{cor:parallelization-of-neural-networks} with $d = \nStates + 2$, $m = 4$, $\nn_1 = \tilde \nn_\symb$, $\nn_2 = \tilde \nn_\move$, $\nn_3 = \nn^\symb_{-1}$, $\nn_4 = \nn^\symb_0$, $\nn_5 = \nn^\symb_1$, $\nn_6 = \nn^\move_{-1}$, $\nn_7 = \nn^\move_0$, $\nn_8 = \nn^\move_1$, $\iota_1 = \iota_2 = \iota_\Delta$ and $\iota_{k} = \iota$ for $k \in \{3,\ldots,8\}$, there exists a neural network $\nn$ such that $\nnDepth{\nn} = 5$, 
        \[
            \nnWidth{\nn} \leq 2 \cdot \max\{3n +5, \nFunctions{\tmnuM}\} + 6 \cdot6 = 2 \max\{3n +5, \nFunctions{\tmnuM}\} + 36,
        \]
        $\nnWeights{\nn} \subseteq \pm \{0,1/4,1,2,3,4\}$, and
        \begin{align*}
            \nn(x,y) &= \begin{pmatrix}
            \tilde \nn_\symb(\proj_{\iota_\Delta}(x,y))\\
            \tilde \nn_\move(\proj_{\iota_\Delta}(x,y))\\
            \nn^\symb_{-1}(\proj_{\iota}(x,y))\\
            \nn^\symb_0(\proj_{\iota}(x,y))\\
            \nn^\symb_1(\proj_{\iota}(x,y))\\
            \nn^\move_{-1}(\proj_{\iota}(x,y))\\
            \nn^\move_0(\proj_{\iota}(x,y))\\
            \nn^\move_1(\proj_{\iota}(x,y))
            \end{pmatrix}
            = \begin{pmatrix}
            \tilde \nn_\symb(x,y)\\
            \tilde \nn_\move(x,y)\\
            \nn^\symb_{-1}(y)\\
            \nn^\symb_0(y)\\
            \nn^\symb_1(y)\\
            \nn^\move_{-1}(y)\\
            \nn^\move_0(y)\\
            \nn^\move_1(y)
            \end{pmatrix},
        \end{align*}
        for every $x \in \R^\nStates$ and $y \in \R^2$. 

        Now, consider $\iota_\Delta: \{1, \dots, \nStates+2\} \to \{1, \dots, \nStates + 2\}, i \mapsto i$, and $\iota: \{1,2\} \to \{1, \dots, \nStates + 2\}, i \mapsto i + \nStates$. By Corollary \ref{cor:parallelization-of-neural-networks} with $d = \nStates + 2$, $m = 4$, $\nn_1 = \tilde \nn_\symb$, $\nn_2 = \nn^\symb_{-1}$, $\nn_3 = \nn^\symb_0$, $\nn_4 = \nn^\symb_1$, $\iota_1 = \iota_\Delta$ and $\iota_2 = \iota_3 = \iota_4 = \iota$, there exists a neural network $\nn$ such that $\nnDepth{\nn} = 5$, 
        $\nnWidth{\nn} \leq \max\{3n +5, \nFunctions{\tmnuM}\} + 6 \cdot 3 = \max\{3n +5, \nFunctions{\tmnuM}\} + 18$,
        $\nnWeights{\nn} \subseteq \pm \{0,1,2,3\}$, and
        \begin{align*}
            \nn(x,y) &= \begin{pmatrix}
            \tilde \nn_\symb(\proj_{\iota_\Delta}(x,y))\\
            \nn^\symb_{-1}(\proj_{\iota}(x,y))\\
            \nn^\symb_0(\proj_{\iota}(x,y))\\
            \nn^\symb_1(\proj_{\iota}(x,y))\\
            \end{pmatrix}
            = \begin{pmatrix}
            \tilde \nn_\symb(x,y)\\
            \nn^\symb_{-1}(y)\\
            \nn^\symb_0(y)\\
            \nn^\symb_1(y)\\
            \end{pmatrix} \in \R^{3 + 2 + 2 + 2} = \R^9,
        \end{align*}
        for every $x \in \R^\nStates$ and $y \in \R^2$. 
        
        % In particular, for $x = \oneHot{\nStates}{\state} \in \R^\nStates$ and $y = \cantorMap(\tape) \in \R^2$, we have
        % \begin{align*}
        %     \nn(x,y) & = \begin{pmatrix}
        %     \tilde \nn_\symb(\oneHot{\nStates},\cantorMap(\tape))\\
        %     \nn^\symb_{-1}(\cantorMap(\tape))\\
        %     \nn^\symb_0(\cantorMap(\tape))\\
        %     \nn^\symb_1(\cantorMap(\tape))\\
        %     \end{pmatrix} = \begin{pmatrix}
        %     \Delta_\symb(\oneHot{\nStates},\cantorMap(\tape))\\
        %     w_{-1}(\cantorMap(\tape))\\
        %     w_0(\cantorMap(\tape))\\
        %     w_1(\cantorMap(\tape))\\
        %     \end{pmatrix},
        % \end{align*}
        
        Now, by Lemma \ref{lem:selection-function}, there exists a neural network $\nn_{g}$ such that $\nnDepth{\nn_g} = 4$, $\nnWidth{\nn_g} = 36$, $\nnWeights{\nn_g} \subseteq \{-1,0,1\}$ and $\nn_g(x,z) = g_{1,3,2}(x,z)$ for every $x \in \R^3$ and $z \in \R^6$. Therefore, by Lemma \ref{lemma:composition-of-neural-networks} applied to $\nn$ and $\nn_g$, there exists a neural network $\tilde \nn_{\text{write}}$ such that $\nnDepth{\tilde \nn_{\text{write}}} = 9$, $\nnWidth{\tilde \nn_{\text{write}}} \leq \max\{3\nStates, \nFunctions{\tmnuM}\} + 23$, $\nnWeights{\tilde \nn_{\text{write}}} \subseteq \pm\{0,1,2,3\}$, and
        \begin{align*}
            \tilde \nn_{\text{write}}(x,y) & = \nn_g(\nn(x,y)) = \nn_g(\tilde \nn_\symb(x,y), \nn^\symb_{-1}(y), \nn^\symb_0(y), \nn^\symb_1(y)),
        \end{align*}
        for every $x \in \R^\nStates$ and $y \in \R^2$. In particular, for $x = \oneHot{n}{q}$ and $y = \cantorMap(\tape)$, we have 
        \begin{align*}
                \tilde \nn_{\text{write}}(\oneHot{n}{q}, \cantorMap(\tape)) & = \nn_g(\tilde \nn_\symb(\oneHot{\nStates},\cantorMap(\tape)), \nn_{-1}(\cantorMap(\tape)), \nn_0(\cantorMap(\tape)), \nn_1(\cantorMap(\tape)))\\
                & = g_{1,3,2}(\Delta_\symb(\oneHot{\nStates},\cantorMap(\tape)), w_{-1}(\cantorMap(\tape)), w_0(\cantorMap(\tape)), w_1(\cantorMap(\tape)))\\
                & = F_{\text{write}}(\oneHot{n}{q}, \cantorMap(\tape)),
        \end{align*}
        for every $q \in \{1,\ldots, \nStates\}$ and $\tape \in \workSymbols^\Z$.
        The analysis for $F_{\text{shift}}$ is analogous, and we get that there exists a neural network $\tilde \nn_{\text{shift}}$ such that $\nnDepth{\tilde \nn_{\text{shift}}} = 9$, $\nnWidth{\tilde \nn_{\text{shift}}} \leq \max\{3\nStates, \nFunctions{\tmnuM}\} + 23$, $\nnWeights{\tilde \nn_{\text{shift}}} \subseteq \pm\{0,1/4,1,2,3,4\}$, and $\tilde \nn_{\text{shift}}(\oneHot{n}{q}, \cantorMap(\tape),\cantorMap(\tape')) = F_{\text{shift}}(\oneHot{n}{q}, \cantorMap(\tape),\cantorMap(\tape'))$ for every $q \in \{1,\ldots, \nStates\}$ and $\tape,\tape' \in \workSymbols^\Z$. Finally, note that
        \[
            F_\tape(x,y,z) = F_{\text{shift}}(x,y,F_{\text{write}}(x,y)).
        \]
        First, consider $\nnId_{n+2,9}$ from Lemma \ref{lemma:identity-neural-network}, and note that $\nnDepth{\nnId_{n+2,9}} = 9$, $\nnWidth{\nnId_{n+2,9}} = 2(\nStates + 2)$ and $\nnWeights{\nnId_{n+2,9}} \subseteq \{-1,0,1\}$, and $\nnId_{n+2,9}(x) = x$ for every $x \in \R^{\nStates + 2}$. By application of Lemma \ref{lemma:parallelization-of-neural-networks} with $d = \nStates + 2 + \neurDim$, $m = 2$, $\nn_1 = \nnId_{n+2,9}$, $\nn_2 = \tilde \nn_{\text{write}}$, and $\iota_1,\iota_2 = \iota : \{1, \dots, \nStates+2\} \to \{1, \dots, \nStates + 2 + \neurDim\}, i \mapsto i$, there exists a neural network $\nn$ such that $\nnDepth{\nn} = 9$, $\nnWidth{\nn} \leq \max\{5\nStates, \nFunctions{\tmnuM}\} + 27$ and $\nnWeights{\nn} \subseteq \pm\{0,1,2,3\}$, and
            \[
                \nn(x,y,z) = (\nnId_{n+2,9}(\proj_\iota(x,y,z)), \tilde \nn_{\text{write}}(\proj_\iota(x,y,z))) = (x, y, \tilde \nn_{\text{write}}(x,y)),
            \]
        for every $x \in \R^{\nStates}$, $y \in \R^2$ and $z \in \R^\neurDim$. Now, by Lemma \ref{lemma:composition-of-neural-networks} applied to $\nn$ and $\tilde \nn_{\text{shift}}$, there exists a neural network $ \nn_\tape$ such that $\nnDepth{ \nn_\tape} = 18$, $\nnWidth{ \nn_\tape} \leq \max\{5\nStates, \nFunctions{\tmnuM}\} + 27$ and $\nnWeights{ \nn_\tape} \subseteq \pm\{0,1/4,1,2,3,4\}$, and
        \[
             \nn_\tape(x,y,z) = \tilde \nn_{\text{shift}}(\nn(x,y,z),\cantorMap(\tape')) = F_{\text{shift}}(x,y,\tilde \nn_{\text{write}}(x,y)) = F_\tape(x,y,z),
        \]
        for every $x \in \R^{\nStates}$, $y \in \R^2$ and $z \in \R^\neurDim$. Therefore, in particular, for $(x,y,z) = (\oneHot{n}{q}; \cantorMap(\tape); \neurState) = \configMap{\tmnuM}(c)$, we have \[ \nn_\tape(\oneHot{n}{q}, \cantorMap(\tape),\cantorMap(\tape')) = F_\tape(\oneHot{n}{q}, \cantorMap(\tape),\cantorMap(\tape'))\] for every $c = (q,\tape,\neurState) \in \Bc_\tmnuM(C)$.
        \item Let $f_1, \ldots, f_{\nFunctions{\tmnuM}}$ be an enumeration of the functions in $\Fc_\tmnuM :=\{\commFunc(\state,\symb) : \state \in \{1,\ldots,\nStates\}, \symb \in \workSymbols\}$. Note that for every $i \in \{1,\ldots, \nFunctions{\tmnuM}\}$, $f_i : \R^\neurDim \to \R^\neurDim$ is either an affine function or the ReLU function, so there exists a neural network $\nn_{f_i}$ such that $\nnDepth{\nn_{f_i}} \leq 2$, $\nnWidth{\nn_{f_i}} = \neurDim$, $\nnWeights{\nn_{f_i}} \subseteq \nnWeights{\tmnuM}$ and $\nn_{f_i}(z) = f_i(z)$ for every $z \in \R^\neurDim$. Moreover, by Lemma \ref{lem:transition-functions-as-neural-networks-2}, there exists a neural network $\nn_\commFunc$ such that $\nnDepth{\nn_\commFunc} = 5$, $\nnWidth{\nn_\commFunc} \leq \max\{3\nStates+5, \nFunctions{\tmnuM}\}$, $\nnWeights{\nn_\commFunc} \subseteq \{-3,-2,-1,0,1,2,3\}$ and $\nn_\commFunc(\oneHot{\nStates},\cantorMap(\tape)) = \Delta_\commFunc(\oneHot{\nStates},\cantorMap(\tape))$ for every $\state \in \{1,\ldots,\nStates\}$ and $\tape \in \workSymbols^\Z$. Now, by Corollary \ref{cor:parallelization-of-neural-networks} with $d = \nStates + 2 + \neurDim$, $m = \nFunctions{\tmnuM}+1$, $\nn_1 = \nn_\commFunc$, $\iota_1 : \{1, \dots, \nStates+2\} \to \{1, \dots, \nStates + 2 + \neurDim\}, i \mapsto i$, $\nn_{k+1} = \nn_{f_k}$ and $\iota_{k+1} = \iota : \{1, \ldots, \neurDim\} \to \{1, \dots, \nStates + 2 + \neurDim\}, i \mapsto i + \nStates + 2$ for every $k \in \{1,\ldots, \nFunctions{\tmnuM}\}$, there exists a neural network $\nn_\commFunc'$ such that $\nnDepth{\nn_\commFunc'} = 5$, $\nnWidth{\nn_\commFunc'} \leq \max\{3\nStates+5, 2\nFunctions{\tmnuM}\} + 2\nFunctions{\tmnuM}\neurDim$, $\nnWeights{\nn_\commFunc'} \subseteq \pm\{0,1,2,3\} \cup \nnWeights{\tmnuM}$ and 
        \[
            \nn_\commFunc'(x,z) = (\nn_\commFunc(\proj_{\iota_1}(x,z)), \nn_{f_1}(\proj_\iota(x,z)), \ldots, \nn_{f_{\nFunctions{\tmnuM}}}(\proj_\iota(x,z))) = (\nn_\commFunc(x), \nn_{f_1}(z), \ldots, \nn_{f_{\nFunctions{\tmnuM}}}(z)),
        \]
        for every $x \in \R^{\nStates + 2}$ and $z \in \R^\neurDim$. In particular, for $x = (\oneHot{n}{q}, \cantorMap(\tape))$ and $z = \neurState$, we have
        \begin{align*}
            \nn_\commFunc'(\oneHot{n}{q},\cantorMap(\tape),\neurState) & = (\nn_\commFunc(\oneHot{n}{q},\cantorMap(\tape)), \nn_{f_1}(\neurState), \ldots, \nn_{f_{\nFunctions{\tmnuM}}}(\neurState))\\
            & = (\Delta_\commFunc(\oneHot{n}{q},\cantorMap(\tape)), f_1(\neurState), \ldots, f_{\nFunctions{\tmnuM}}(\neurState)),
        \end{align*}
        for every $q \in \{1,\ldots, \nStates\}$, $\tape \in \workSymbols^\Z$ and $\neurState \in \R^\neurDim$. Now, by Lemma \ref{lem:selection-function}, there exists a neural network $\nn_g$ such that $\nnDepth{\nn_g} = 4$, $\nnWidth{\nn_g} = 4\nFunctions{\tmnuM}(\neurDim+1)$, $\nnWeights{\nn_g} \subseteq \pm\{0,1,C\}$ and $\nn_g(x,z) = g_{C,\nFunctions{\tmnuM},\neurDim}(x,z)$ for every $x \in \R$ and $z \in \R^{\nFunctions{\tmnuM}\neurDim}$. Therefore, by Lemma \ref{lemma:composition-of-neural-networks} applied to $\nn_\commFunc'$ and $\nn_g$, there exists a neural network $ \nn_\commFunc$ such that $\nnDepth{ \nn_\commFunc} = 9$, $\nnWidth{ \nn_\commFunc} \leq \max\{3\nStates, 2\nFunctions{\tmnuM}\neurDim\} + 2\nFunctions{\tmnuM}\neurDim + 5$ and $\nnWeights{ \nn_\commFunc} \subseteq \pm\{0,1,2,3,C\} \cup \pm\nnWeights{\tmnuM}$, and
        \begin{align*}
             \nn_\commFunc(x,z) & = \nn_g(\nn_\commFunc'(x,z)) = \nn_g(\nn_\commFunc(x), \nn_{f_1}(z), \ldots, \nn_{f_{\nFunctions{\tmnuM}}}(z)),
        \end{align*}
        for every $x \in \R^{\nStates + 2}$ and $z \in \R^\neurDim$. In particular, for $(x,z) = \configMap{\tmnuM}(c)$ for some $c := (q,\tape,\neurState) \in \Bc_\tmnuM(C)$, we have
        \begin{align*}
            \nn_\commFunc(\configMap{\tmnuM}(c)) =  \nn_\commFunc(\oneHot{n}{q},\cantorMap(\tape), \neurState) & = \nn_g(\nn_\commFunc(\oneHot{n}{q},\cantorMap(\tape)), \nn_{f_1}(\neurState), \ldots, \nn_{f_{\nFunctions{\tmnuM}}}(\neurState))\\
            & = g_{C,\nFunctions{\tmnuM},\neurDim}(\Delta_\commFunc(\oneHot{n}{q},\cantorMap(\tape)), f_1(\neurState), \ldots, f_{\nFunctions{\tmnuM}}(\neurState))\\
            & \overset{\eqref{eq:definition-of-F-neural-state}}= F_{\neurState,C}(\configMap{\tmnuM}(c)).
        \end{align*}
    \end{enumerate}
    To conclude the proof, we apply Lemma \ref{lemma:parallelization-of-neural-networks} with $d = \nStates + 2 + \neurDim$, $m = 3$, $\nn_1 = \nn_q$, $\nn_2 = \nn_\tape$, $\nn_3 = \nn_\commFunc$, and $\iota_1,\iota_2,\iota_3 = \iota : \{1, \dots, \nStates+2+ \neurDim\} \to \{1, \dots, \nStates + 2 + \neurDim\}, i \mapsto i$, there exists a neural network $\nn$ such that $\nnDepth{\nn} = 18$, $\nnWidth{\nn} \leq \max\{3\nStates, \nFunctions{\tmnuM},\nStates + \neurDim\} + 5 + \max\{5\nStates, \nFunctions{\tmnuM}\} + 27 + \max\{3\nStates, 2\nFunctions{\tmnuM}\neurDim\} + 2\nFunctions{\tmnuM}\neurDim + 5 \leq \max\{5\nStates, \nStates + 2\nFunctions{\tmnuM}\neurDim\} + 2\nFunctions{\tmnuM}\neurDim + 37$ and $\nnWeights{\nn} \subseteq \pm\{0,1/4,1,2,3,4,C\} \cup \pm\nnWeights{\tmnuM}$, and
    \begin{align*}
        \nn(x,y,z) & = (\nn_q(\proj_{\iota}(x,y,z)), \nn_\tape(\proj_{\iota}(x,y,z)), \nn_\commFunc(\proj_{\iota}(x,y,z)))\\
        & = (\nn_q(x,y,z), \nn_\tape(x,y,z), \nn_\commFunc(x,y,z)),
    \end{align*}
    for every $x \in \R^\nStates$, $y \in \R^2$ and $z \in \R^\neurDim$. In particular, for $c = (q,\tape,\neurState) \in \Bc_\tmnuM(C)$, we have
    \begin{align*}
        \nn(\configMap{\tmnuM}(c)) & = (\nn_q(\oneHot{n}{q}), \nn_\tape(\oneHot{n}{q},\cantorMap(\tape),\neurState), \nn_\commFunc(\oneHot{n}{q},\cantorMap(\tape),\neurState))\\
        & = (q, F_\tape(\oneHot{n}{q},\cantorMap(\tape),\neurState), F_{\neurState,C}(\oneHot{n}{q},\cantorMap(\tape),\neurState)) = F_\tmnuM(c).
    \end{align*}
    This concludes the proof.
\end{proof}

Note that $\Scale$ only uses the functions $\neurState \mapsto \neurState$ and $\neurState \mapsto 2\neurState$ to update its neural state, so $\nFunctions{\Scale} = 3$ (including the function $\ReLU$), and $\tmnuWeights{\Scale} = \{1,2\}$. 

Note that in state 1, the \TMNU uses either the function $\neurState \to (\neurState_2,0)$ or $\neurState \to \neurState$, and in State 2 it uses either the function $\neurState \to \left(\frac{1}{2} \neurState_1, \neurState_2 + \neurState_1\right)$ or $\neurState \to \left(\frac{1}{2} \neurState_1, \neurState_2\right)$. Hence, we have $\nFunctions{\Contr^+} = 5$, and $\tmnuWeights{\Contr^+} = \{0,1/2,-1/2, 1\}$. 

Note that $\tmnuM_{\Rc}$ uses exactly the functions $\neurState \mapsto A_x \neurState_{1:d} + b_h$, $\neurState \mapsto A_h \neurState + b_h$, $\ReLU$ and $\neurState \mapsto (A_o, 0) \neurState + (b_o,0)$, so $\nFunctions{\tmnuM_{\Rc}} = 4$ and $\tmnuWeights{\tmnuM_{\Rc}} = \nnWeights{\Rc} \cup \{0\}$.

By a close look to the proofs in Paper Clemens, we have $\nnWeights{\Rc^\times} \subseteq \{0,1/2,-1/2,1,-1\}$, and $\tmnuWeights{\tmnuM_{\Rc^\times}} \subseteq \{0,1/2,-1/2,1\}$.

By inspection of the definition above, we get that $ \nFunctions{\UpPoly} = 1 + 6 + 1 + 3 + 2 = 13, \nnWidth{\UpPoly} = 16, \text{ and } \nnWeights{\UpPoly} \subseteq \{0,1/2,-1/2,1,-1\}$. 

By inspection of the procedure above, we get that $\nFunctions{\Poly} =12$, $\nnWidth{\Poly} = 16$, and $\nnWeights{\Poly} \subseteq \{0,1/2,-1/2,1,-1,2\}$.

By inspection of the procedure above, we get that $\nFunctions{\Continuous} = 14$, and $\nnWeights{\Continuous} = \{0,1/2,-1/2,1,-1,2\}$.